\documentclass[10pt,twocolumn,letterpaper]{article}

\usepackage{iccv}
\usepackage{times}
\usepackage{epsfig}
\usepackage{graphicx}
\usepackage{amsmath}
\usepackage{amssymb}
\usepackage{multirow}
\usepackage{bbding}
\usepackage{color}

\usepackage[breaklinks=true,bookmarks=false]{hyperref}

\iccvfinalcopy 

\def\iccvPaperID{****} 

\ificcvfinal\fi

\begin{document}

\title{DVIS: Decoupled Video Instance Segmentation Framework}

\author{Tao Zhang$^{1}${\qquad}Xingye Tian$^{2}${\qquad}Yu Wu$^{1}${\qquad}Shunping Ji$^{1}$\thanks{ Corresponding author.}\\
Xuebo Wang$^{2}${\qquad}Yuan Zhang$^{2}${\qquad}Pengfei Wan$^{2}$ \vspace{3mm}\\
$^{1}$Wuhan University\qquad$^{2}$Y-tech, Kuaishou Technology 
}

\maketitle
\ificcvfinal\thispagestyle{empty}\fi

\begin{abstract}
Video instance segmentation (VIS) is a critical task with diverse applications, including autonomous driving and video editing. Existing methods often underperform on complex and long videos in real world, primarily due to two factors. Firstly, offline methods are limited by the tightly-coupled modeling paradigm, which treats all frames equally and disregards the interdependencies between adjacent frames. Consequently, this leads to the introduction of excessive noise during long-term temporal alignment. Secondly, online methods suffer from inadequate utilization of temporal information. To tackle these challenges, we propose a decoupling strategy for VIS by dividing it into three independent sub-tasks: segmentation, tracking, and refinement. The efficacy of the decoupling strategy relies on two crucial elements: 1) attaining precise long-term alignment outcomes via frame-by-frame association during tracking, and 2) the effective utilization of temporal information predicated on the aforementioned accurate alignment outcomes during refinement. We introduce a novel referring tracker and temporal refiner to construct the \textbf{D}ecoupled \textbf{VIS} framework (\textbf{DVIS}). DVIS achieves new SOTA performance in both VIS and VPS, surpassing the current SOTA methods by 7.3 AP and 9.6 VPQ on the OVIS and VIPSeg datasets, which are the most challenging and realistic benchmarks. Moreover, thanks to the decoupling strategy, the referring tracker and temporal refiner are super light-weight (only 1.69\% of the segmenter FLOPs), allowing for efficient training and inference on a single GPU with 11G memory. The code is available at \href{https://github.com/zhang-tao-whu/DVIS}{https://github.com/zhang-tao-whu/DVIS}.
\end{abstract}

\begin{figure}[t]
\begin{center}
\includegraphics[width=1.0\linewidth]{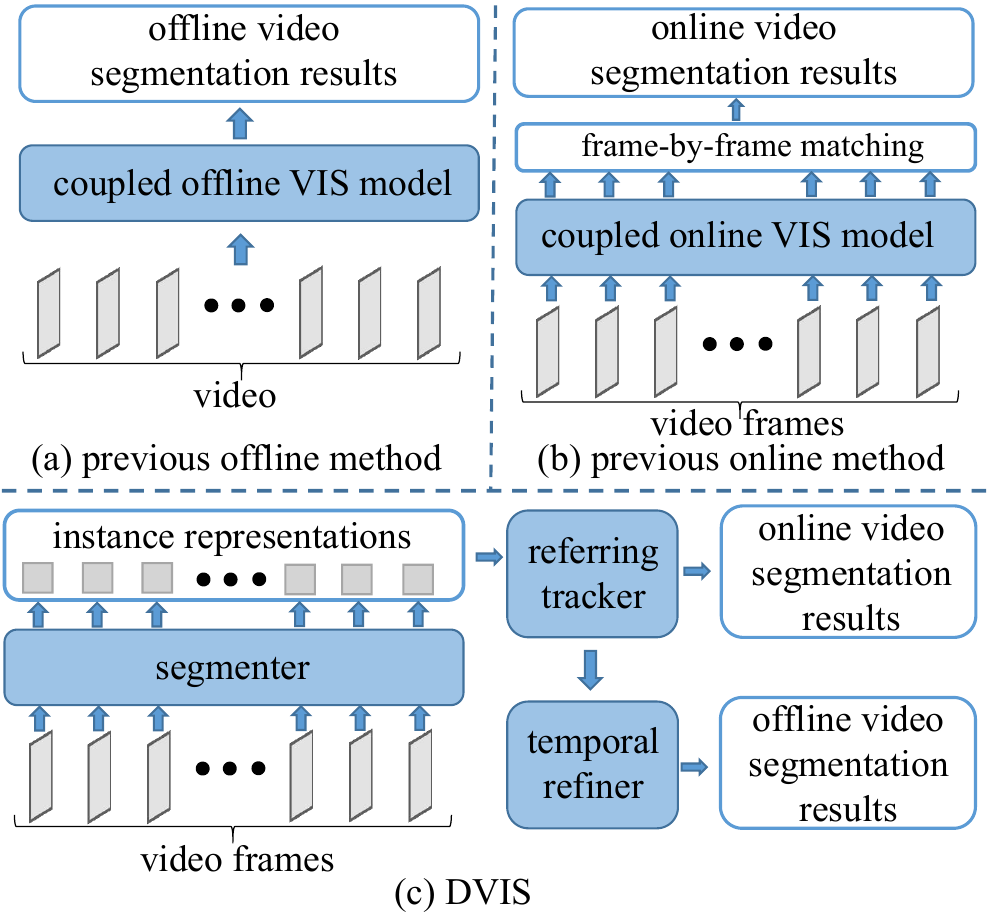}\vspace{-2mm}
\end{center}
\caption{\textbf{Pipelines of previous offline (a), online (b), and proposed DVIS (c) frameworks.} Unlike previous methods that rely on tightly coupled networks, DVIS consists of independent components, including a segmenter, a referring tracker, and a temporal refiner.}\vspace{-1mm}
\label{fig:pipelines}
\end{figure}
\section{Introduction}

Video Instance Segmentation (VIS) is a critical computer vision task that involves identifying, segmenting, and tracking all interested instances in a video simultaneously. This task was first introduced in \cite{masktrackrcnn}. The importance of VIS lies in its ability to facilitate many downstream computer vision applications, such as online autonomous driving and offline video editing.

Previous studies \cite{ifc,mask2former,seqformer,vita} have demonstrated successful performance validation on videos with short durations and simple  scenes \cite{masktrackrcnn}. However, in real-world scenarios, videos often present highly complex scenes, severe instance occlusions, and prolonged durations \cite{ovis}. As a result, these approaches \cite{ifc,mask2former,seqformer,vita} have exhibited poor performance on videos \cite{ovis} that are more representative of real-world scenarios.

We believe that the fundamental reason for the failure of the aforementioned methods \cite{ifc, mask2former, seqformer, vita} lies in the assumption that a coupled network can effectively predict the video segmentation results for any video, irrespective of its length, scene complexity, or instance occlusion levels. In the case of lengthy videos (\textit{e.g.} 100 seconds and 500 frames), with intricate scenes, the same instance may exhibit significant variations in position, shape, and size between the first and last frames \cite{ovis}. Even for experienced humans, accurately associating the same instance in two frames that are separated by a considerable interval is challenging without observing its gradual transformation trajectory over time. Therefore, the alignment/tracking difficulty is significantly increased in complex scenarios and lengthy videos, and even cutting-edge methods such as \cite{mask2formervis} face challenges in achieving convergence \cite{minvis}.

To tackle the aforementioned challenges, we propose to decouple the VIS task into three sub-tasks that are independent of video length and complexity: segmentation, tracking, and refinement. Segmentation aims to extract all appearing objects and obtain their representations from a single frame. Tracking aims to link the same object between adjacent frames. Refinement utilizes all temporal information of the object to optimize both segmentation and association results. Thus we have our decoupled VIS framework, as illustrated in Figure \ref{fig:pipelines} (c). It contains three separate and independent components, \textit{i.e.}, a segmenter, a tracker, and a refiner. Given the extensive research on the segmenter in the field of image instance segmentation, our focus is to design an effective tracker for robustly associating objects across adjacent frames and a refiner for improving the quality of segmentation and tracking. 

To achieve effective instance association, we propose the following principles: (1) encourage sufficient interaction between instance representations of adjacent frames to fully exploit their similarity for better association. (2) avoid mixing their information during the interaction process to prevent introducing indistinguishable noise that may interfere with the association results. Current SOTA methods, such as \cite{idol,minvis}, violate principle 1 by utilizing heuristic algorithms to match adjacent frame instance representations without any interaction, resulting in a significant performance gap compared to our method. While \cite{genvis,rovis} achieve interaction between instance representations of adjacent frames by passing instance representations, they violate principle 2. Following both principles, we designed the Referring Cross Attention (RCA) module, which serves as the core component of our highly effective referring tracker. RCA is a modified version of standard cross-attention \cite{detr} that introduces identification to avoid the blending of instance representations in consecutive frames and efficiently utilize their similarities.
We further propose a novel temporal refiner that leverages 1D convolution and self-attention to effectively integrate temporal information, and cross-attention to correct instance representations. 

An decoupled VIS framework, called DVIS, is then naturally constructed by combining the segmenter, the referring tracker, and the temporal refiner. DVIS achieves new SOTA performance on all the VIS datasets, surpassing previous SOTA method \cite{idol} by 7.3 AP on the most challenging OVIS dataset \cite{ovis}. Additionally, DVIS can be seamlessly extended to other video segmentation tasks, such as video panoptic segmentation (VPS) \cite{vpsnet}, without any modification. DVIS also achieves new SOTA performance on the video panoptic segmentation dataset VIPSeg~\cite{clippanofcn}, surpassing previous SOTA method \cite{tarvis} by 9.6 VPQ. DVIS achieved \textbf{1st place} in the VPS Track of the PVUW challenge at CVPR 2023.

Our decoupling strategy not only significantly improves the performance of video segmentation, but also dramatically reduces hardware resource requirements. Specifically, our proposed tracker and refiner operate exclusively on the instance representations output by the segmenter, avoiding the significant computational cost associated with interacting with image features. As a result, the computation cost of the tracker and refiner is negligible (only 5.18\%/1.69\% of the segmenter with R50/Swin-L backbone). Thanks to the decoupling design of the VIS task and framework, the tracker and refiner can be trained separately while keeping other components frozen. These advantages allow DVIS to be trained on a single GPU with 11G memory.

In summary , our main contributions are:
\begin{itemize}
\item We investigate the failure reasons of current methods on complex and lengthy real-world videos, and we address these challenges by introducing a novel decoupling strategy for VIS, which involves decomposing it into three decoupled sub-tasks: segmentation, tracking, and refinement.
\item Following the decoupling strategy, we propose DVIS, which includes a simple yet effective referring tracker and temporal refiner to produce precision alignment results and efficiently utilize temporal information, respectively.
\item DVIS achieves new SOTA performance in both VIS and VPS, as validated on five major benchmarks: OVIS \cite{ovis}, YouTube-VIS \cite{masktrackrcnn} 2019, 2021, and 2022, as well as VIPSeg \cite{clippanofcn}. Notably, DVIS significantly reduces the resources required for video segmentation, enabling efficient training and inference on a single GPU with 11G memory.
\end{itemize}
\section{Related Works}

\textbf{Online Video Instance Segmentation.} Most mainstream online VIS methods follow a pipeline of segmenting and associating instances. MaskTrack R-CNN \cite{masktrackrcnn} incorporates a tracking head based on \cite{maskrcnn} and associates instances in adjacent frames using multiple cues such as similarity score, semantic consistency, spatial correlation, and detection confidence. \cite{sipmask} replaces the segmenter in the above pipeline with a one-stage instance segmentation network. \cite{crossvis} proposes a crossover learning scheme that segments the same instances in another frame using the instance features of the current frame. With stronger segmenters and the widespread application of transformers in vision tasks \cite{mask2former,detr}, recent works such as \cite{minvis,idol} have achieved outstanding performance. \cite{minvis} proposes a minimal VIS framework based on \cite{mask2former} that achieves instance association by measuring the similarity between the same instances in adjacent frames. \cite{idol} introduces contrastive learning in VIS to obtain a more discriminative instance representation. \cite{rovis,genvis} completely remove heuristic matching algorithms by delivering instance representations and modeling inter-frame association. Inspired by \cite{efficientvis, genvis, minvis, TrackFormer}, DVIS also performs tracking based on instance representations, which significantly reduces memory requirements. Our proposed DVIS introduces a novel component called the referring tracker, which models inter-frame association by denoising current instance representations with the help of previous frame instance representations.
\begin{figure*}[t]
\begin{center}
\includegraphics[width=0.98\linewidth]{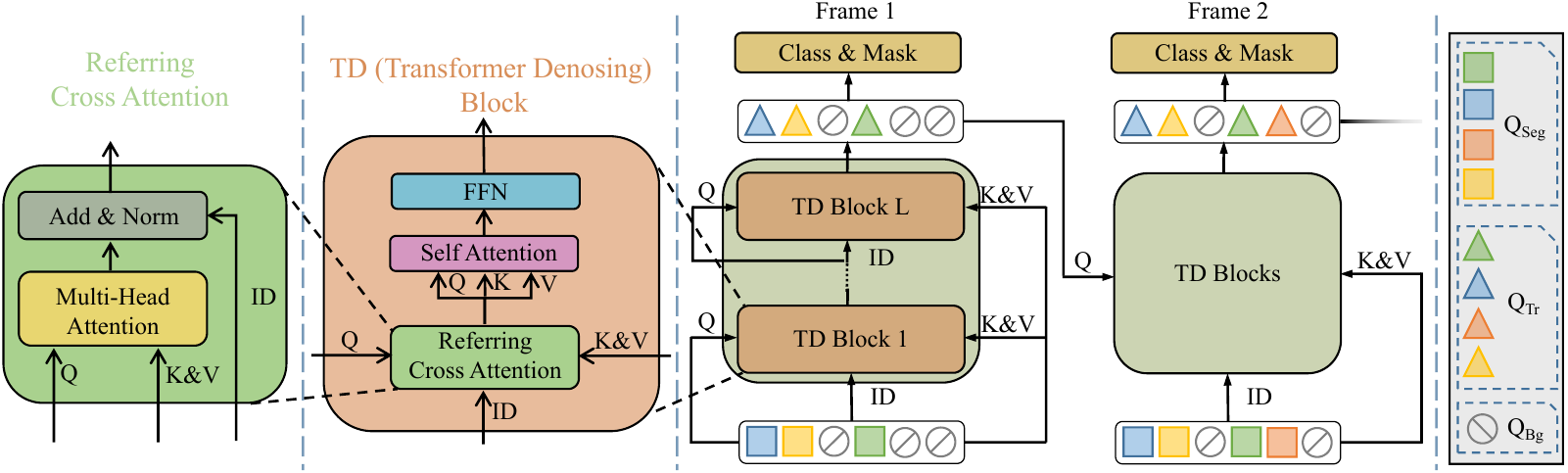}\vspace{-2mm}
\end{center}
\caption{\textbf{The framework of the referring tracker.} The instance representations output by the segmenter ($Q_{seg}$) and referring tracker ($Q_{Tr}$) are represented by squares and triangles, respectively. Instances with the same ID are assigned the same color.}\vspace{-3mm}
\label{fig:online_tracker}
\end{figure*}

\textbf{Offline Video Instance Segmentation.} Previous offline video instance segmentation (VIS) methods have used various approaches to model the spatio-temporal representations of instances in the video. In \cite{stemseg}, instance spatio-temporal embeddings are modeled using a 3D CNN. The first transformer-based VIS architecture proposed in \cite{vistr} uses learnable initial embeddings for each instance of each frame, making it challenging to model instances with complex motion trajectories. \cite{ifc} introduces inter-frame communication, which reduces computation and memory overhead while improving performance. By directly modeling a video-level representation for each instance, \cite{mask2formervis} achieves impressive results. \cite{seqformer} constructs a VIS framework based on deformable attention \cite{deformabledetr} to separate temporal and spatial interactions between instance representations and videos. To significantly reduce memory consumption and enable offline methods to handle long videos, \cite{vita} constructs the video-level instance representation from the instance representations of each frame. While \cite{genvis} implements a semi-online VIS framework by replacing frames with clips, no significant gains were observed compared to the online version. The current SOTA methods for VIS have been demonstrated to overlook the importance of the refinement sub-task. Specifically, the refinement process has been neglected by \cite{idol,minvis,rovis,genvis}, while \cite{ifc,mask2formervis,seqformer,vita} exhibit a lack of clear separation between refinement and other aspects of the segmentation and tracking sub-tasks. Our proposed DVIS achieves SOTA performance by decoupling the VIS task and designing an efficient temporal refiner to fully utilize the information of the overall video.
\section{Method}
By reflecting on and summarizing the shortcomings of \cite{minvis,vita}, we have proposed DVIS, a novel decoupled framework for VIS that consists of three independent components: a segmenter, a referring tracker, and a temporal refiner, illustrated in Figure \ref{fig:pipelines}(c). Specifically, we use Mask2Former \cite{mask2former} as the segmenter in DVIS. The referring tracker is introduced in Section \ref{sec:online}, while the temporal refiner is presented in Section \ref{sec:offline}.

\subsection{Referring Tracker}\label{sec:online}
The referring tracker models the inter-frame association as a referring denoising task. The referring cross-attention is the core component of the referring tracker that effectively utilizes the similarity between instance representations of adjacent frames while avoiding their mixture.

\textbf{Architecture.} Figure \ref{fig:online_tracker} illustrates the architecture of the referring tracker. It takes in the instance queries $\{Q_{seg}^{i} | i \in [1, T]\}$ generated by the segmenter and outputs the instance queries $\{Q_{Tr}^{i} | i \in [1, T]\}$ corresponding to the instances in the previous frame for the current frame, where $T$ is the length of the video. Firstly, the hungarian matching algorithm \cite{hungarian} is employed to match $Q_{seg}$ of adjacent frames, as is done in \cite{minvis}:
\begin{equation}
  \left\{
  \begin{aligned}
  \tilde{Q}_{seg}^{i} & = \text{Hungarian}(\tilde{Q}_{seg}^{i-1},Q_{seg}^{i}), \quad i\in[2,T] \\
  \tilde{Q}_{seg}^{i} & = Q_{seg}^{i}, \quad i=1
  \end{aligned}
  \right.
\label{eq:1},
\end{equation}
where $\tilde{Q}_{seg}$ is the matched intance queries of the segmenter. The hungarian matching algorithm is not strictly necessary and omitting it results in only a slight performance degradation, as shown in Section \ref{sec:ablation}. $\tilde{Q}_{seg}$ can be considered as the tracking result with noise and serves as the initial query for the referring tracker. To denoise the initial query $\tilde{Q}_{seg}^{i}$ of the current frame, the online tracker uses the denoised instance queries $Q_{Tr}^{i-1}$ from the previous frame as a reference.

The objective of the referring tracker is to refine the initial value with noise, which may contain incorrect tracking results, and produce accurate tracking results. The referring tracker comprises a sequence of $L$ transformer denoising blocks, each of which consists of a referring cross-attention, a standard self-attention, and a feedforward network (FFN).

The referring cross-attention (RCA) is a crucial component of the denoising block, designed to capture the correlation between the current frame and its historical frames. Since the instance representations in adjacent frames are highly similar but differ in position, shape, size, etc., using the previous frame's instance representation as the initial instance representation for the current frame (as done by \cite{rovis, genvis}) can introduce ambiguous information that makes the denoising task more difficult. RCA overcomes this issue by introducing identification (ID), while still effectively utilizing the similarity between the query (Q) and key (K) to generate the correct output. As shown in Figure \ref{fig:online_tracker}, RCA is inspired by \cite{referring} and differs only slightly from the standard cross-attention:
\begin{equation}
RCA(ID,Q,K,V)=ID+MHA(Q,K,V)
\label{eq:2}.
\end{equation}
$MHA$ refers to Multi-Head Attention \cite{transformer}, while $ID$, $Q$, $K$, and $V$ denote identification, query, key, and value, respectively. 

Finally, the denoised instance query $Q_{Tr}$ is utilized as an input for the class head and mask head, which produce the category and mask coefficient output, respectively.

\textbf{Losses.} The referring tracker tracks instances frame by frame, and as such, the network is supervised using a loss function that aligns with this paradigm. Specifically, the instance label and prediction $\hat{y}_{Tr}$ are only matched on the frame where the instance first appears. To expedite convergence during the early training phase, the prediction of the frozen segmenter $\hat{y}_{seg}$ is used for matching instead of the referring tracker's prediction.
\begin{equation}
  \left\{
  \begin{aligned}
  & \hat{\sigma}=\arg \underset{\sigma}{\min} \ \ \sum^{N}_{i=1} \ \mathcal{L}_{match}(y_{i}^{f(i)},\hat{y}_{\sigma(i)}^{f(i)}) \\
  & \hat{y} = \hat{y}_{Tr} \ \ if \ \ Iter \geq \frac{Max\_Iter}{2} \ \ else \ \  \hat{y}_{seg}
  \end{aligned}
  \right.
\label{eq:3},
\end{equation}
where $f(i)$ represents the frame in which the $i$-th instance first appears. $\mathcal{L}_{match}(y_{i}^{f(i)},\hat{y}_{\sigma(i)}^{f(i)})$ is a pair-wise matching cost, as used in \cite{mask2formervis}, between the ground truth $y$ and the prediction $\hat{y}$ having index $\sigma(i)$ on the $f(i)$ frame.

The loss function $\mathcal{L}$ is exactly the same as that in \cite{mask2formervis}.
\begin{equation}
\mathcal{L}_{Tr} = \sum^{T}_{t=1} \ \sum^{N}_{i=1} \mathcal{L}(y_{i}^{t}, {\hat{y}}_{\hat{\sigma}(i)}^{t})
\label{eq:4}.
\end{equation}

\subsection{Temporal Refiner}\label{sec:offline}
\begin{figure}[t]
\begin{center}
\includegraphics[width=1.00\linewidth]{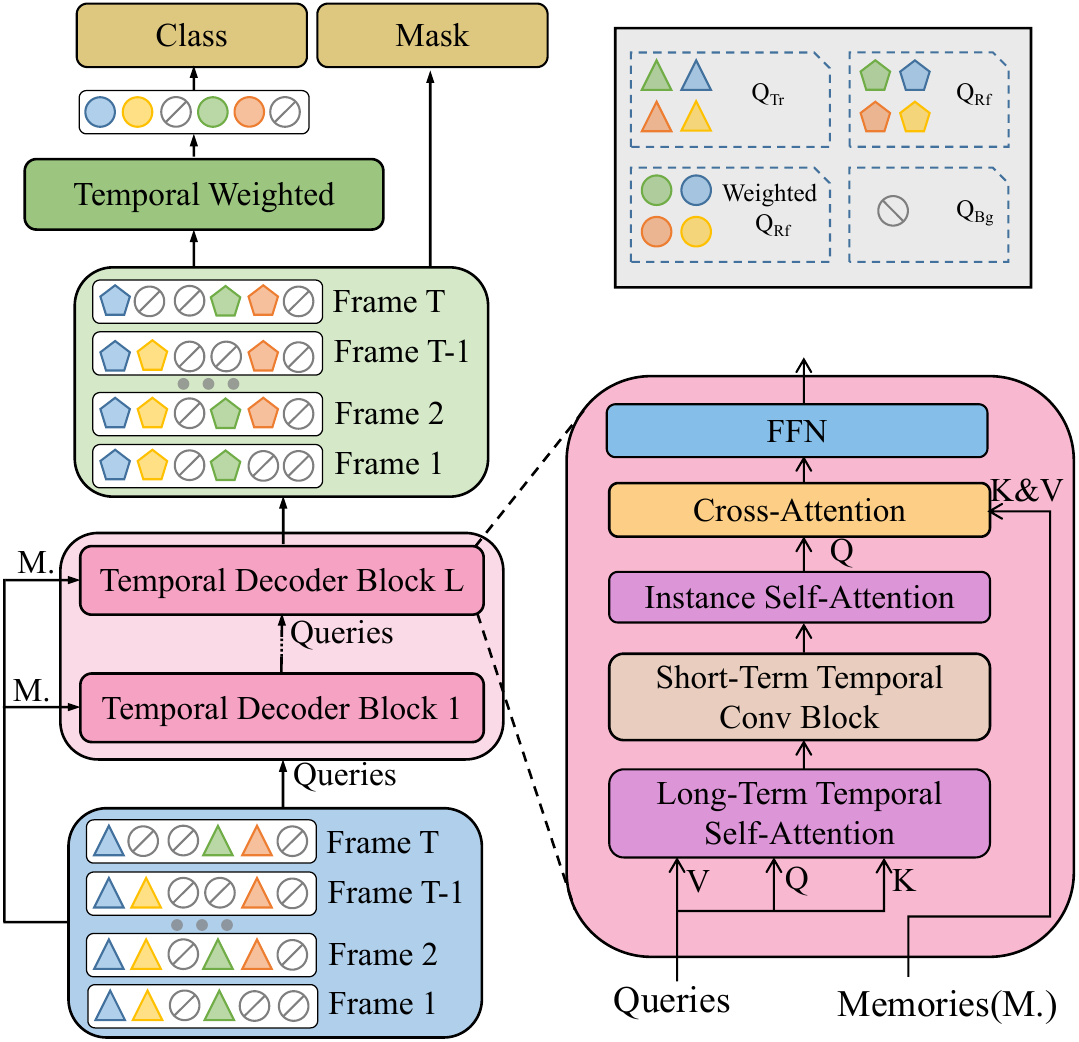}\vspace{-2mm}
\end{center}
\caption{\textbf{The framework of the temporal refiner.} Instance representations for each frame ($Q_{Rf}$) are denoted by pentagons, while the instance representations for the entire video ($\hat{Q}_{Rf}$) are denoted by circles. Different colors indicate different instance IDs.}\vspace{-3mm}
\label{fig:offline_tracker}
\end{figure}
The failure of previous offline video instance segmentation methods can mainly be attributed to the challenge of effectively leveraging temporal information in highly coupled networks. Additionally, previous online video instance segmentation methods lacked a refinement step. To address these issues, we developed an independent temporal refiner to effectively utilize information from the entire video and refine the output of the referring tracker.

\textbf{Architecture.} Figure \ref{fig:offline_tracker} shows the architecture of the temporal refiner. It takes the instance query $Q_{Tr}$ output from the referring tracker as input and outputs the instance query $Q_{Rf}$ after fully aggregating the overall information of the video. The temporal refiner is composed of $L$ temporal decoder blocks that are cascaded together. Each temporal decoder block consists of two main components, namely the short-term temporal convolution block and the long-term temporal attention block. The short-term temporal convolution block exploits motion information while the long-term temporal attention block exploits information from the entire video. These blocks are implemented using 1D convolution and standard self-attention, respectively, and both operate in the time dimension.  

Lastly, the mask coefficients for each instance in each frame are predicted by the mask head based on the refined instance query $Q_{Rf}$. The class head predicts the category and score for each instance across the entire video, using the temporal weighting of $Q_{Rf}$. The temporal weighting process can be defined as follows:
\begin{equation}
\hat{Q}_{Rf} = \sum^{T}_{t=1} SoftMax(Linear(Q_{Rf}^{t}))Q_{Rf}^{t}
\label{eq:5},
\end{equation}
where $\hat{Q}_{Rf}$ is the temporal weighting of $Q_{Rf}$.

\textbf{Losses.} The same matching cost and loss functions as \cite{mask2formervis} are used to supervise the temporal refiner during training. The segmenter and referring tracker are frozen during training, and therefore the referring tracker's prediction results are used for matching in the early training phase to guide the network towards faster convergence.
\begin{equation}
  \left\{
  \begin{aligned}
  & \hat{\sigma}=\arg \underset{\sigma}{\min} \ \ \sum^{N}_{i=1} \ \mathcal{L}_{match}(y_{i},\hat{y}_{\sigma(i)}) \\
  & \hat{y} = \hat{y}_{Rf} \ \ if \ \ Iter \geq \frac{Max\_Iter}{2} \ \ else \ \  \hat{y}_{Tr}
  \end{aligned}
  \right.
\label{eq:6},
\end{equation}
where $\hat{y}_{Rf}$ is the prediction result of the temporal refiner. The loss function is:
\begin{equation}
\mathcal{L}_{Rf} = \sum^{N}_{i=1} \mathcal{L}(y_{i}, {\hat{y}}_{\hat{\sigma}(i)})
\label{eq:7}.
\end{equation}
\section{Experiments}
We evaluate the performance of DVIS for VIS on the OVIS \cite{ovis}, YouTube VIS 2019, 2021, and 2022 \cite{masktrackrcnn} datasets, and for VPS on the VIPSeg \cite{clippanofcn} dataset. In Appendix, the descriptions of these datasets can be found in Section \ref{sec:datasets}, while implementation details, including network training and inference settings, are provided in Section \ref{sec:implement}.
\begin{table}[t]
\centering
\setlength{\tabcolsep}{0.5mm}
\begin{tabular}{l|l|l|ccccc}
	\multirow{2}{*}{} & \multicolumn{2}{c|}{\multirow{2}{*}{Method}} & \multicolumn{5}{c}{OVIS} \\
	~ & \multicolumn{2}{c|}{~} & AP &  AP$_{\rm 50}$ & AP$_{\rm 75}$ & AR$_{\rm 1}$ & AR$_{\rm 10}$  \\
	\hline
	\multirow{21}{*}{\rotatebox{90}{Online}} & \multirow{12}{*}{\rotatebox{90}{ResNet50}}& MaskTrack R-CNN \cite{masktrackrcnn}  & 10.8 & 25.3 & 8.5 & 7.9 & 14.9 \\
	~ & ~ & CMaskTrack R-CNN \cite{ovis} & 15.4 & 33.9 & 13.1 & 9.3 & 20.0 \\
	~ & ~ & CrossVIS \cite{crossvis}& 14.9 & 32.7 & 12.1 & 10.3 & 19.8 \\
	~ & ~ & VISOLO \cite{visolo}& 15.3 & 31.0 &  13.8 & 11.1 & 21.7 \\
	~ & ~ &  MinVIS \cite{minvis}& 25.0 & 45.5 & 24.0 & 13.9 & 29.7 \\
	~ & ~ & MinVIS$^{\dag}$ \cite{minvis} & 26.4 & 49.6 & 25.2 & 13.3 & 31.1 \\
	~ & ~ & IDOL \cite{idol}& 28.2 & 51.0 & 28.0 & 14.5 & \textbf{38.6} \\
	~ & ~ & IDOL$^{\dag}$ \cite{idol} & 30.2 & 51.3 & 30.0 & 15.0 & 37.5 \\ 
	~ & ~ & ROVIS \cite{rovis}& 30.2 & 53.9 & 30.1 & 13.6 & 36.3 \\
	~ & ~ & Ours & 30.2 & \textbf{55.0} & 30.5 & 14.5 & 37.3 \\
	~ & ~ & Ours$^{\dag}$ & \textbf{31.0} & 54.8 & \textbf{31.9} & \textbf{15.2} & 37.6 \\
	\cline{2-8}
	~ & \multirow{11}{*}{\rotatebox{90}{Swin-L}} & MinVIS \cite{minvis} & 39.4 & 61.5 & 41.3 & 18.1 & 43.3 \\
	~ & ~ & MinVIS$^{\dag}$ \cite{minvis} & 41.6 & 65.2 & 42.8 & 19.3 & 45.1 \\
	~ & ~ & IDOL \cite{idol}& 40.0 & 63.1 & 40.5 & 17.6 & 46.4 \\
	~ & ~ & IDOL$^{\dag}$ \cite{idol} & 42.6 & 65.7 & 45.2 & 17.9 & 49.6 \\
	~ & ~ & ROVIS \cite{rovis}& 41.6 & 65.0 & 42.9 & 18.7 & 46.9 \\
	~ & ~ & ROVIS$^{\dag}$ \cite{rovis} & 42.6 & 64.7 & 42.6 & 18.4 & 49.1 \\
	~ & ~ & GenVis$^{\star}$ \cite{genvis} & 45.2 & 69.1 & 48.4 & 19.1 & 48.6 \\
	~ & ~ & Ours & 45.9 & 71.1 & 48.3 & 18.5 & 51.5 \\
	~ & ~ & Ours$^{\dag}$ & \textbf{47.1} & \textbf{71.9} & \textbf{49.2} & \textbf{19.4} & \textbf{52.5} \\
	\hline 
	\multirow{12}{*}{\rotatebox{90}{Offline}} & \multirow{7}{*}{\rotatebox{90}{ResNet50}} & IFC \cite{ifc} & 13.1 & 27.8 & 11.6 & 9.4 & 23.9 \\
	~ & ~ & SeqFormer \cite{seqformer}& 15.1 & 31.9 & 13.8 & 10.4 & 27.1 \\
	~ & ~ & Mask2Former-VIS \cite{mask2formervis}& 17.3 & 37.3 & 15.1 & 10.5 & 23.5 \\
	~ & ~ & VITA$^{\star}$ \cite{vita} & 19.6 & 41.2 & 17.4 & 11.7 & 26.0 \\
	~ & ~ & Ours & 33.8 & \textbf{60.4} & \textbf{33.5} & 15.3 & 39.5 \\
	~ & ~ & Ours$^{\dag}$ & \textbf{34.1} & 59.8 & 32.3 & \textbf{15.9} & \textbf{41.1} \\
	\cline{2-8}
	~ & \multirow{7}{*}{\rotatebox{90}{Swin-L}} & VITA$^{\star}$ \cite{vita} & 27.7 & 51.9 & 24.9 & 14.9 & 33.0 \\
	~ & ~ & Mask2Former-VIS \cite{mask2formervis}& 25.8 & 46.5 & 24.4 & 13.7 & 32.2 \\
	~ & ~ & GenVIS$^{\star}$ \cite{genvis} & 45.4 & 69.2 & 47.8 & 18.9 & 49.0 \\
    ~ & ~ & MDQE$^{\dag}$ \cite{mdqe} & 42.6 & 67.8 & 44.3 & 18.3 & 46.5 \\
	~ & ~ & Ours & 48.6 & 74.7 & 50.5 & 18.8 & 53.8 \\
	~ & ~ & Ours$^{\dag}$ & \textbf{49.9} & \textbf{75.9} & \textbf{53.0} & \textbf{19.4} & \textbf{55.3} \\	
	\hline
 \end{tabular}
 \caption{\textbf{Results on the OVIS validation set.} ${\dag}$ denotes training and evaluation at 720px. ${\star}$ denotes using COCO pseudo videos. The best metrics in each group are bolded.}\vspace{-6mm}
 \label{tab:ovis}
\end{table}
\begin{table*}[t]
\centering
\setlength{\tabcolsep}{2.0mm}
\vspace{-4mm}
\begin{tabular}{l|l|l|ccccc|ccccc}
	\multirow{2}{*}{} & \multirow{2}{*}{Method} & \multirow{2}{*}{Backbone} &  \multicolumn{5}{|c}{Youtube-VIS 2019} &  \multicolumn{5}{|c}{Youtube-VIS 2021} \\
	~ & ~ & ~ & AP &  AP$_{\rm 50}$ & AP$_{\rm 75}$ &  AR$_{\rm 1}$ & AR$_{\rm 10}$ & AP &  AP$_{\rm 50}$ & AP$_{\rm 75}$ &  AR$_{\rm 1}$ & AR$_{\rm 10}$ \\
	\hline
	\multirow{10}{*}{\rotatebox{90}{Online}} & MaskTrack R-CNN \cite{masktrackrcnn} & ResNet-50 & 30.3 & 51.1 & 32.6 & 31.0 & 35.5 & 28.6 & 48.9 & 29.6 & 26.5 & 33.8 \\
	~ & SipMask \cite{sipmask} & ResNet-50 & 33.7 & 54.1 & 35.8 & 35.4 & 40.1 & 31.7 & 52.5 & 34.0 & 30.8 & 37.8 \\
	~ & CrossVIS \cite{crossvis}& ResNet-50 & 36.3 & 56.8 & 38.9 & 35.6 & 40.7 & 34.2 & 54.4 & 37.9  & 30.4 & 38.2\\
	~ & VISOLO \cite{visolo}& ResNet-50 & 38.6 & 56.3 & 43.7 & 35.7 & 42.5 & 36.9 & 54.7 & 40.2 & 30.6 & 40.9 \\
	~ & MinVIS \cite{minvis}& ResNet-50 & 47.4 & 69.0 & 52.1 & 45.7 & 55.7 & 44.2 & 66.0 & 48.1 & 39.2 & 51.7 \\
	~ & IDOL \cite{idol}& ResNet-50 & 49.5 & \textbf{74.0} & 52.9 & \textbf{47.7} & 58.7 & 43.9 & 68.0 & 49.6 & 38.0 & 50.9 \\
	~ & Ours & ResNet-50 & \textbf{51.2} & 73.8 & \textbf{57.1} & 47.2 & \textbf{59.3} & \textbf{46.4} & \textbf{68.4} & \textbf{49.6} & \textbf{39.7} & \textbf{53.5} \\
	\cline{2-13}
	~ & MinVIS \cite{minvis}& Swin-L & 61.6 & 83.3 & 68.6 & 54.8 & 66.6 & 55.3 & 76.6 & 62.0 & 45.9 & 60.8 \\
	~ & IDOL \cite{idol}& Swin-L & \textbf{64.3} & \textbf{87.5} & \textbf{71.0} & 55.6 & \textbf{69.1} & 56.1 & \textbf{80.8} & 63.5 & 45.0 & 60.1 \\
	~ & Ours & Swin-L & 63.9 & 87.2 & 70.4 & \textbf{56.2} & 69.0 & \textbf{58.7} & 80.4 & \textbf{66.6} & \textbf{47.5} & \textbf{64.6}\\ 
	\hline
	\multirow{11}{*}{\rotatebox{90}{Offline}} & EfficientVIS \cite{efficientvis} & ResNet-50 & 37.9 & 59.7 & 43.0 & 40.3 & 46.6 & 34.0 & 57.5 & 37.3 & 33.8 & 42.5 \\
	~ & IFC \cite{ifc}& ResNet-50 & 41.2 & 65.1 & 44.6 & 42.3 & 49.6 & 35.2 & 55.9 & 37.7 & 32.6 & 42.9 \\
	~ & Mask2Former-VIS \cite{mask2formervis}& ResNet-50 & 46.4 & 68.0 & 50.0 & - & - & 40.6 & 60.9 & 41.8 & - & -\\
	~ & SeqFormer \cite{seqformer}& ResNet-50 & 47.4 & 69.8 & 51.8 & 45.5 & 54.8 & 40.5 & 62.4 & 43.7 & 36.1 & 48.1 \\
	~ & VITA \cite{vita}& ResNet-50 & 49.8 & 72.6 & 54.5 & \textbf{49.4} & \textbf{61.0} & 45.7 & 67.4 & 49.5 & \textbf{40.9} & 53.6 \\
	~ & Ours & ResNet-50 & \textbf{52.6} & \textbf{76.5} & \textbf{58.2} & 47.4 & 60.4 & \textbf{47.4} & \textbf{71.0} & \textbf{51.6} & 39.9 & \textbf{55.2} \\
	\cline{2-13}
	~ & SeqFormer \cite{seqformer}& Swin-L & 59.3 & 82.1 & 66.4 & 51.7 & 64.4 & 51.8 & 74.6 & 58.2 & 42.8 & 58.1 \\
	~ & Mask2Former-VIS \cite{mask2formervis}& Swin-L & 60.4 & 84.4 & 67.0 & - & - & 52.6 & 76.4 & 57.2 & - & - \\
	~ & VITA \cite{vita}& Swin-L & 63.0 & 86.9 & 67.9 & 56.3 & 68.1 & 57.5 & 80.6 & 61.0 & 47.7 & 62.6 \\
	~ & Ours & Swin-L & \textbf{64.9} & \textbf{88.0} & \textbf{72.7} & \textbf{56.5} & \textbf{70.3} & \textbf{60.1} & \textbf{83.0} & \textbf{68.4} & \textbf{47.7} & \textbf{65.7} \\
\hline
 \end{tabular}
 \caption{\textbf{Results on the validation set of YouTube-VIS 2019 \& 2021.} The best metrics in each group are bolded.}\vspace{-5mm}
 \label{tab:ytvis}
\end{table*}
\begin{table}[t]
\centering
\begin{tabular}{l|l|ccccc}
	\multirow{2}{*}{} & \multicolumn{1}{c|}{\multirow{2}{*}{Method}} & \multicolumn{5}{c}{YouTube-VIS 2022} \\
	~ & \multicolumn{1}{c|}{~} & AP &  AP$_{\rm 50}$ & AP$_{\rm 75}$ & AR$_{\rm 1}$ & AR$_{\rm 10}$  \\
	\hline
	\multirow{3}{*}{\rotatebox{90}{Swin-L}} & VITA \cite{vita}  & 41.1 & 63.0 & 44.0 & 39.3 & 44.3 \\
	~ & MinVIS \cite{minvis} &  33.1 & 54.8 & 33.7 & 29.5 & 36.6\\
	~ &  Ours & \textbf{45.9} & 69.0 & \textbf{48.8} & 37.2 & \textbf{51.8} \\
\hline
 \end{tabular}\vspace{1mm}
 \caption{\textbf{Results on the YouTube-VIS 2022 long videos.} The best metrics in each group are bolded.}\vspace{-4mm}
 \label{tab:ytvis22}
\end{table}
\begin{table}[t]
\centering
\setlength{\tabcolsep}{0.4mm}
\begin{tabular}{l|l|cccc}
	\multirow{2}{*}{} & \multicolumn{1}{c|}{\multirow{2}{*}{Method}} & \multicolumn{4}{c}{VIPSeg} \\
	~ & \multicolumn{1}{c|}{~} & VPQ &  VPQ$^{\rm Th}$ & VPQ$^{\rm St}$ & STQ   \\
	\hline
	\multirow{9}{*}{R50} & VPSNet \cite{vpsnet}  & 14.0 & 14.0 & 14.2 & 20.8 \\
	~ & VPSNet-SiamTrack \cite{siamtrack} &  17.2 & 17.3 & 17.3 & 21.1 \\
	~ & VIP-Deeplab \cite{vipdeeplab} &  16.0 & 12.3 & 18.2 & 22.0 \\
	~ & Clip-PanoFCN \cite{clippanofcn} & 22.9 & 25.0 & 20.8 & 31.5 \\
	~ & Video K-Net \cite{videoknet} & 26.1 & - & - & 31.5 \\
	~ & TarVIS \cite{tarvis} & 33.5 & 39.2 & 28.5 & \textbf{43.1} \\
    ~ & Tube-Link \cite{tubelink} & 39.2 & - & - & 39.5 \\
    ~ & Video-kMax \cite{videokmax} & 38.2 & - & - & 39.9 \\
	~ & Ours & \textbf{43.2} & \textbf{43.6} & \textbf{42.8} & 42.8 \\
	\hline
	\multirow{2}{*}{Swin-L} & TarVIS \cite{tarvis} & 48.0 & 58.2 & 39.0 & 52.9 \\
	~ &  Ours & \textbf{57.6} & \textbf{59.9} & \textbf{55.5} & \textbf{55.3}\\
\hline
 \end{tabular}
 \caption{\textbf{Results on the VIPSeg dataset.} The best metrics in each group are bolded.}\vspace{-6mm}
 \label{tab:vipseg}
\end{table}

\subsection{Main Results}
We compare DVIS with current SOTA online and offline VIS methods on the OVIS, YouTube-VIS 2019, 2021, and 2022 datasets. When compared with online methods, DVIS will discard the temporal refiner as it utilizes information from future frames, in order to maintain a fair comparison. The results are reported in Tables \ref{tab:ovis}, \ref{tab:ytvis}, and \ref{tab:ytvis22}, respectively. We adopt MinVIS \cite{minvis} as our baseline because DVIS is essentially identical to MinVIS after removing the referring tracker and temporal refiner. We also compare DVIS with current SOTA methods for VPS, and the results are shown in Table \ref{tab:vipseg}. The visualization of DVIS's prediction results on these datasets is available in Figures \ref{fig:ovis demo}, \ref{fig:ytvis demo}, and \ref{fig:vipseg demo} of the Appendix.

\textbf{Performance on the OVIS Dataset.} In online mode, DVIS achieves 31.0 AP with ResNet50 and 47.1 AP with Swin-L on the OVIS validation set, outperforming the baseline MinVIS \cite{minvis} by 4.6 AP and 5.5 AP, respectively. The referring tracker has shown significant performance gains, especially for medium and heavily occluded objects, as discussed in Section \ref{sec:ablation}. DVIS outperforms the current SOTA online VIS methods IDOL \cite{idol} and RO-VIS \cite{rovis} by 4.5 AP. This demonstrates the successful design of the referring tracker for robust tracking results particularly in heavily occluded scenarios.

In offline mode, DVIS achieves 34.1 AP with ResNet50 and 49.9 AP with Swin-L on the OVIS validation set, surpassing DVIS running in online mode by 3.1 AP and 2.8 AP, respectively. More impressively, DVIS outperforms the baseline MinVIS by 8.3 AP. Additionally, DVIS surpasses the previous pure offline VIS methods Mask2Former-VIS \cite{mask2formervis} and VITA \cite{vita} by 24.1 AP and 22.2 AP, respectively. Thus, DVIS achieves a new SOTA performance, demonstrating the superiority of the decoupled framework in complex scenarios compared to the previous coupled framework.

\textbf{Performance on the YouTube-VIS 2019 and 2021 Datasets.} In online mode, DVIS achieved 51.2 AP with ResNet50 and 63.9 AP with Swin-L on the YouTube-VIS 2019 validation set, outperforming MinVIS \cite{minvis} by 3.8 AP and 2.3 AP, respectively. For YouTube-VIS 2021 validation set, DVIS achieved 46.4 AP with ResNet50 and 58.7 AP with Swin-L in online mode, outperforming MinVIS \cite{minvis} by 2.2 AP and 3.4 AP, respectively. On the YouTube-VIS datasets, DVIS running in online mode shows comparable performance with the current SOTA method IDOL \cite{idol}.

In offline mode, DVIS achieved 52.6 AP with ResNet50 and 64.9 AP with Swin-L on the YouTube-VIS 2019 validation set, outperforming DVIS running in online mode by 1.4 AP and 1.0 AP. Similarly, on the YouTube-VIS 2021 validation set, DVIS achieved 47.4 AP with ResNet50 and 60.1 AP with Swin-L, outperforming DVIS (online mode) by 1.0 AP and 1.4 AP. DVIS achieves a new SOTA performance on the YouTube-VIS 2019 and 2021 datasets.

\textbf{Performance on the YouTube-VIS 2022 Dataset.} DVIS achieves a new SOTA performance of 52.8 AP on the validation set of the YouTube-VIS 2022 dataset, with 59.6 AP on short videos and 45.9 AP on long videos. Since the short videos of the YouTube-VIS 2022 dataset largely overlap with the YouTube-VIS 2021 dataset, we compare the performance of DVIS with other methods only on long videos, as shown in Table \ref{tab:ytvis22}. DVIS outperforms the baseline method MinVIS \cite{minvis} by 12.8 AP and the current SOTA method VITA \cite{vita} by 4.8 AP.

\textbf{Performance on the VIPSeg Dataset.} DVIS achieves 43.2 VPQ and 57.6 VPQ on the VIPSeg validation set when using ResNet50 and Swin-L backbones, respectively, surpassing the current SOTA VPS method TarVIS \cite{tarvis} by 9.7 VPQ and 9.6 VPQ. These results demonstrate the outstanding performance of DVIS on video panoptic segmentation (VPS) and its potential to achieve SOTA performance on all video segmentation tasks.

\subsection{Ablation Experiments}\label{sec:ablation}
Ablation experiments were conducted on the OVIS dataset, with DVIS evaluated using ResNet50 and input resized to 360p unless otherwise specified.

\begin{table}[t]
\setlength{\tabcolsep}{0.8mm}
\centering
\begin{tabular}{l|cccc}
	 & AP$_{\rm all}$ &  AP$_{\rm l}$ & AP$_{\rm m}$ & AP$_{\rm h}$  \\
	\hline
	baseline & 41.6 & 64.8 & 49.0 & 20.5 \\
	+Tracker & 47.1\textcolor{green}{(+5.5)} & 64.7\textcolor{red}{(-0.1)} & 54.2\textcolor{green}{(+5.2)} & 24.8\textcolor{green}{(+4.3)} \\
	+Refiner & 49.9\textcolor{green}{(+8.3)} & 67.1\textcolor{green}{(+2.3)} & 56.0\textcolor{green}{(+7.0)} & 29.8\textcolor{green}{(+9.4)} \\
\hline
 \end{tabular}
\caption{\textbf{Ablation study of the proposed components.} The baseline is MinVIS \cite{minvis}. All models use Swin-L as the backbone and are evaluated on the OVIS validation set with 720p input. AP$_{\rm l}$, AP$_{\rm m}$ and AP$_{\rm h}$ refer to the AP of the light, medium, and heavily occluded instances, respectively.}\vspace{-1mm}
 \label{tab:comp}
\end{table}
\begin{figure*}[t]
\begin{center}
\vspace{-5mm}\includegraphics[width=1.02\linewidth]{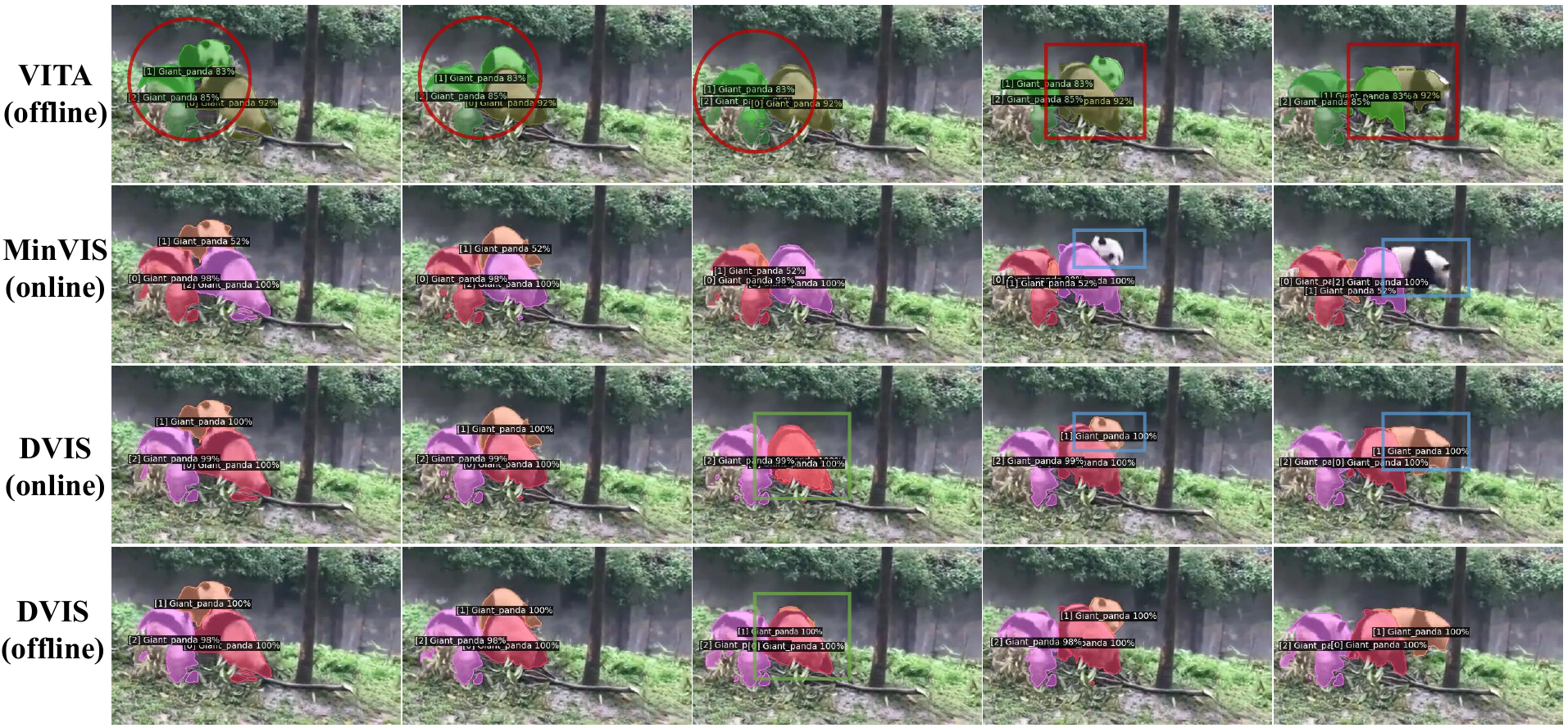}
\end{center}\vspace{-5mm}
\caption{\textbf{Visualization results comparing DVIS with current SOTA online and offline VIS methods.} VITA shows poor segmentation quality (highlighted with red circles) and tracking stability (highlighted with red rectangles). The referring tracker demonstrates strong tracking ability (highlighted with blue rectangles), while the temporal refiner effectively utilizes contextual information from previous and future frames (highlighted with green rectangles).}\vspace{-1mm}
\label{fig:comp demos}
\end{figure*}
\textbf{Effectiveness of Referring Tracker and Temporal Refiner.} We conducted ablation experiments on the OVIS dataset to evaluate the effectiveness of the referring tracker and temporal refiner. The results of the experiments are presented in Table \ref{tab:comp}. Our findings indicate that the referring tracker leads to significant performance gains when processing medium and heavily occluded objects, resulting in an increase of 5.2 AP$_{\rm m}$ and 4.3 AP$_{\rm h}$, respectively. However, there is a slight decrease of 0.1 AP$_{\rm l}$ in the case of lightly occluded objects, indicating that the improvement in the referring tracker is primarily in tracking quality rather than segmentation quality. We further illustrate our findings by presenting an instance of a completely occluded panda with ID 1 in the third frame, which is tracked well by DVIS but not by MinVIS \cite{minvis}, in Figure \ref{fig:comp demos}.

The temporal refiner leads to performance gains in both segmentation quality and tracking quality, with improvements of 2.3 AP$_{\rm l}$, 7.0 AP$_{\rm m}$, and 9.4 AP$_{\rm h}$ across the board. The temporal refiner effectively utilizes the entire video information, leading to more significant improvements for heavily occluded objects, as demonstrated in Figure \ref{fig:comp demos} where the green rectangles highlight a highly occluded panda. Despite this challenge, the temporal refiner produces accurate segmentation results, while the referring tracker fails due to its inability to leverage the full video information.
\begin{table}[t]
\centering
\begin{tabular}{l|ccccc}
	Initial Value & AP &  AP$_{50}$ & AP$_{75}$ & AR$_{1}$ & AR$_{10}$  \\
	\hline
	Zero & 28.9 & 52.5 & 27.9 & 14.7 & 35.7 \\ %
	$Q_{Tr}^{pre}$ & 28.3 & 50.1 & 27.3 & 14.5 & 33.8\\ %
	$Q_{seg}$ &  29.8 & 54.3 & 28.3 & 14.8 & 36.5 \\ %
	Matched $Q_{seg}$ & \textbf{30.5} & \textbf{54.7} & \textbf{30.1} & \textbf{15.0} & \textbf{36.5}\\ %
\hline
 \end{tabular}
 \caption{\textbf{Ablation study of the initial instance representation in the referring tracker.} $Q_{Tr}^{pre}$ denotes the instance representation in the previous frame, and $Q_{seg}$ denotes the instance representation output by the segmenter.}\vspace{-3mm}
 \label{tab:noisy init}
\end{table}

\textbf{Initial Instance Representation of Referring Tracker.} Our proposed referring tracker represents the VIS as referring denoising, making it crucial to select an appropriate initial value with noise. We evaluate the performance with different initial values and report the results in Table \ref{tab:noisy init}. The best performance is achieved when using the $Q_{seg}$ obtained by matching with the hungarian algorithm as the initial value. When zero is used as the initial value, the denoising task becomes a more challenging reconstruction problem, leading to a drop of 1.6 AP. Using the $Q_{Tr}$ of the previous frame as the initial value results in a 2.2 AP performance degradation, as it contains too much interference information. The network also performs well when using the unmatched $Q_{seg}$, where the initial values of the instance queries of each frame are linked by the learnable prior information, demonstrating the robustness of the referring tracker.

\begin{table}[t]
\centering
\begin{tabular}{l|ccccc}
	Cross Attn Type & AP &  AP$_{50}$ & AP$_{75}$ & AR$_{1}$ & AR$_{10}$  \\
	\hline
	Standard & 2.9 & 5.0 & 2.8 & 2.4 & 3.4 \\
	Referring & \textbf{30.5} & \textbf{54.7} & \textbf{30.1} & \textbf{15.0} & \textbf{36.5} \\
\hline
 \end{tabular}
 \caption{\textbf{Ablation study on the type of cross-attention in the referring tracker.}}\vspace{-2mm}
 \label{tab:cross attn}
\end{table}

\textbf{Referring Cross-Attention.} The referring cross-attention is a crucial component of the referring tracker, responsible for linking historical frames with the current frame. We evaluated the importance of the referring cross-attention by comparing it to the standard cross-attention, where $ID$ is set to $Q$ in Equation \ref{eq:2}. The results in Table \ref{tab:cross attn} demonstrate that replacing the referring cross-attention with the standard cross-attention leads to an extreme drop in performance. This finding highlights the critical role of inter-frame associations modeled by the referring cross-attention in the success of the referring tracker.

\begin{table}[t]
\centering
\setlength{\tabcolsep}{1.4mm}
\begin{tabular}{l|ccccc}
	 & AP &  AP$_{50}$ & AP$_{75}$ & AR$_{1}$ & AR$_{10}$  \\
	\hline
	Baseline & 32.2 & 57.9 & 31.3 & 15.1 & 38.7 \\
	w/o Long-term Attn. & 31.0 & 56.2 & 29.3 & 14.9 & 37.8 \\
	w/o Short-term Conv. & 31.8 & 56.6 & 30.2 & 14.8 & 37.9 \\ 
	w/o Cross Attn. & 30.6 & 55.5 & 28.7 & 14.8 & 36.6 \\
\hline
 \end{tabular}
 \caption{\textbf{Ablation study on the components of the temporal decoder block.} Attn. denotes attention and Conv. denotes convolutional.``w/o" refer to without.}\vspace{-3mm}
 \label{tab:offline decoder}
\end{table}

\textbf{Impact of Different Components of Temporal Decoder Block.} To evaluate the impact of different components of the temporal decoder block, we conducted experiments by removing each component individually and reporting the corresponding performance in Table \ref{tab:offline decoder}. Our results show that removing long-term temporal self-attention led to a performance degradation of 1.2 AP. Although the function of the long-term attention overrides the function of the short-term convolution, removing the short-term convolution still resulted in a performance degradation of 0.4 AP, suggesting that it is beneficial for utilizing information in adjacent frames. Moreover, the removal of cross-attention resulted in a significant drop of 1.6 AP since incorrect instance queries cannot be efficiently corrected without it, even though information from different frames can still be utilized.

\textbf{Performance of DVIS in Semi-Online Mode.} In real-world scenarios, videos are often of infinite length, making it impossible to run VIS models in pure offline mode. We conduct experiments to measure the performance difference between semi-online and offline modes, as shown in Table \ref{tab:clip length} of Appendix. When videos are cut into clips of length 1 as input to DVIS, i.e., no other frame information available for the current frame, the performance was only comparable to that of DVIS without temporal refiner. However, as the clip length increased, the performance of the semi-online mode gradually approached that of the pure offline mode, and achieved comparable performance after the clip length exceeded 80 frames (33.8 $vs.$ 33.8).


\textbf{Computational Cost.} The computational cost of DVIS components was measured by evaluating the parameters, MACs, and inference time of the segmenter, referring tracker, and temporal refiner. Table \ref{tab:consumption} presents the results. When using Mask2Former with ResNet50 and Swin-L as the segmenter, the referring tracker and temporal refiner combined only accounted for 5.18\% and 1.69\% of the segmenter's computation, respectively. This demonstrates that the referring tracker and temporal refiner can efficiently achieve VIS with almost negligible computational cost.
\begin{table}[t]
\centering
\setlength{\tabcolsep}{0.7mm}
\vspace{-6mm}
\begin{tabular}{l|c|c|ccc}
	Component & Inp. & N$_{\rm Q}$ & Params(M) & MACs(G) & Time(ms) \\
	\hline
	M2F(R50) & 480p & 100 & 43.95 & 103.73 & 48.10 \\
	Tracker & 480p & 100 & 9.68 & 1.68 & 7.63 \\
	Refiner & 480p & 100 & 14.41 & 3.69 & 1.11 \\
	\hline
	M2F(SwinL) & 720p & 200 & 215.30 & 851.00 & 275.19 \\
	Tracker & 720p & 200 & 9.68 & 5.13 & 7.97 \\
	Refiner & 720p & 200 & 14.41 & 9.27 & 2.00 \\
\hline
 \end{tabular}
  \caption{\textbf{Computational cost of DVIS components.} M2F refers to the Mask2Former used as the segmenter of DVIS. Inp. denotes the size of the input video, and N$_{\rm Q}$ denotes the number of queries. The inference time per frame is measured on a 1080Ti GPU.}\vspace{-4mm}
 \label{tab:consumption}
\end{table}

\section{Conclusion}\vspace{-2mm}
In this paper, we propose DVIS, a decoupled VIS framework that separates the VIS task into three sub-tasks: segmentation, tracking, and refinement. Our contributions are three-fold: 1) we decouple the VIS task and introduce the DVIS framework, 2) we propose the referring tracker, which enhances tracking robustness by modeling inter-frame associations as referring denoising, and 3) we propose the temporal refiner, which utilizes information from the entire video to refine segmentation results, a capability that was missing in previous methods. Our results show that DVIS achieves SOTA performance on all VIS datasets, outperforming all existing methods, supporting the effectiveness of our decoupling standpoint and the design of DVIS. Additionally, DVIS's SOTA performance on VPS demonstrates its potential and versatility. We believe that DVIS will serve as a strong and fundamental baseline, and our decoupling insights will inspire future works in both online and offline VIS.

{\small
\bibliographystyle{ieee_fullname}
\bibliography{ref}

\begin{thebibliography}{10}\itemsep=-1pt

\bibitem{tarvis}
Ali Athar, Alexander Hermans, Jonathon Luiten, Deva Ramanan, and Bastian Leibe.
\newblock Tarvis: A unified approach for target-based video segmentation.
\newblock {\em arXiv preprint arXiv:2301.02657}, 2023.

\bibitem{stemseg}
Ali Athar, Sabarinath Mahadevan, Aljosa Osep, Laura Leal-Taix{\'e}, and Bastian
  Leibe.
\newblock Stem-seg: Spatio-temporal embeddings for instance segmentation in
  videos.
\newblock In {\em Computer Vision--ECCV 2020: 16th European Conference,
  Glasgow, UK, August 23--28, 2020, Proceedings, Part XI 16}, pages 158--177.
  Springer, 2020.

\bibitem{sipmask}
Jiale Cao, Rao~Muhammad Anwer, Hisham Cholakkal, Fahad~Shahbaz Khan, Yanwei
  Pang, and Ling Shao.
\newblock Sipmask: Spatial information preservation for fast image and video
  instance segmentation.
\newblock In {\em Computer Vision--ECCV 2020: 16th European Conference,
  Glasgow, UK, August 23--28, 2020, Proceedings, Part XIV 16}, pages 1--18.
  Springer, 2020.

\bibitem{detr}
Nicolas Carion, Francisco Massa, Gabriel Synnaeve, Nicolas Usunier, Alexander
  Kirillov, and Sergey Zagoruyko.
\newblock End-to-end object detection with transformers.
\newblock In {\em Computer Vision--ECCV 2020: 16th European Conference,
  Glasgow, UK, August 23--28, 2020, Proceedings, Part I 16}, pages 213--229.
  Springer, 2020.

\bibitem{mask2formervis}
Bowen Cheng, Anwesa Choudhuri, Ishan Misra, Alexander Kirillov, Rohit Girdhar,
  and Alexander~G Schwing.
\newblock Mask2former for video instance segmentation.
\newblock {\em arXiv preprint arXiv:2112.10764}, 2021.

\bibitem{mask2former}
Bowen Cheng, Ishan Misra, Alexander~G Schwing, Alexander Kirillov, and Rohit
  Girdhar.
\newblock Masked-attention mask transformer for universal image segmentation.
\newblock In {\em Proceedings of the IEEE/CVF Conference on Computer Vision and
  Pattern Recognition}, pages 1290--1299, 2022.

\bibitem{visolo}
Su~Ho Han, Sukjun Hwang, Seoung~Wug Oh, Yeonchool Park, Hyunwoo Kim, Min-Jung
  Kim, and Seon~Joo Kim.
\newblock Visolo: Grid-based space-time aggregation for efficient online video
  instance segmentation.
\newblock In {\em Proceedings of the IEEE/CVF Conference on Computer Vision and
  Pattern Recognition}, pages 2896--2905, 2022.

\bibitem{maskrcnn}
Kaiming He, Georgia Gkioxari, Piotr Doll{\'a}r, and Ross Girshick.
\newblock Mask r-cnn.
\newblock In {\em Proceedings of the IEEE international conference on computer
  vision}, pages 2961--2969, 2017.

\bibitem{genvis}
Miran Heo, Sukjun Hwang, Jeongseok Hyun, Hanjung Kim, Seoung~Wug Oh, Joon-Young
  Lee, and Seon~Joo Kim.
\newblock A generalized framework for video instance segmentation.
\newblock {\em arXiv preprint arXiv:2211.08834}, 2022.

\bibitem{vita}
Miran Heo, Sukjun Hwang, Seoung~Wug Oh, Joon-Young Lee, and Seon~Joo Kim.
\newblock Vita: Video instance segmentation via object token association.
\newblock {\em arXiv preprint arXiv:2206.04403}, 2022.

\bibitem{minvis}
De-An Huang, Zhiding Yu, and Anima Anandkumar.
\newblock Minvis: A minimal video instance segmentation framework without
  video-based training.
\newblock {\em arXiv preprint arXiv:2208.02245}, 2022.

\bibitem{ifc}
Sukjun Hwang, Miran Heo, Seoung~Wug Oh, and Seon~Joo Kim.
\newblock Video instance segmentation using inter-frame communication
  transformers.
\newblock {\em Advances in Neural Information Processing Systems},
  34:13352--13363, 2021.

\bibitem{vpsnet}
Dahun Kim, Sanghyun Woo, Joon-Young Lee, and In~So Kweon.
\newblock Video panoptic segmentation.
\newblock In {\em Proceedings of the IEEE/CVF Conference on Computer Vision and
  Pattern Recognition}, pages 9859--9868, 2020.

\bibitem{hungarian}
Harold~W Kuhn.
\newblock The hungarian method for the assignment problem.
\newblock {\em Naval research logistics quarterly}, 2(1-2):83--97, 1955.

\bibitem{mdqe}
Minghan Li, Shuai Li, Wangmeng Xiang, and Lei Zhang.
\newblock Mdqe: Mining discriminative query embeddings to segment occluded
  instances on challenging videos.
\newblock In {\em Proceedings of the IEEE/CVF Conference on Computer Vision and
  Pattern Recognition}, pages 10524--10533, 2023.

\bibitem{tubelink}
Xiangtai Li, Haobo Yuan, Wenwei Zhang, Guangliang Cheng, Jiangmiao Pang, and
  Chen~Change Loy.
\newblock Tube-link: A flexible cross tube baseline for universal video
  segmentation.
\newblock {\em arXiv preprint arXiv:2303.12782}, 2023.

\bibitem{videoknet}
Xiangtai Li, Wenwei Zhang, Jiangmiao Pang, Kai Chen, Guangliang Cheng, Yunhai
  Tong, and Chen~Change Loy.
\newblock Video k-net: A simple, strong, and unified baseline for video
  segmentation.
\newblock In {\em Proceedings of the IEEE/CVF Conference on Computer Vision and
  Pattern Recognition}, pages 18847--18857, 2022.

\bibitem{adamw}
Ilya Loshchilov and Frank Hutter.
\newblock Decoupled weight decay regularization.
\newblock {\em arXiv preprint arXiv:1711.05101}, 2017.

\bibitem{TrackFormer}
Tim Meinhardt, Alexander Kirillov, Laura Leal-Taixe, and Christoph
  Feichtenhofer.
\newblock Trackformer: Multi-object tracking with transformers.
\newblock In {\em Proceedings of the IEEE/CVF conference on computer vision and
  pattern recognition}, pages 8844--8854, 2022.

\bibitem{clippanofcn}
Jiaxu Miao, Xiaohan Wang, Yu Wu, Wei Li, Xu Zhang, Yunchao Wei, and Yi Yang.
\newblock Large-scale video panoptic segmentation in the wild: A benchmark.
\newblock In {\em Proceedings of the IEEE/CVF Conference on Computer Vision and
  Pattern Recognition}, pages 21033--21043, 2022.

\bibitem{ovis}
Jiyang Qi, Yan Gao, Yao Hu, Xinggang Wang, Xiaoyu Liu, Xiang Bai, Serge
  Belongie, Alan Yuille, Philip~HS Torr, and Song Bai.
\newblock Occluded video instance segmentation: A benchmark.
\newblock {\em International Journal of Computer Vision}, 130(8):2022--2039,
  2022.

\bibitem{vipdeeplab}
Siyuan Qiao, Yukun Zhu, Hartwig Adam, Alan Yuille, and Liang-Chieh Chen.
\newblock Vip-deeplab: Learning visual perception with depth-aware video
  panoptic segmentation.
\newblock In {\em Proceedings of the IEEE/CVF Conference on Computer Vision and
  Pattern Recognition}, pages 3997--4008, 2021.

\bibitem{referring}
Seonguk Seo, Joon-Young Lee, and Bohyung Han.
\newblock Urvos: Unified referring video object segmentation network with a
  large-scale benchmark.
\newblock In {\em Computer Vision--ECCV 2020: 16th European Conference,
  Glasgow, UK, August 23--28, 2020, Proceedings, Part XV 16}, pages 208--223.
  Springer, 2020.

\bibitem{videokmax}
Inkyu Shin, Dahun Kim, Qihang Yu, Jun Xie, Hong-Seok Kim, Bradley Green, In~So
  Kweon, Kuk-Jin Yoon, and Liang-Chieh Chen.
\newblock Video-kmax: A simple unified approach for online and near-online
  video panoptic segmentation.
\newblock {\em arXiv preprint arXiv:2304.04694}, 2023.

\bibitem{transformer}
Ashish Vaswani, Noam Shazeer, Niki Parmar, Jakob Uszkoreit, Llion Jones,
  Aidan~N Gomez, {\L}ukasz Kaiser, and Illia Polosukhin.
\newblock Attention is all you need.
\newblock {\em Advances in neural information processing systems}, 30, 2017.

\bibitem{vistr}
Yuqing Wang, Zhaoliang Xu, Xinlong Wang, Chunhua Shen, Baoshan Cheng, Hao Shen,
  and Huaxia Xia.
\newblock End-to-end video instance segmentation with transformers.
\newblock In {\em Proceedings of the IEEE/CVF conference on computer vision and
  pattern recognition}, pages 8741--8750, 2021.

\bibitem{siamtrack}
Sanghyun Woo, Dahun Kim, Joon-Young Lee, and In~So Kweon.
\newblock Learning to associate every segment for video panoptic segmentation.
\newblock In {\em Proceedings of the IEEE/CVF Conference on Computer Vision and
  Pattern Recognition}, pages 2705--2714, 2021.

\bibitem{seqformer}
Junfeng Wu, Yi Jiang, Song Bai, Wenqing Zhang, and Xiang Bai.
\newblock Seqformer: Sequential transformer for video instance segmentation.
\newblock In {\em Computer Vision--ECCV 2022: 17th European Conference, Tel
  Aviv, Israel, October 23--27, 2022, Proceedings, Part XXVIII}, pages
  553--569. Springer, 2022.

\bibitem{idol}
Junfeng Wu, Qihao Liu, Yi Jiang, Song Bai, Alan Yuille, and Xiang Bai.
\newblock In defense of online models for video instance segmentation.
\newblock In {\em Computer Vision--ECCV 2022: 17th European Conference, Tel
  Aviv, Israel, October 23--27, 2022, Proceedings, Part XXVIII}, pages
  588--605. Springer, 2022.

\bibitem{efficientvis}
Jialian Wu, Sudhir Yarram, Hui Liang, Tian Lan, Junsong Yuan, Jayan Eledath,
  and Gerard Medioni.
\newblock Efficient video instance segmentation via tracklet query and
  proposal.
\newblock In {\em Proceedings of the IEEE/CVF Conference on Computer Vision and
  Pattern Recognition}, pages 959--968, 2022.

\bibitem{masktrackrcnn}
Linjie Yang, Yuchen Fan, and Ning Xu.
\newblock Video instance segmentation.
\newblock In {\em Proceedings of the IEEE/CVF International Conference on
  Computer Vision}, pages 5188--5197, 2019.

\bibitem{crossvis}
Shusheng Yang, Yuxin Fang, Xinggang Wang, Yu Li, Chen Fang, Ying Shan, Bin
  Feng, and Wenyu Liu.
\newblock Crossover learning for fast online video instance segmentation.
\newblock In {\em Proceedings of the IEEE/CVF International Conference on
  Computer Vision}, pages 8043--8052, 2021.

\bibitem{rovis}
Zitong Zhan, Daniel McKee, and Svetlana Lazebnik.
\newblock Robust online video instance segmentation with track queries.
\newblock {\em arXiv preprint arXiv:2211.09108}, 2022.

\bibitem{deformabledetr}
Xizhou Zhu, Weijie Su, Lewei Lu, Bin Li, Xiaogang Wang, and Jifeng Dai.
\newblock Deformable detr: Deformable transformers for end-to-end object
  detection.
\newblock {\em arXiv preprint arXiv:2010.04159}, 2020.

\end{thebibliography}
}

\clearpage
\appendix
\begin{center}{\bf \Large Appendix}\end{center}\vspace{-2mm}
\renewcommand{\thetable}{\Roman{table}}
\renewcommand{\thefigure}{\Roman{figure}}
\setcounter{table}{0}
\setcounter{figure}{0}

\section{Datasets}\label{sec:datasets}

\textbf{OVIS.} OVIS \cite{ovis} is a new and most challenging VIS dataset with 25 object categories. It consists of 607/140/154 videos for training/validation/testing, which are longer and have more instance annotations compared to YouTube-VIS \cite{masktrackrcnn}. OVIS also includes a significant number of videos with objects exhibiting severe occlusion, complex motion trajectories, and rapid deformation, making it more representative of real-world scenarios. Thus, OVIS serves as an ideal benchmark for evaluating the performance of various VIS methods.

\textbf{YouTube-VIS 2019, 2021 and 2022.} YouTube-VIS 2019 is a large-scale dataset for VIS that includes 2,238/302/343 videos for training/validation/testing, respectively. To improve the quality of the dataset, YouTube-VIS 2021 was created by expanding the videos and refining the annotations, and it includes 2,985/421/453 videos for training/validation/testing, respectively. Both YouTube-VIS 2019 and 2021 contain 40 object categories, although there are differences in the category labels. The YouTube-VIS 2022 dataset uses the same training set as YouTube-VIS 2021, but includes long videos in the validation and test sets.

\textbf{VIPSeg.} VIPSeg \cite{clippanofcn} is a large-scale dataset for panoptic segmentation in the wild, featuring a broad range of real-world scenarios and 124 categories, including 58 things’ and 66 stuff’s classes. It provides 3,536 videos and 84,750 frames, of which 2,806/343/387 videos are allocated for training/validation/testing, respectively.

\section{Implementation Details}\label{sec:implement}
\textbf{Settings.} DVIS employs \cite{mask2former} as the segmenter. The referring tracker uses 6 transformer denoising blocks, while the temporal refiner uses 6 temporal decoder blocks. These settings are consistent across all datasets.

\textbf{Training.} We use the AdamW optimizer \cite{adamw} with a base learning rate of 1e-4 and weight decay of 5e-2. The segmenter is initialized with weights trained by MinVIS \cite{minvis} and is frozen during training. For the YouTube-VIS 2019, 2021, and 2022 datasets, we use COCO pseudo videos \cite{seqformer} for jointly training DVIS. For the OVIS and VIPSeg datasets, no additional data is used. We first train the referring tracker and then freeze it to train the temporal refiner. Both the referring tracker and the temporal refiner are trained for 20,000/40,000/20,000 iterations on the OVIS/YouTube-VIS/VIPSeg datasets, respectively, with a learning rate reduction factor of 10 at 14,000/28,000/14,000 iterations. The referring tracker and temporal refiner are trained on 8 GPUs each with 5 and 21 continuous frames, respectively.

\begin{table}[t]
\centering
\setlength{\tabcolsep}{1.8mm}
\begin{tabular}{l|c|ccccc}
	~ & Length & AP &  AP$_{\rm 50}$ & AP$_{\rm 75}$ & AR$_{\rm 1}$ & AR$_{\rm 10}$  \\
	\hline
	\multirow{4}{*}{Tracker} & 3 & 28.3 & 52.6 & 26.6 & 14.6 & 34.7 \\
	~ & 5 & \textbf{29.5} & \textbf{54.6} & \textbf{29.5} & 14.1 & 36.3 \\
	~ & 11 & 29.3 & 52.3 & 27.5 & 13.7 & \textbf{36.7} \\
	~ & 21 & 29.4 & 52.9 & 27.7 & \textbf{14.5} & 36.1 \\
	\hline
	\multirow{4}{*}{Refiner} & 5 & 32.1 & 57.8 & 31.7 & 15.5 & 38.4 \\
	~ & 11 & 32.9 & 57.9 & 33.2 & 15.4 & 39.2 \\
	~ & 21 & \textbf{33.8} & \textbf{60.4} & \textbf{33.5} & 15.3 & 39.5 \\
	~ & 25 & 33.6 & 60.2 & 32.6 & \textbf{15.7} & \textbf{40.3}\\
\hline
 \end{tabular}
 \caption{\textbf{Ablation about the length of the training video clips.}}
 \label{tab:training length}
\end{table}

\begin{table}[t]
\centering
\begin{tabular}{l|ccccc}
	Clip Length & AP &  AP$_{50}$ & AP$_{75}$ & AR$_{1}$ & AR$_{10}$  \\
	\hline
	1 & 30.0 & 54.5 & 29.4 & 14.7 & 35.7 \\
	10 & 31.7 & 56.8 & 31.5 & 15.0 & 37.8 \\
	20 & 32.5 & 58.0 & 31.6 & 15.1 & 38.3 \\
	40 & 32.8 & 58.5 & 32.6 & 14.9 & 38.3 \\
	80 & 33.8 & 60.0 & 33.6 & 15.2 & 39.3 \\
 	120 & 33.8 & 60.3 & \textbf{33.7} & 15.2 & \textbf{39.6} \\
	$L_{\text{video}}$ & \textbf{33.8} & \textbf{60.4} & 33.5 & \textbf{15.3} & 39.5 \\
\hline
 \end{tabular}
 \caption{\textbf{Ablation study on the length of video clips for semi-online inference.} $L_{\text{video}}$ denotes the length of the video.}
 \label{tab:clip length}
\end{table}

\textbf{Inference.} Unless otherwise specified, the input video is downscaled to 360p/480p when using a ResNet50/Swin-L backbone during inference.

\section{Additional Ablation Experiments}
\textbf{Impact of Training Video Clip Length on Performance.} We investigate the impact of training video clip length on the performance of DVIS, and present the results in Table \ref{tab:training length}. We observe that the performance of the tracker and temporal refiner improves with increasing training video clip length, but stabilizes when the clip length exceeds 5/21.

\section{Limitations and Future Works}
In this work, we present DVIS, which has achieved SOTA performance on all VIS datasets, particularly on occluded videos. We also demonstrated the feasibility of using DVIS in semi-online mode, where comparable performance to pure offline mode can be achieved when the clip length exceeds 80. However, there remains a significant gap between real-world videos and the videos in current VIS datasets. Real-world videos are potentially infinitely long and contain an infinite number of instances, which is not currently handled well by DVIS. To address this, we plan to design unique processing mechanisms that can handle new and disappearing instances to ensure that DVIS can perfectly handle real-world videos with infinite instances and infinite lengths. This will be a focus of our future work.

\begin{figure*}[t]

\begin{minipage}[c]{1.00\linewidth}
\includegraphics[width=0.163\linewidth]{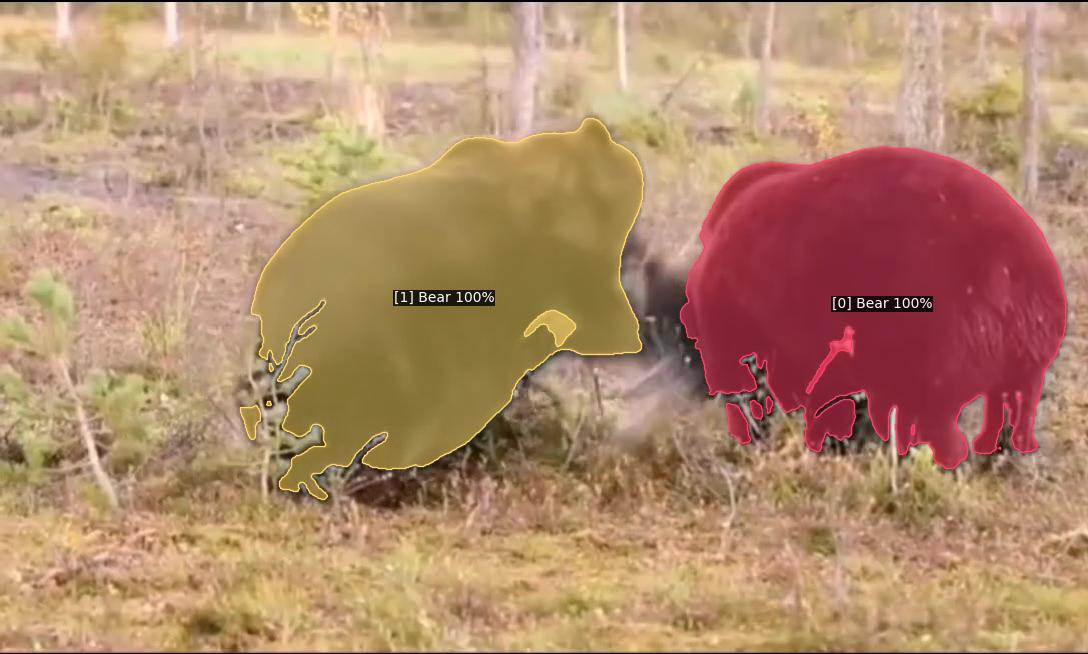}
\includegraphics[width=0.163\linewidth]{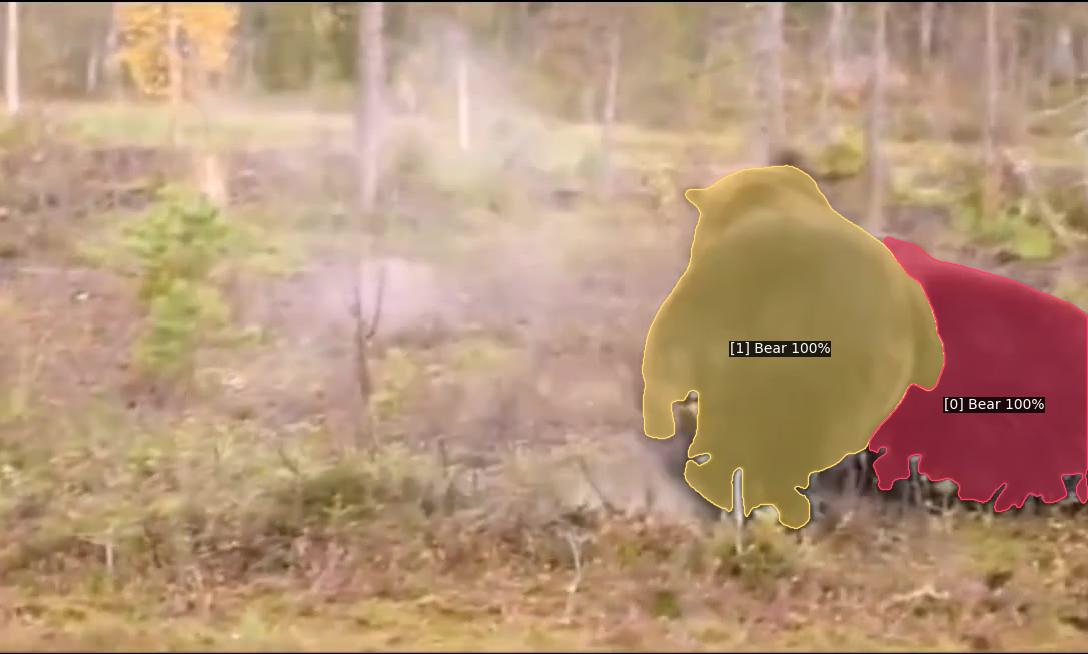}
\includegraphics[width=0.163\linewidth]{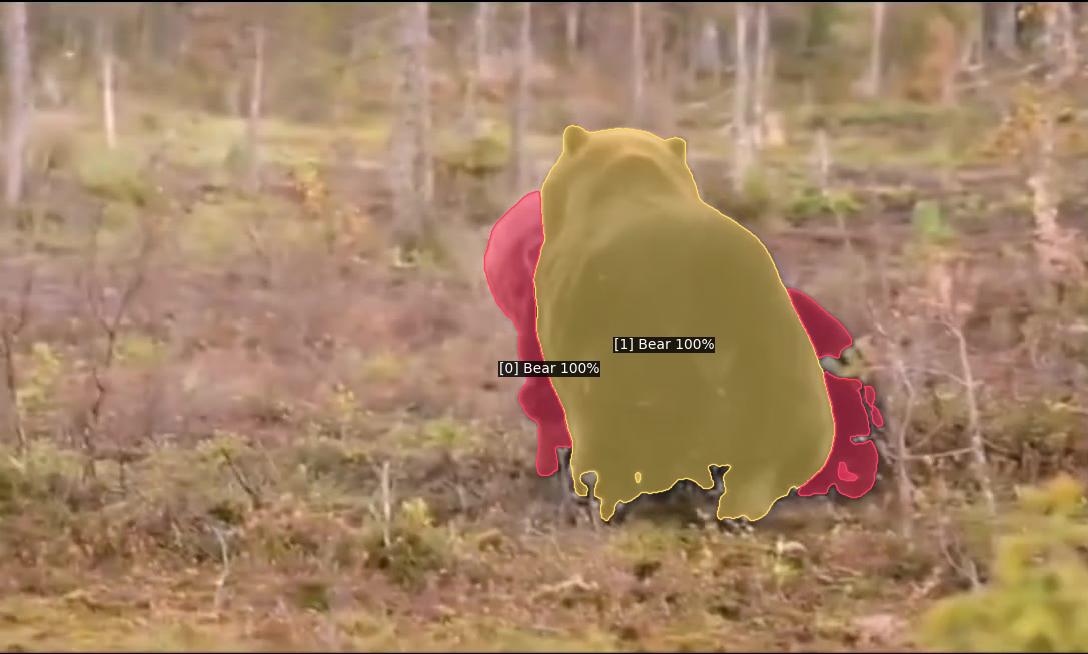}
\includegraphics[width=0.163\linewidth]{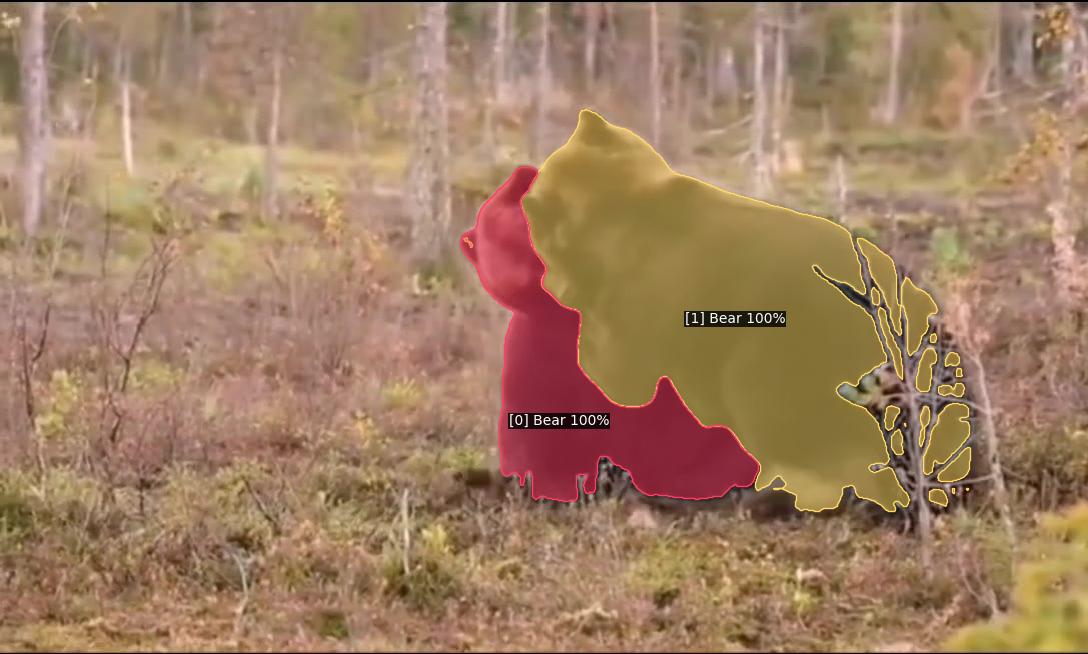}
\includegraphics[width=0.163\linewidth]{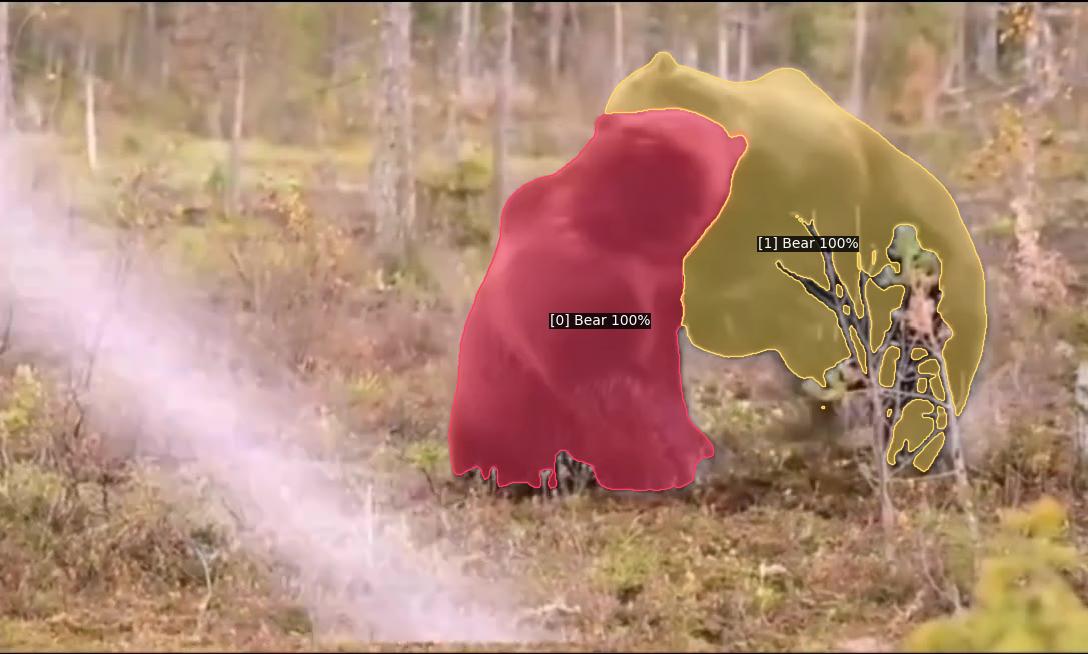}
\includegraphics[width=0.163\linewidth]{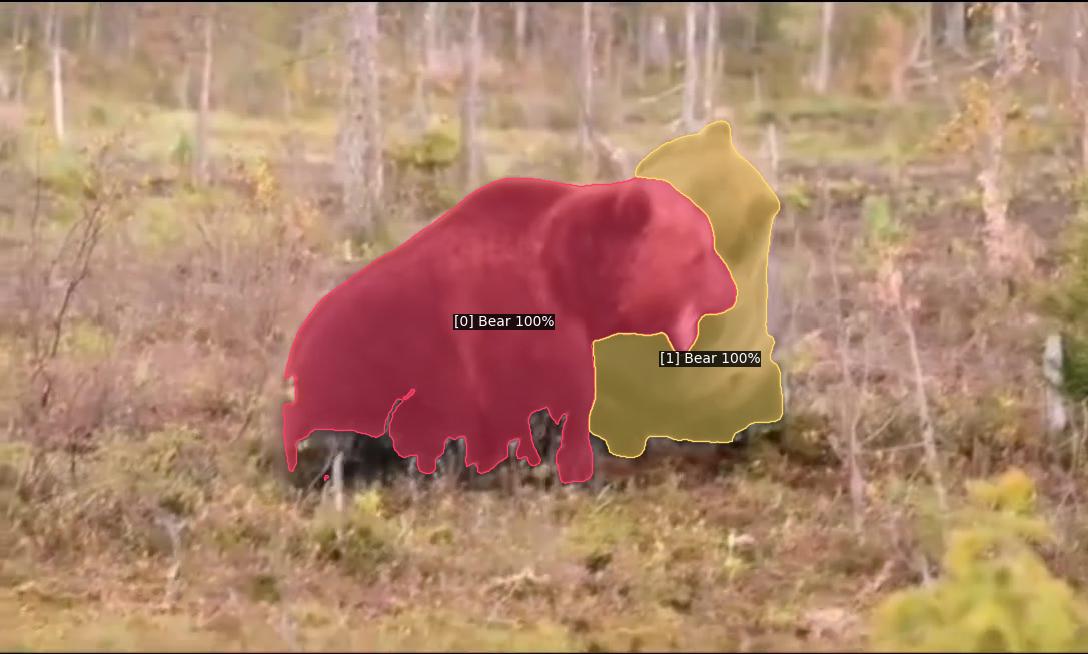}
\end{minipage}\hfill
\begin{minipage}[c]{1.0\linewidth}
\includegraphics[width=0.163\linewidth]{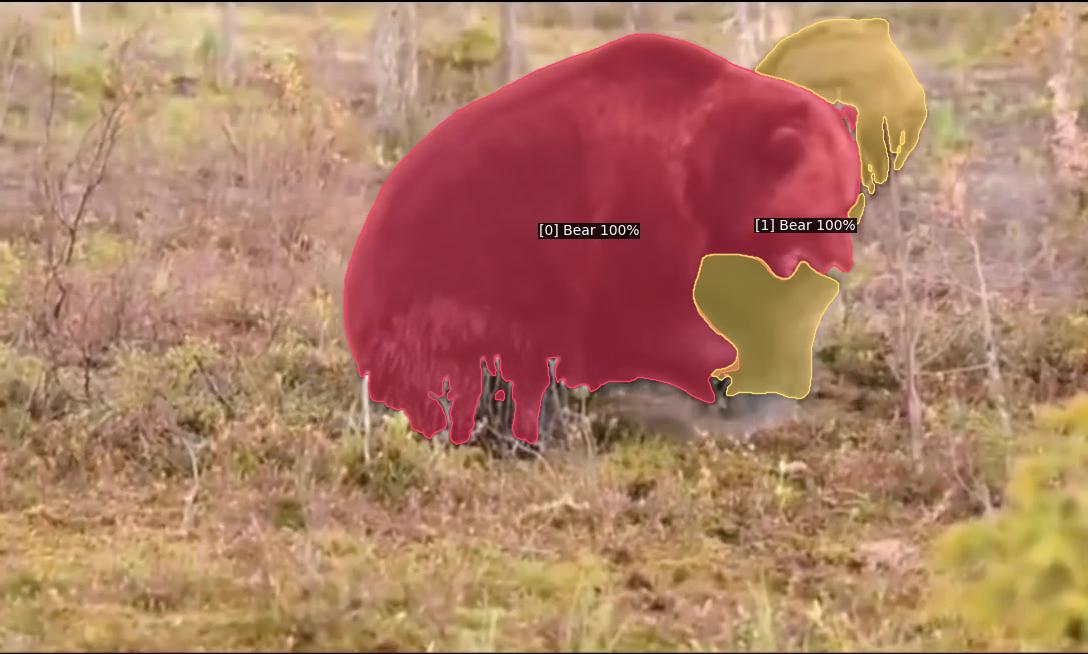}
\includegraphics[width=0.163\linewidth]{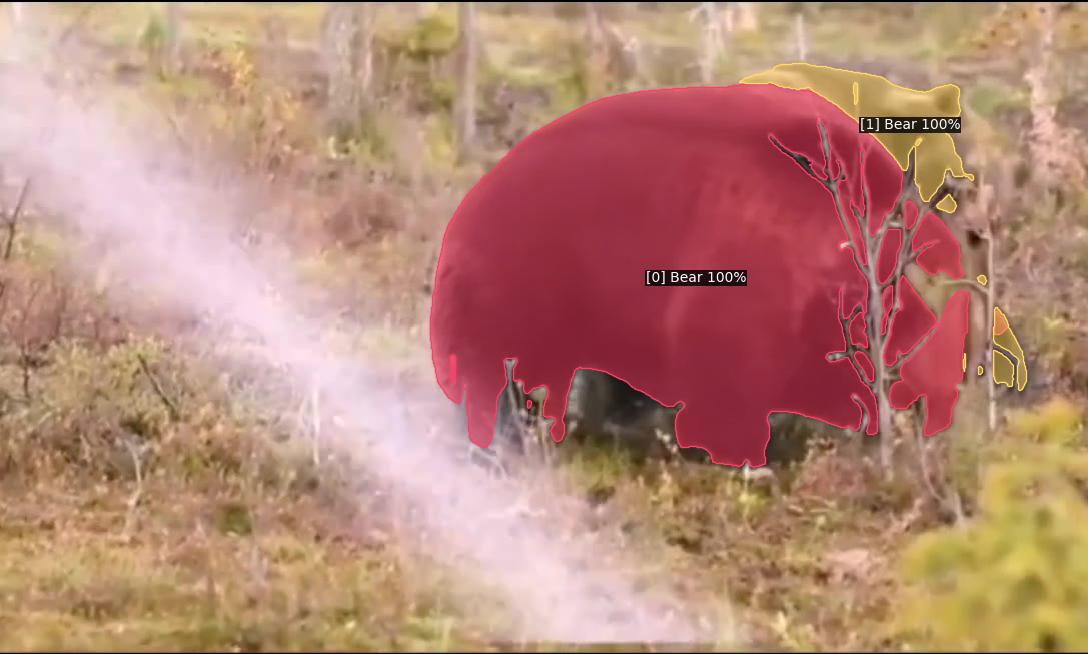}
\includegraphics[width=0.163\linewidth]{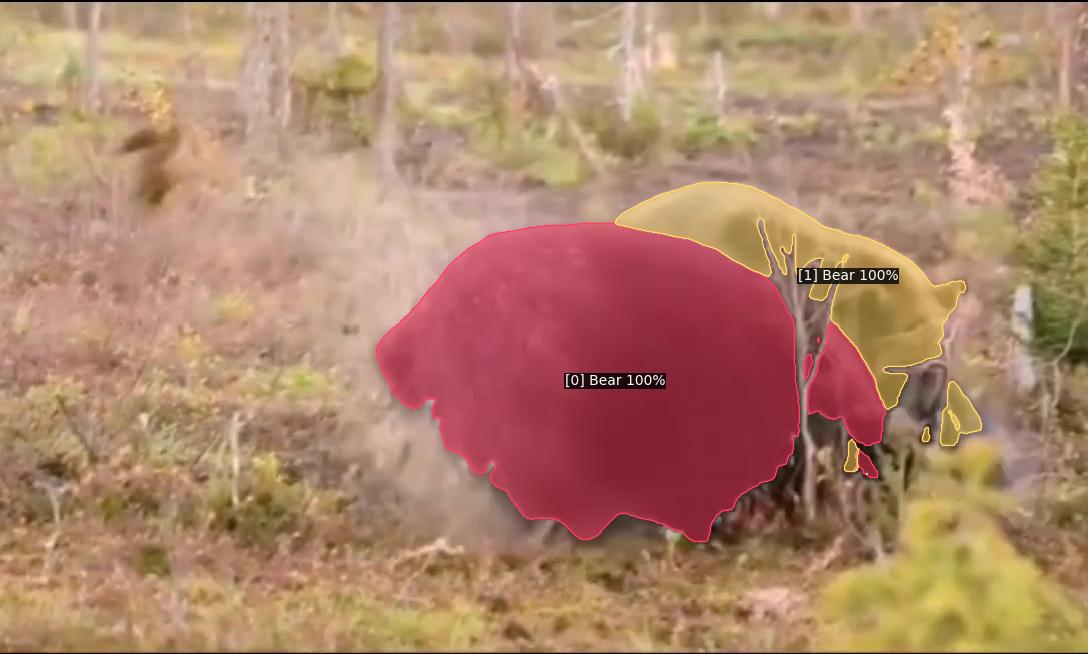}
\includegraphics[width=0.163\linewidth]{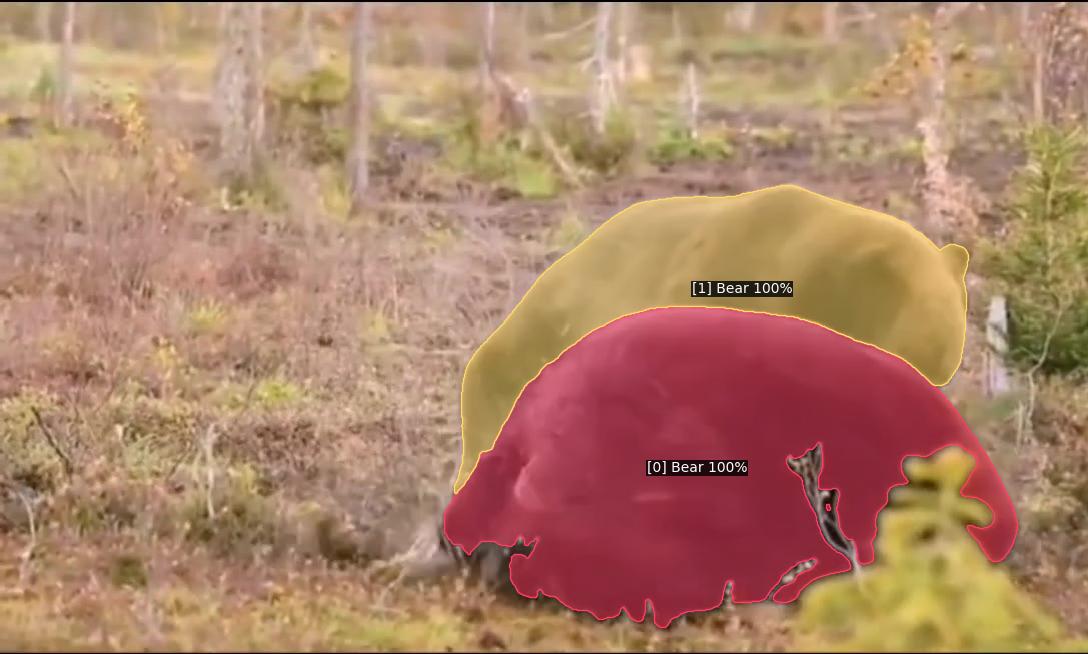}
\includegraphics[width=0.163\linewidth]{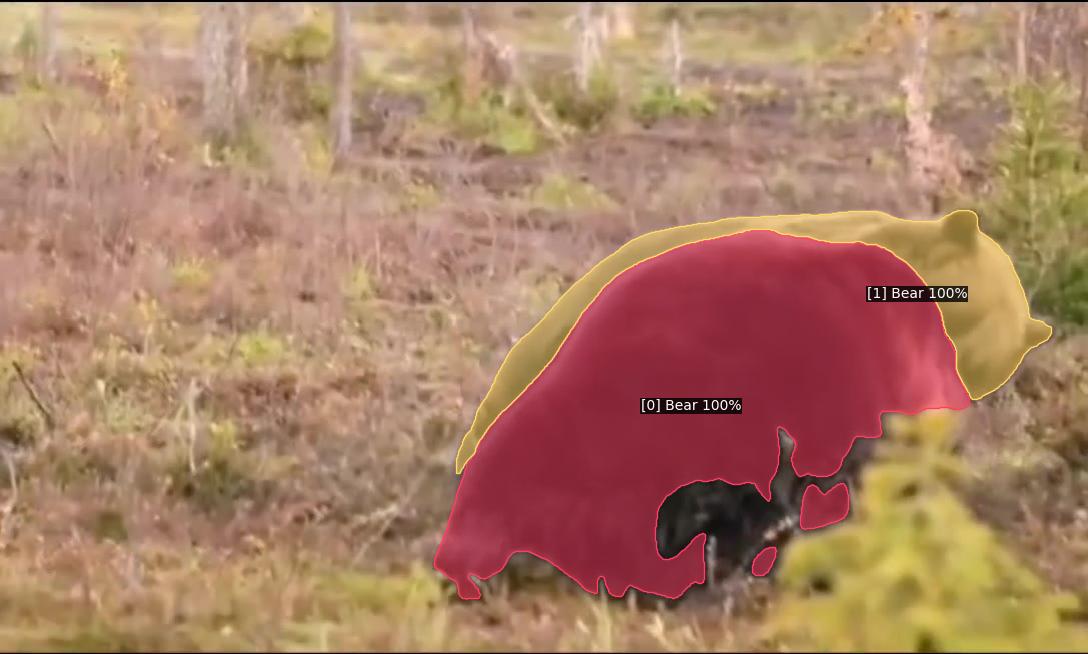}
\includegraphics[width=0.163\linewidth]{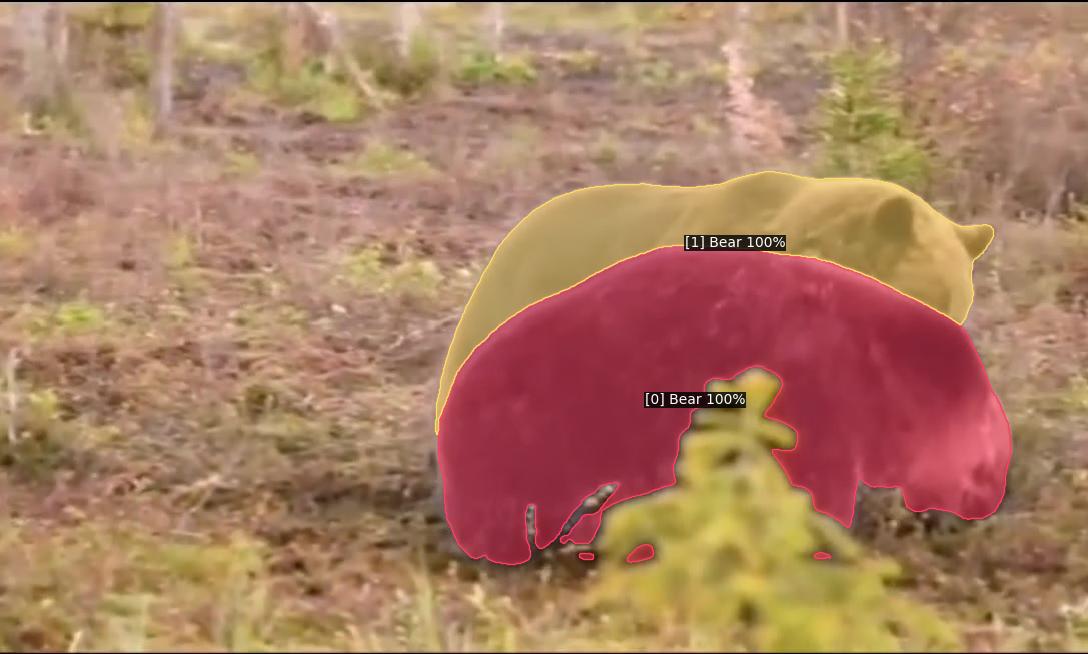}
\end{minipage}\hfill\vspace{1mm}

\begin{minipage}[c]{1.00\linewidth}
\includegraphics[width=0.163\linewidth]{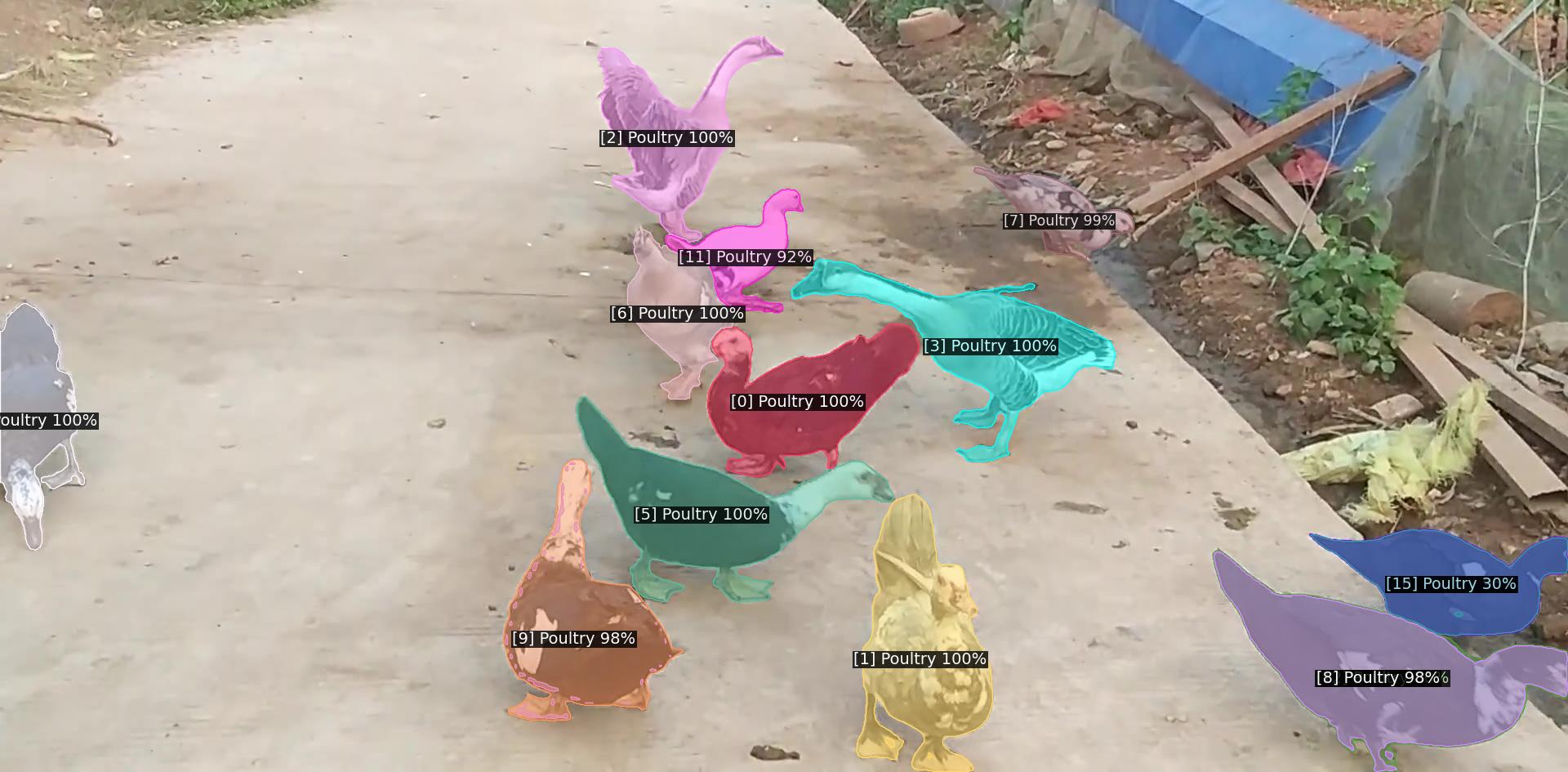}
\includegraphics[width=0.163\linewidth]{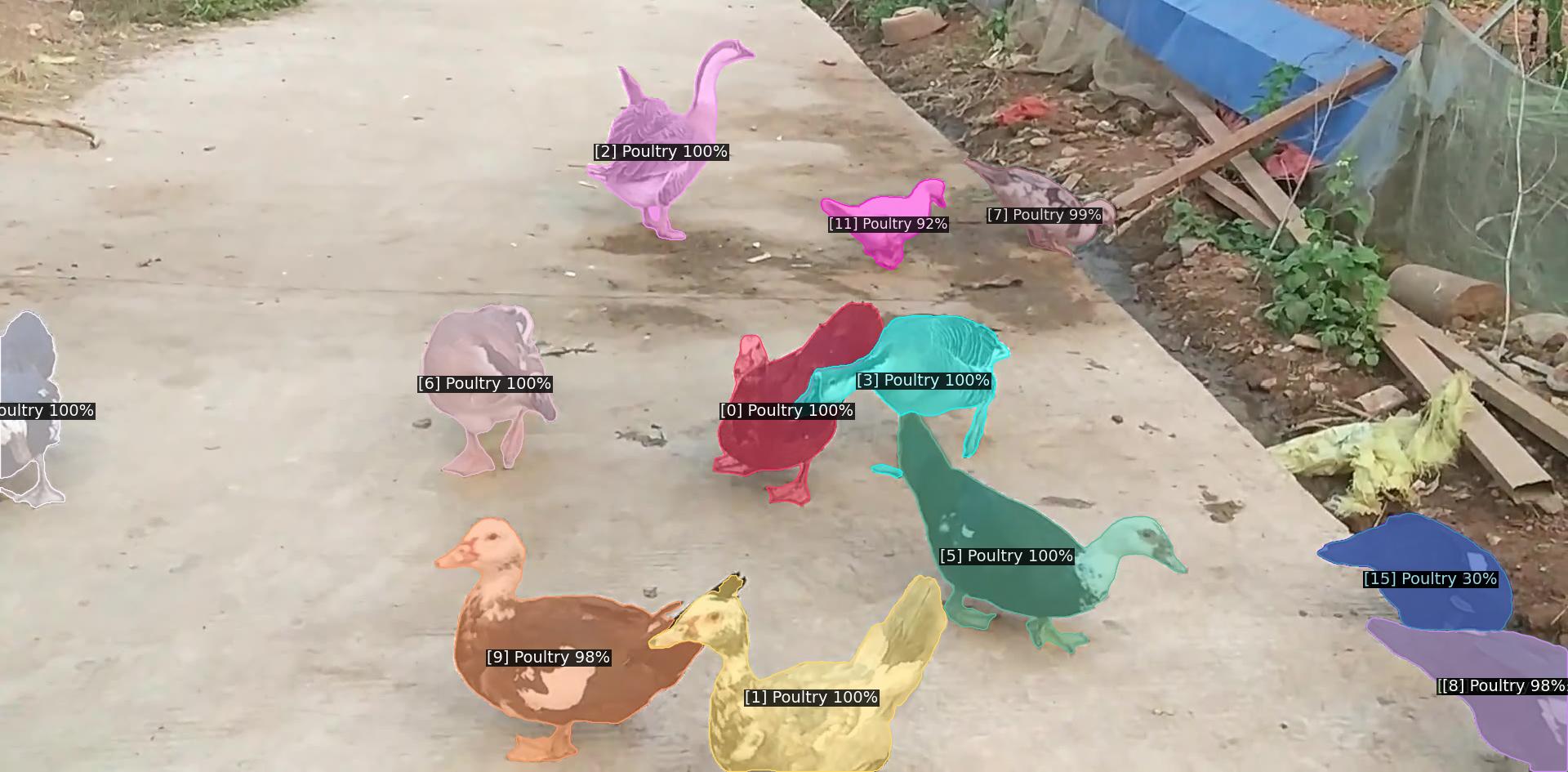}
\includegraphics[width=0.163\linewidth]{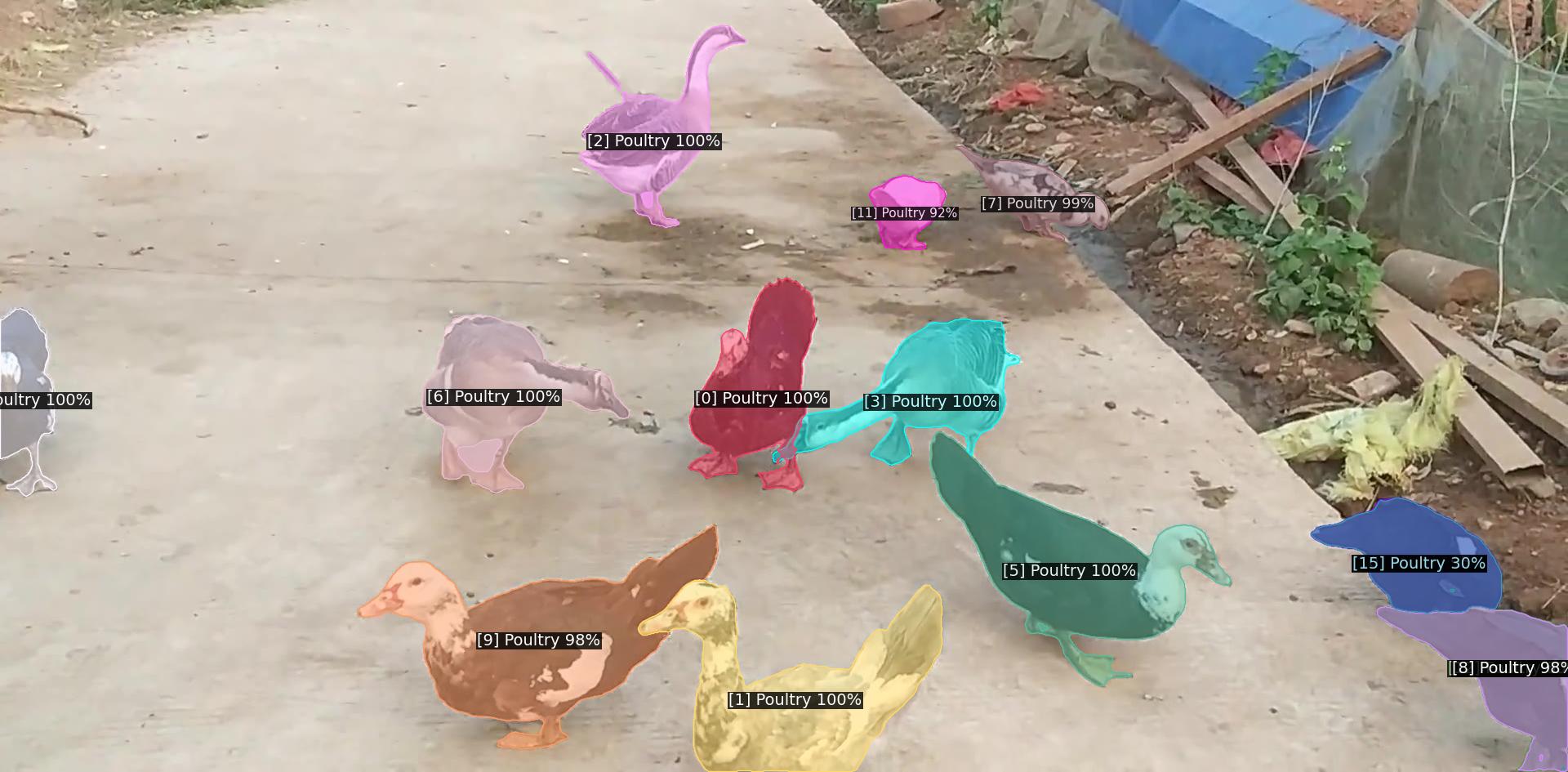}
\includegraphics[width=0.163\linewidth]{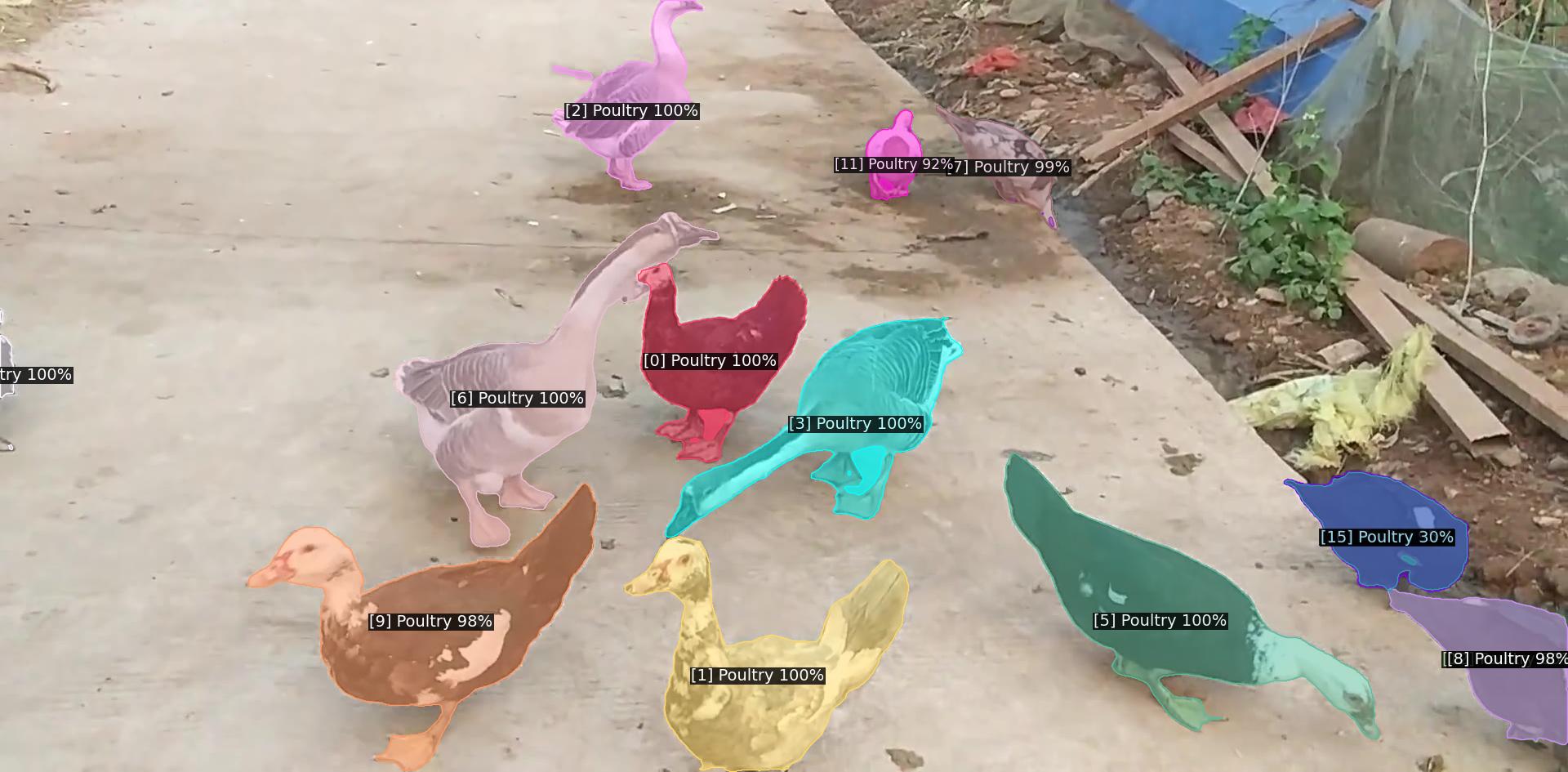}
\includegraphics[width=0.163\linewidth]{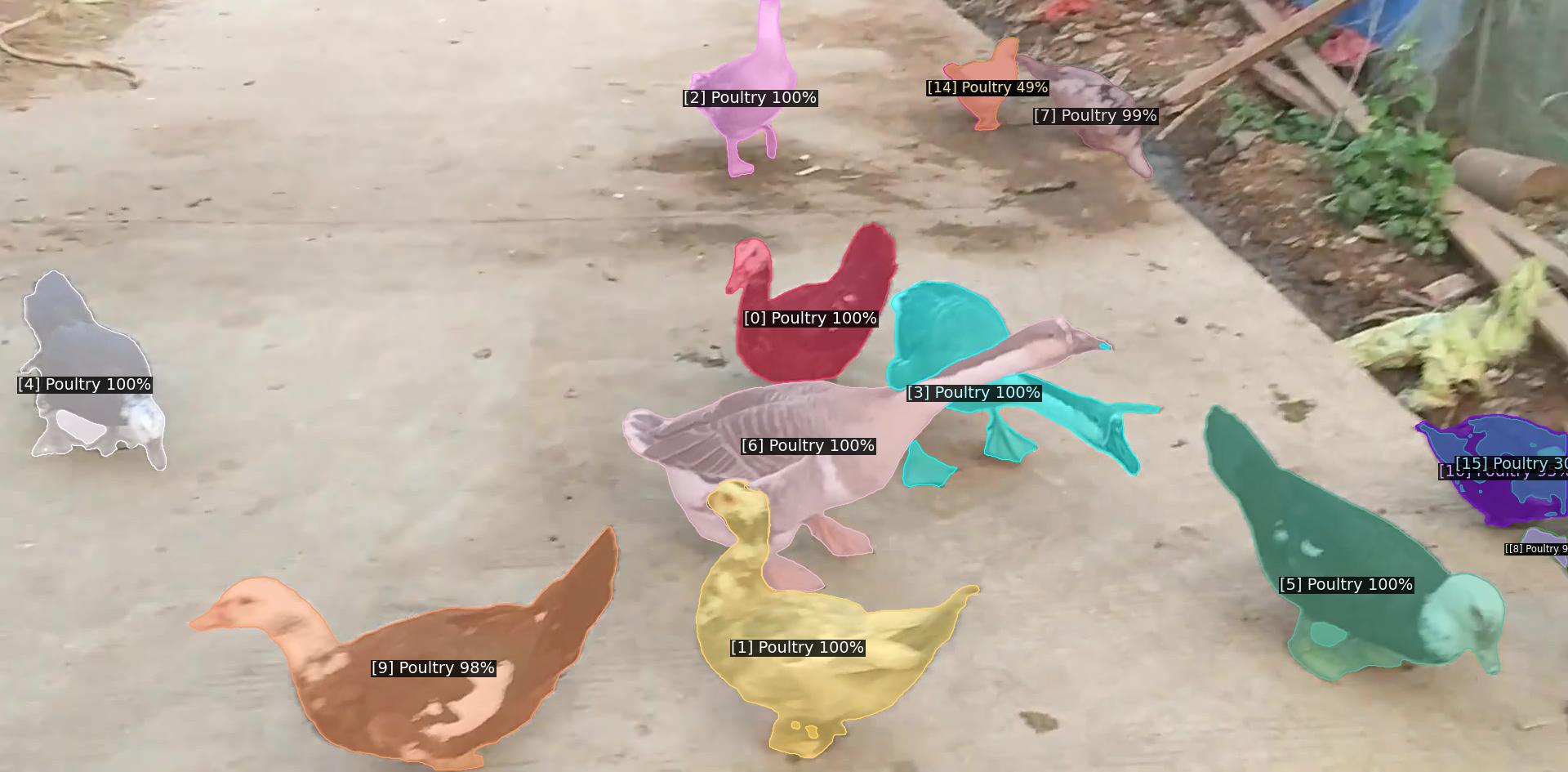}
\includegraphics[width=0.163\linewidth]{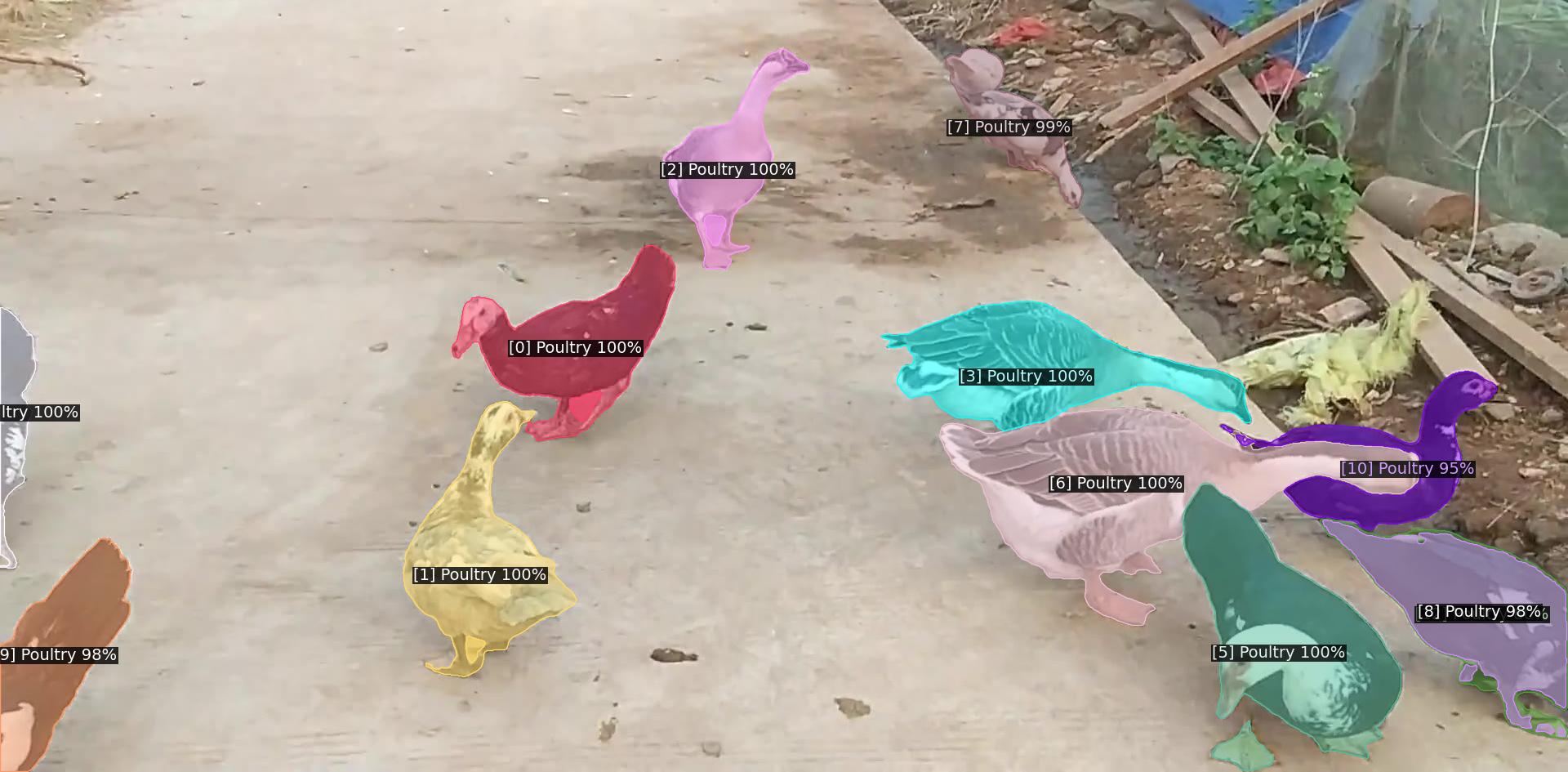}
\end{minipage}\hfill
\begin{minipage}[c]{1.0\linewidth}
\includegraphics[width=0.163\linewidth]{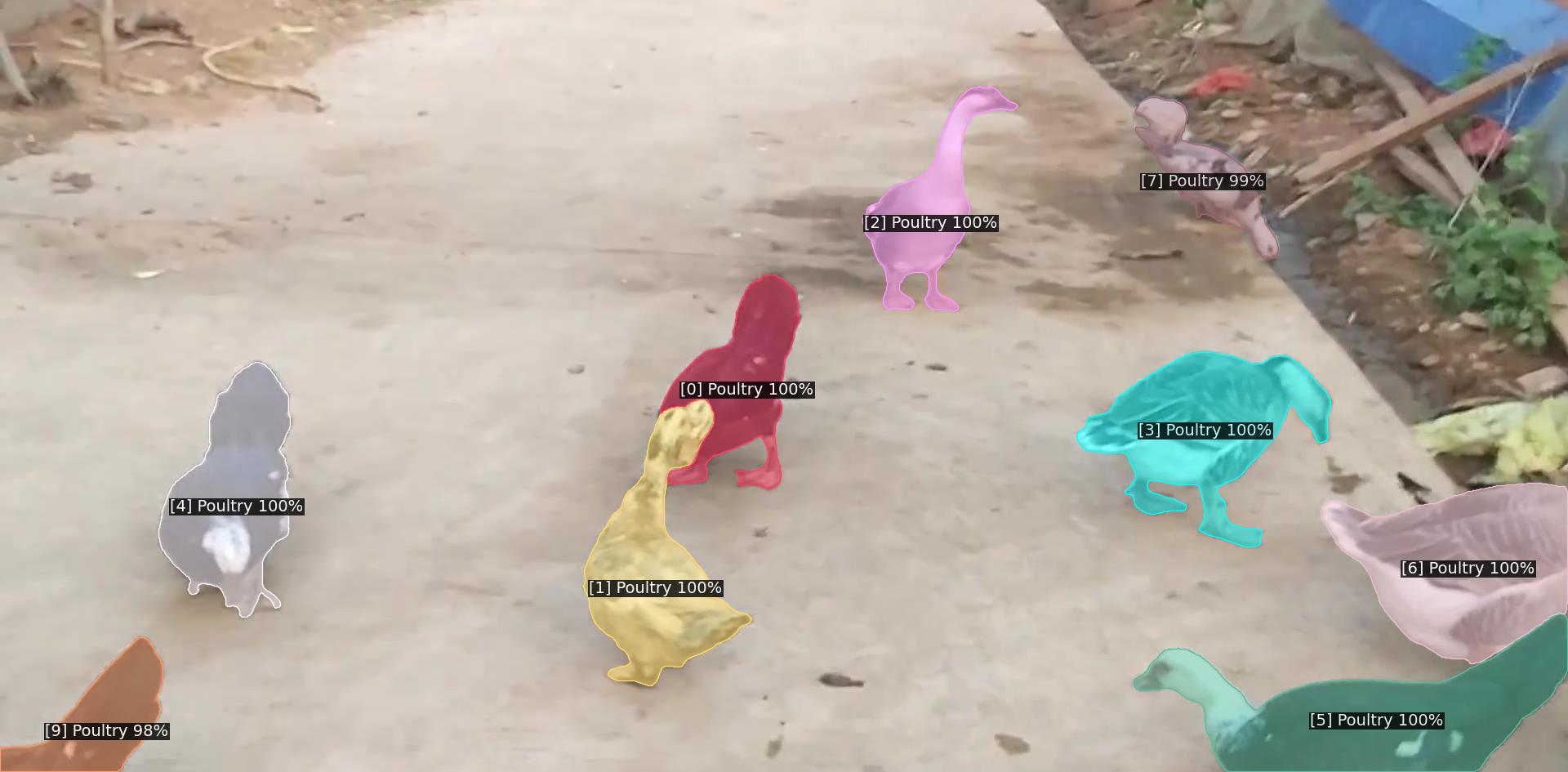}
\includegraphics[width=0.163\linewidth]{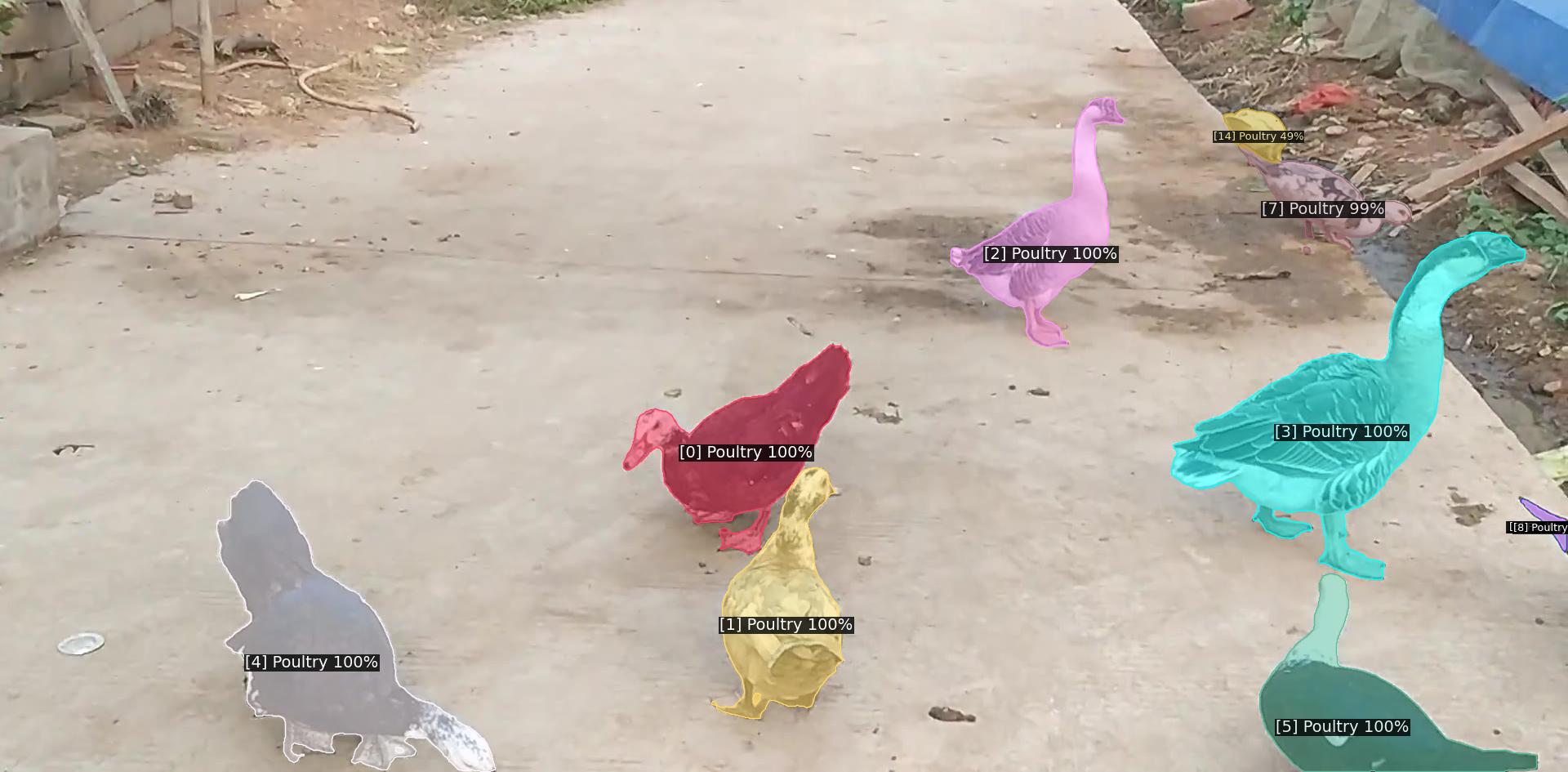}
\includegraphics[width=0.163\linewidth]{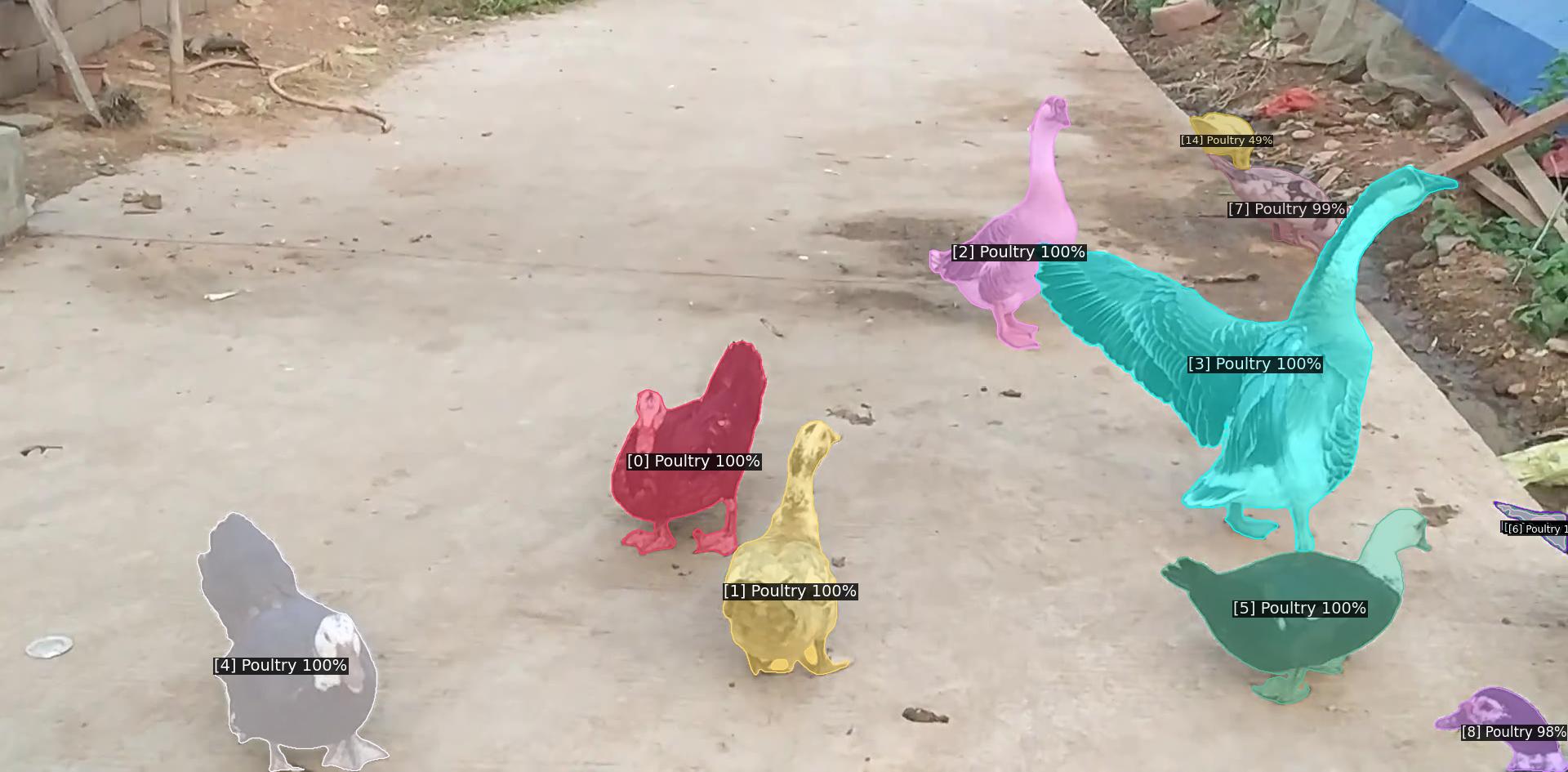}
\includegraphics[width=0.163\linewidth]{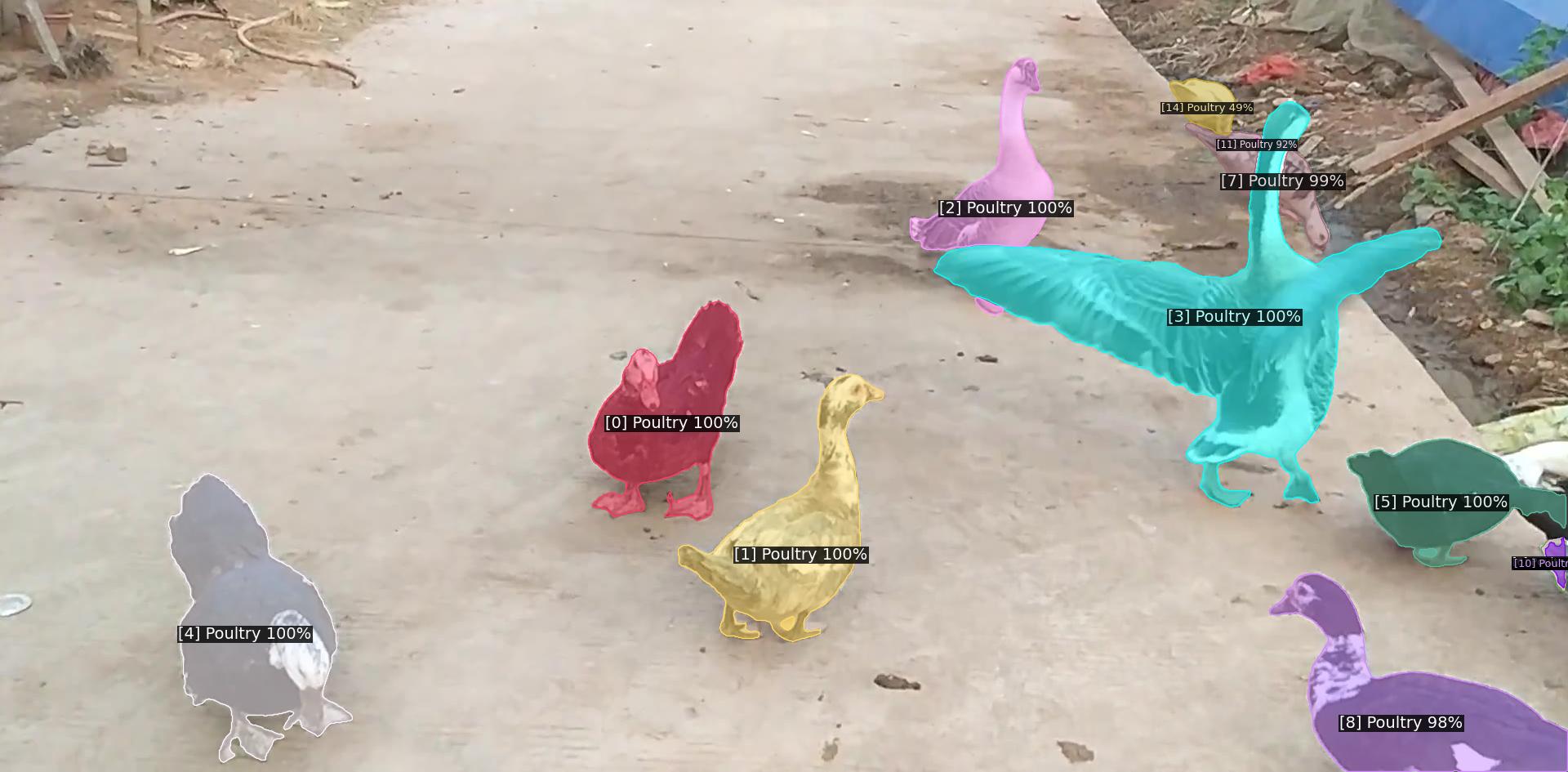}
\includegraphics[width=0.163\linewidth]{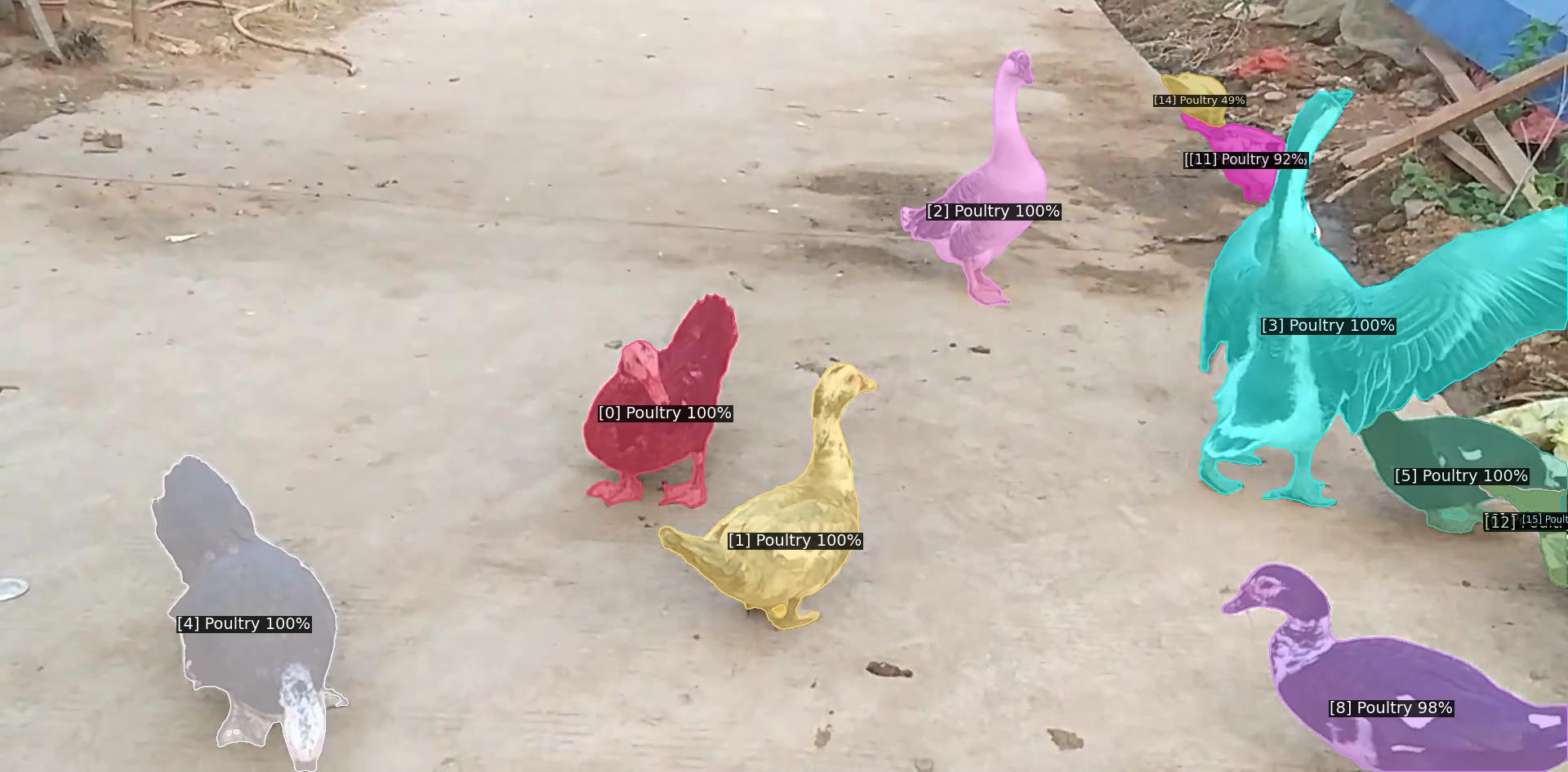}
\includegraphics[width=0.163\linewidth]{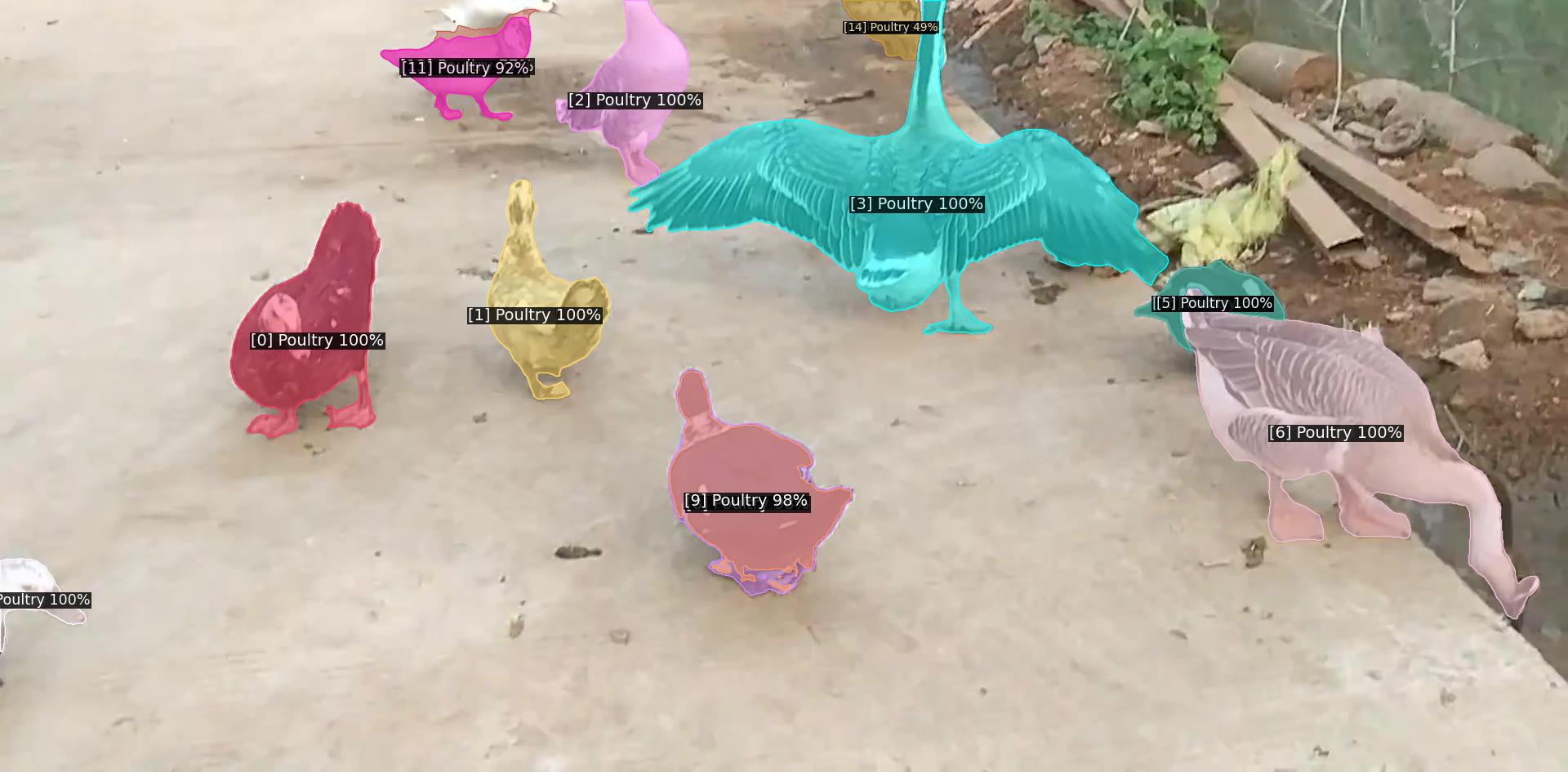}
\end{minipage}\hfill\vspace{1mm}

\begin{minipage}[c]{1.00\linewidth}
\includegraphics[width=0.163\linewidth]{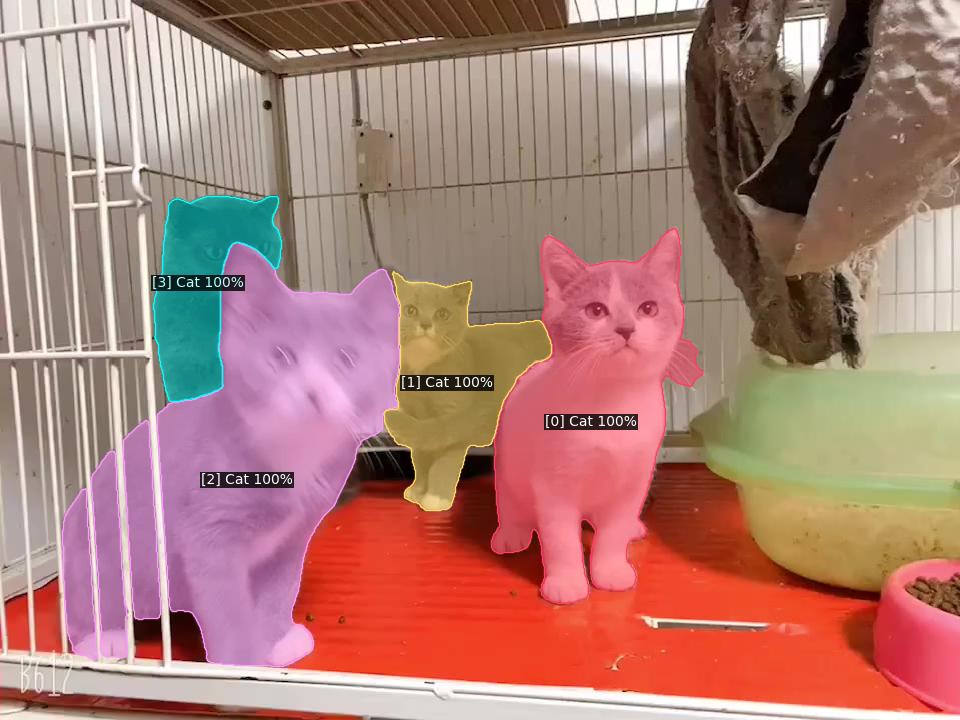}
\includegraphics[width=0.163\linewidth]{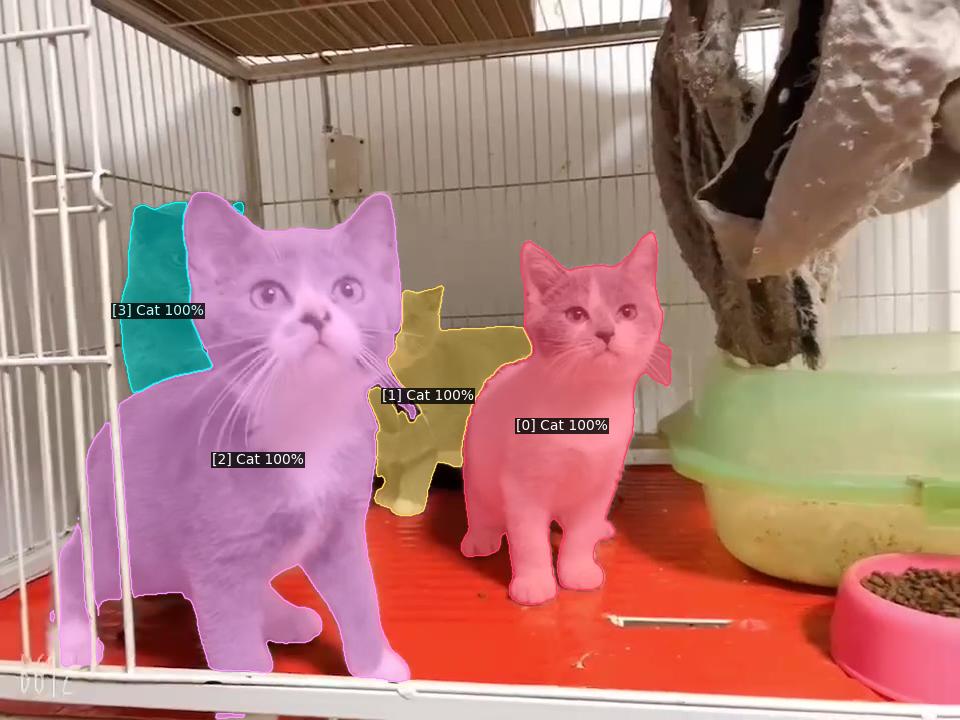}
\includegraphics[width=0.163\linewidth]{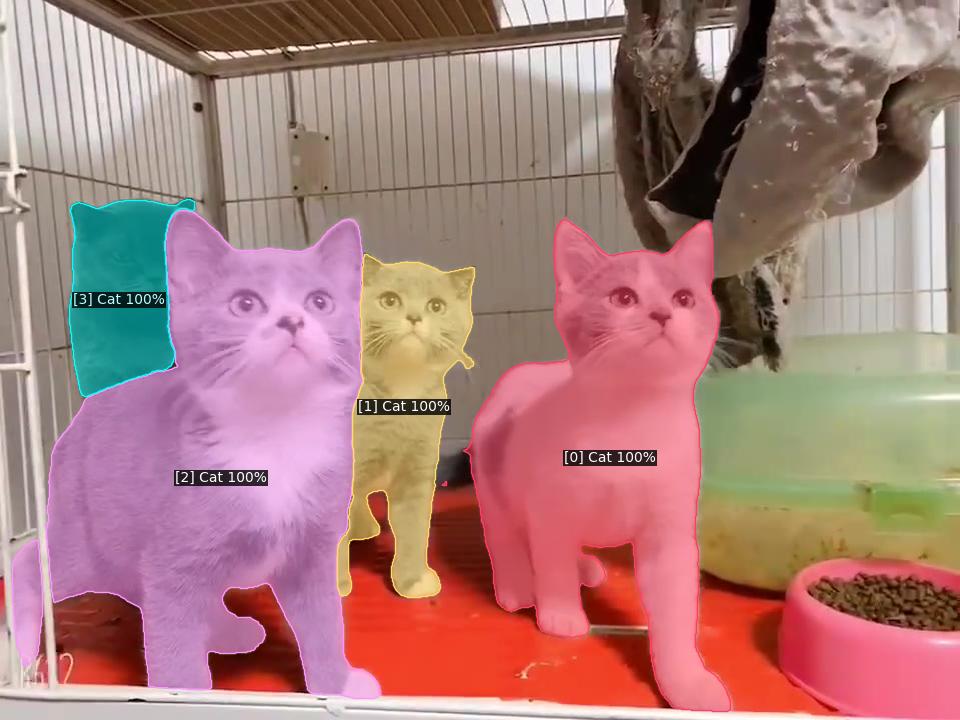}
\includegraphics[width=0.163\linewidth]{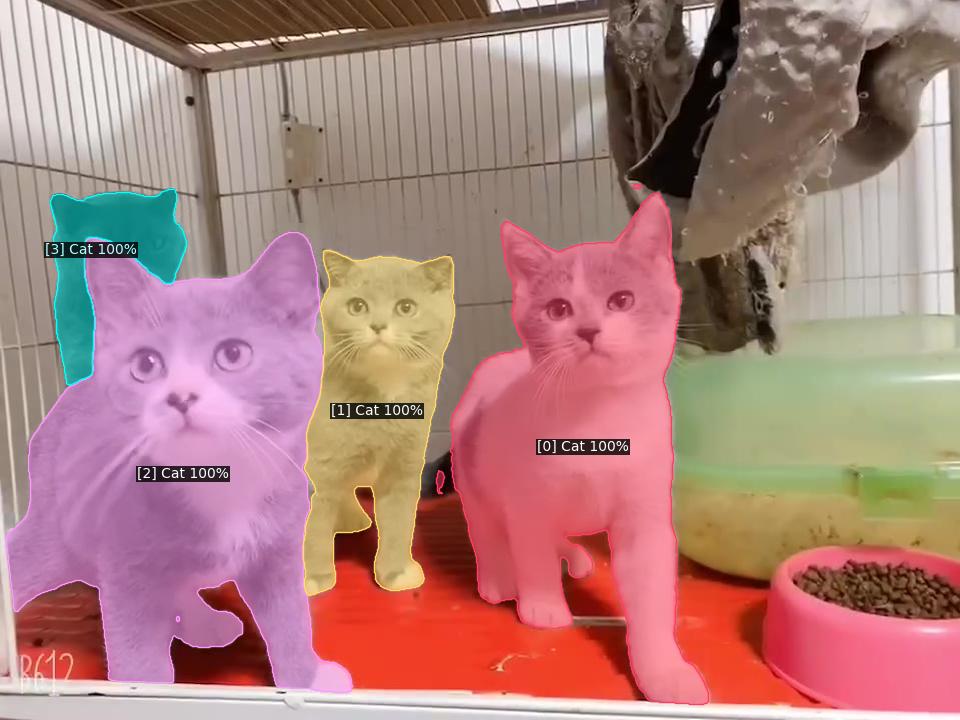}
\includegraphics[width=0.163\linewidth]{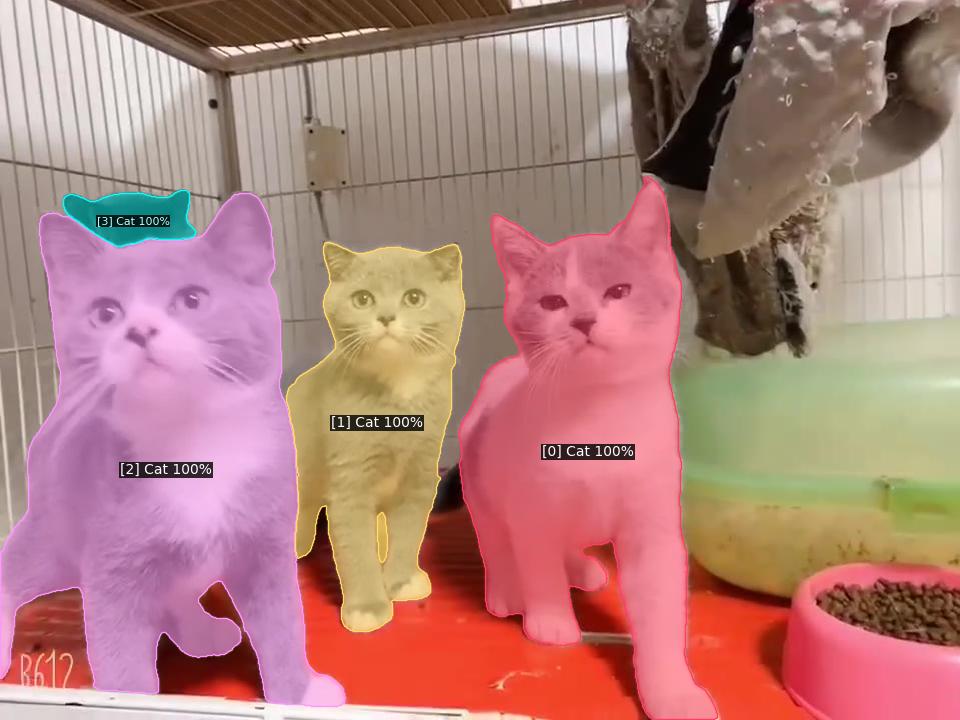}
\includegraphics[width=0.163\linewidth]{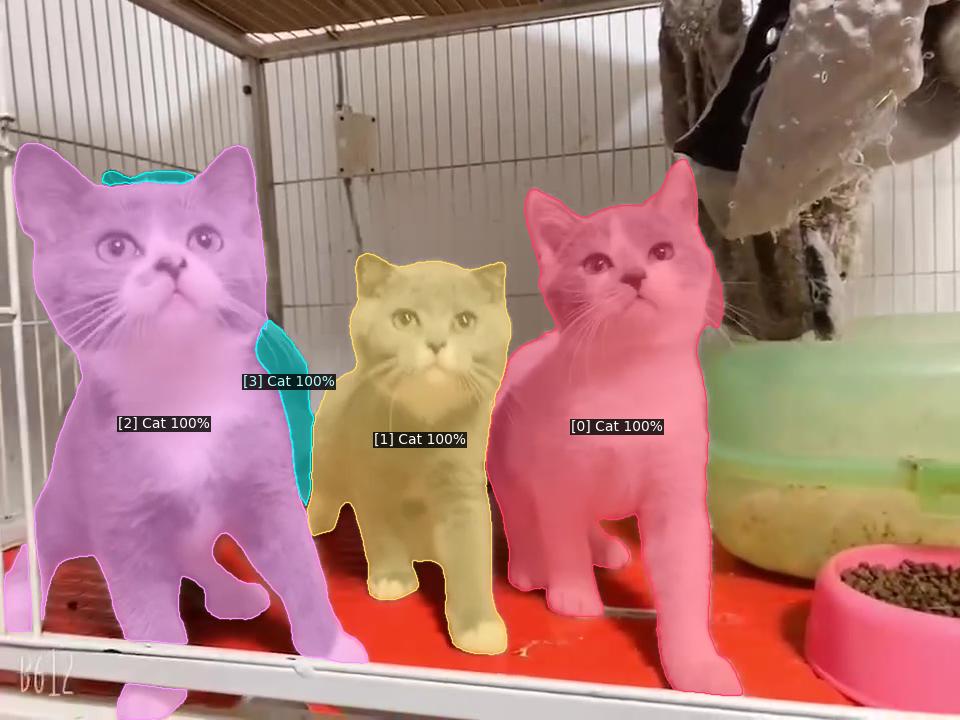}
\end{minipage}\hfill
\begin{minipage}[c]{1.0\linewidth}
\includegraphics[width=0.163\linewidth]{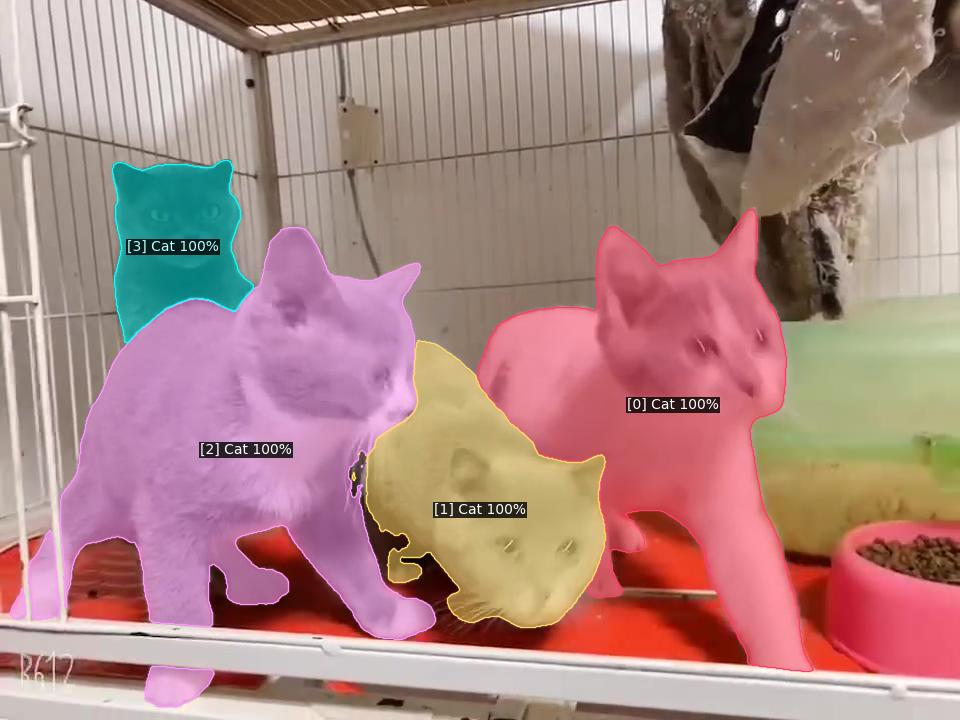}
\includegraphics[width=0.163\linewidth]{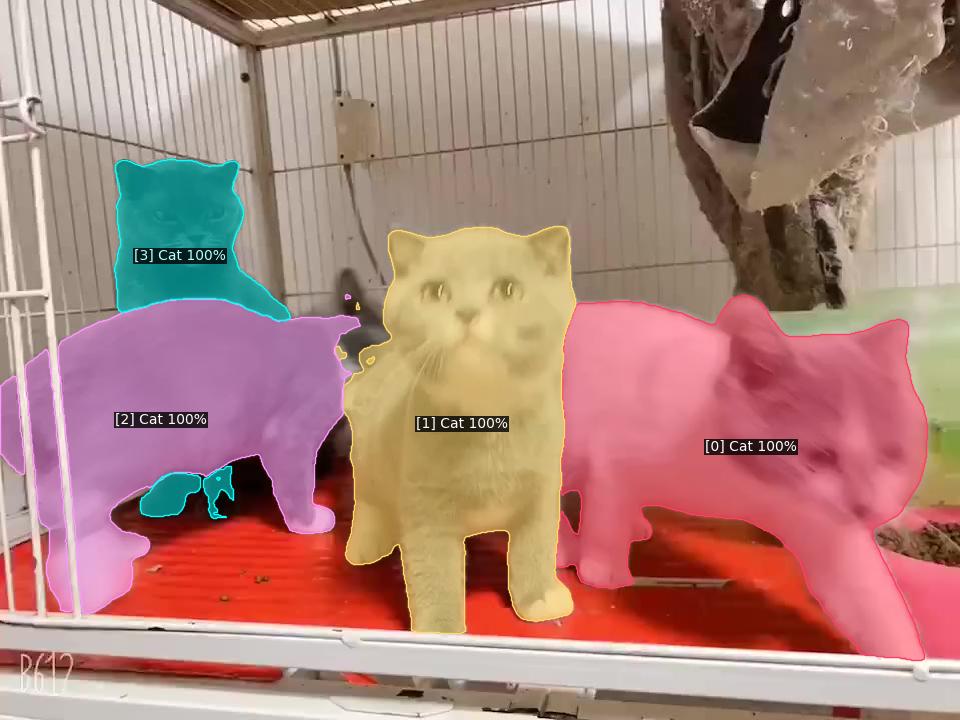}
\includegraphics[width=0.163\linewidth]{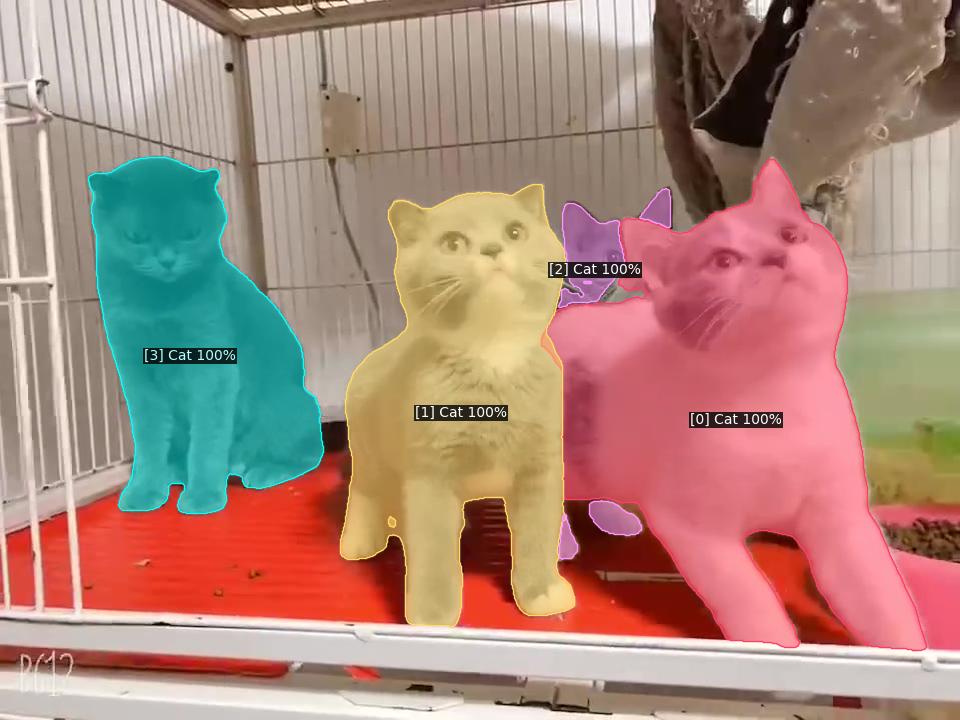}
\includegraphics[width=0.163\linewidth]{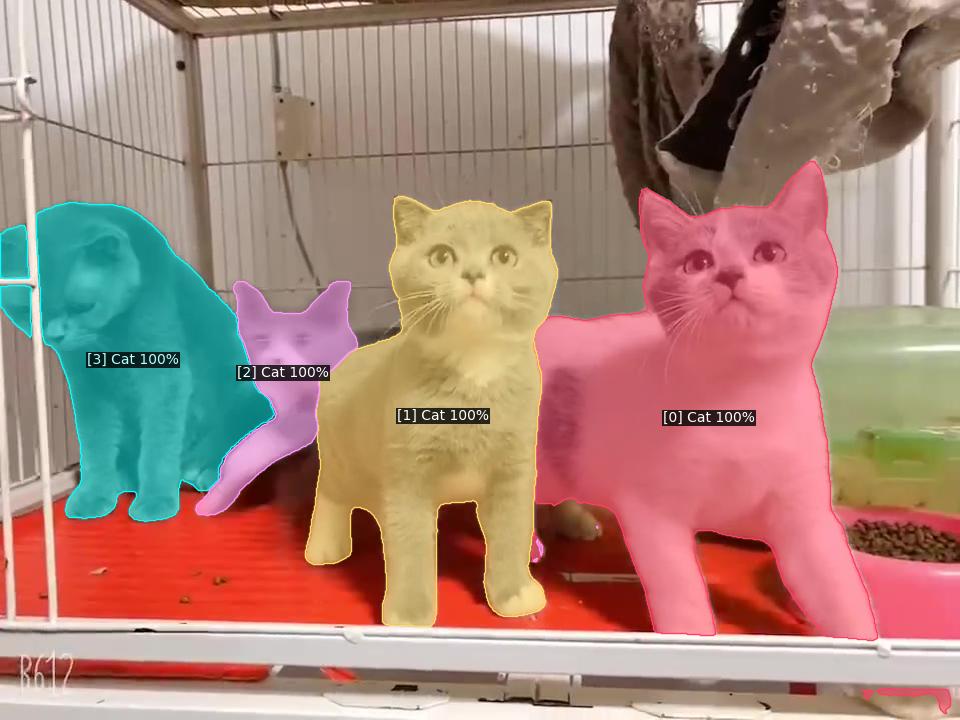}
\includegraphics[width=0.163\linewidth]{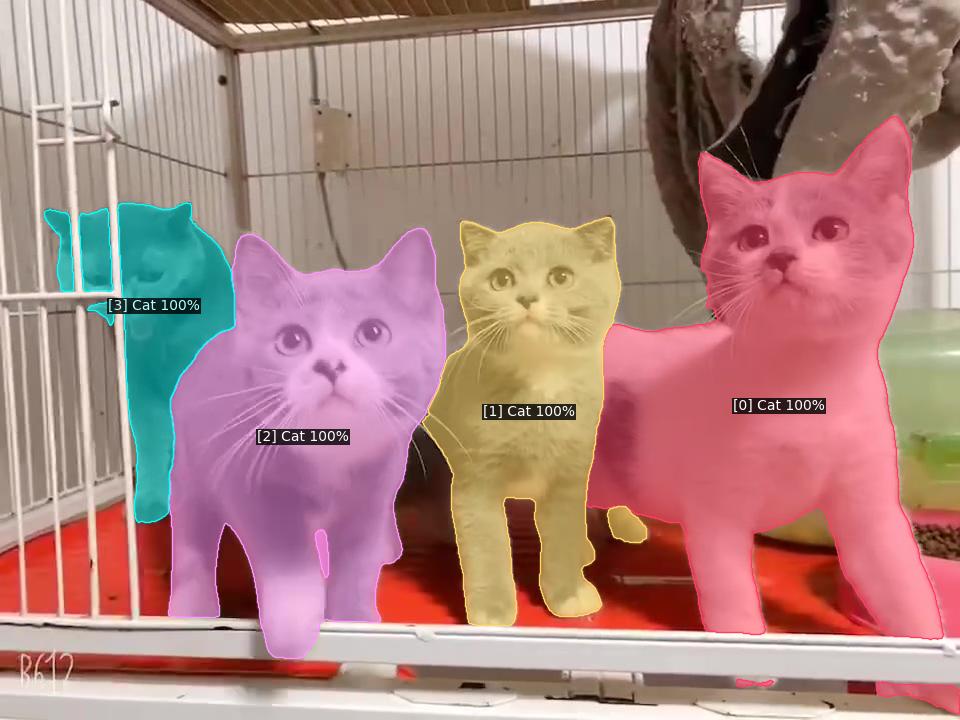}
\includegraphics[width=0.163\linewidth]{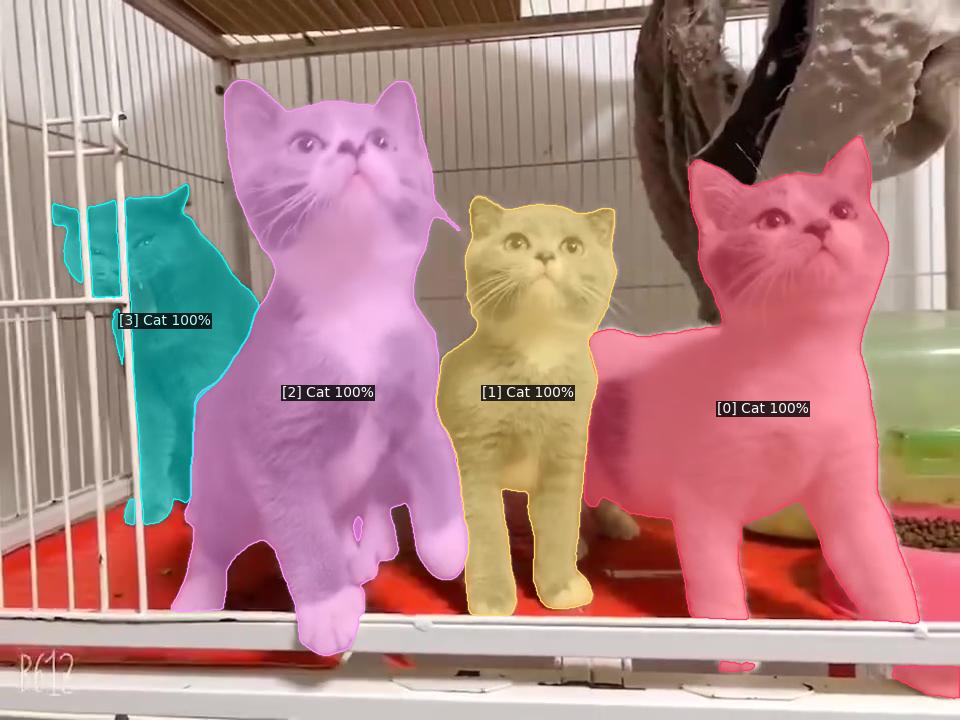}
\end{minipage}\hfill\vspace{1mm}

\begin{minipage}[c]{1.00\linewidth}
\includegraphics[width=0.163\linewidth]{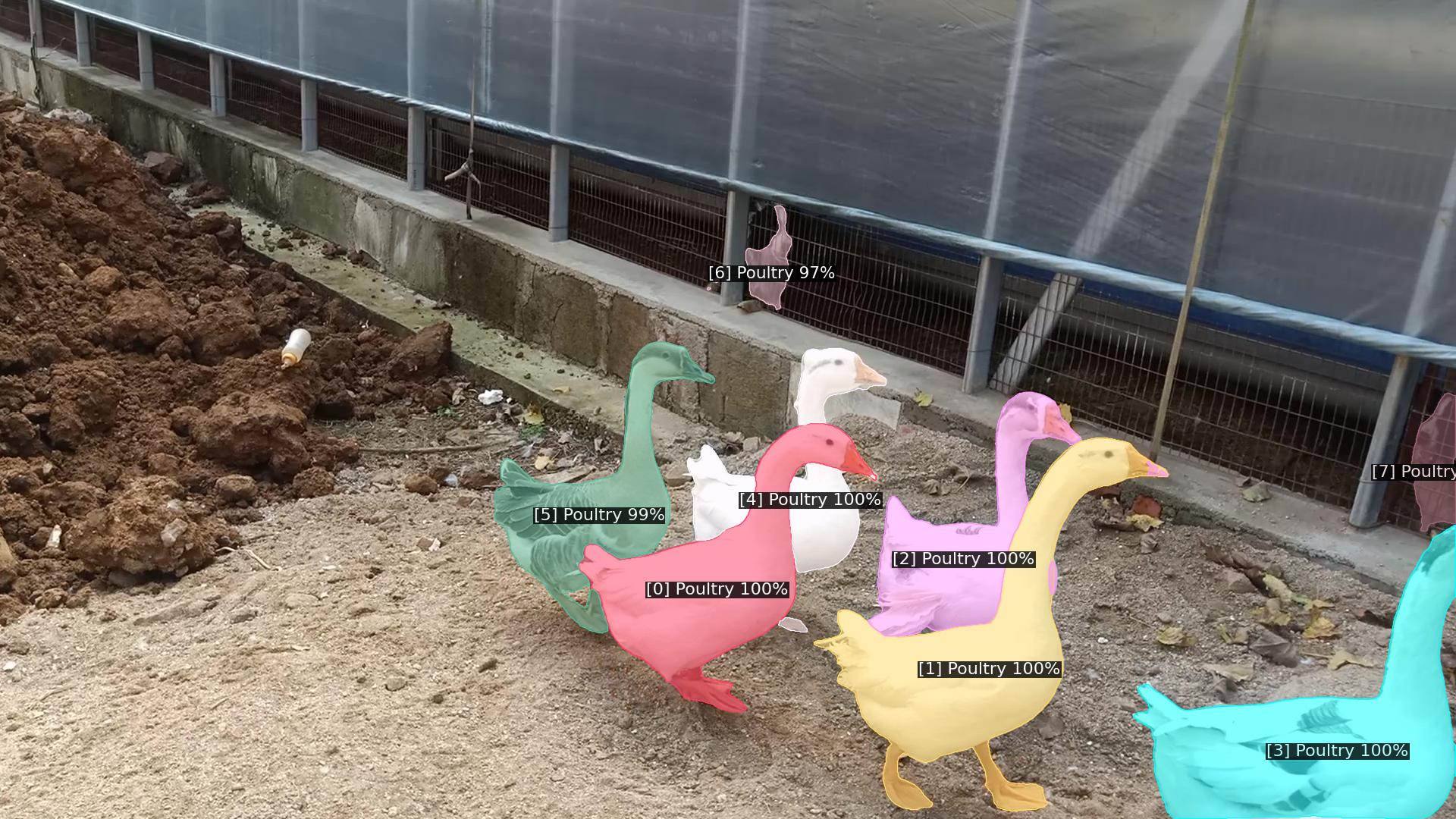}
\includegraphics[width=0.163\linewidth]{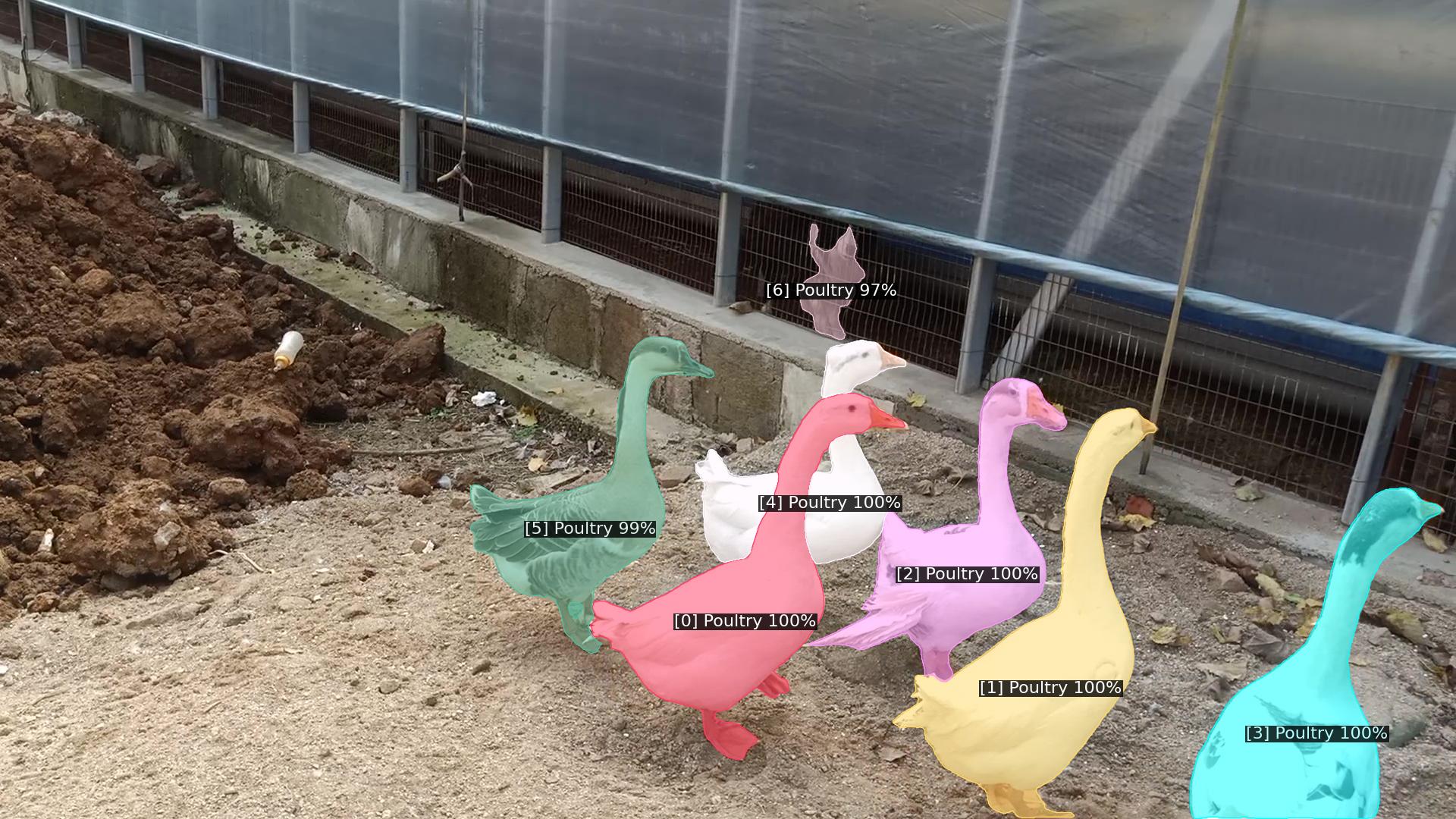}
\includegraphics[width=0.163\linewidth]{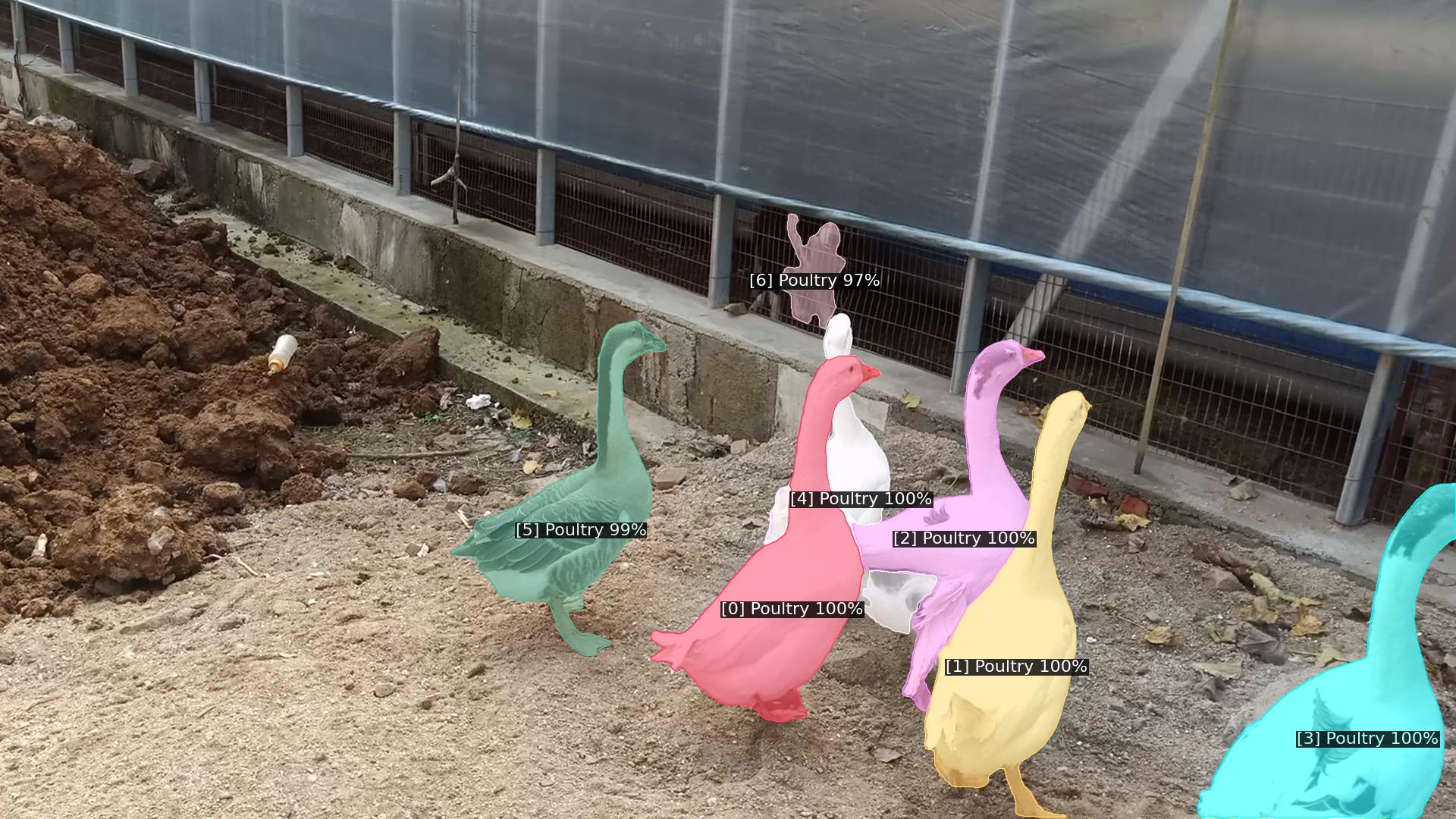}
\includegraphics[width=0.163\linewidth]{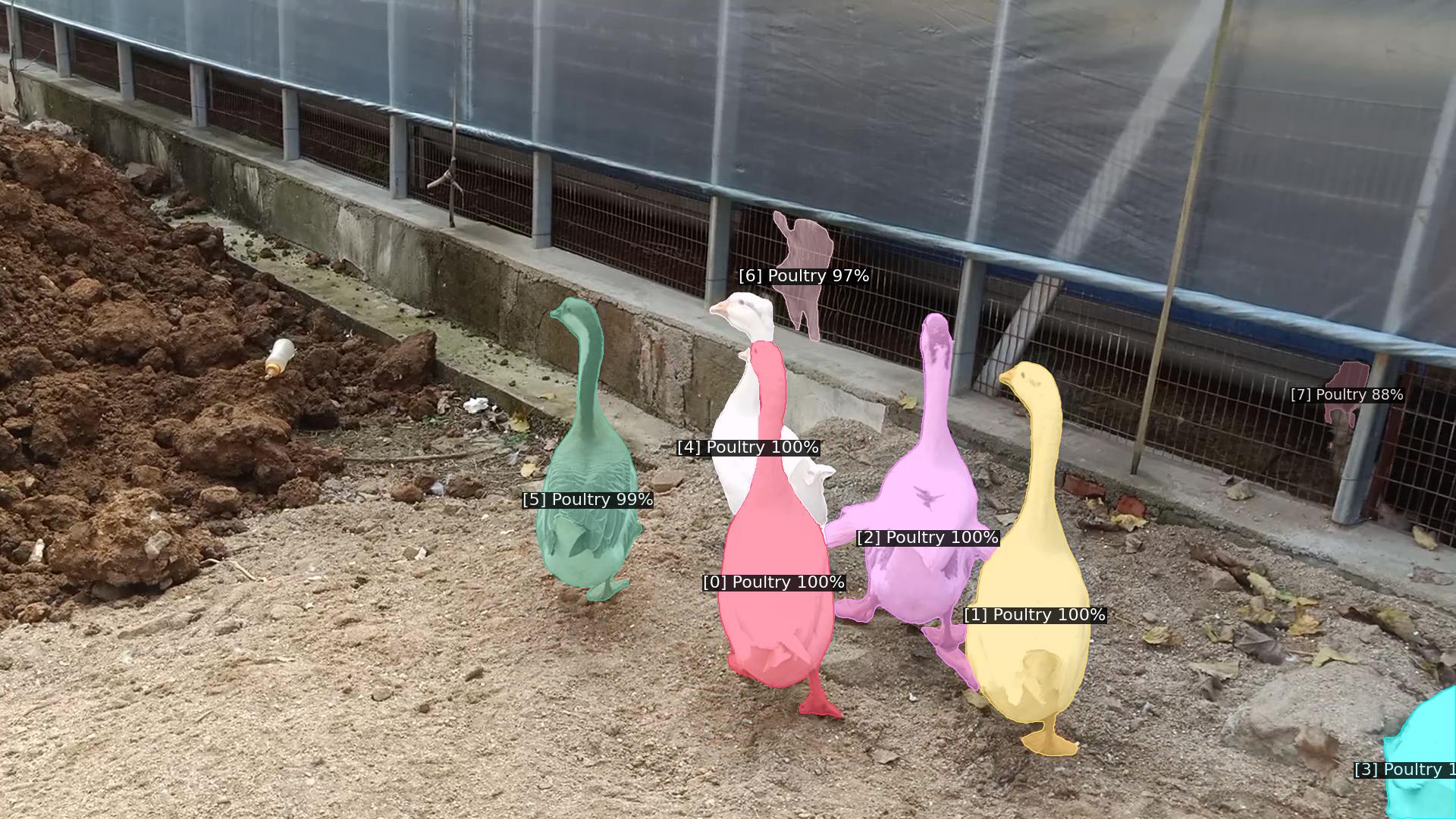}
\includegraphics[width=0.163\linewidth]{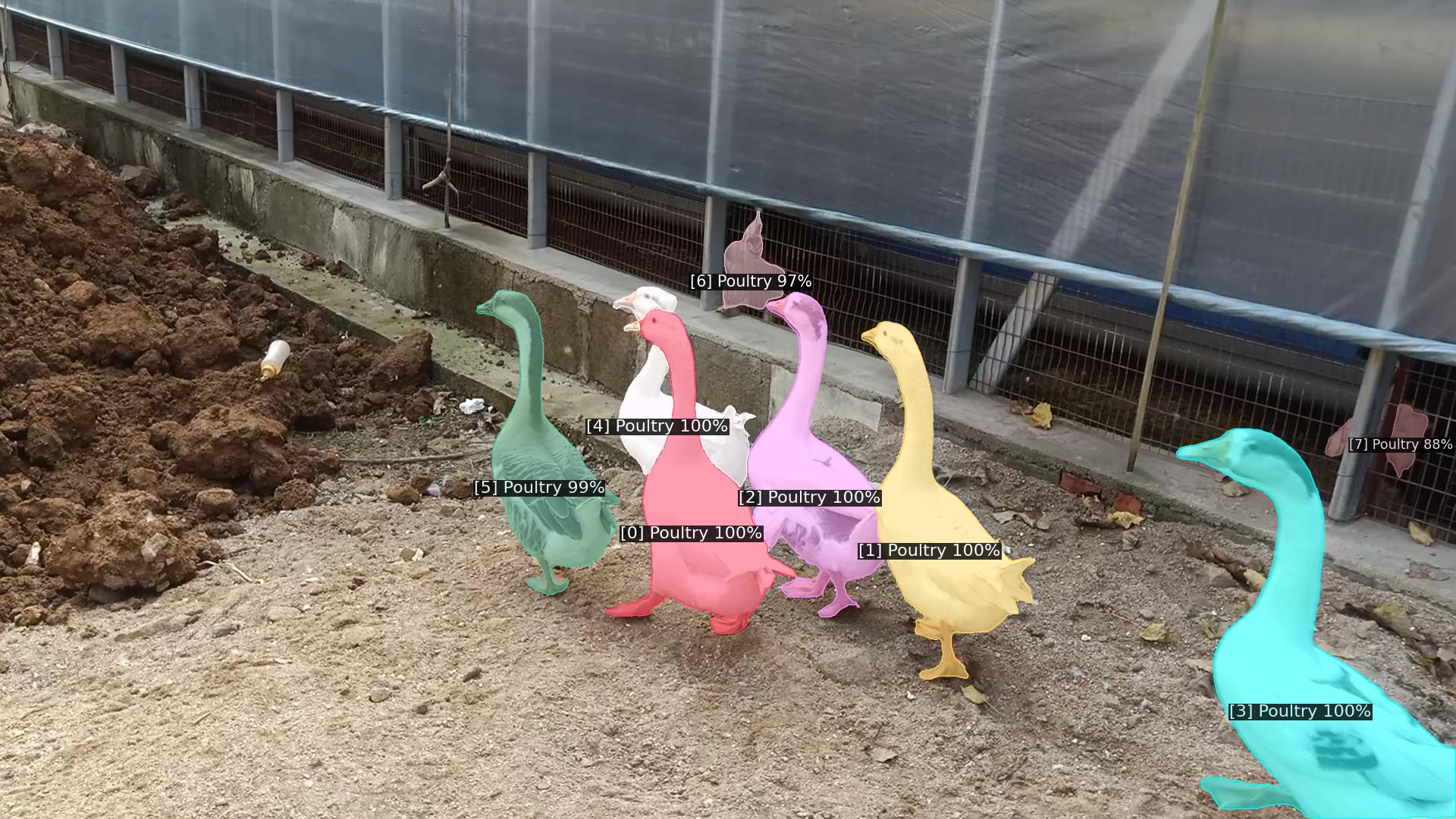}
\includegraphics[width=0.163\linewidth]{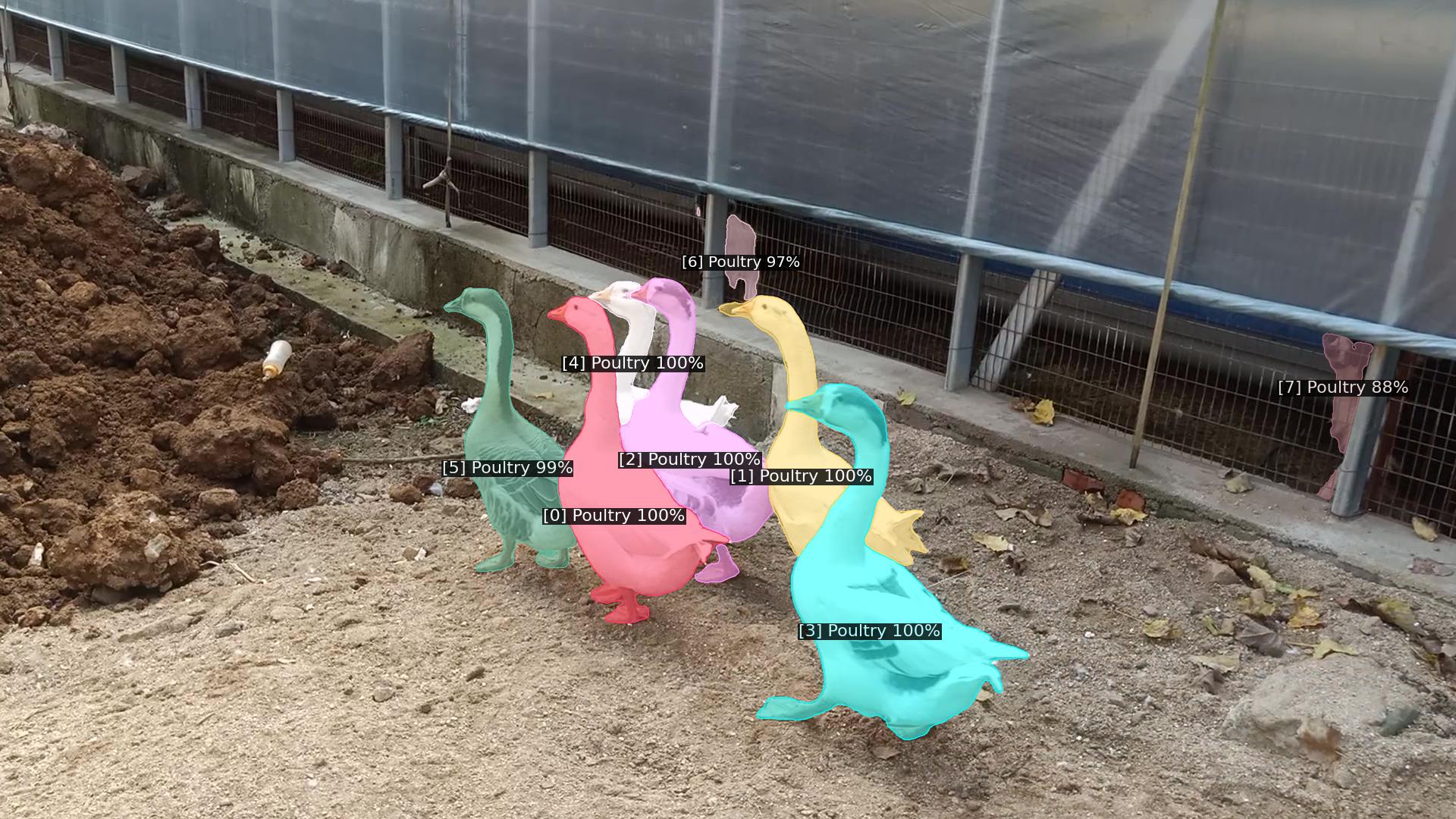}
\end{minipage}\hfill
\begin{minipage}[c]{1.0\linewidth}
\includegraphics[width=0.163\linewidth]{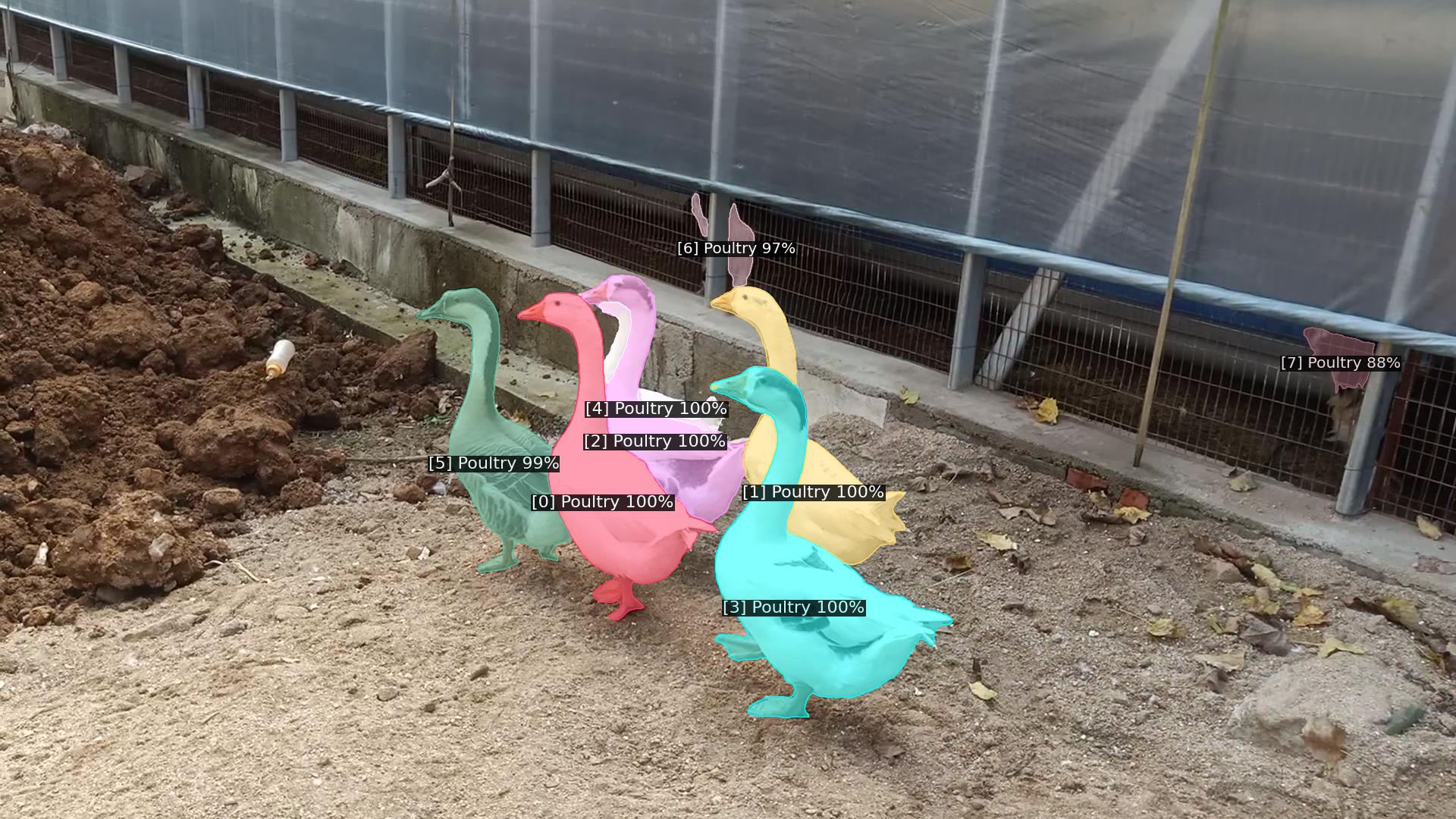}
\includegraphics[width=0.163\linewidth]{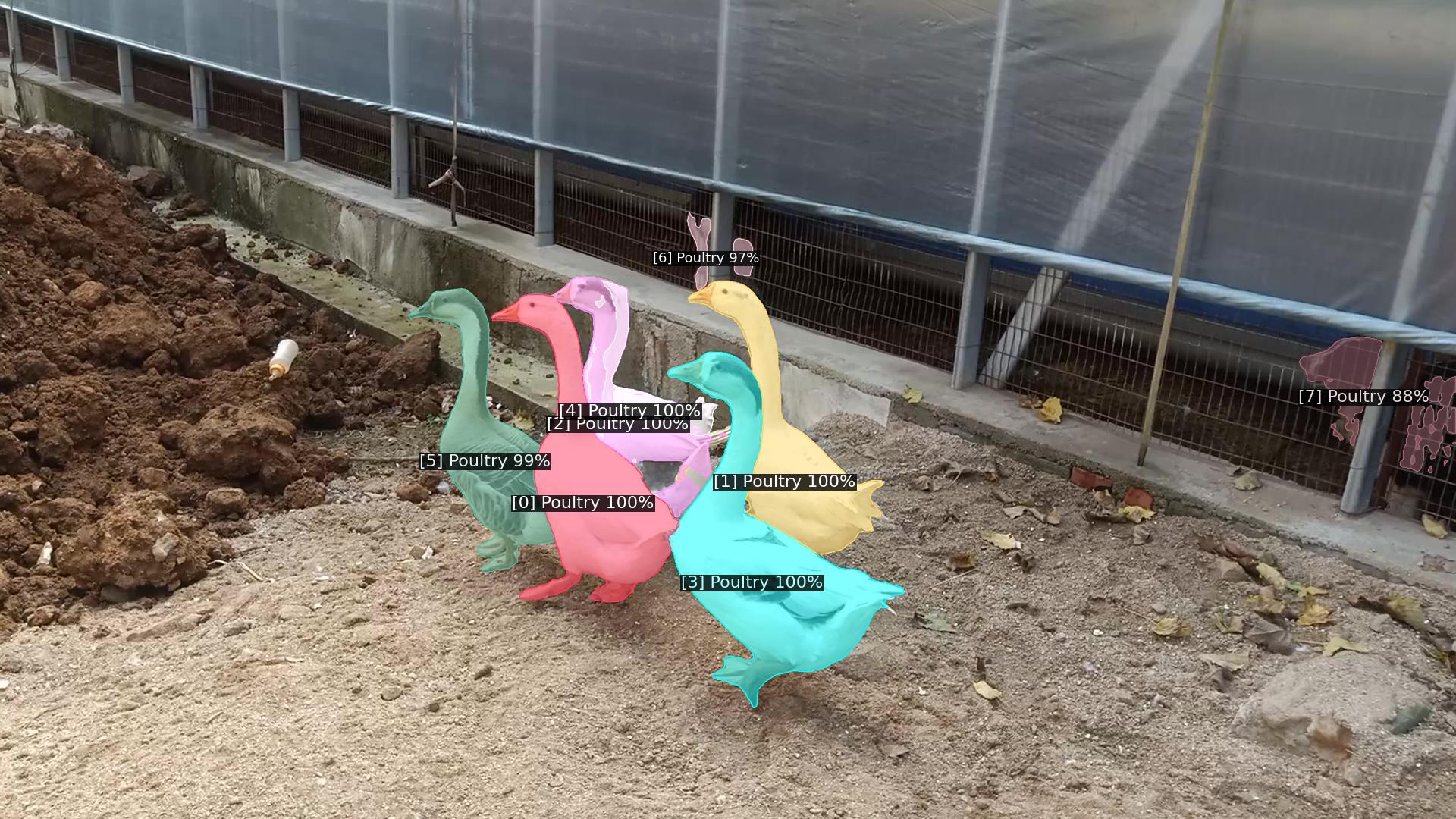}
\includegraphics[width=0.163\linewidth]{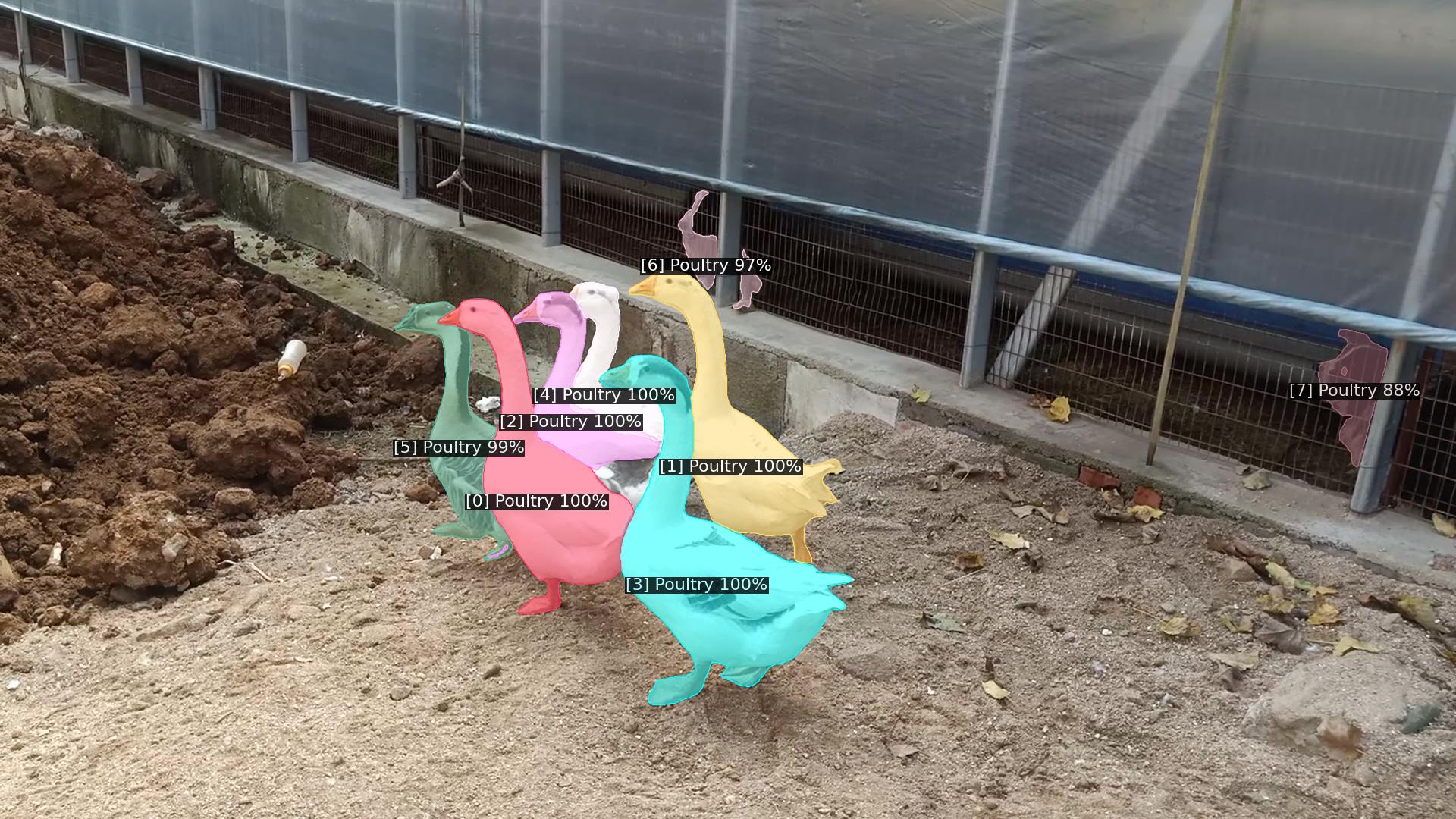}
\includegraphics[width=0.163\linewidth]{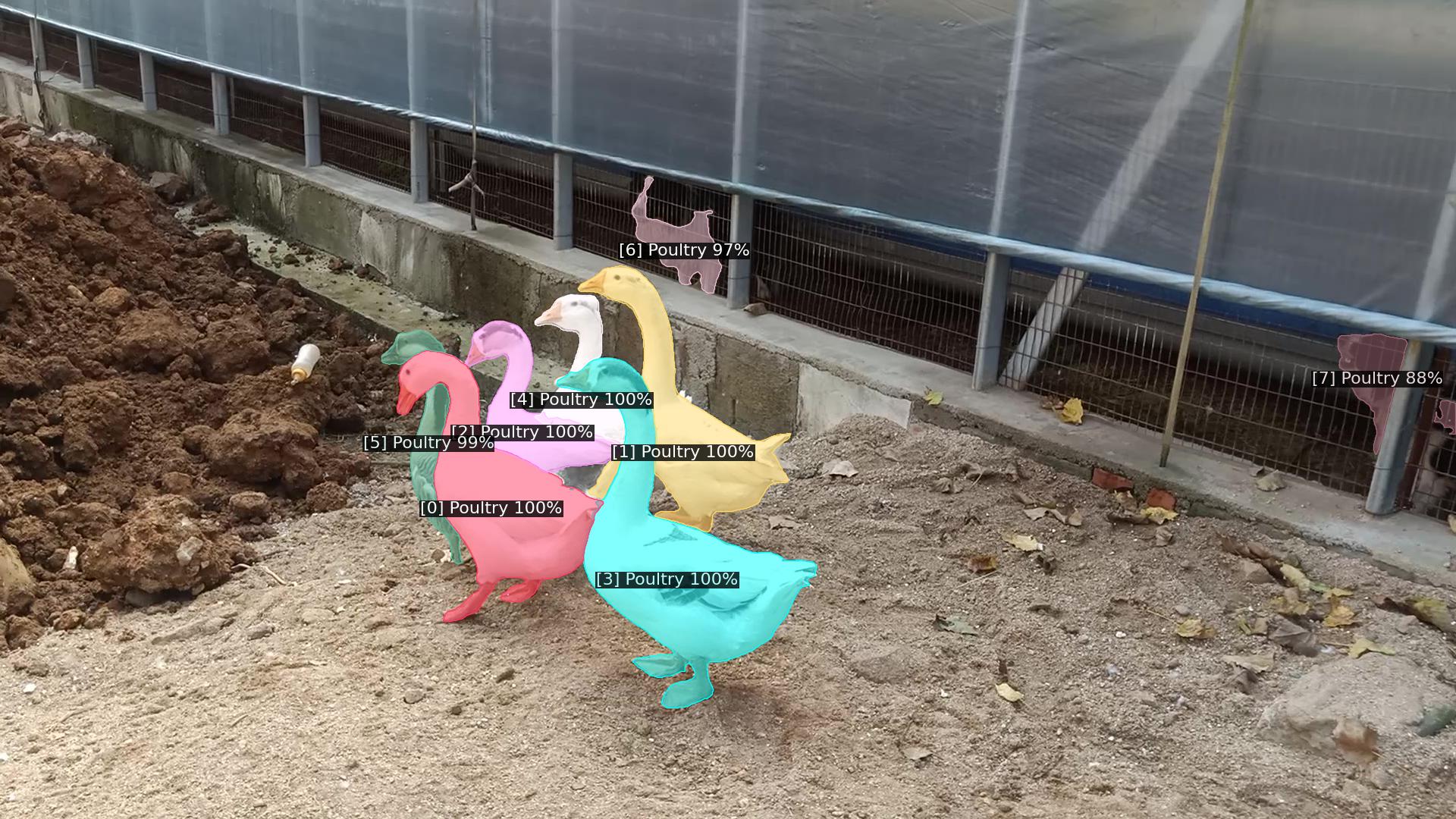}
\includegraphics[width=0.163\linewidth]{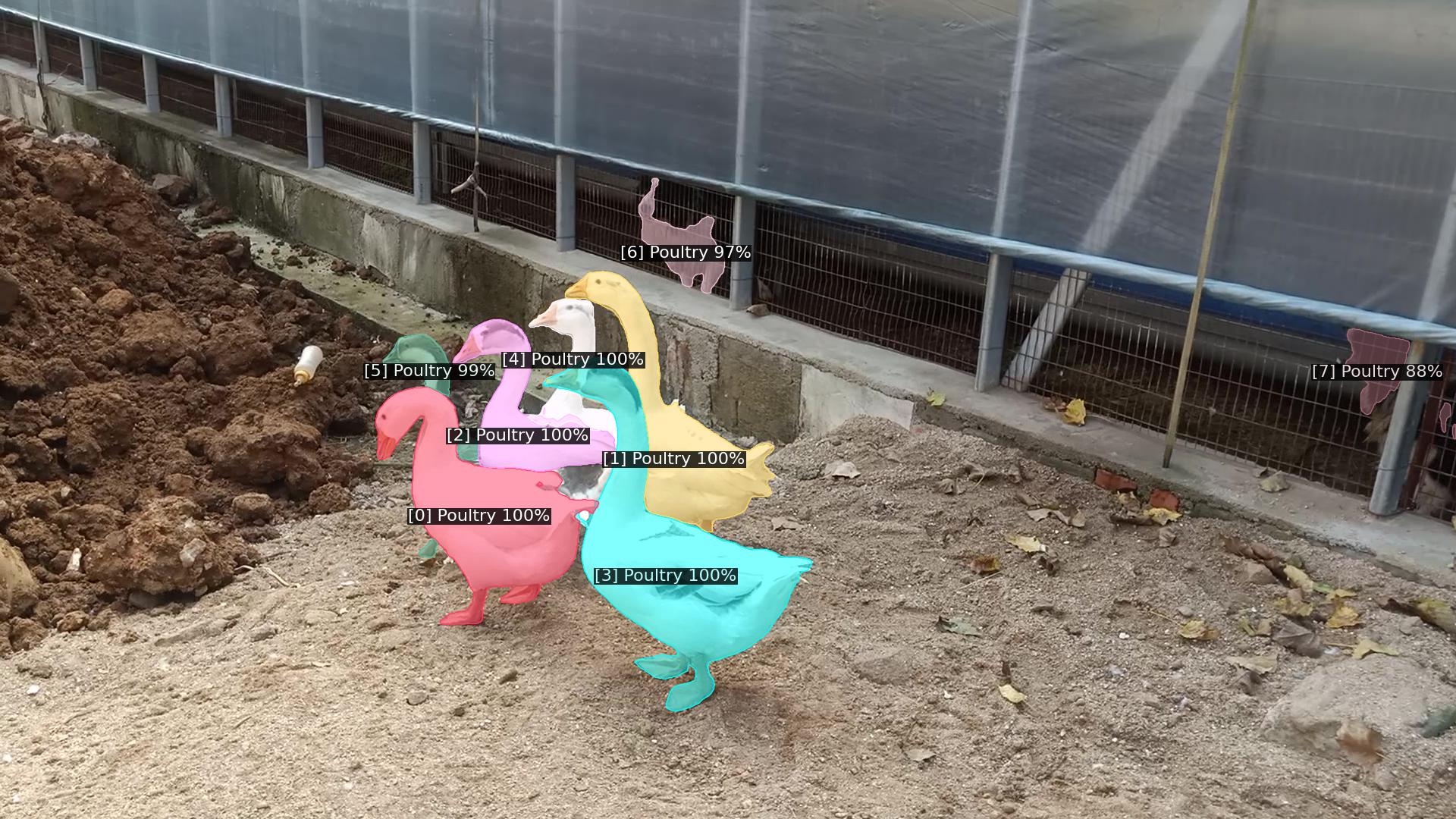}
\includegraphics[width=0.163\linewidth]{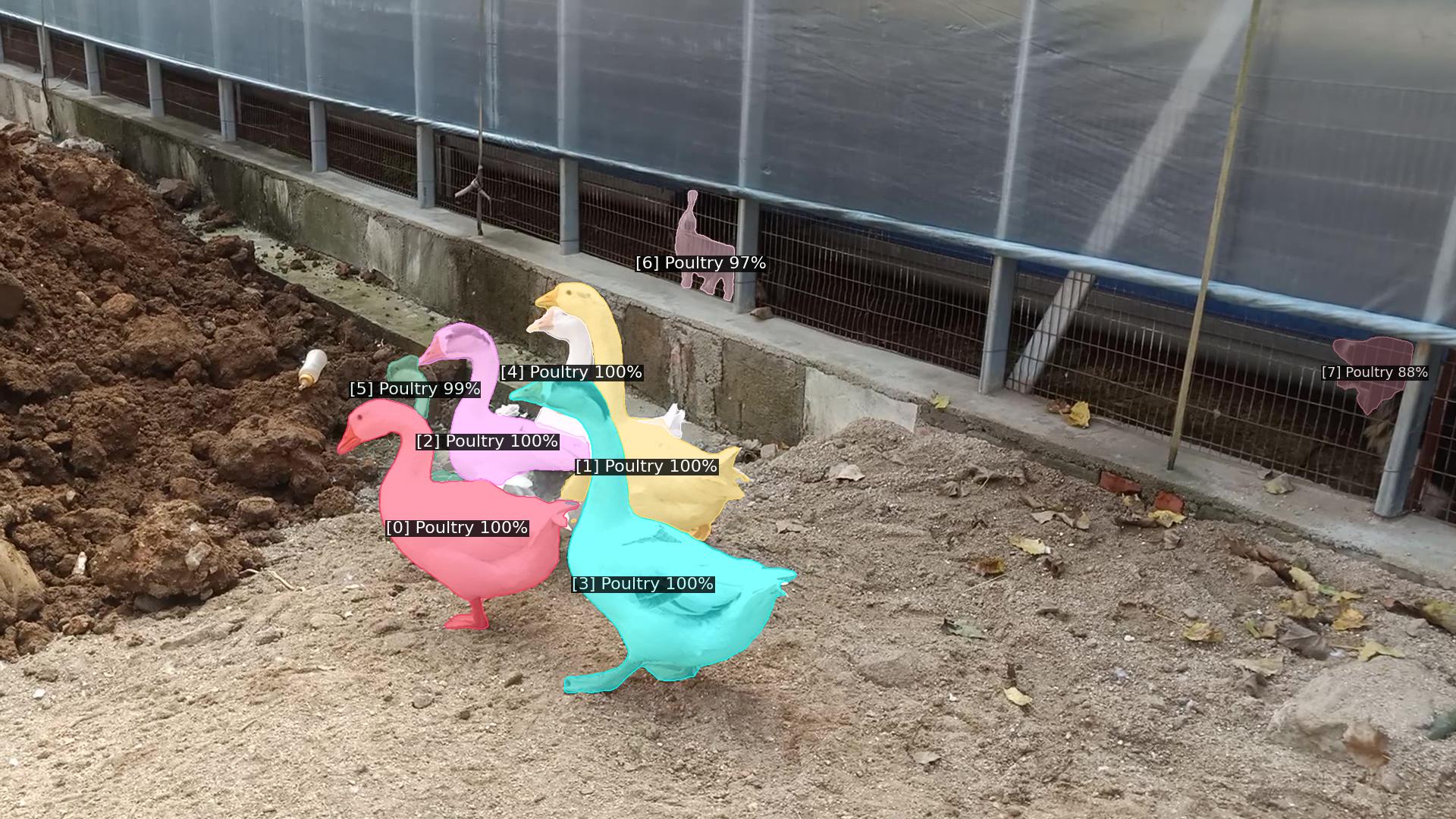}
\end{minipage}\hfill\vspace{1mm}

\caption{\textbf{Visualization results obtained on the OVIS dataset.}}
\label{fig:ovis demo}
\end{figure*}

\begin{figure*}[t]

\begin{minipage}[c]{1.00\linewidth}
\includegraphics[width=0.163\linewidth]{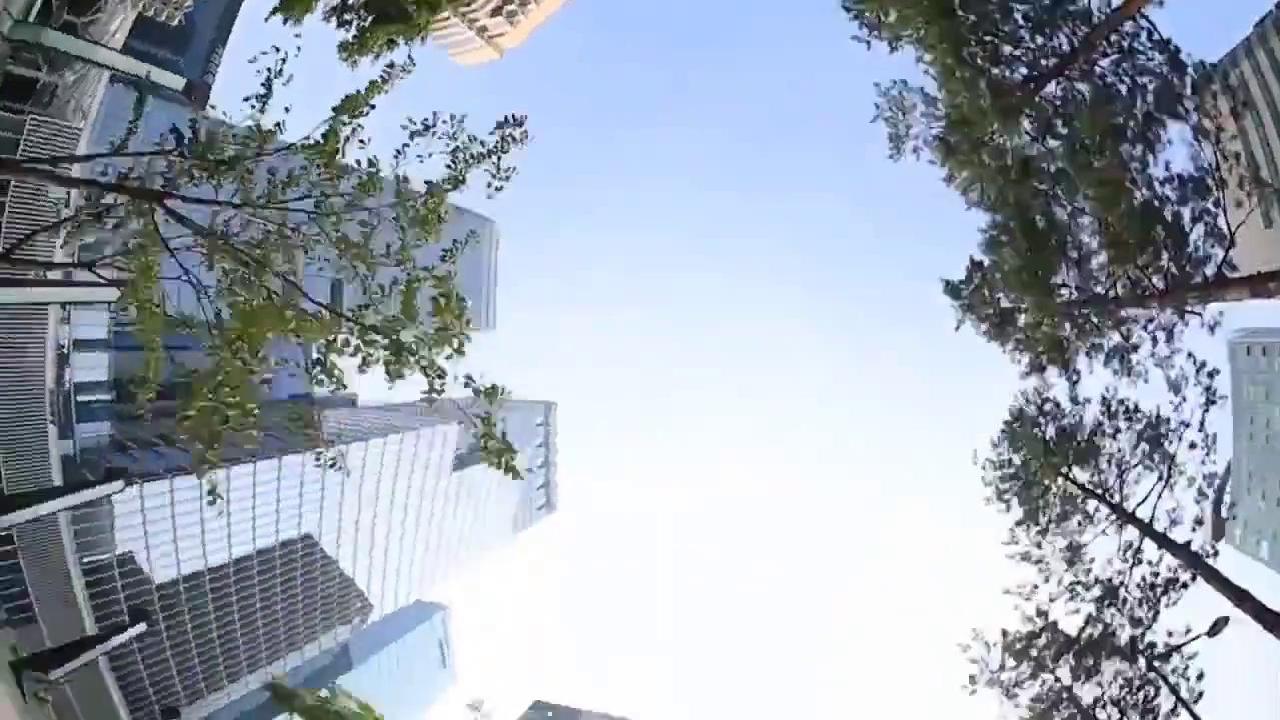}
\includegraphics[width=0.163\linewidth]{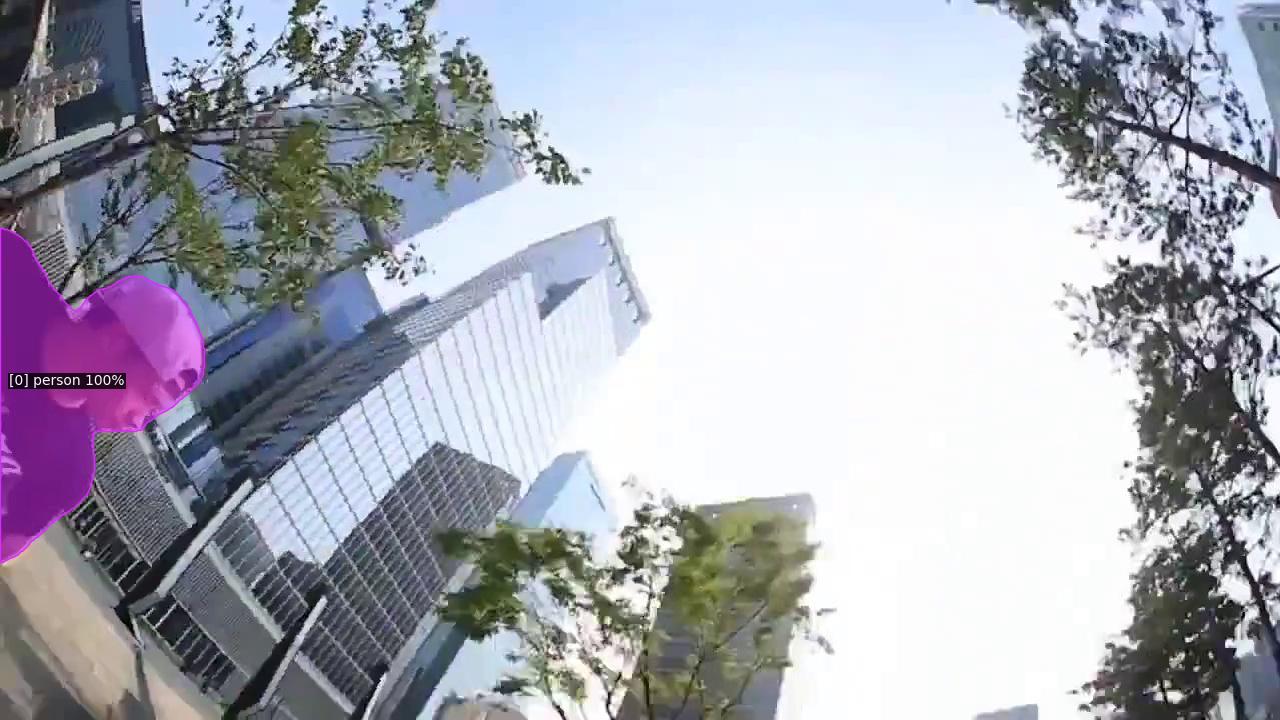}
\includegraphics[width=0.163\linewidth]{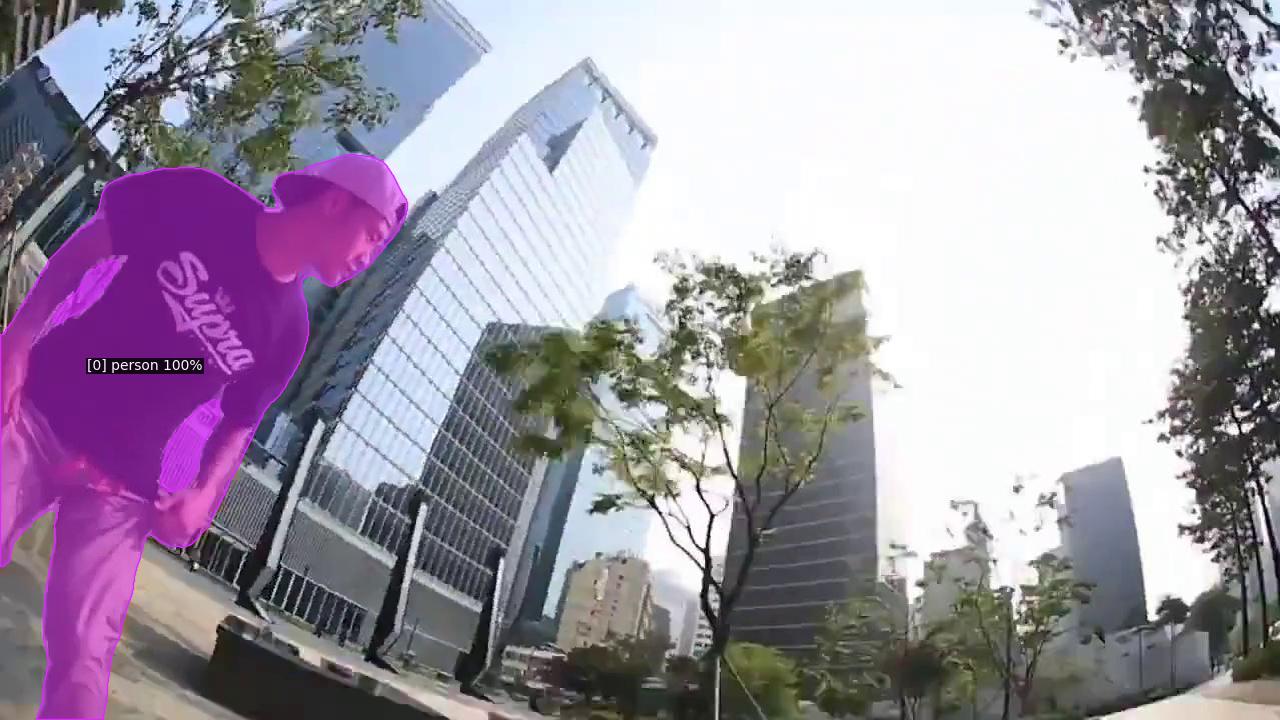}
\includegraphics[width=0.163\linewidth]{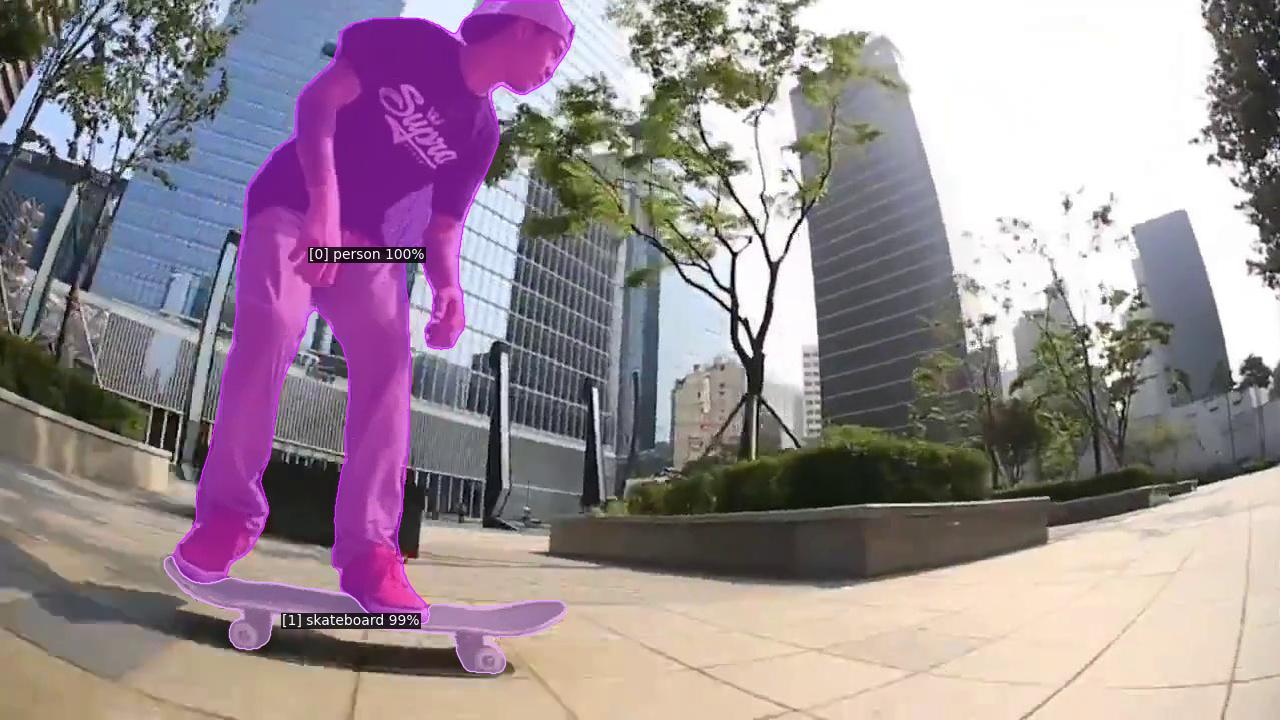}
\includegraphics[width=0.163\linewidth]{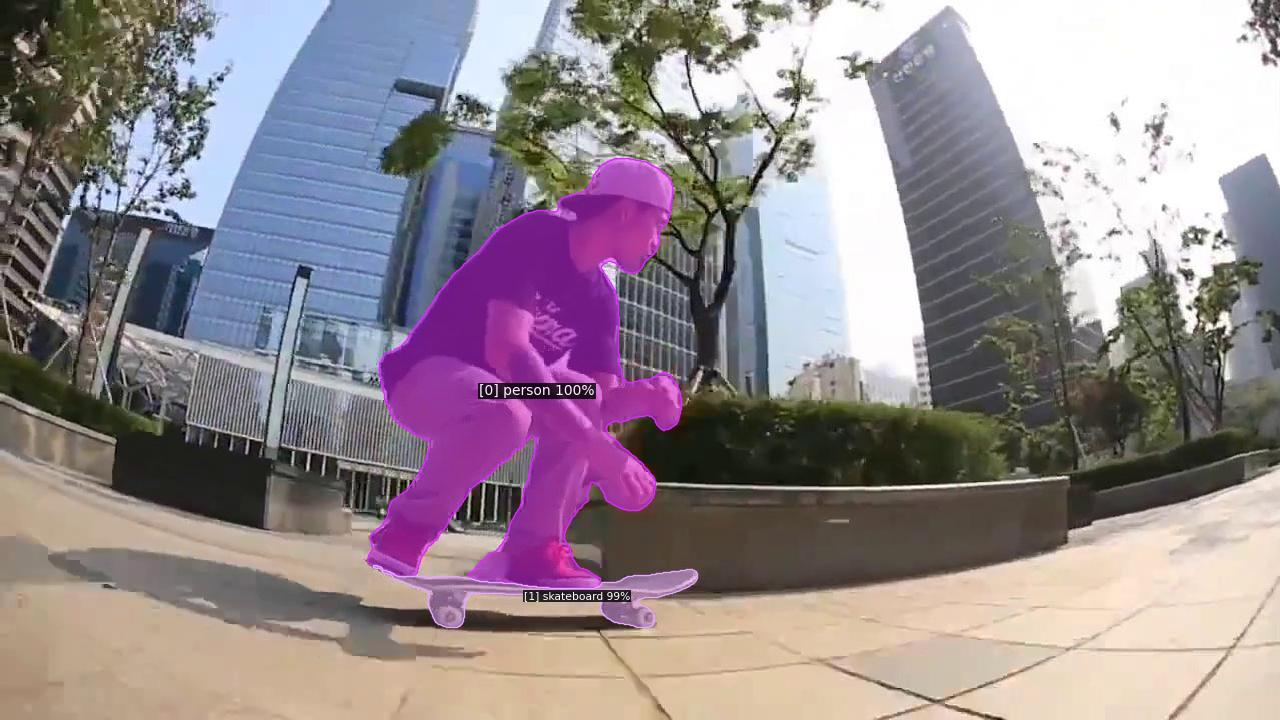}
\includegraphics[width=0.163\linewidth]{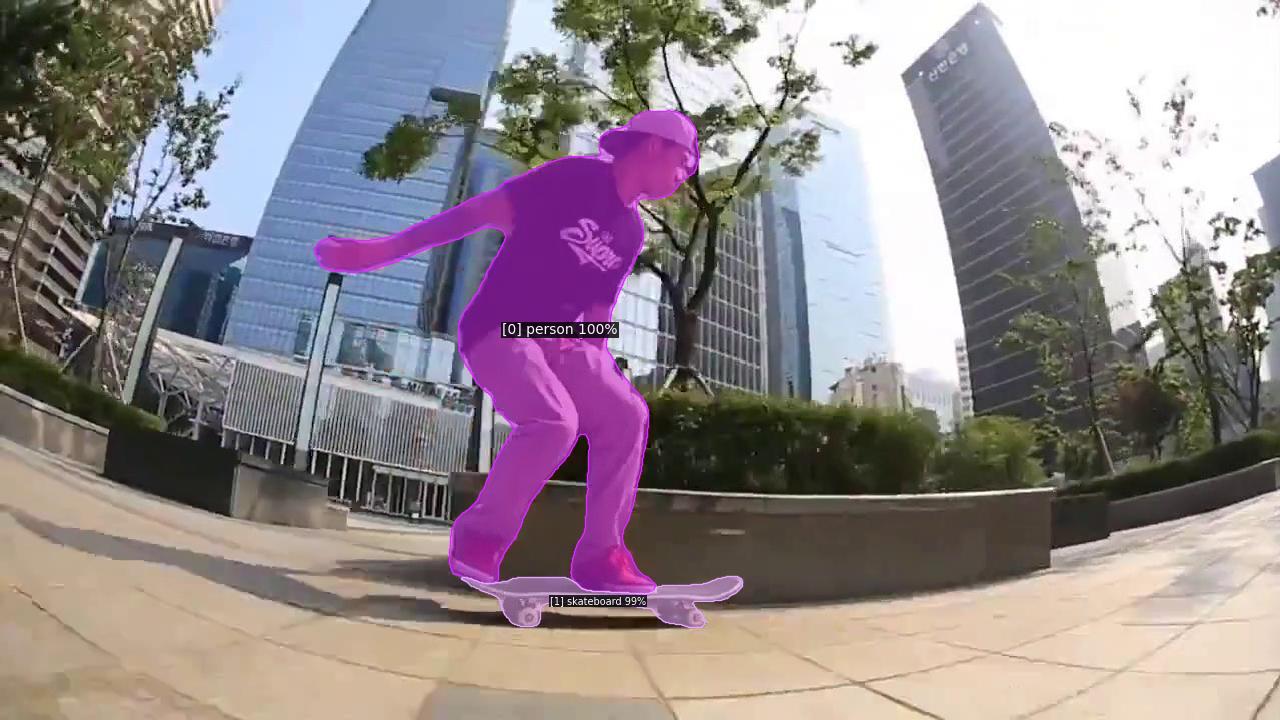}
\end{minipage}\hfill
\begin{minipage}[c]{1.0\linewidth}
\includegraphics[width=0.163\linewidth]{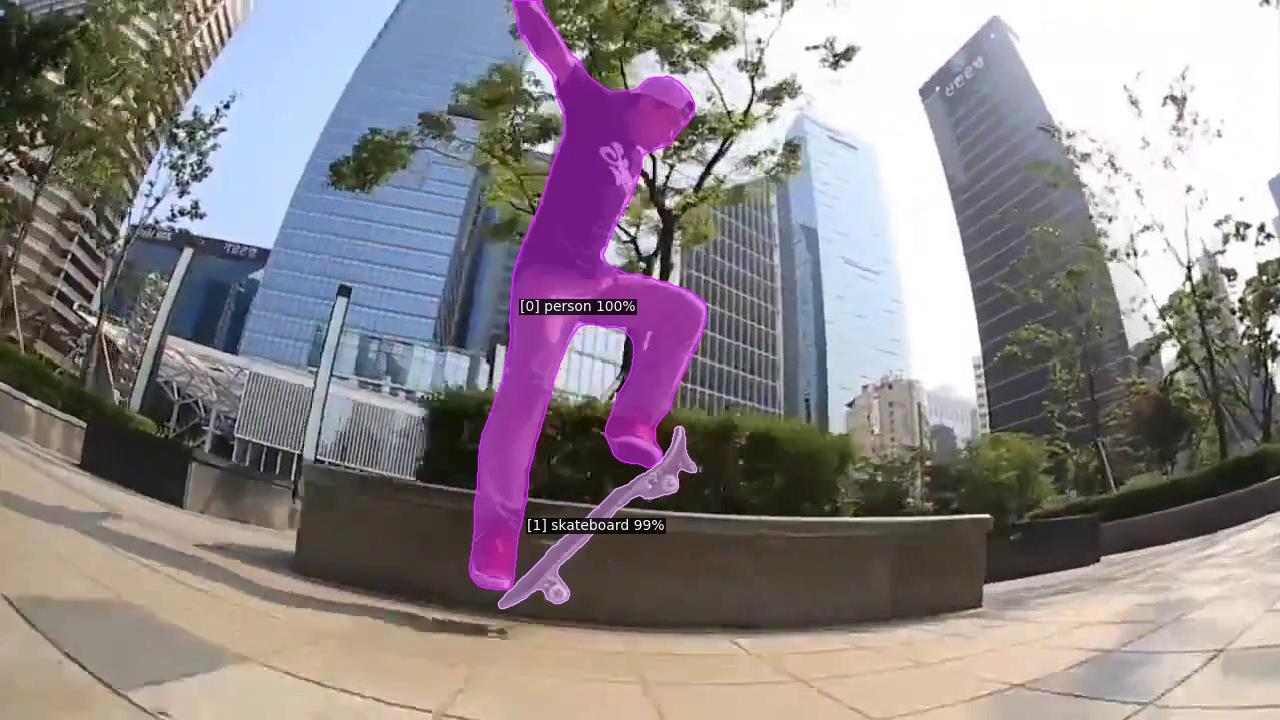}
\includegraphics[width=0.163\linewidth]{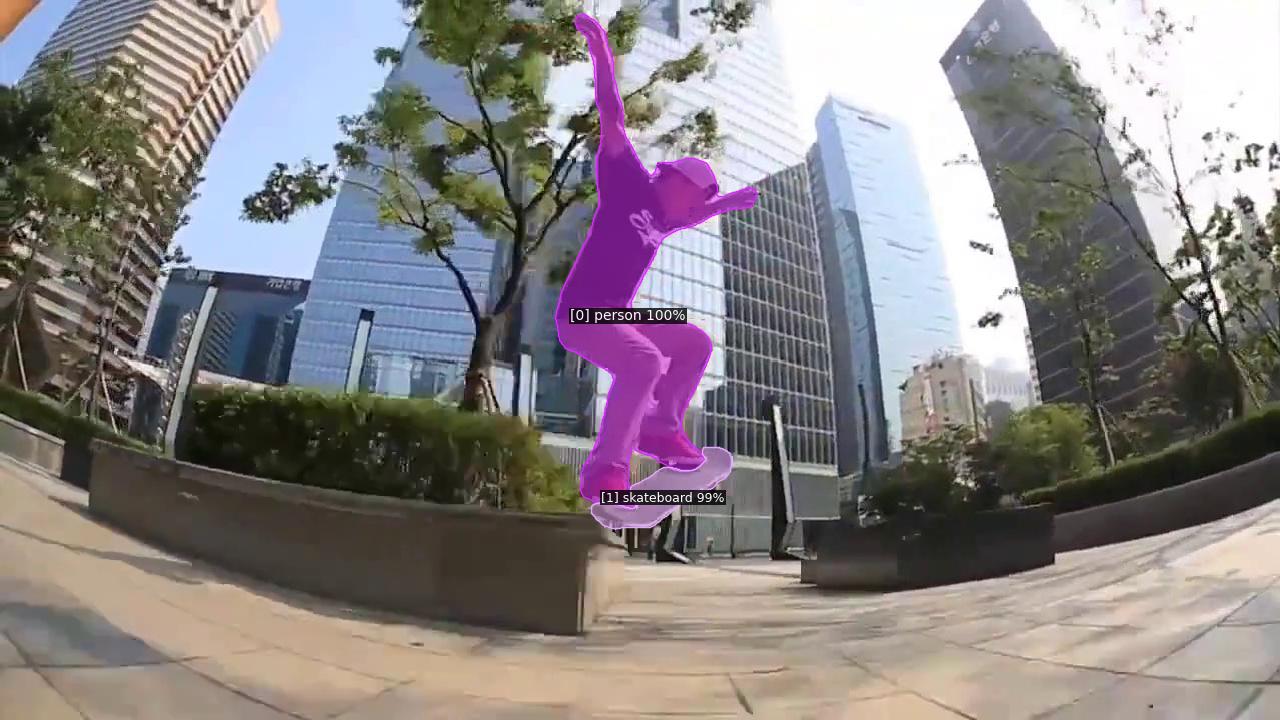}
\includegraphics[width=0.163\linewidth]{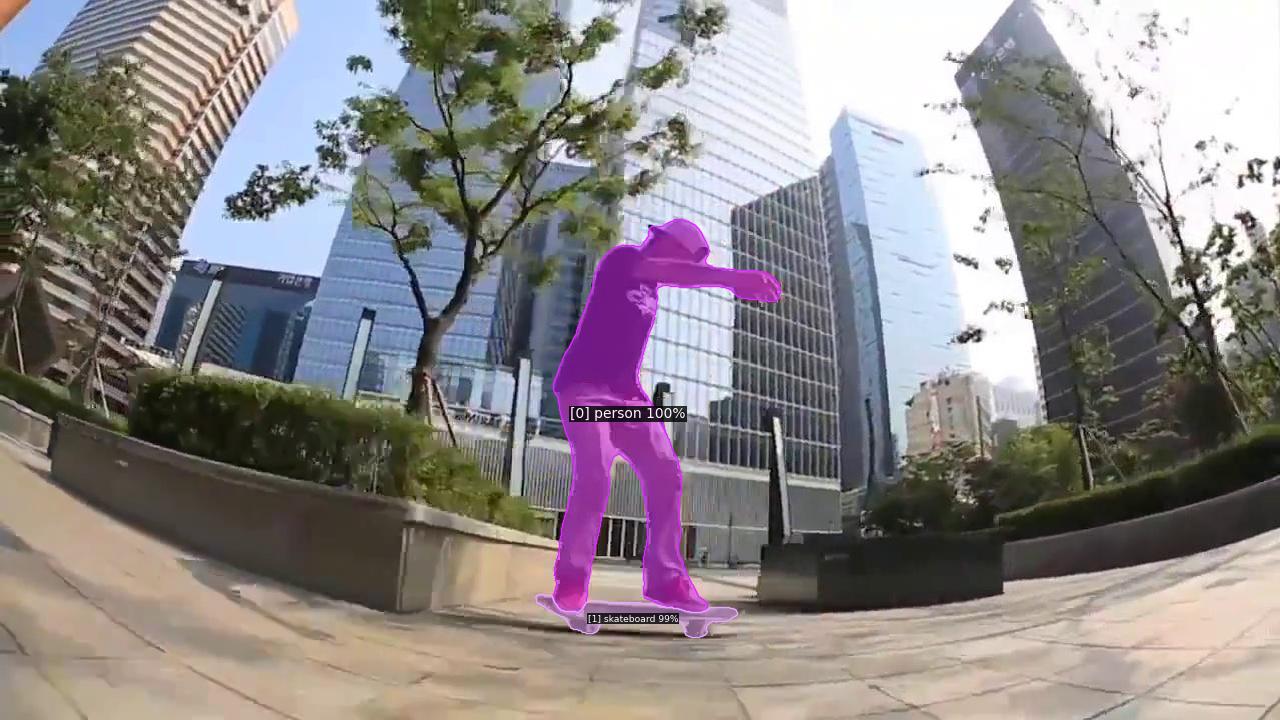}
\includegraphics[width=0.163\linewidth]{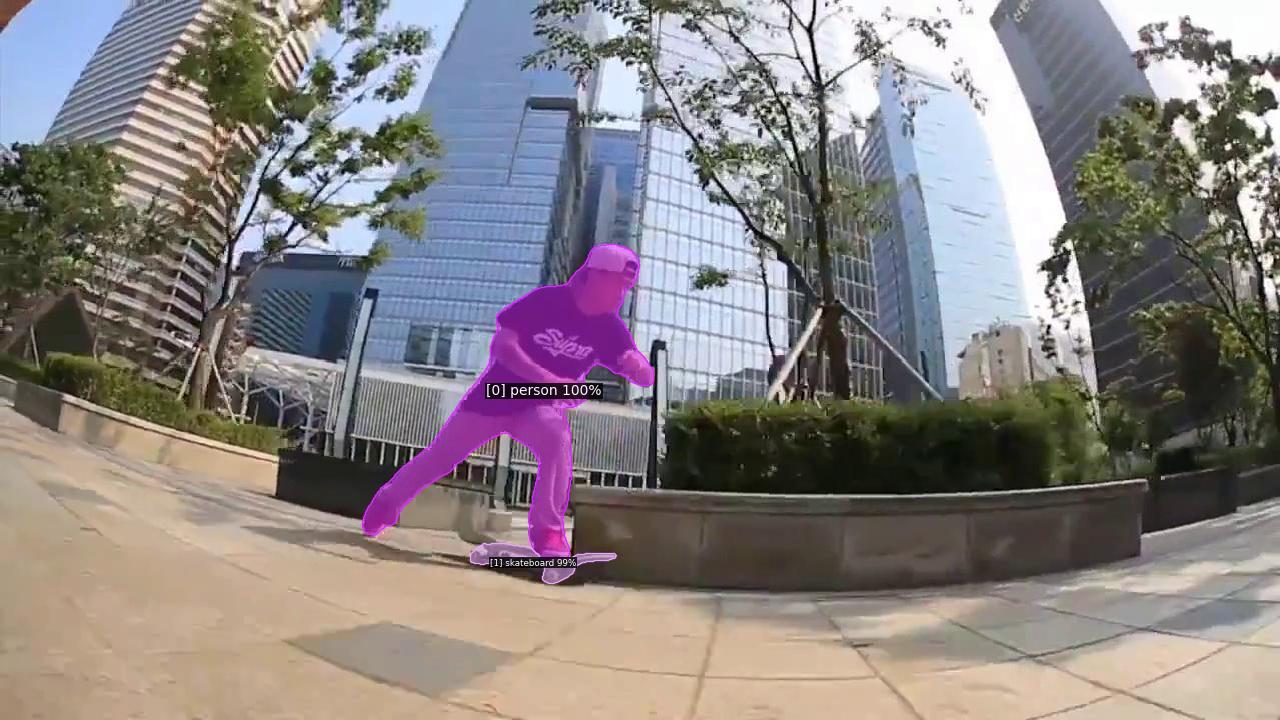}
\includegraphics[width=0.163\linewidth]{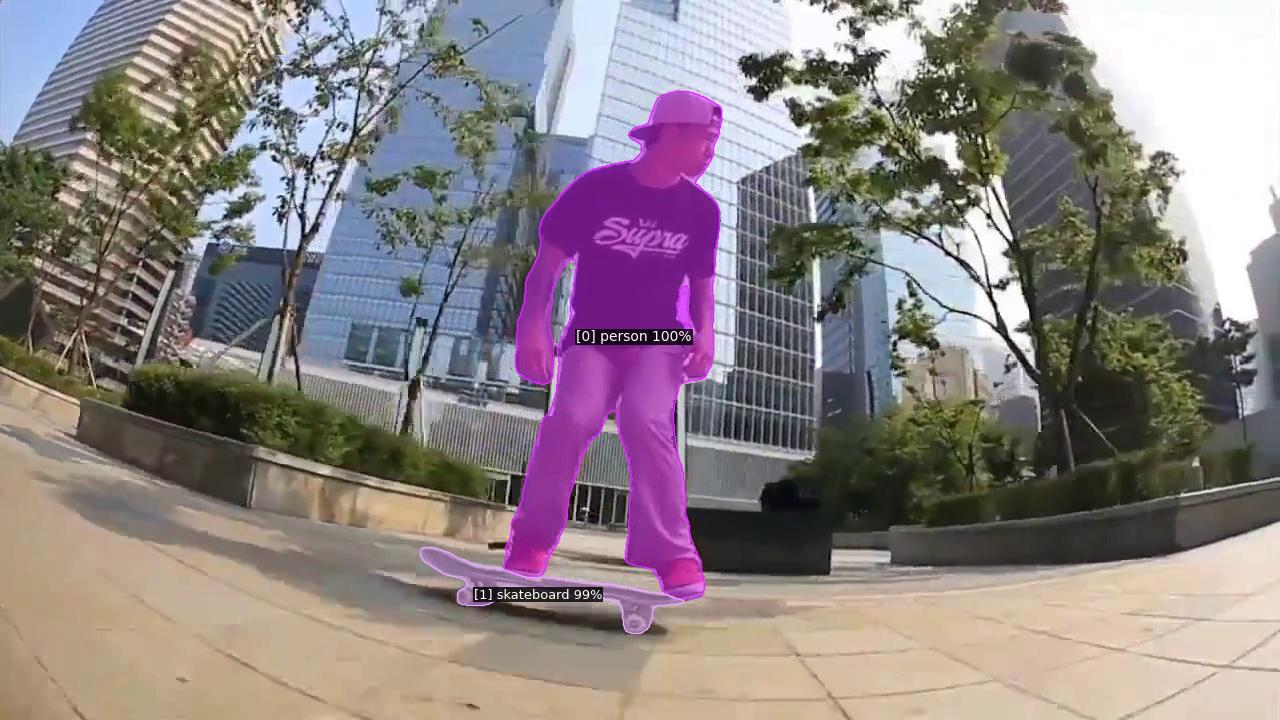}
\includegraphics[width=0.163\linewidth]{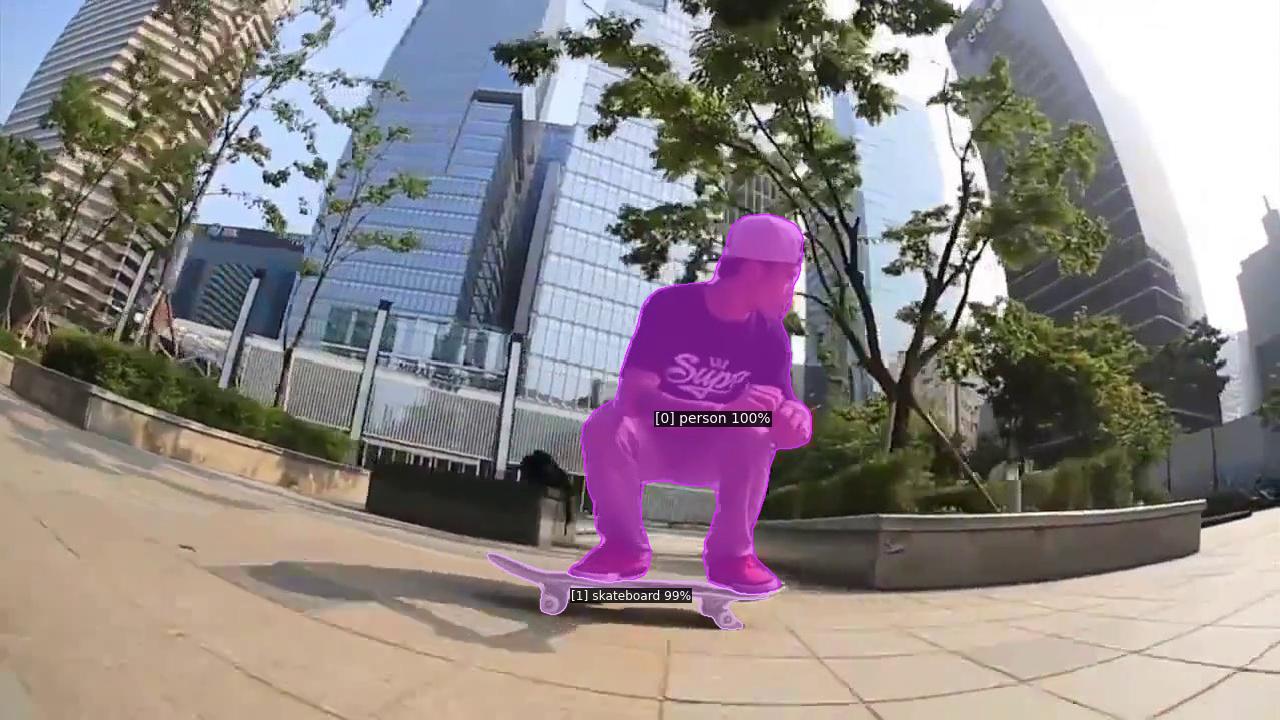}
\end{minipage}\hfill\vspace{1mm}

\begin{minipage}[c]{1.00\linewidth}
\includegraphics[width=0.163\linewidth]{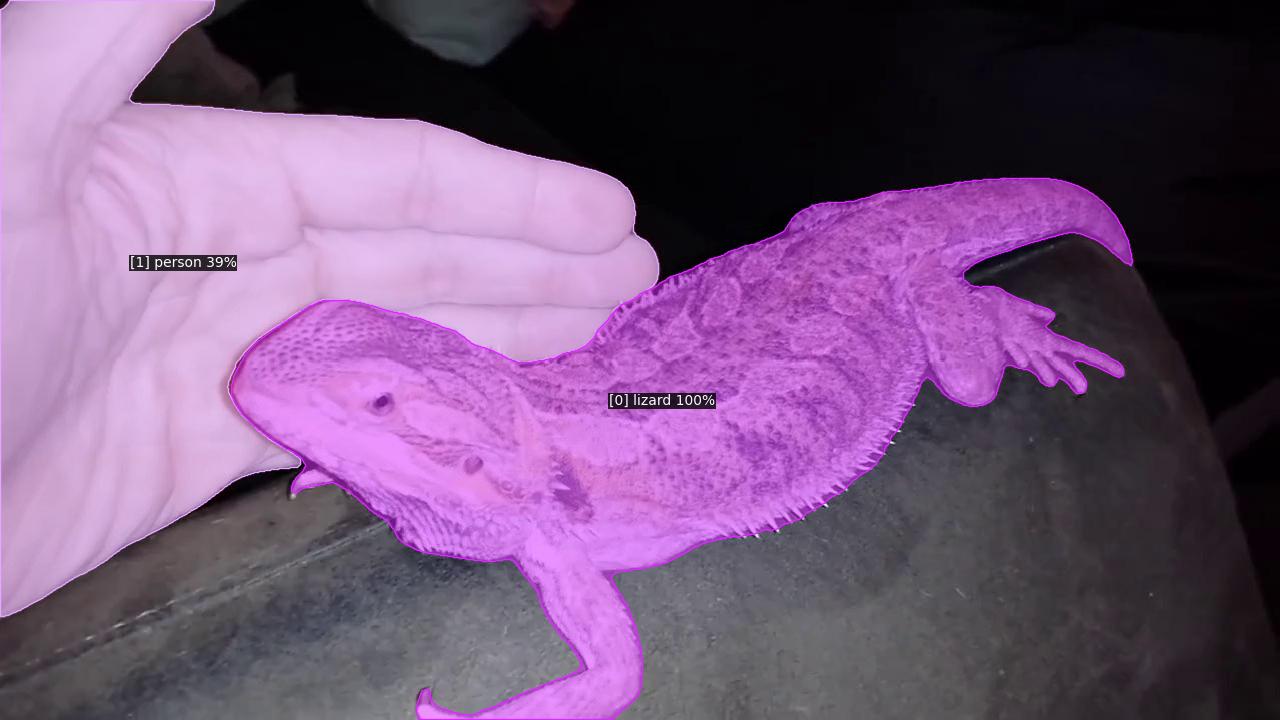}
\includegraphics[width=0.163\linewidth]{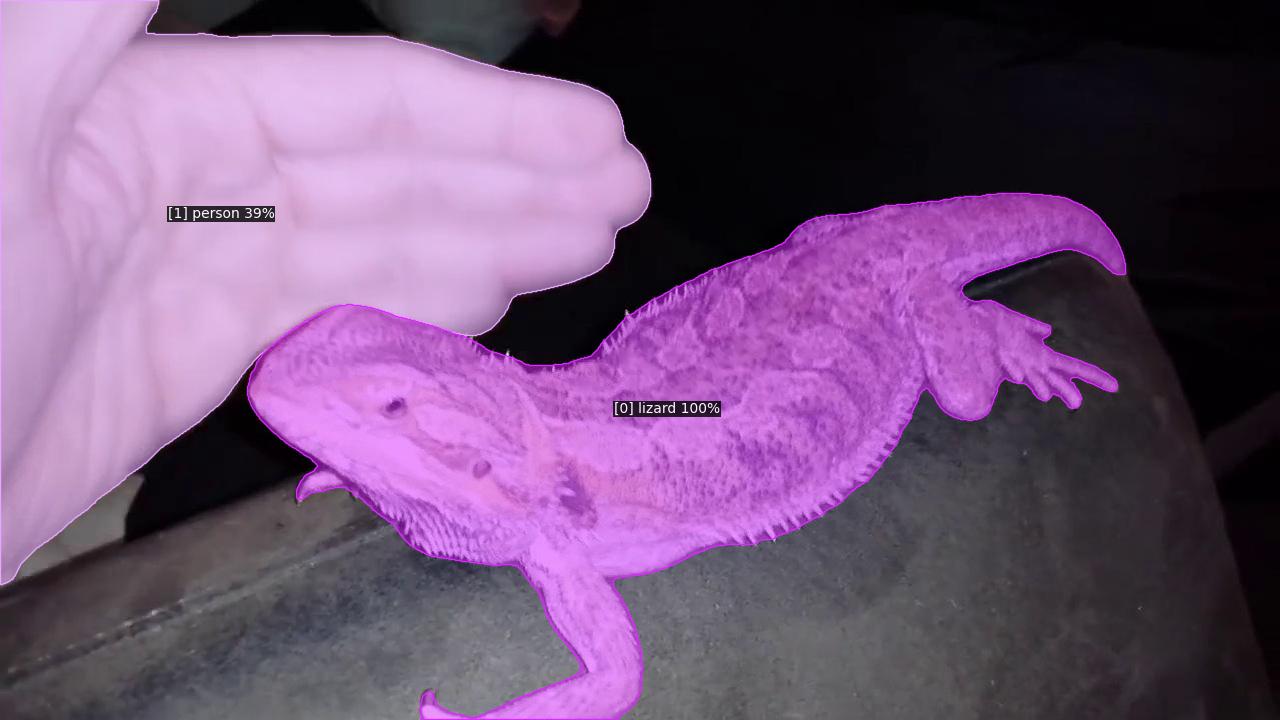}
\includegraphics[width=0.163\linewidth]{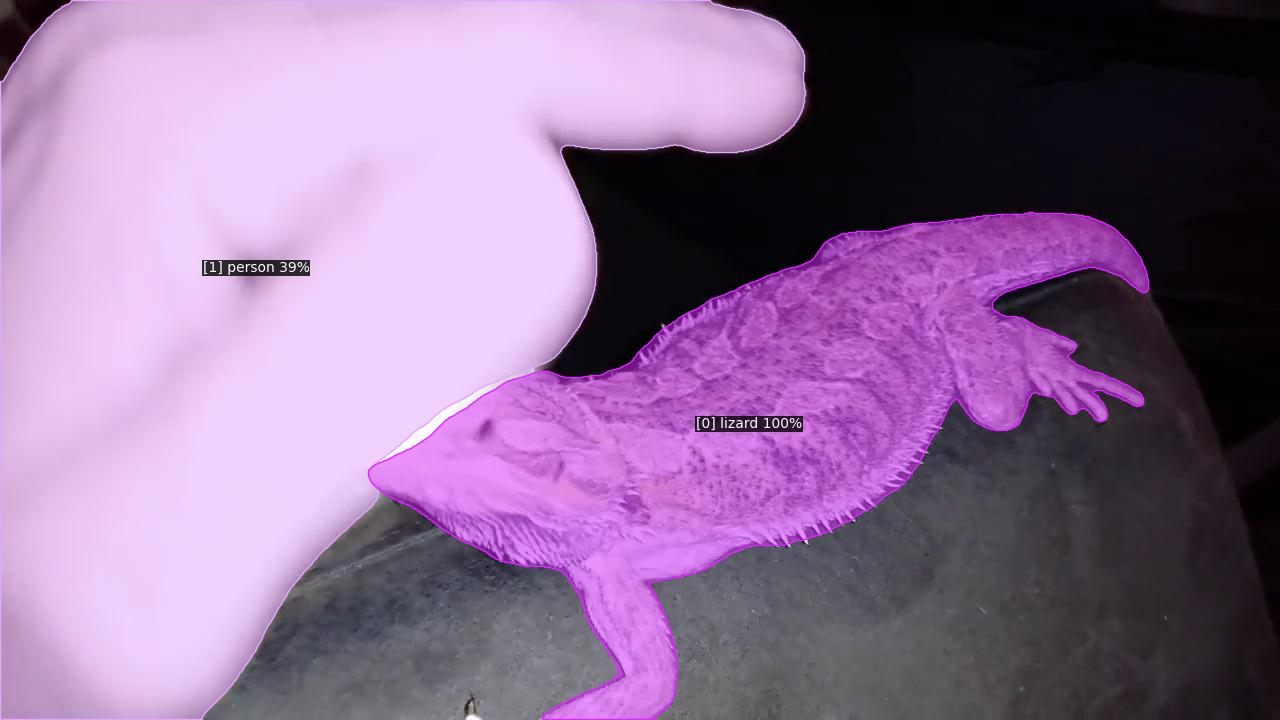}
\includegraphics[width=0.163\linewidth]{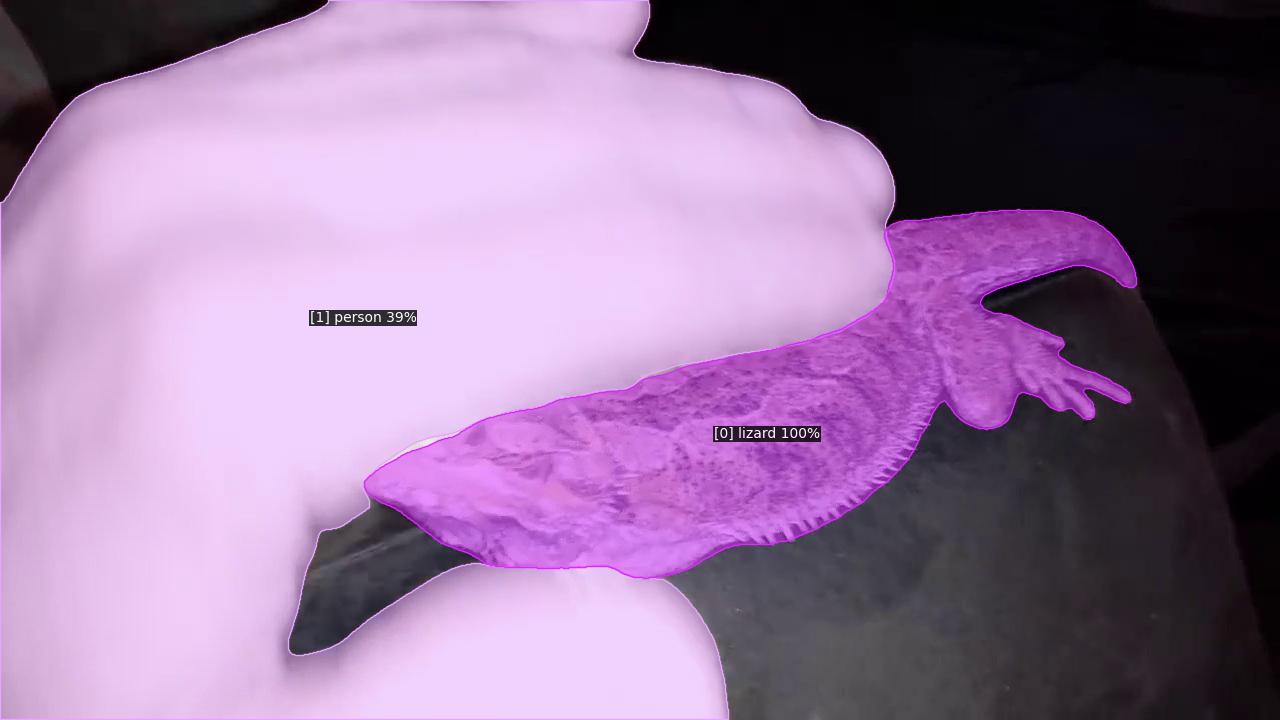}
\includegraphics[width=0.163\linewidth]{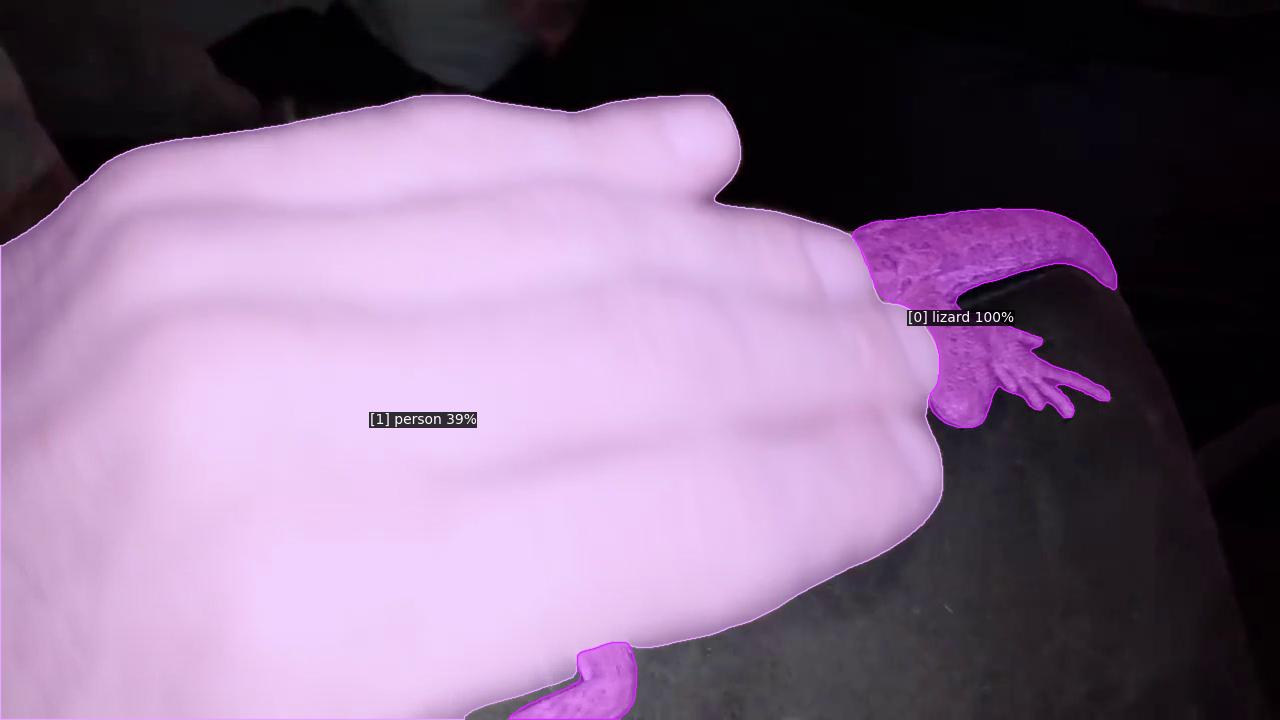}
\includegraphics[width=0.163\linewidth]{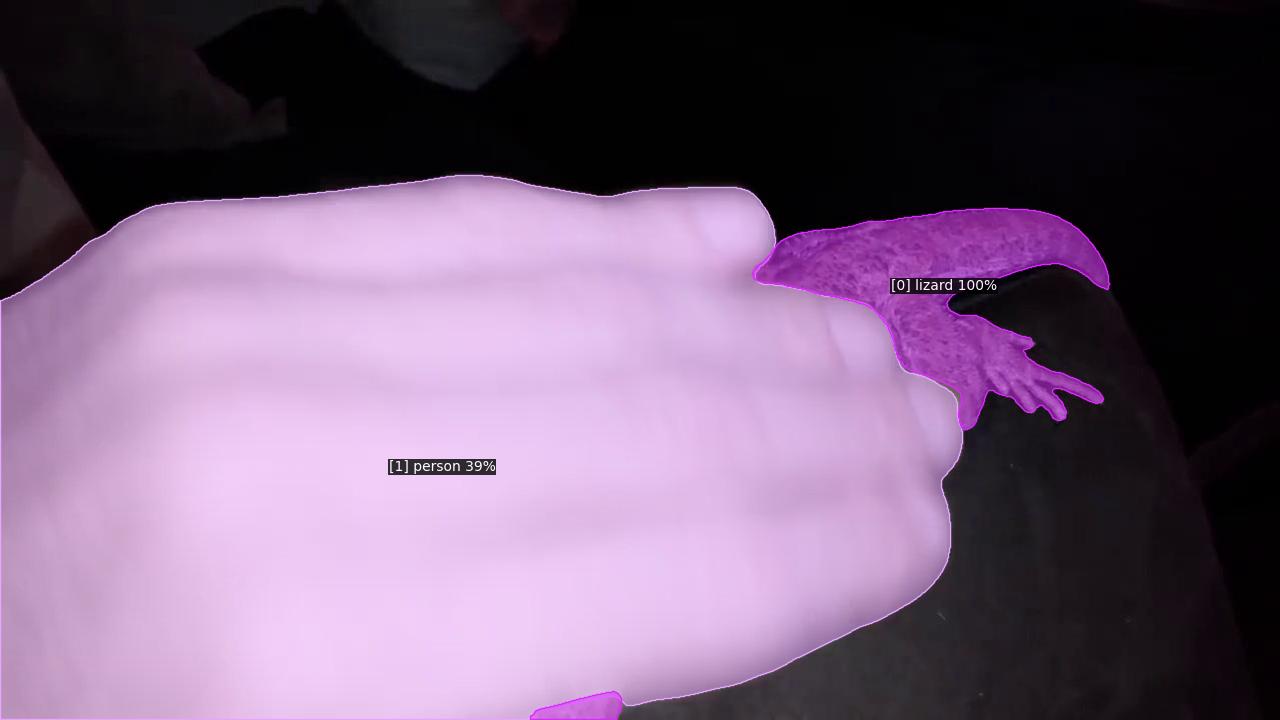}
\end{minipage}\hfill
\begin{minipage}[c]{1.0\linewidth}
\includegraphics[width=0.163\linewidth]{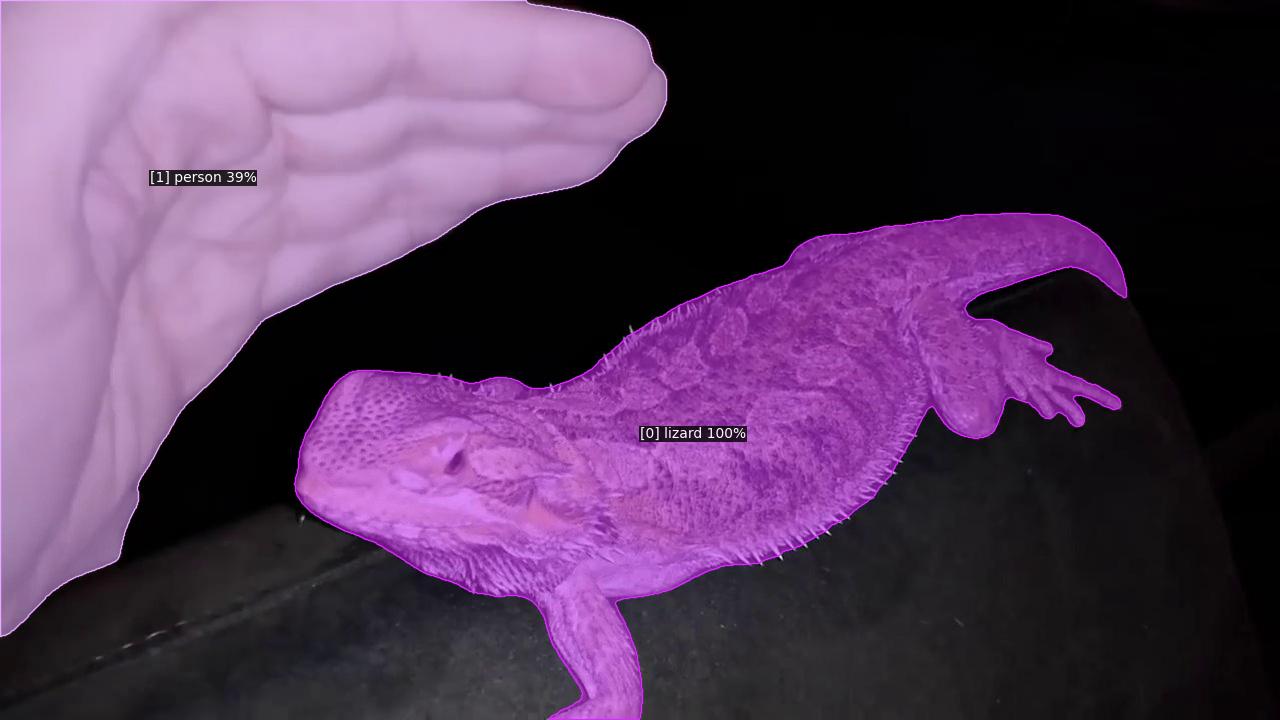}
\includegraphics[width=0.163\linewidth]{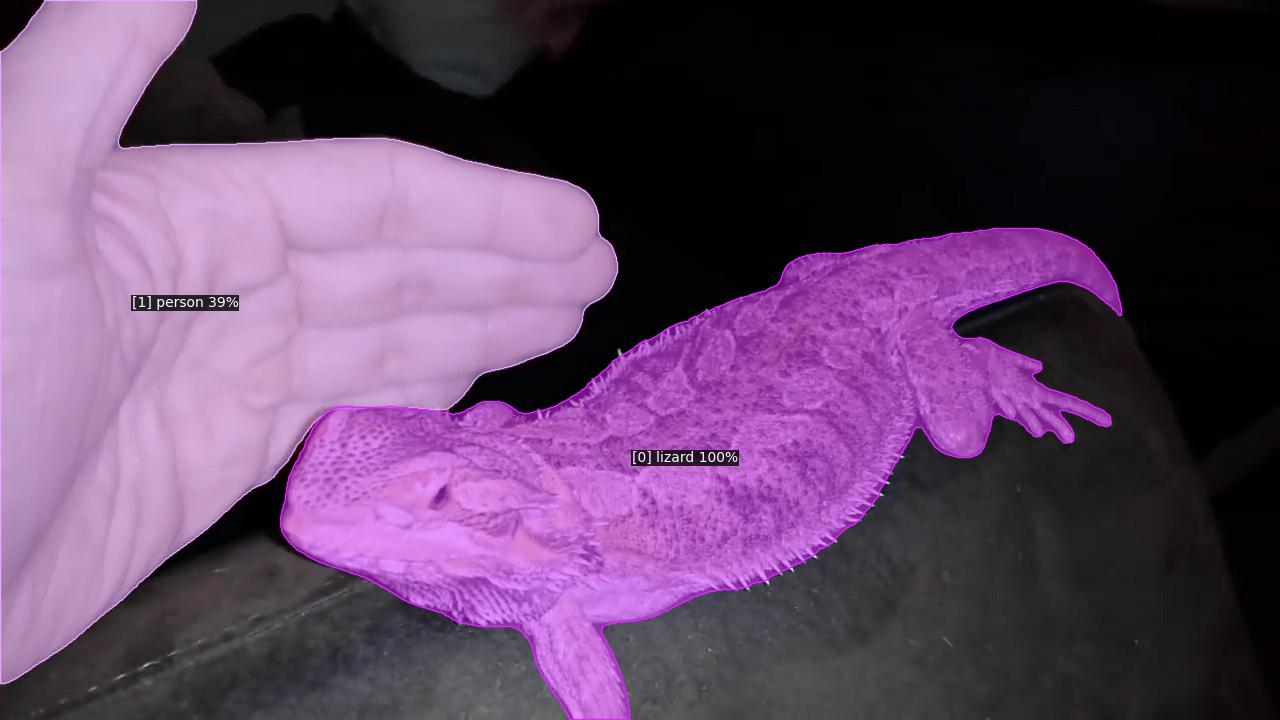}
\includegraphics[width=0.163\linewidth]{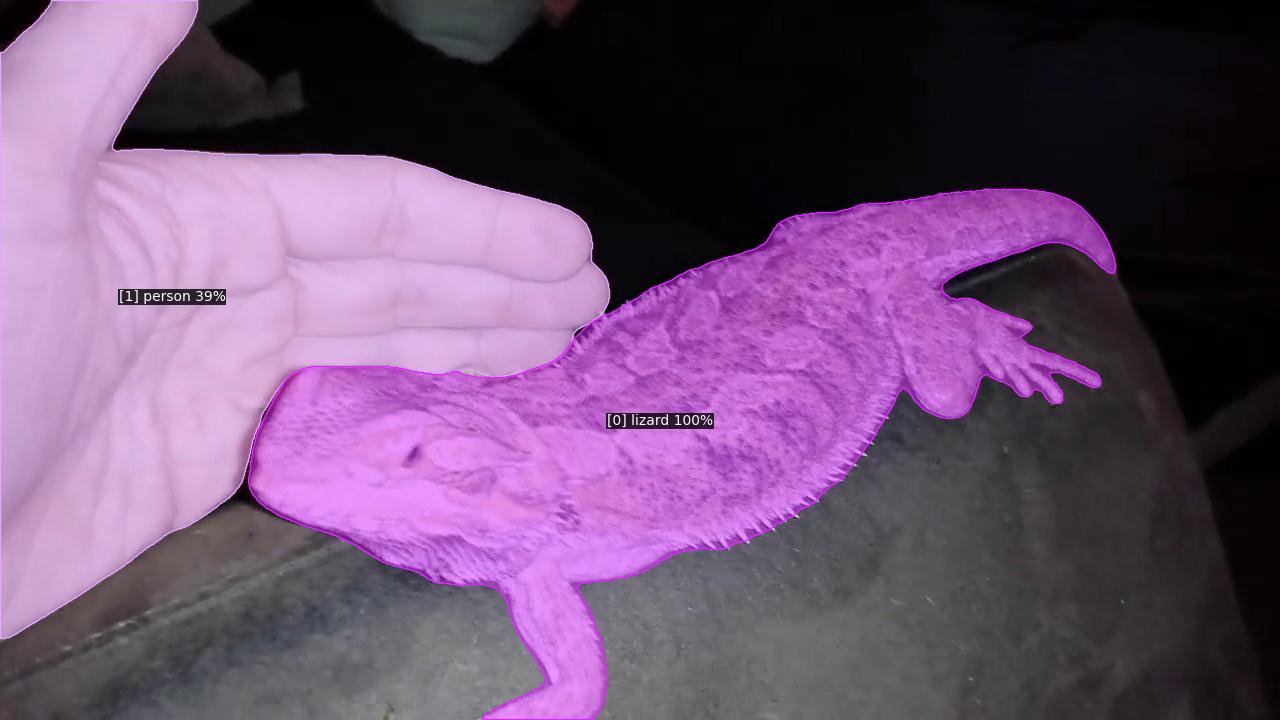}
\includegraphics[width=0.163\linewidth]{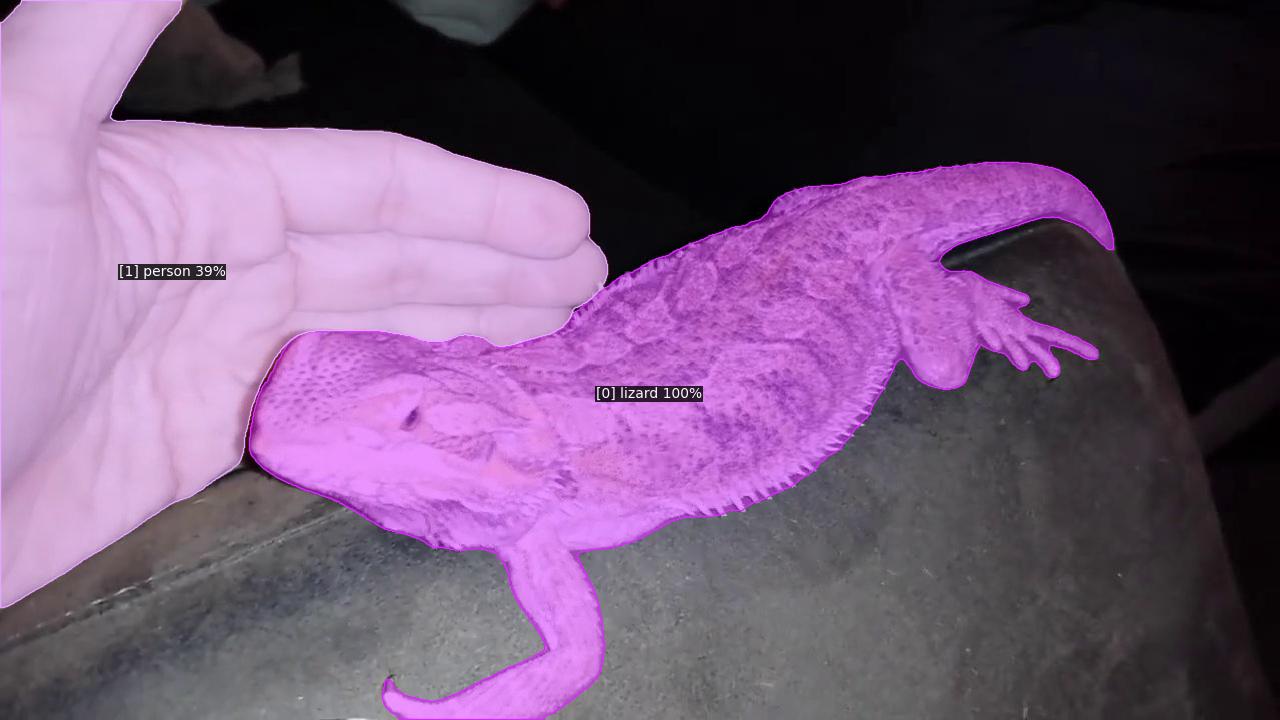}
\includegraphics[width=0.163\linewidth]{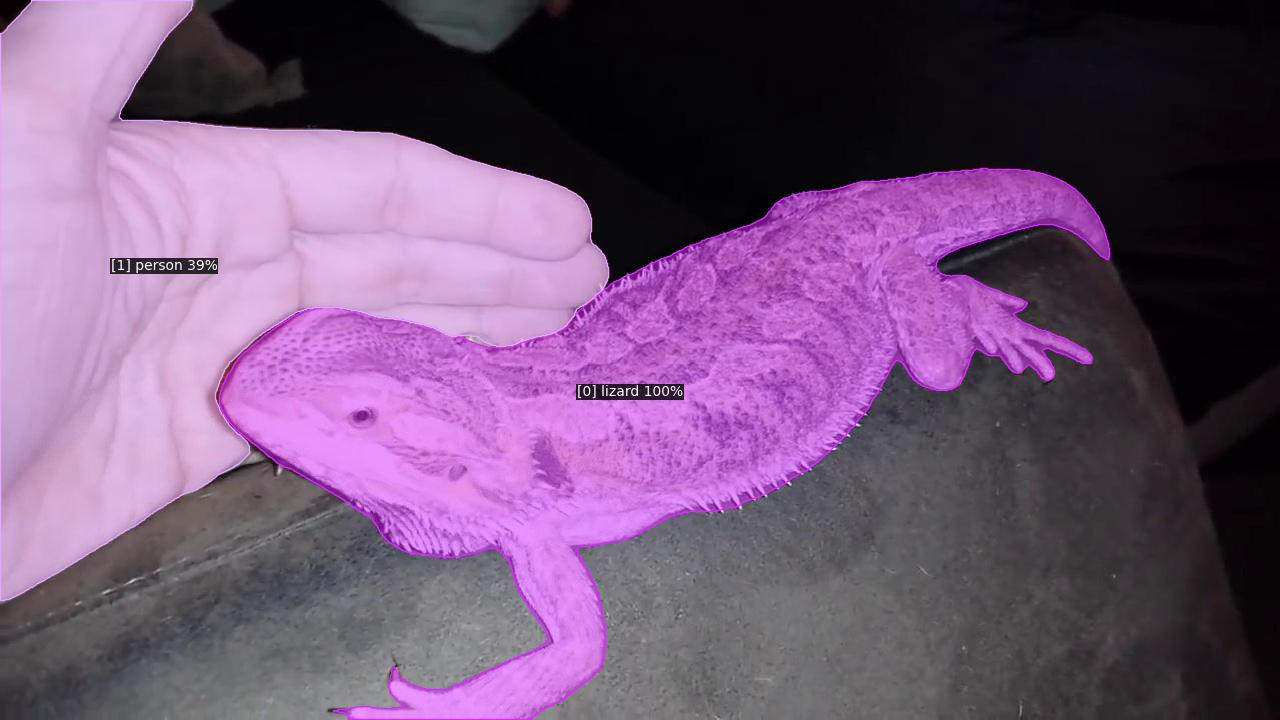}
\includegraphics[width=0.163\linewidth]{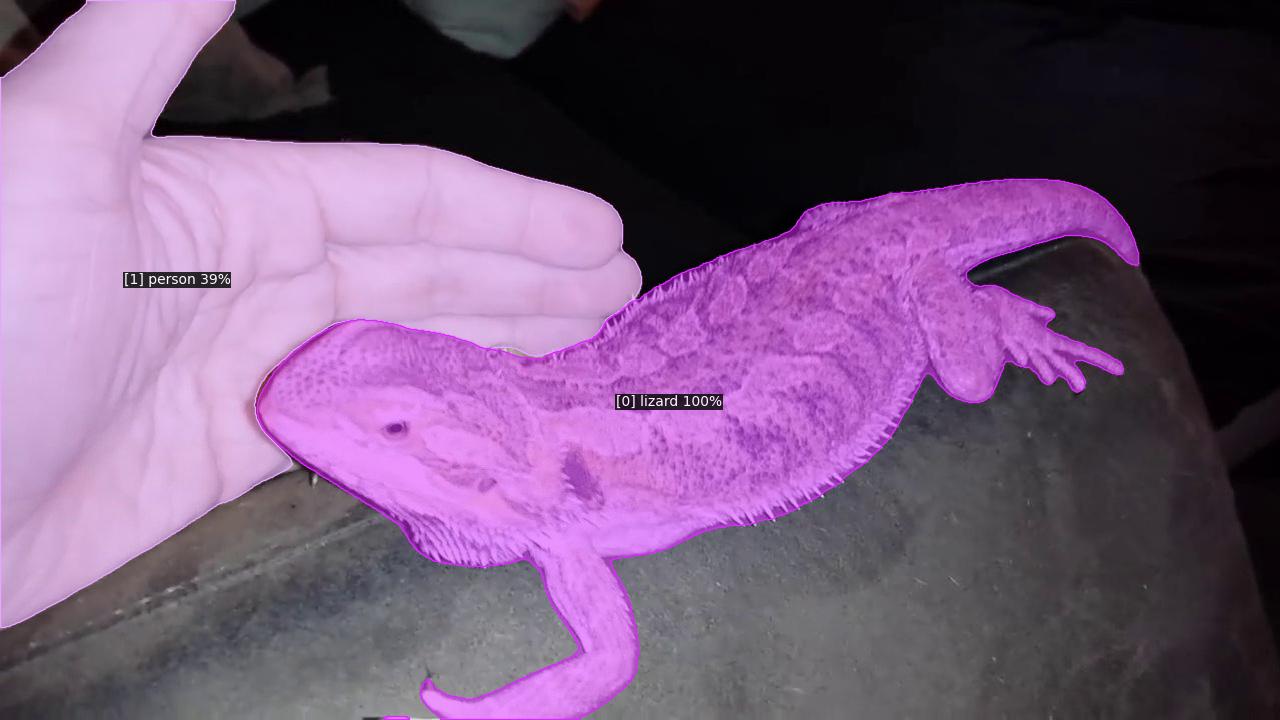}
\end{minipage}\hfill\vspace{1mm}

\begin{minipage}[c]{1.00\linewidth}
\includegraphics[width=0.163\linewidth]{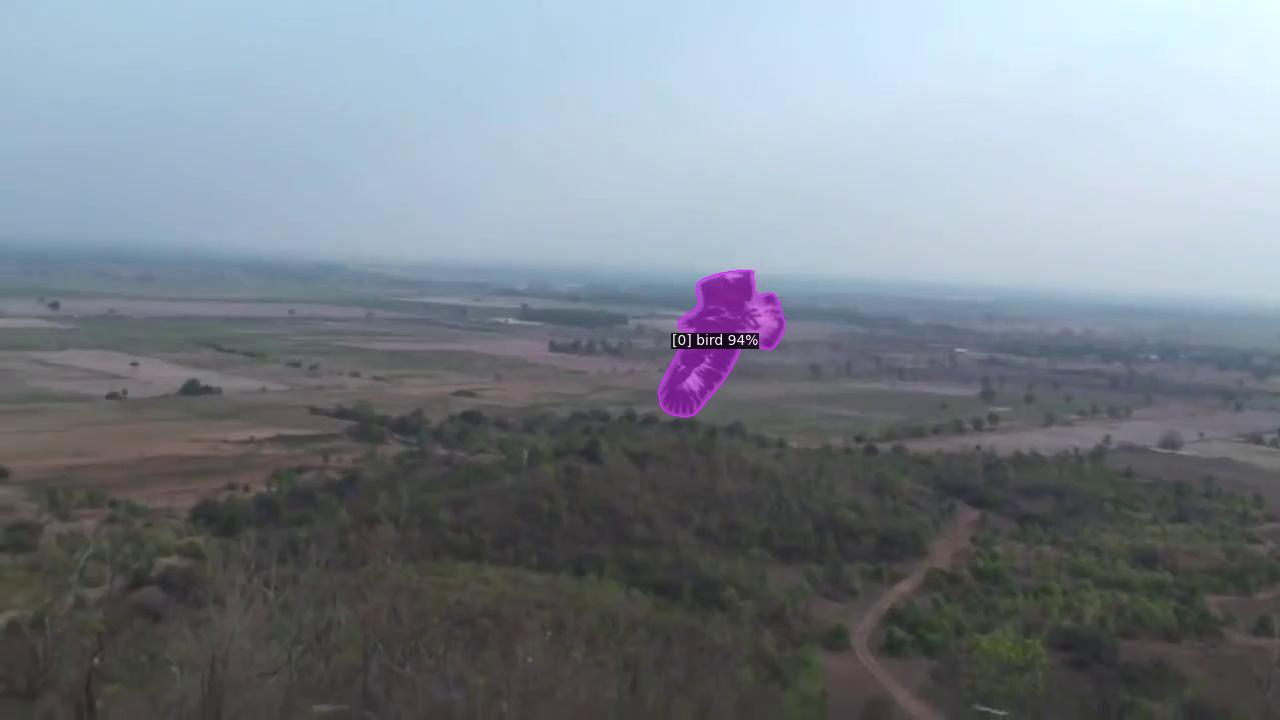}
\includegraphics[width=0.163\linewidth]{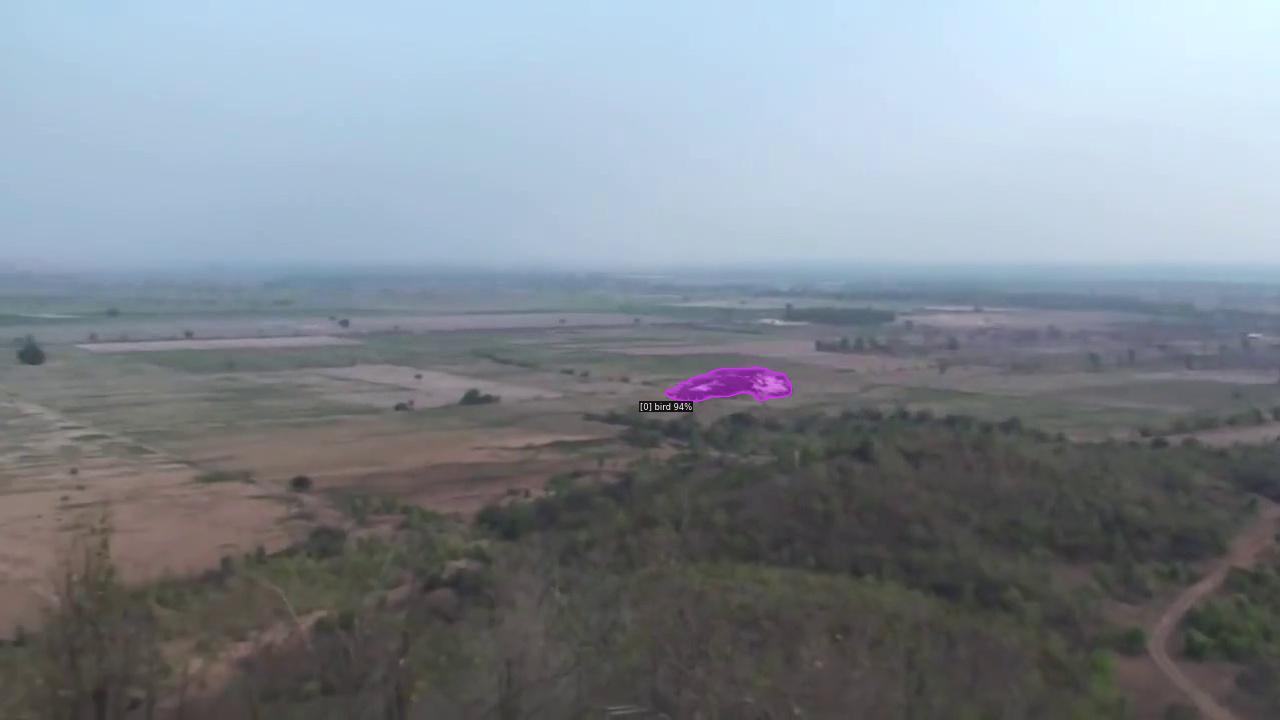}
\includegraphics[width=0.163\linewidth]{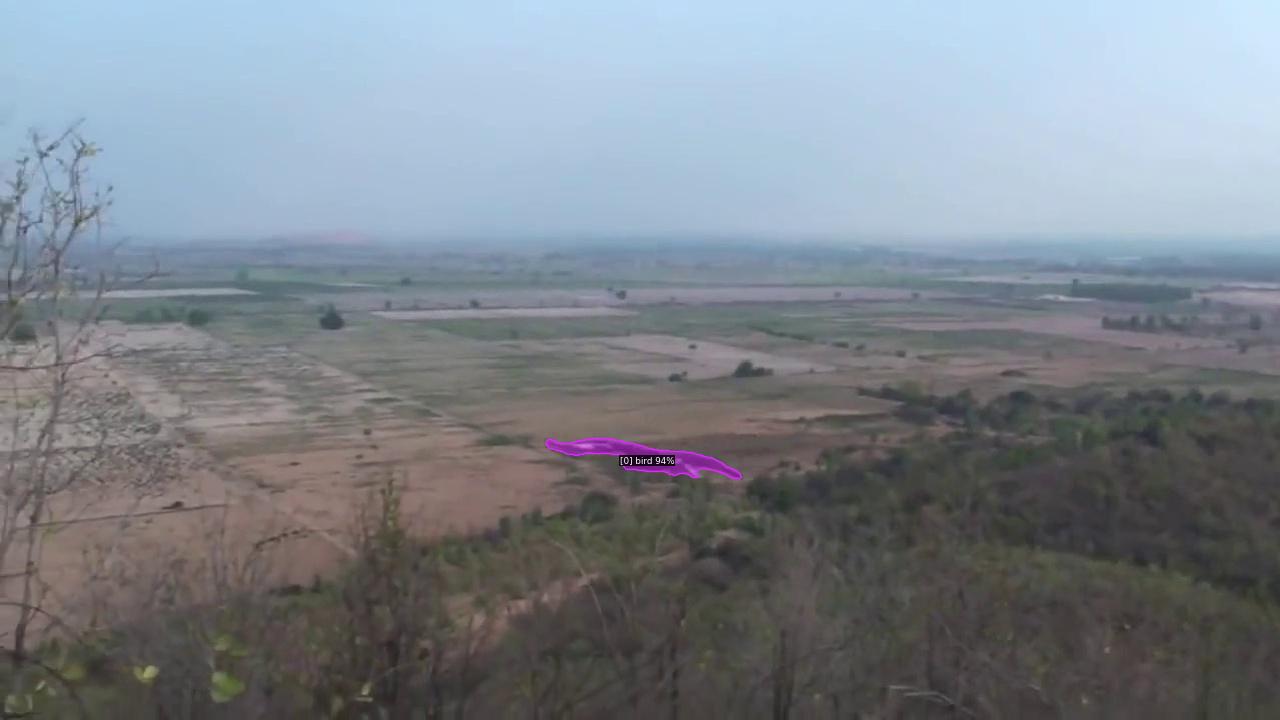}
\includegraphics[width=0.163\linewidth]{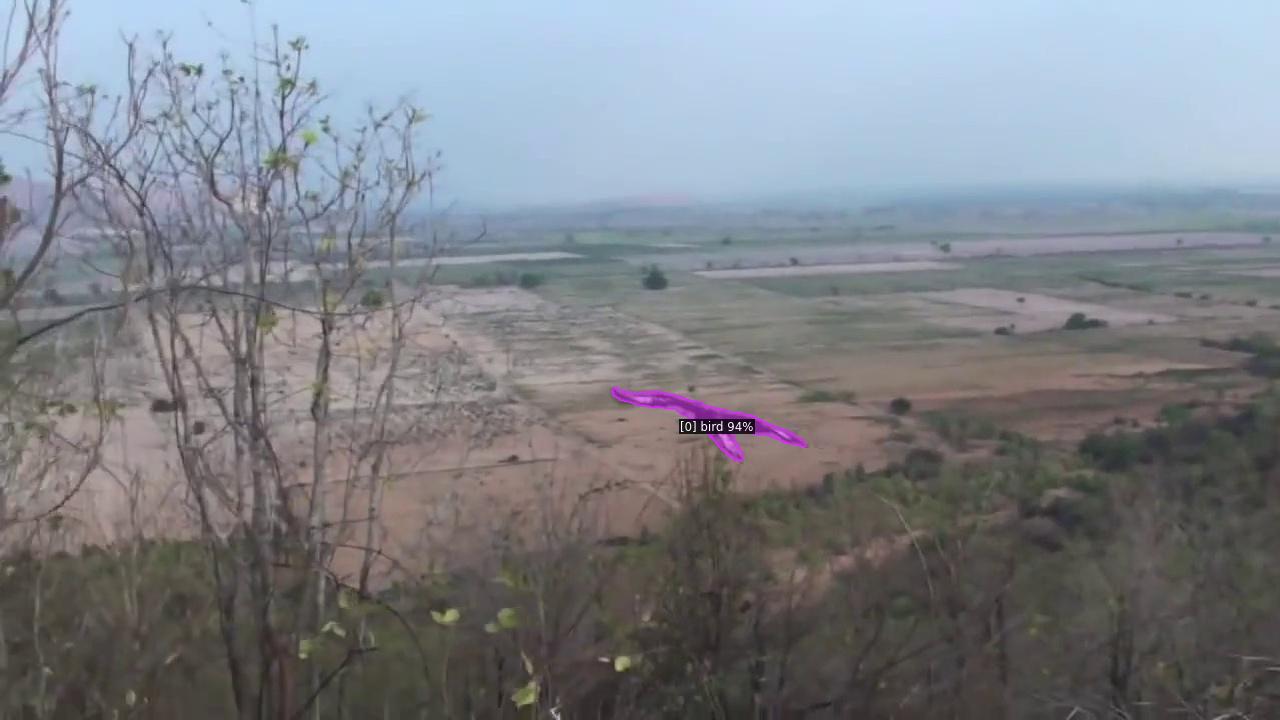}
\includegraphics[width=0.163\linewidth]{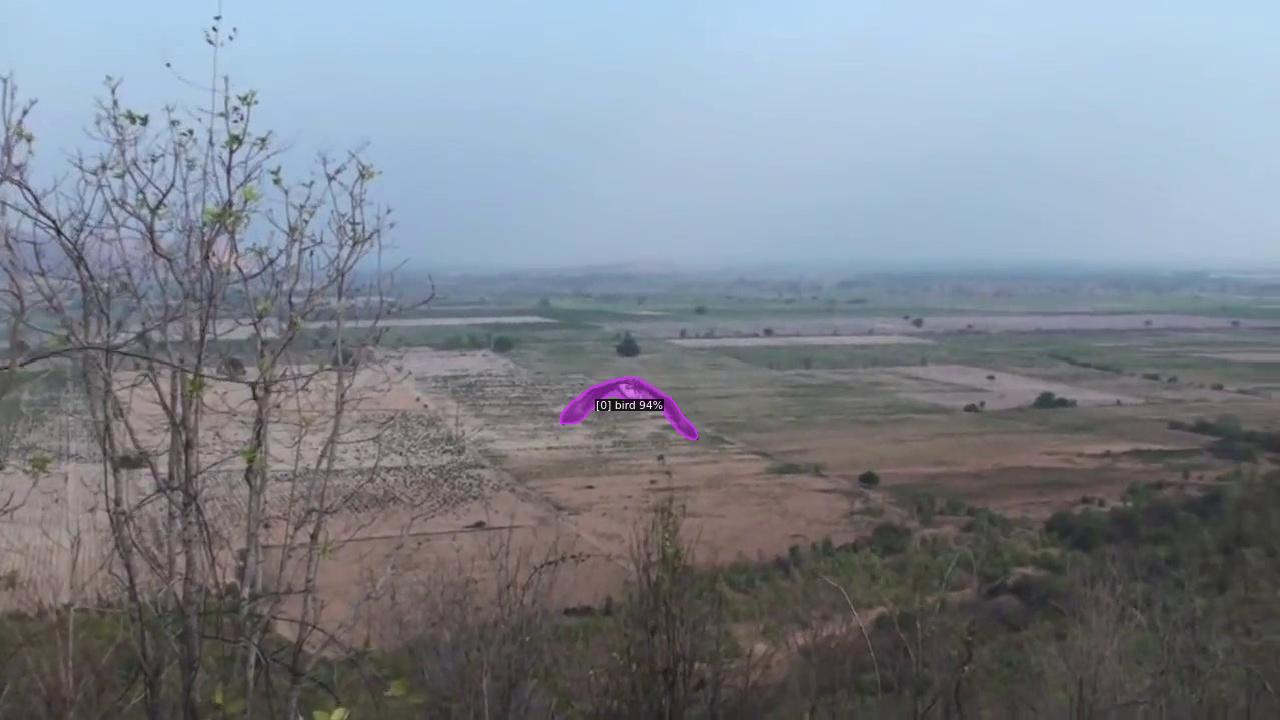}
\includegraphics[width=0.163\linewidth]{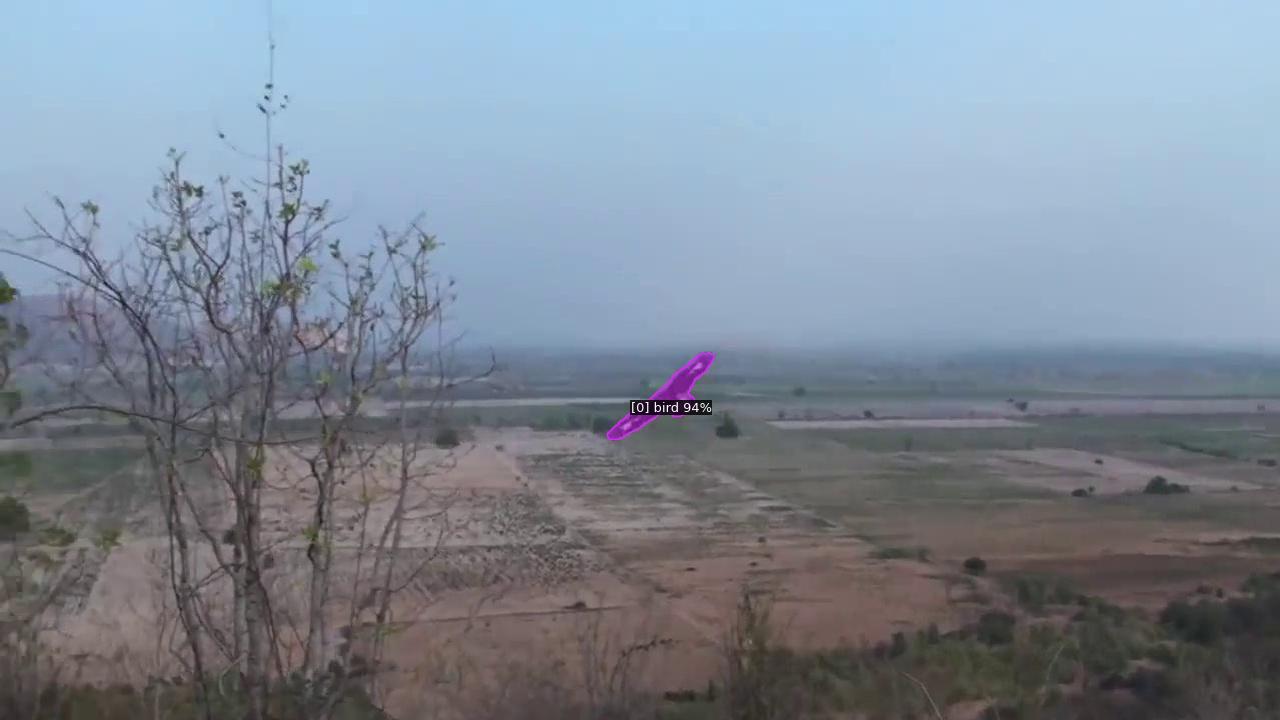}
\end{minipage}\hfill
\begin{minipage}[c]{1.0\linewidth}
\includegraphics[width=0.163\linewidth]{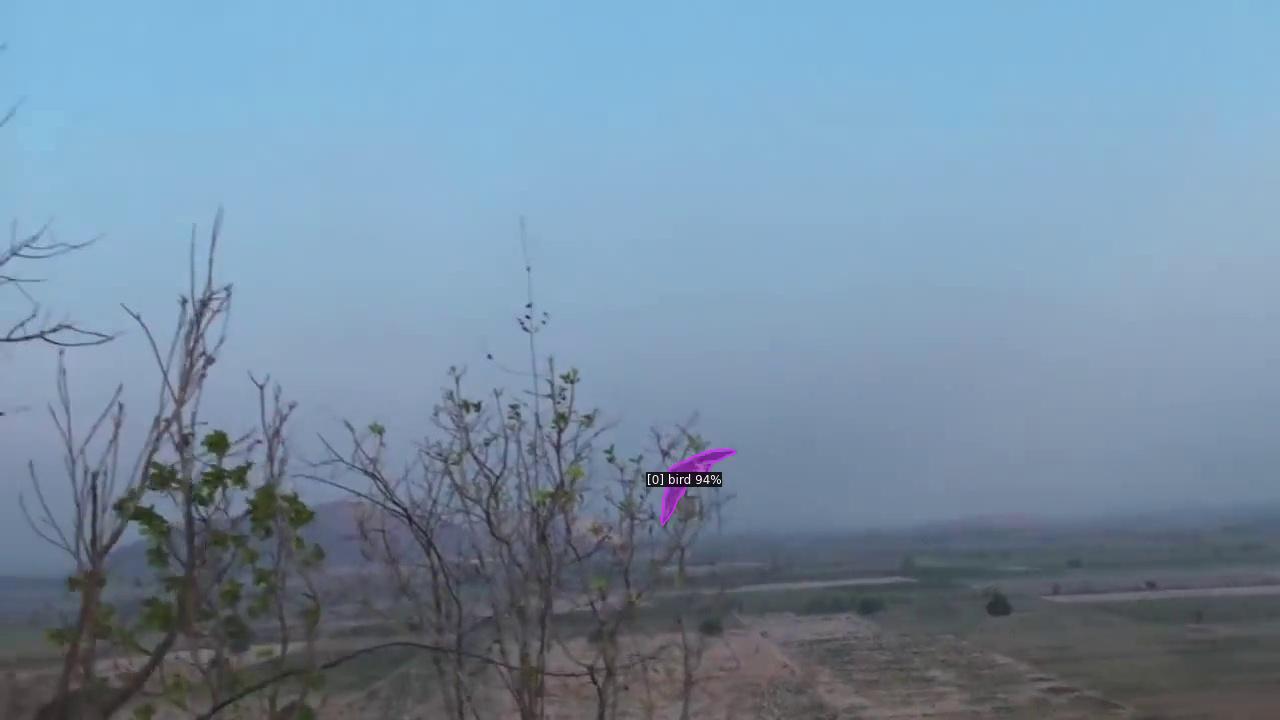}
\includegraphics[width=0.163\linewidth]{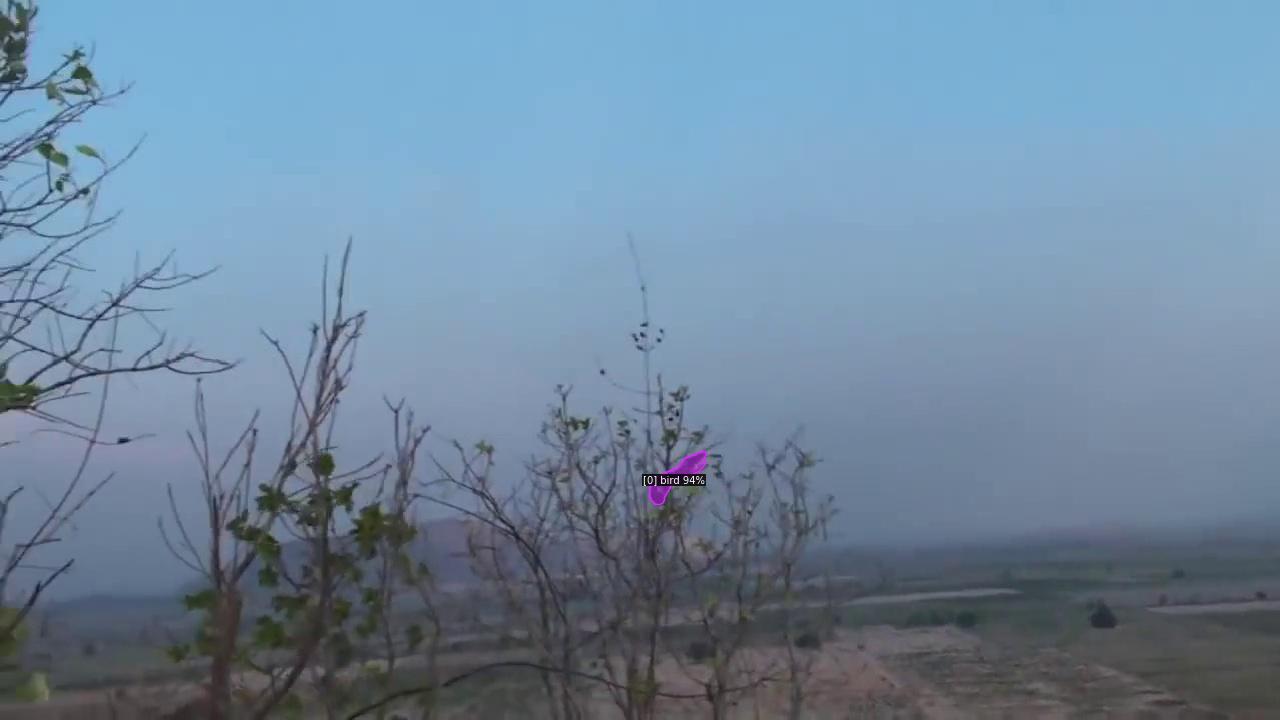}
\includegraphics[width=0.163\linewidth]{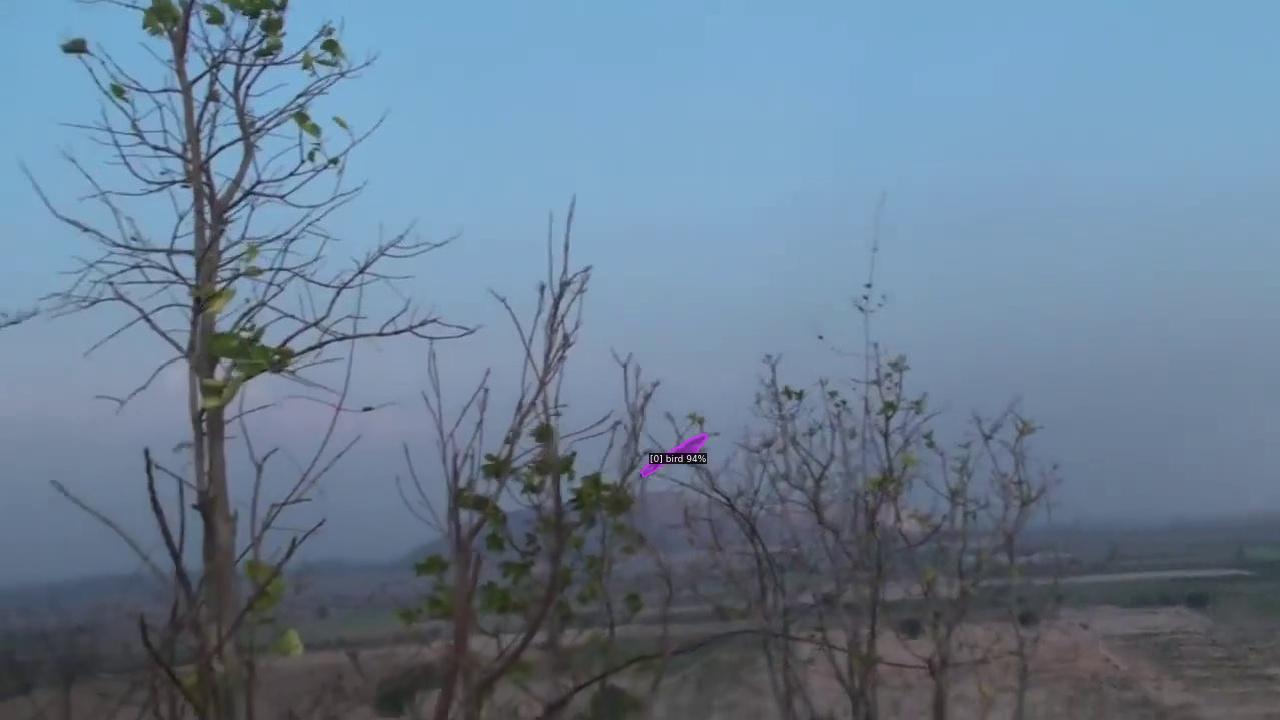}
\includegraphics[width=0.163\linewidth]{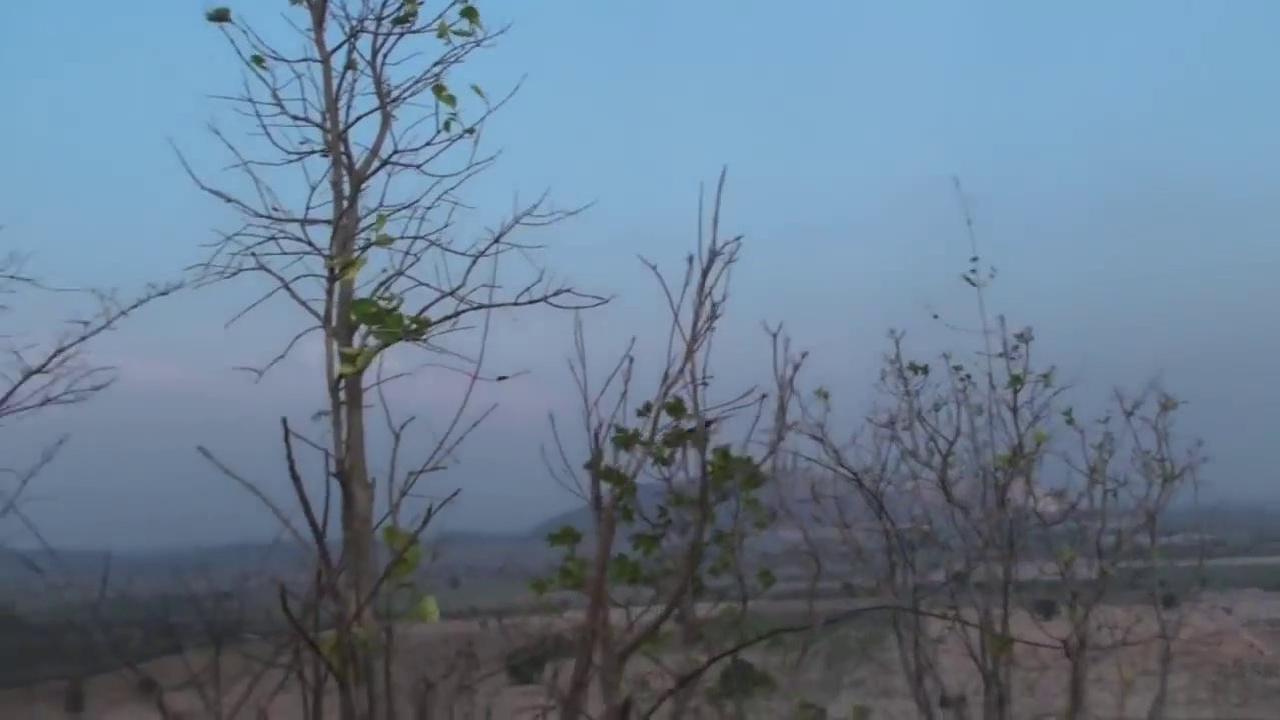}
\includegraphics[width=0.163\linewidth]{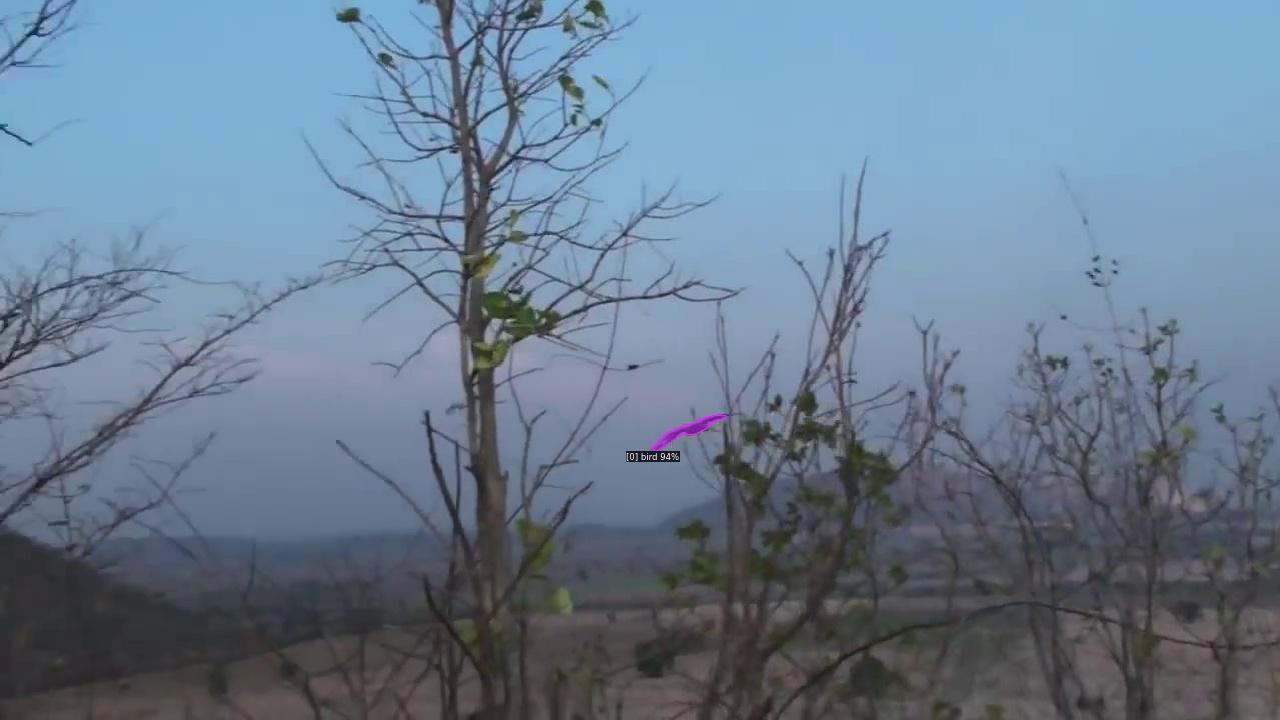}
\includegraphics[width=0.163\linewidth]{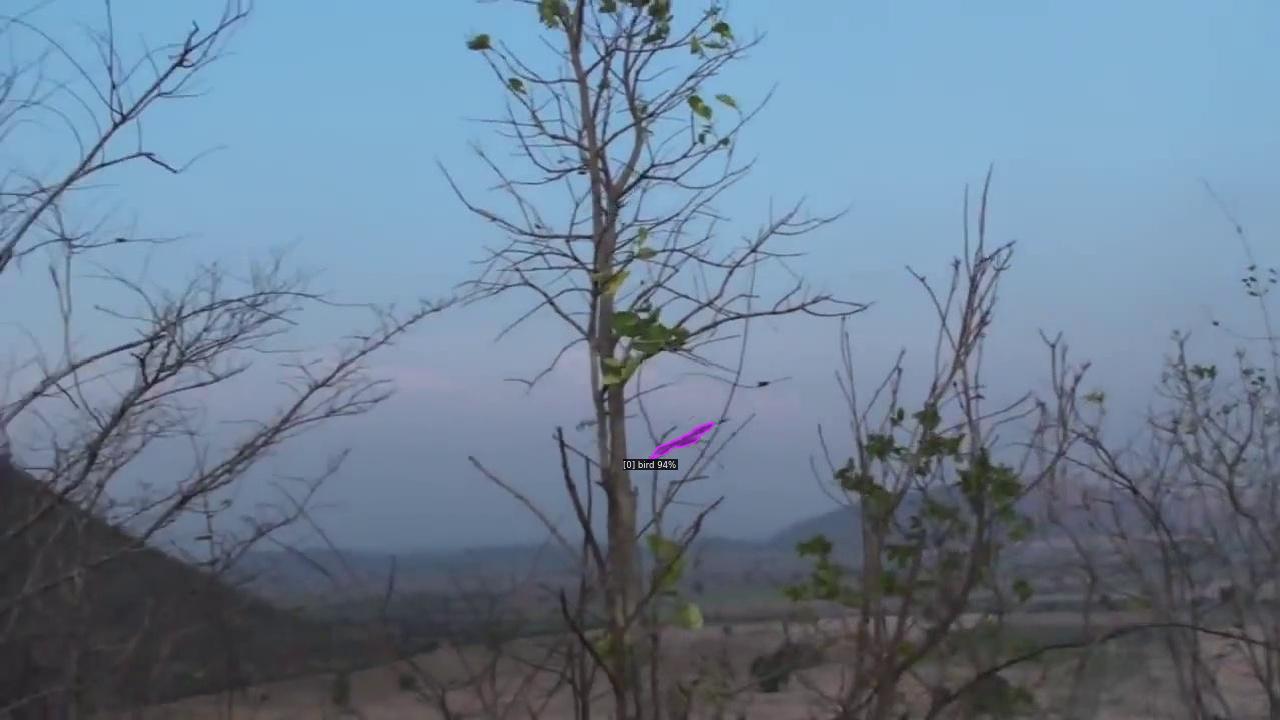}
\end{minipage}\hfill\vspace{1mm}

\begin{minipage}[c]{1.00\linewidth}
\includegraphics[width=0.163\linewidth]{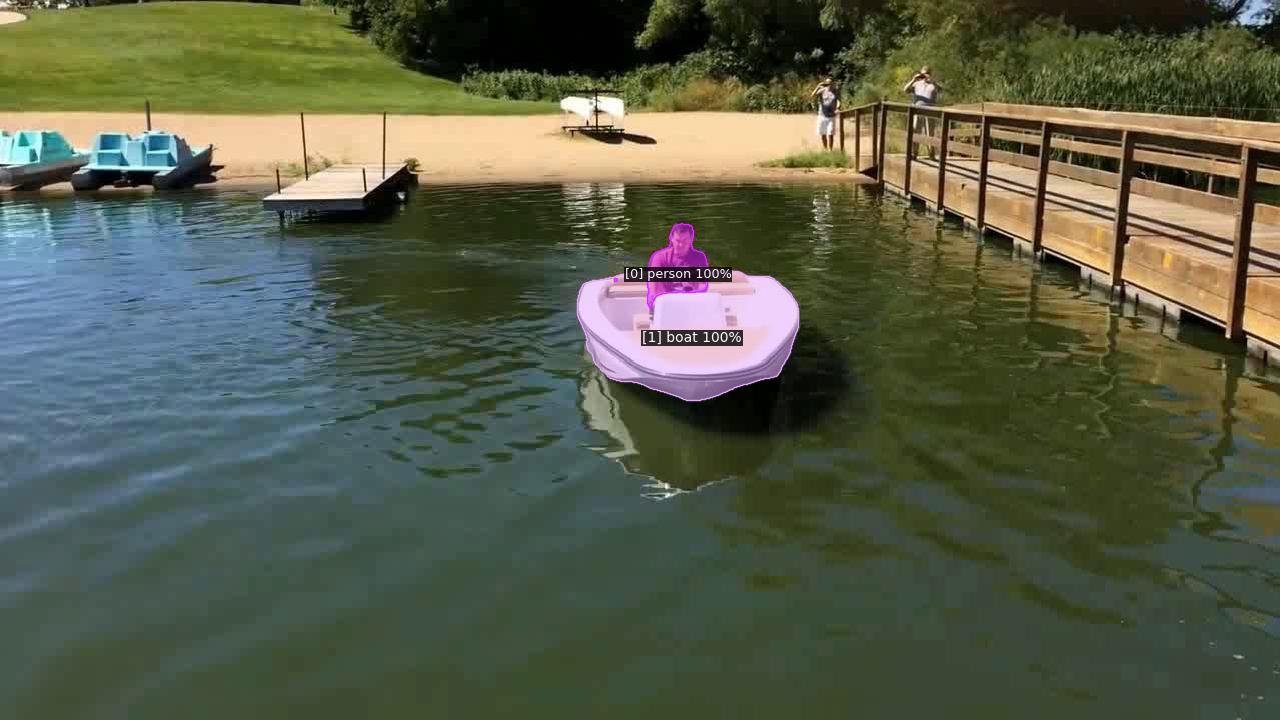}
\includegraphics[width=0.163\linewidth]{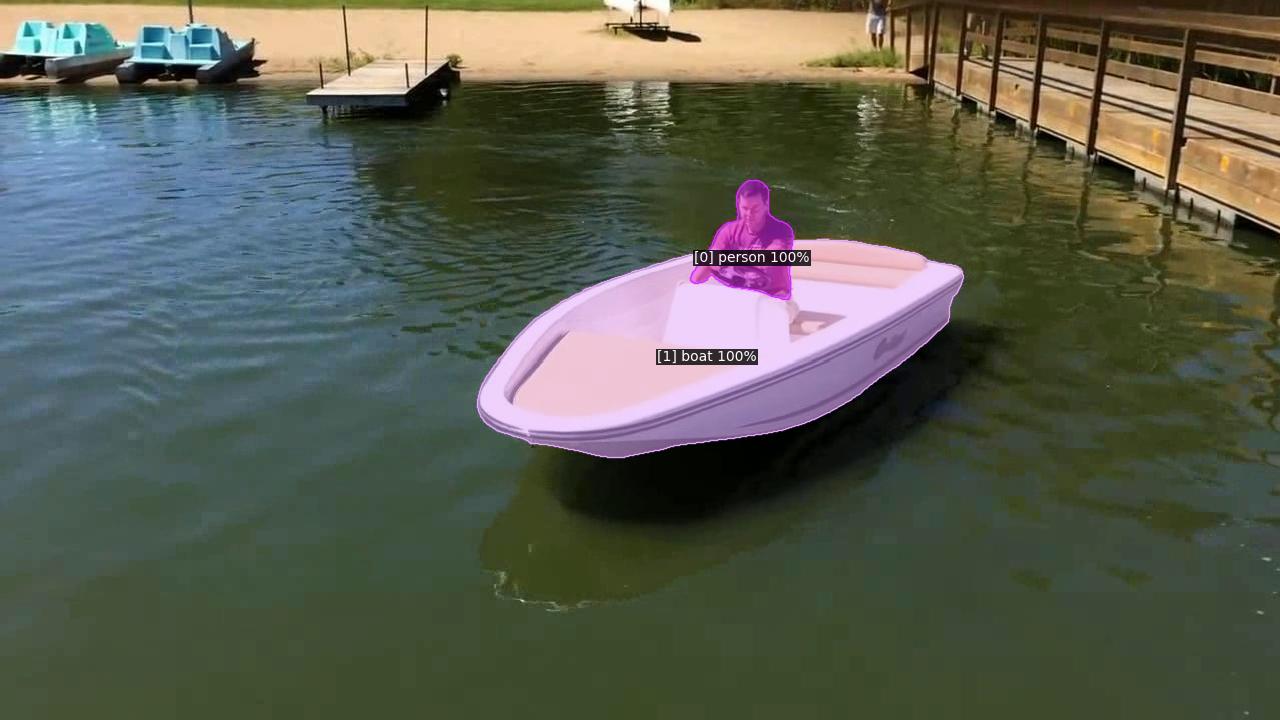}
\includegraphics[width=0.163\linewidth]{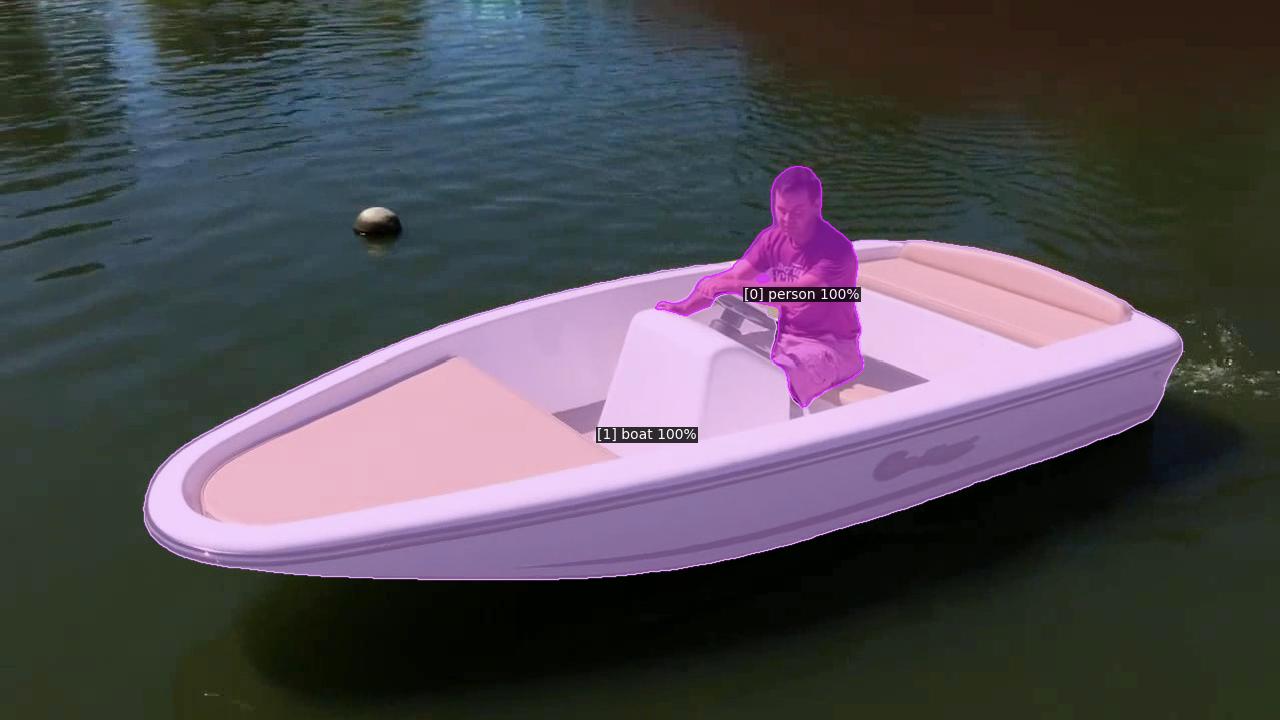}
\includegraphics[width=0.163\linewidth]{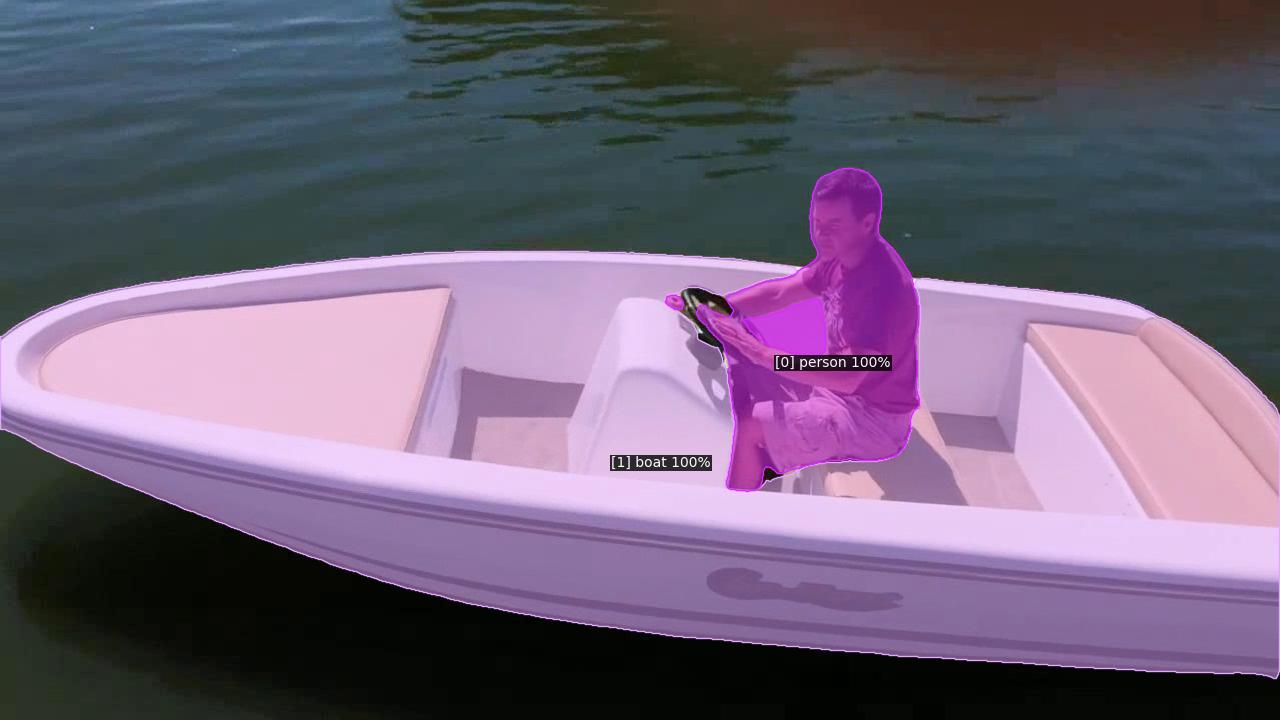}
\includegraphics[width=0.163\linewidth]{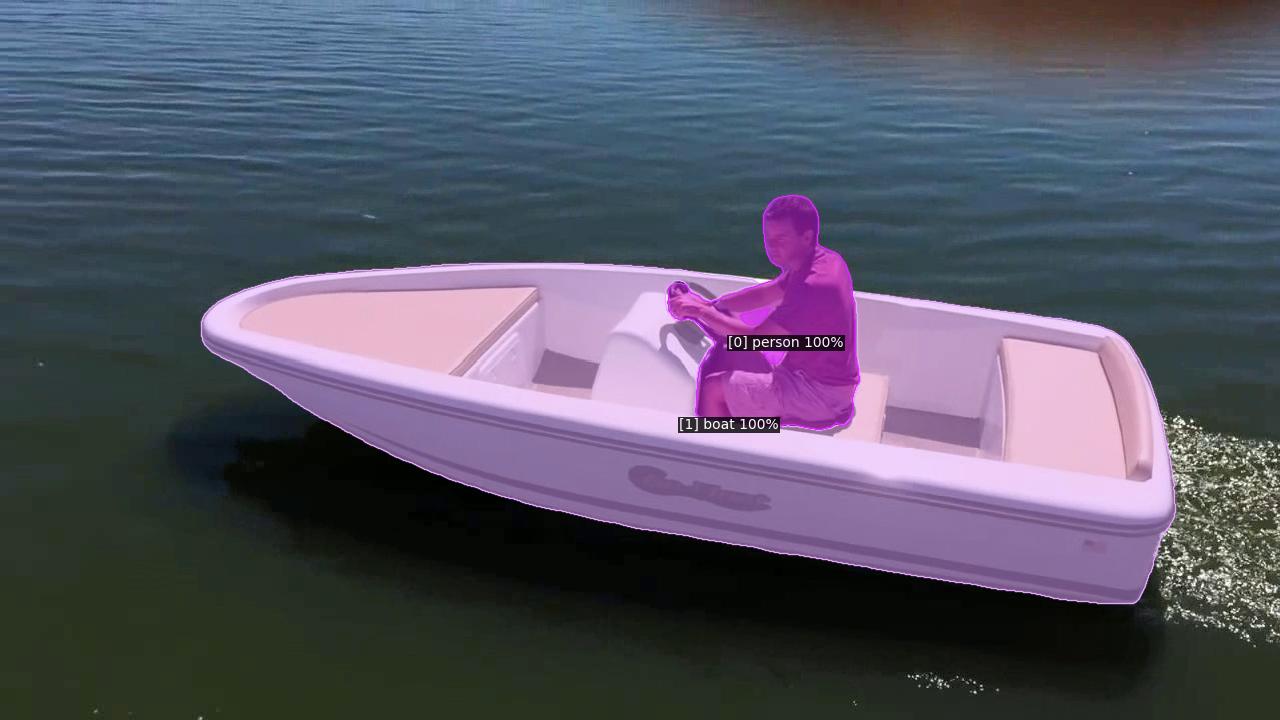}
\includegraphics[width=0.163\linewidth]{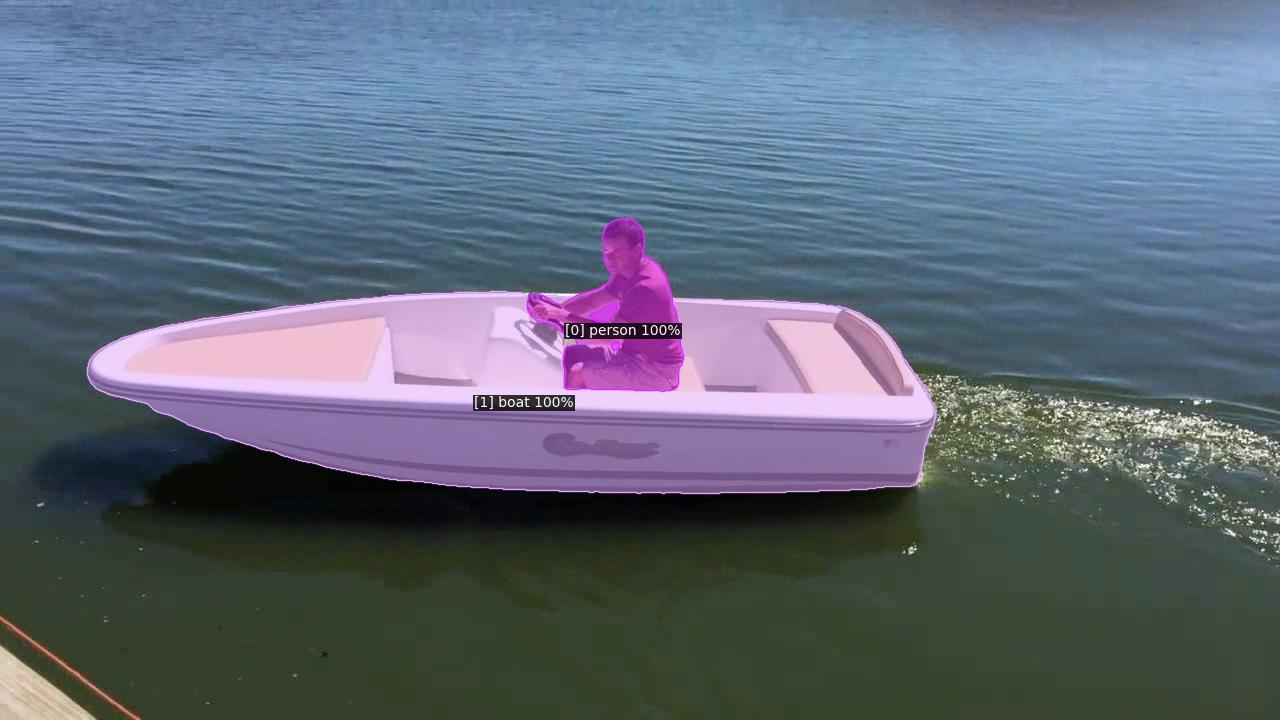}
\end{minipage}\hfill
\begin{minipage}[c]{1.0\linewidth}
\includegraphics[width=0.163\linewidth]{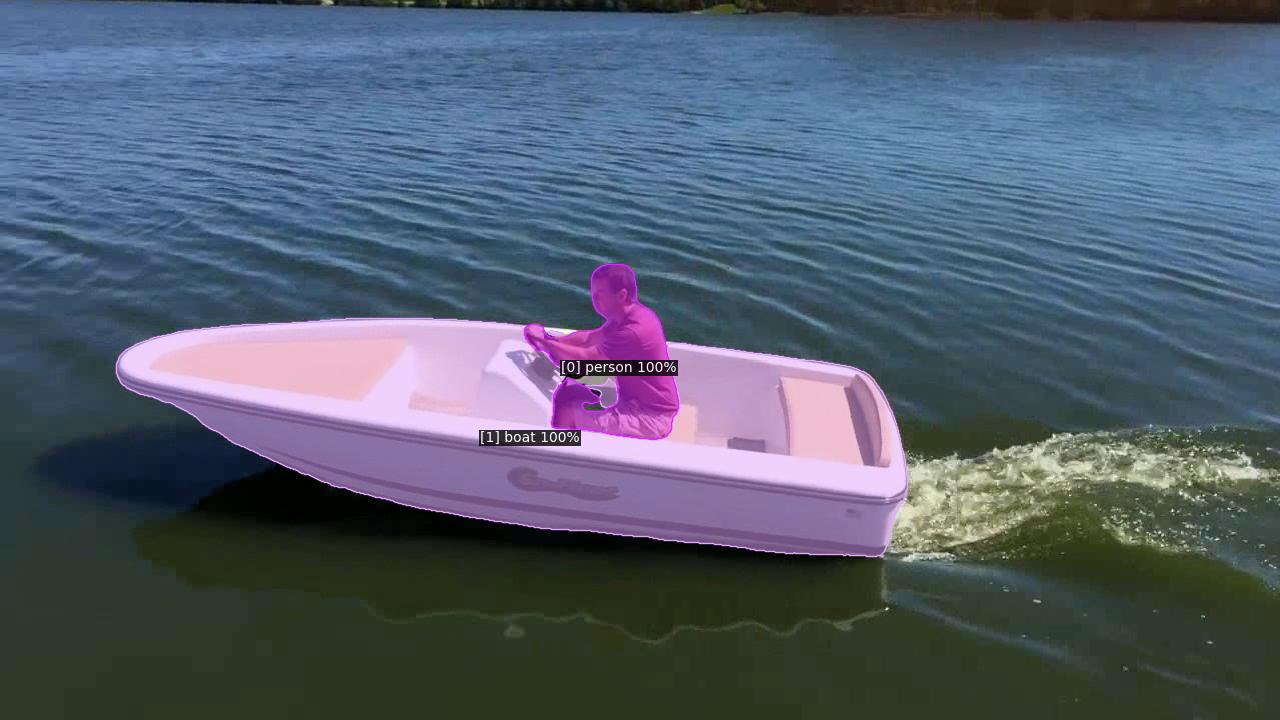}
\includegraphics[width=0.163\linewidth]{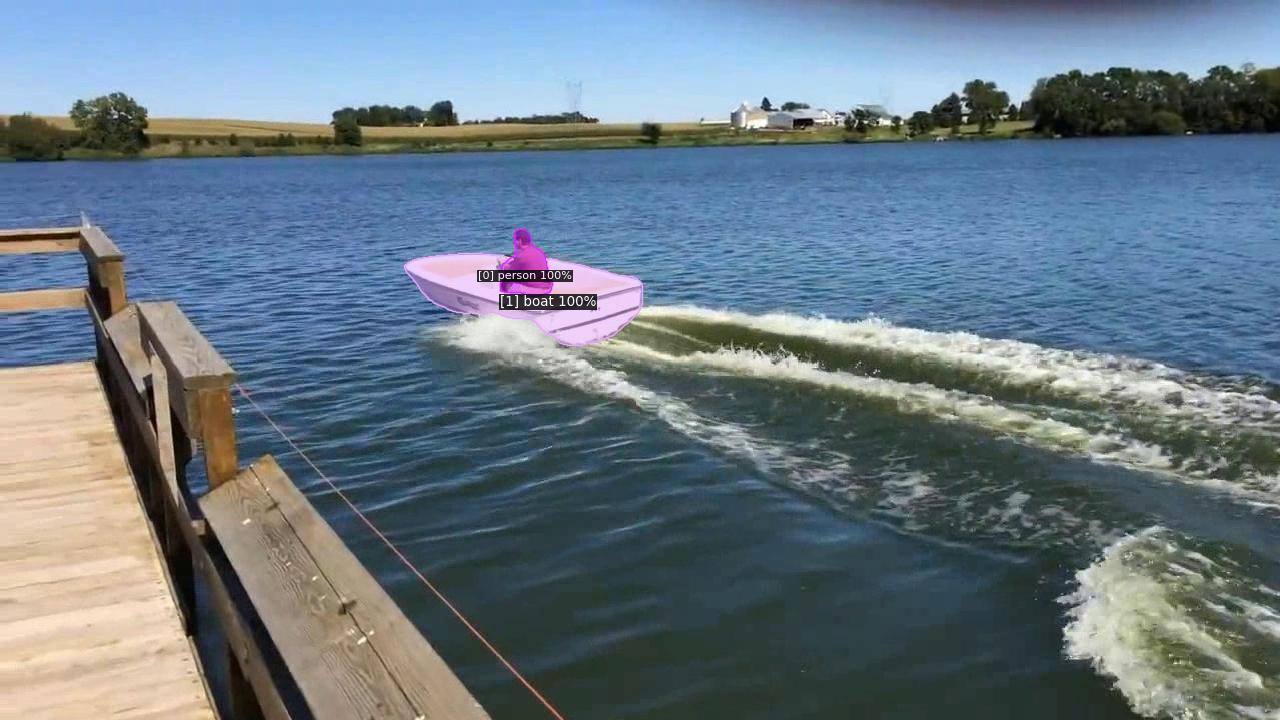}
\includegraphics[width=0.163\linewidth]{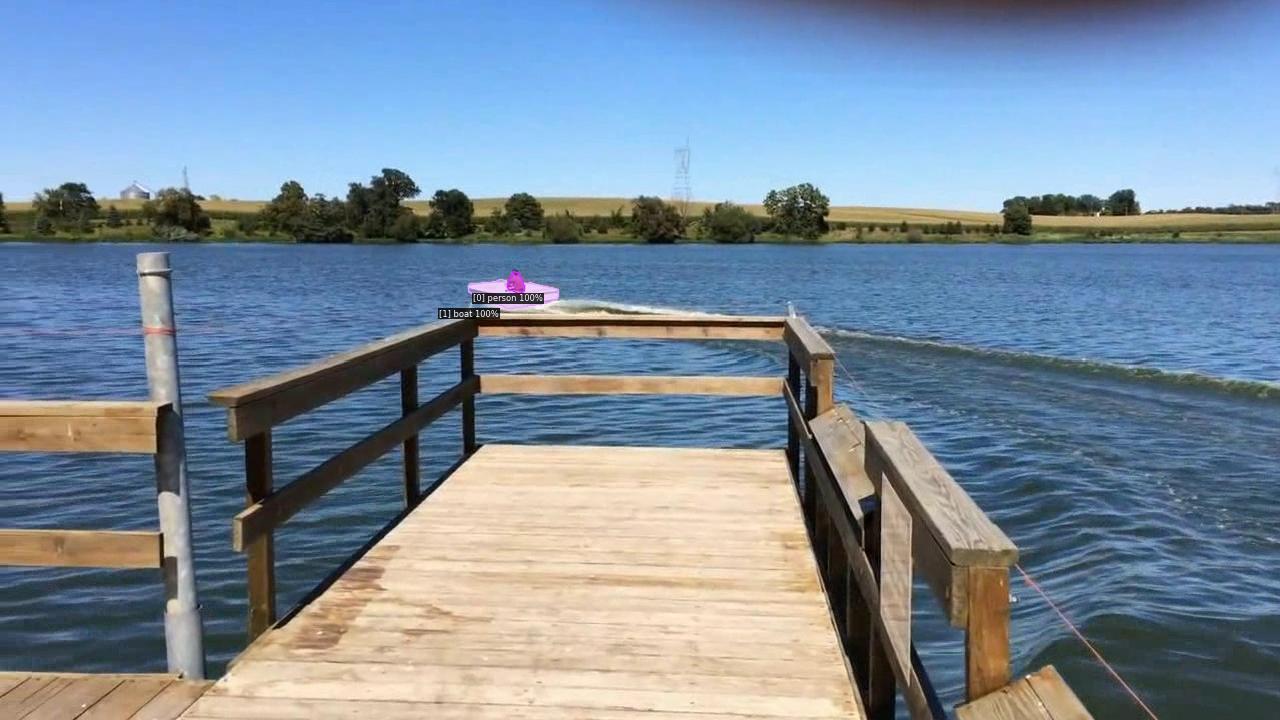}
\includegraphics[width=0.163\linewidth]{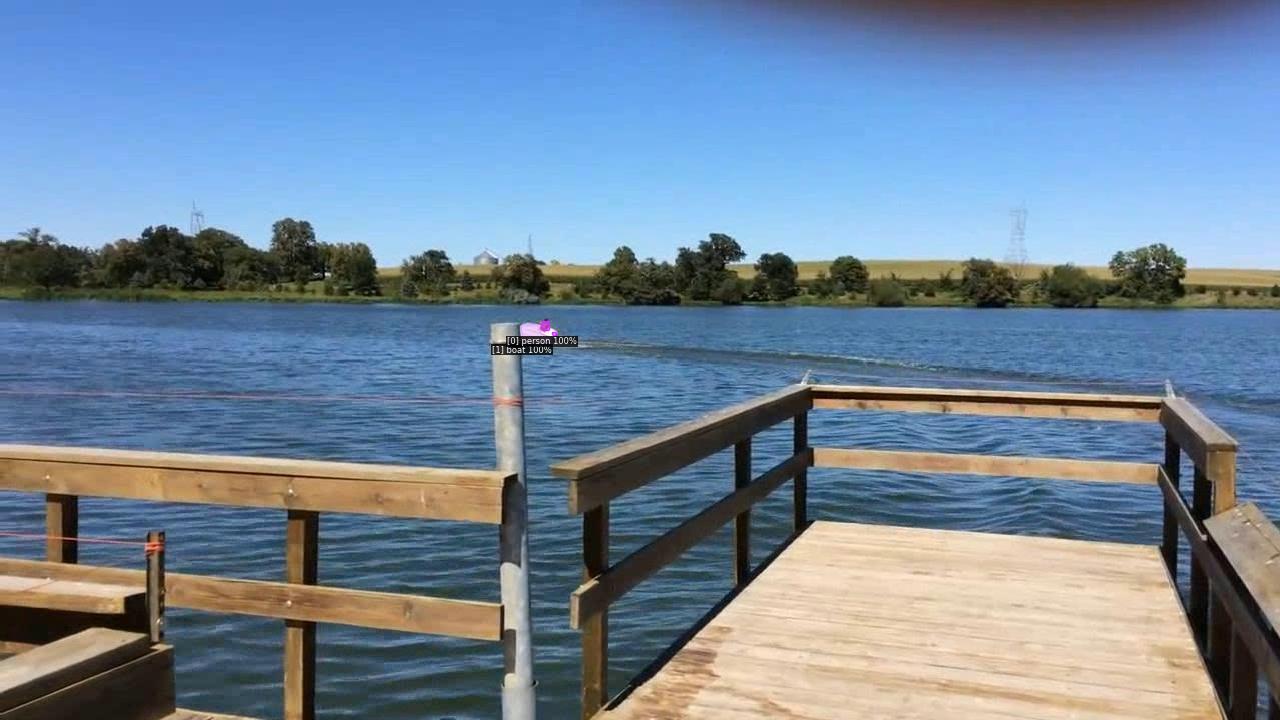}
\includegraphics[width=0.163\linewidth]{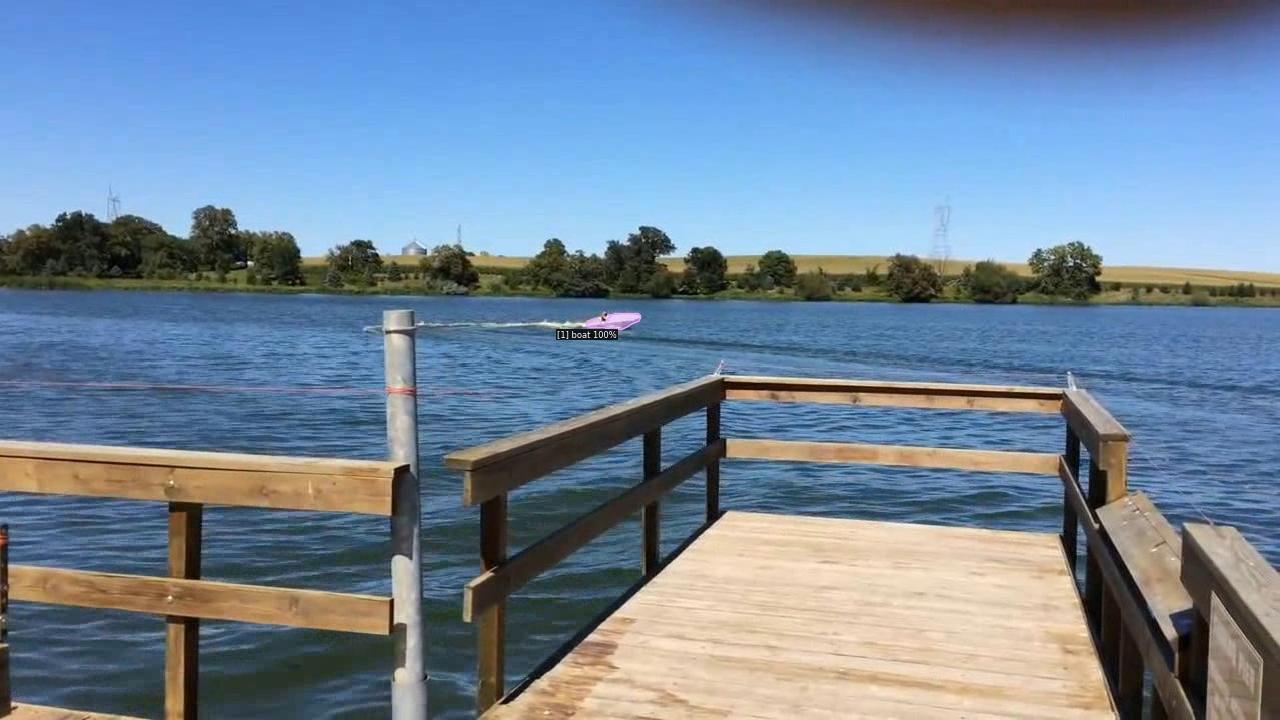}
\includegraphics[width=0.163\linewidth]{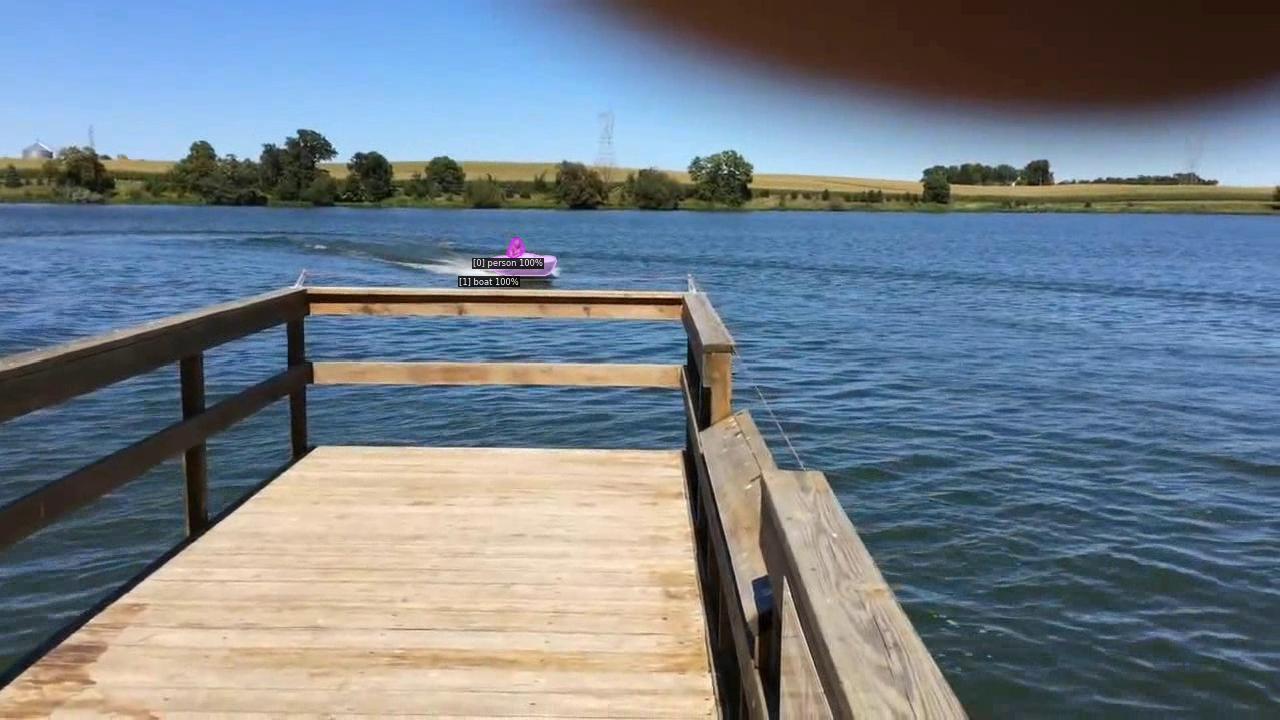}
\end{minipage}\hfill\vspace{1mm}

\caption{\textbf{Visualization results obtained on the YouTube-VIS datasets.}}
\label{fig:ytvis demo}
\end{figure*}

\begin{figure*}[t]
\begin{minipage}[c]{1.00\linewidth}
\includegraphics[width=0.163\linewidth]{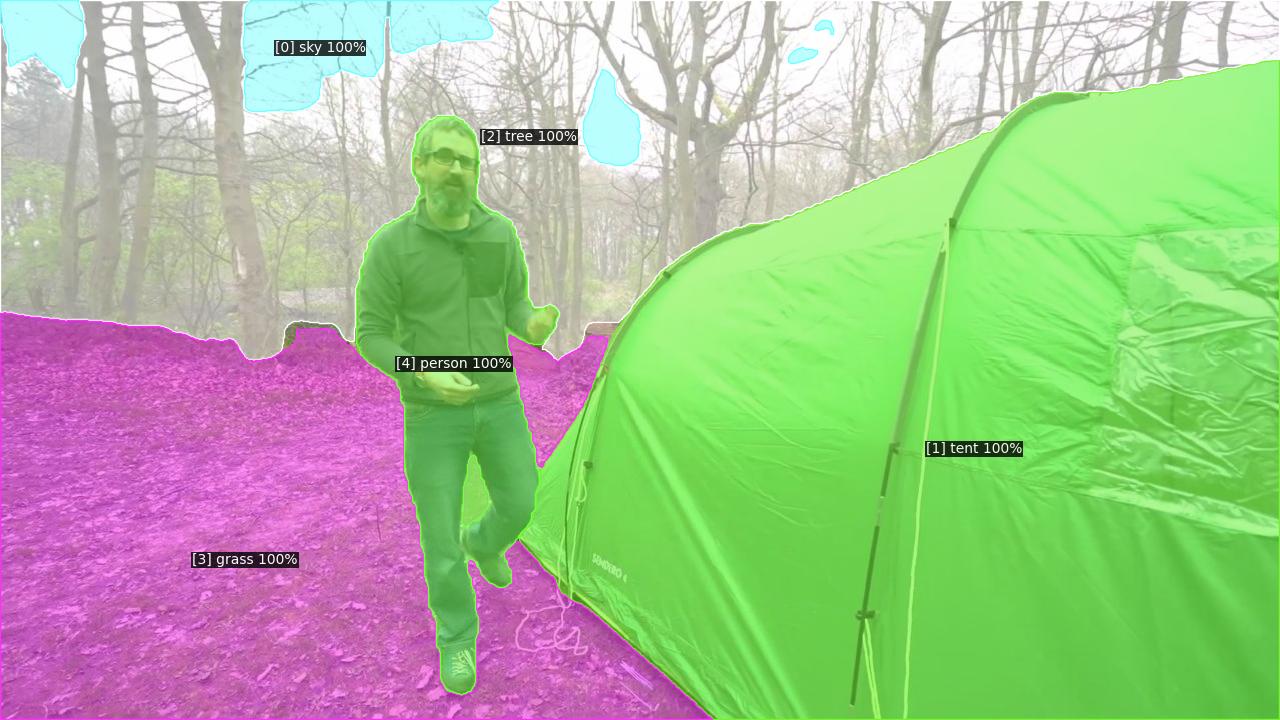}
\includegraphics[width=0.163\linewidth]{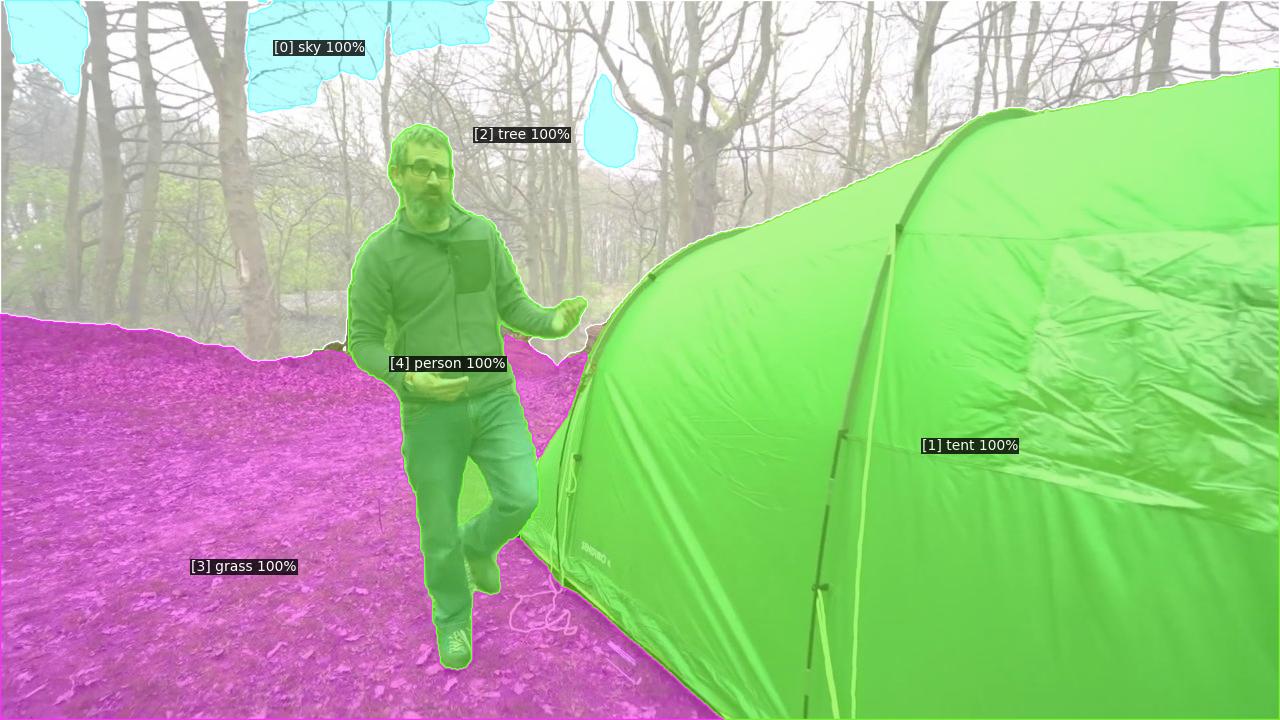}
\includegraphics[width=0.163\linewidth]{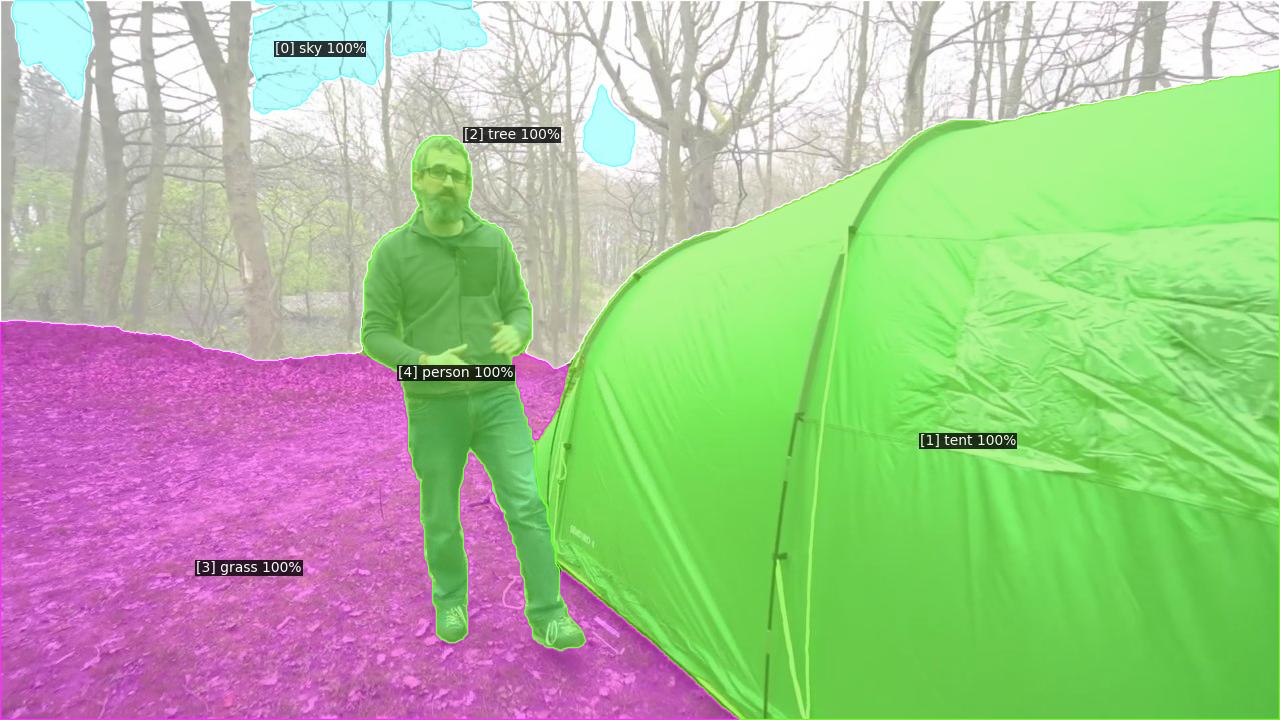}
\includegraphics[width=0.163\linewidth]{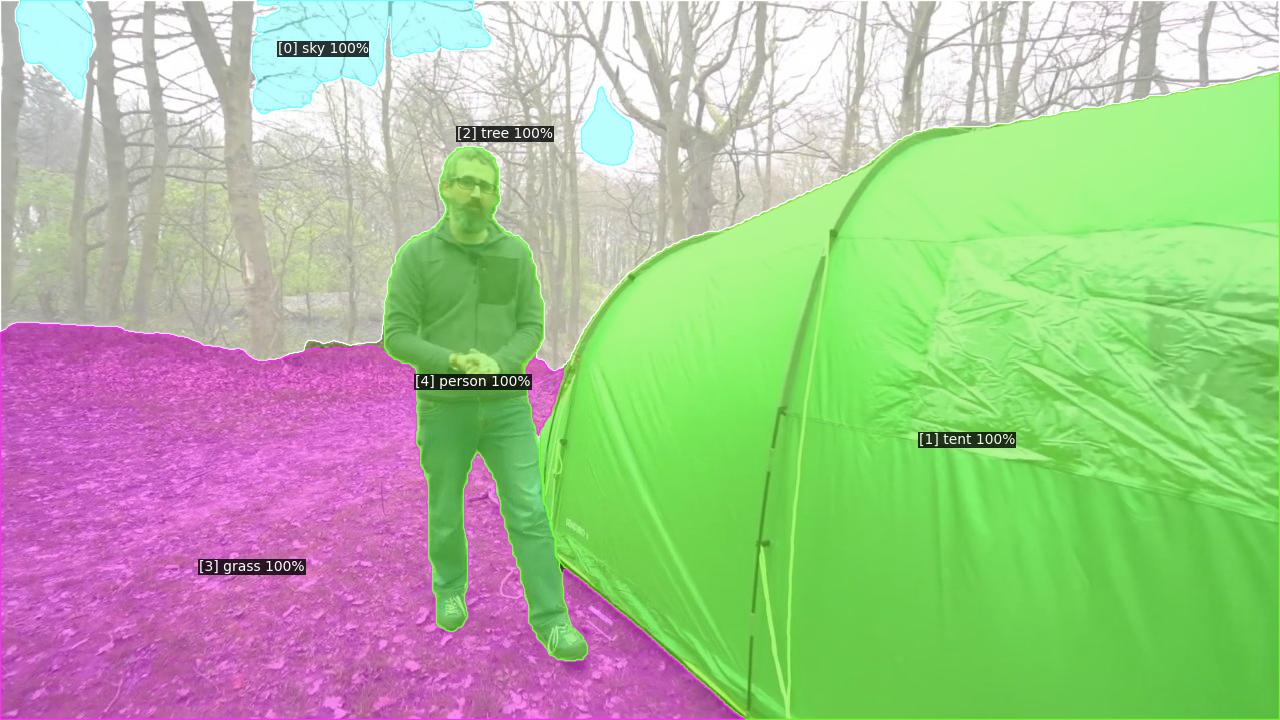}
\includegraphics[width=0.163\linewidth]{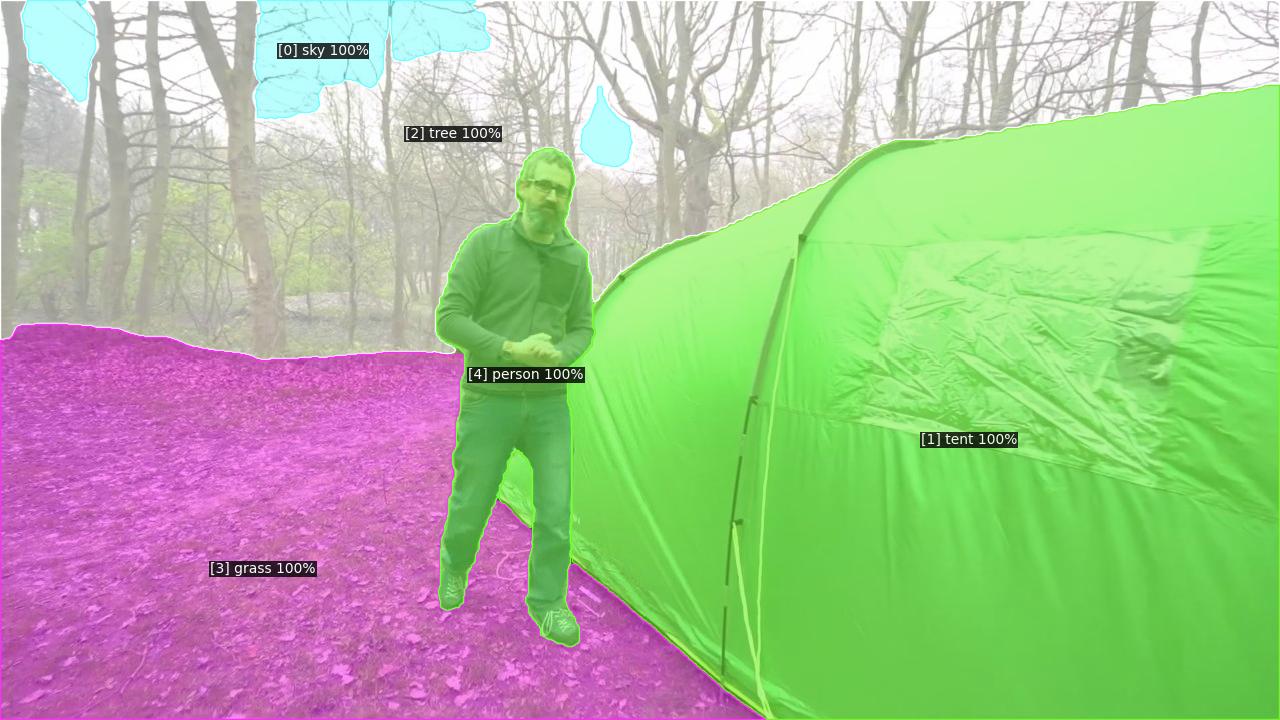}
\includegraphics[width=0.163\linewidth]{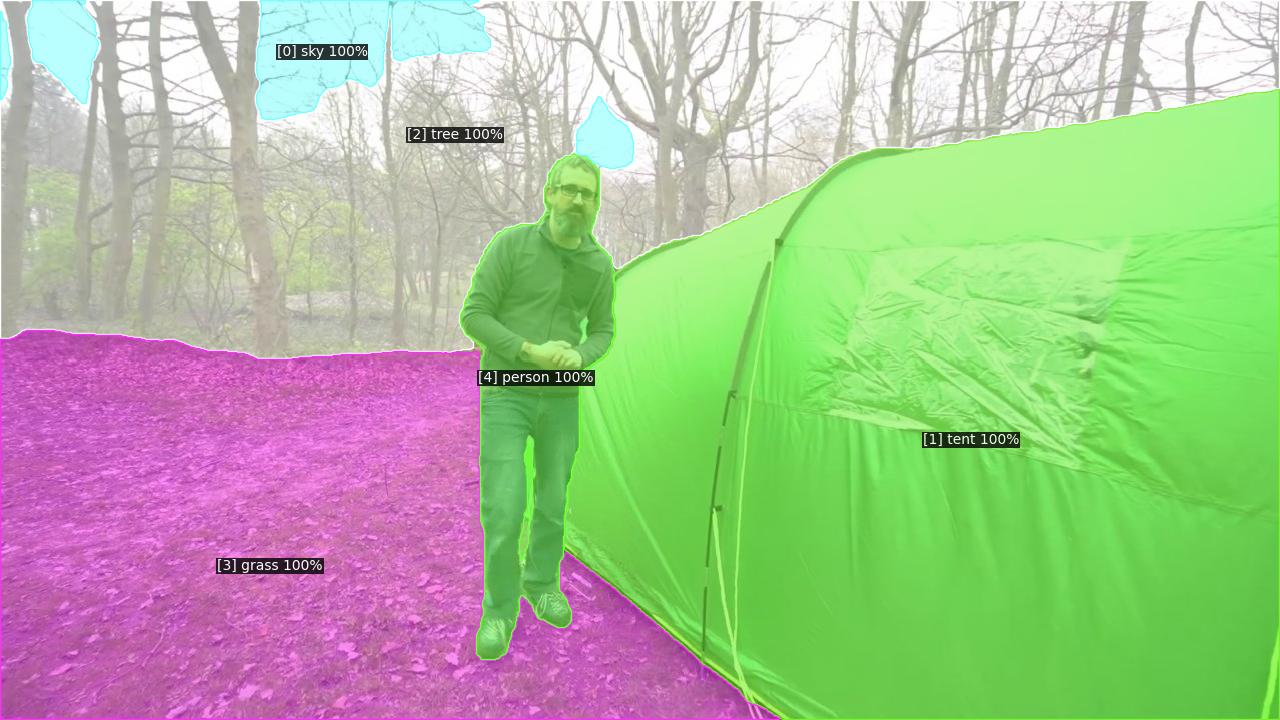}
\end{minipage}\hfill
\begin{minipage}[c]{1.0\linewidth}
\includegraphics[width=0.163\linewidth]{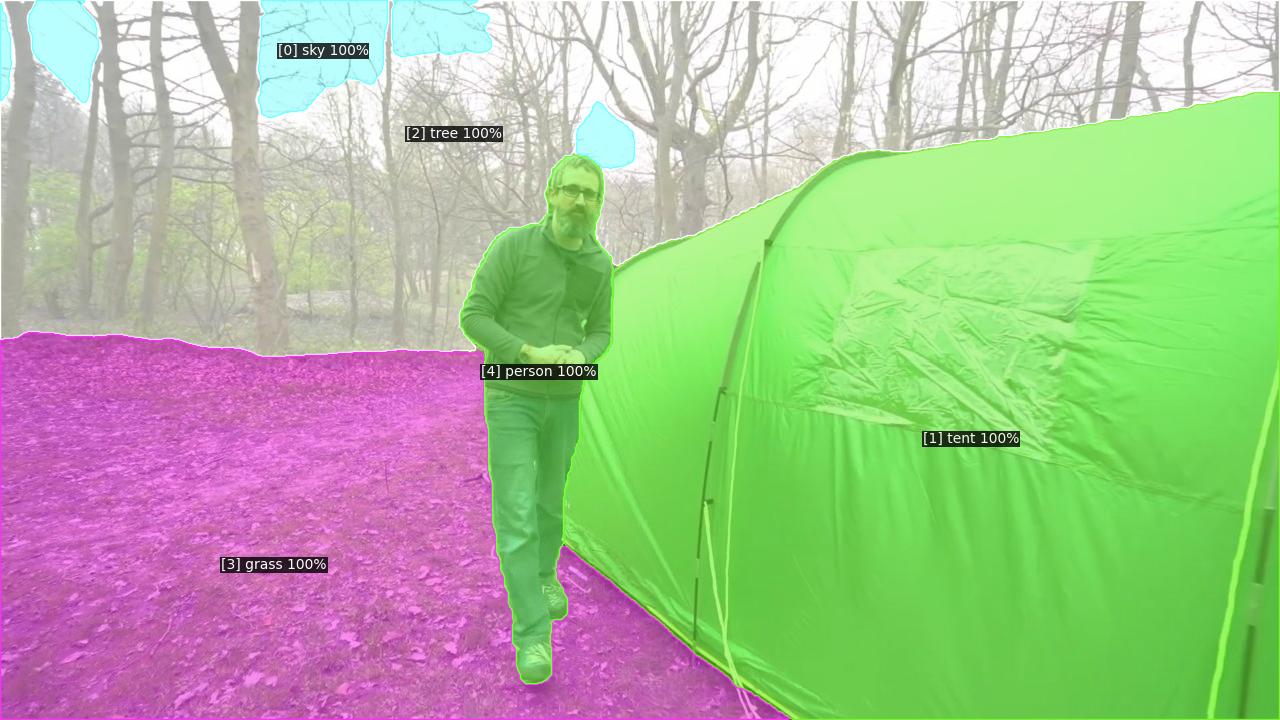}
\includegraphics[width=0.163\linewidth]{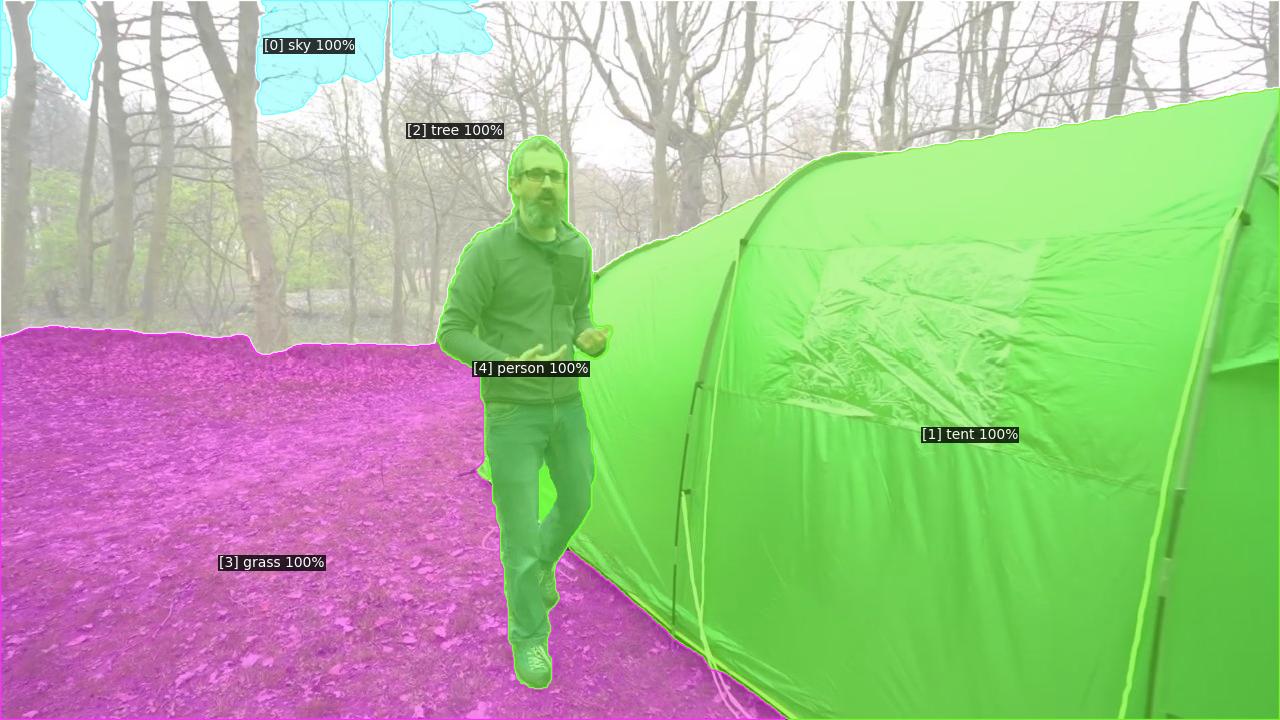}
\includegraphics[width=0.163\linewidth]{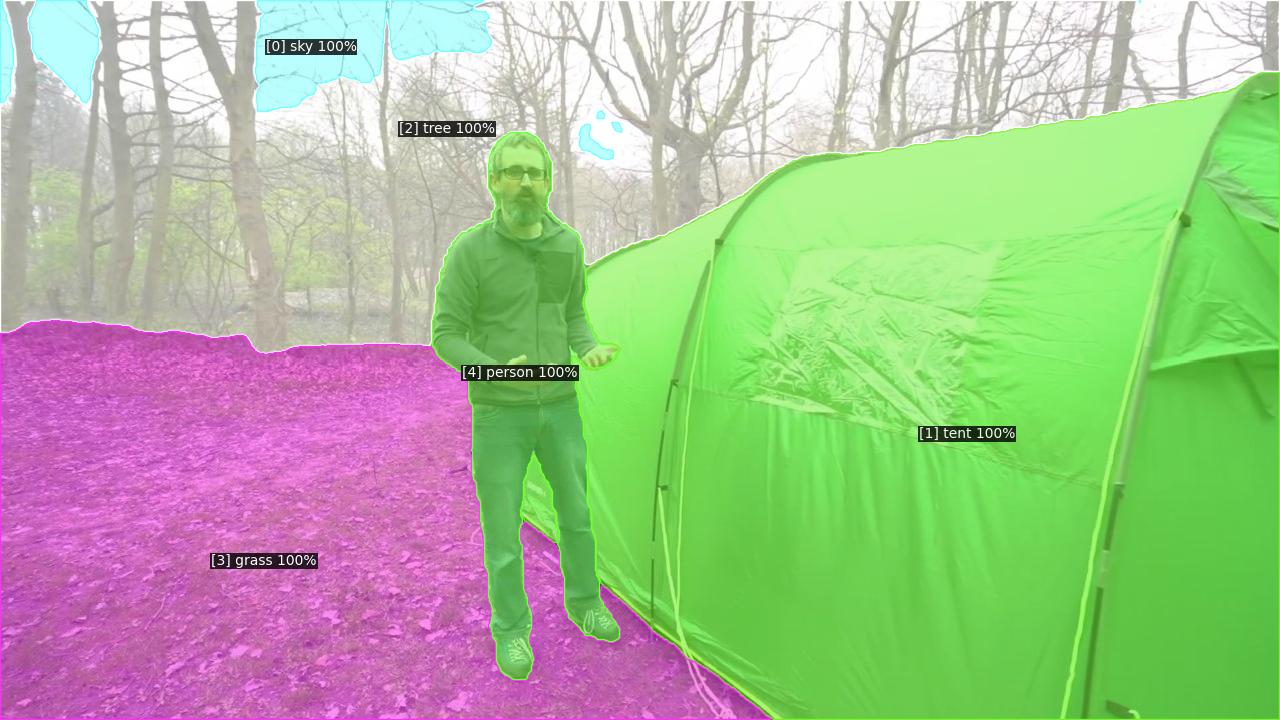}
\includegraphics[width=0.163\linewidth]{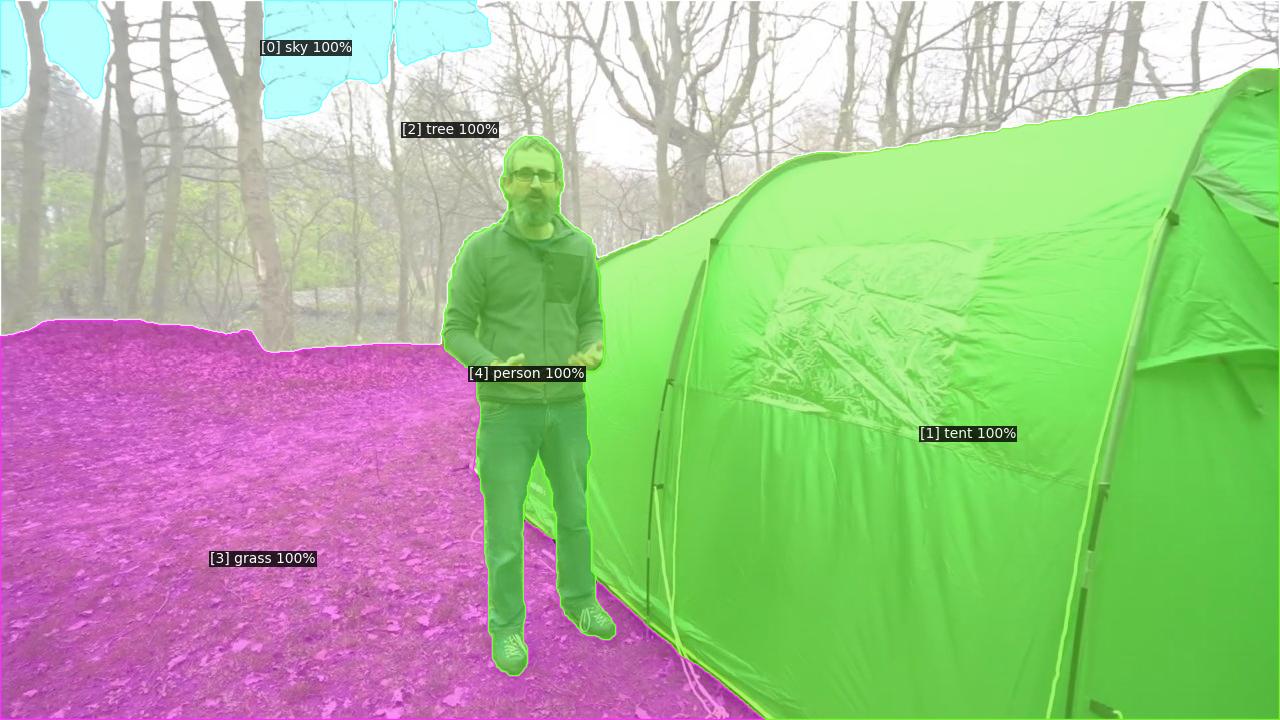}
\includegraphics[width=0.163\linewidth]{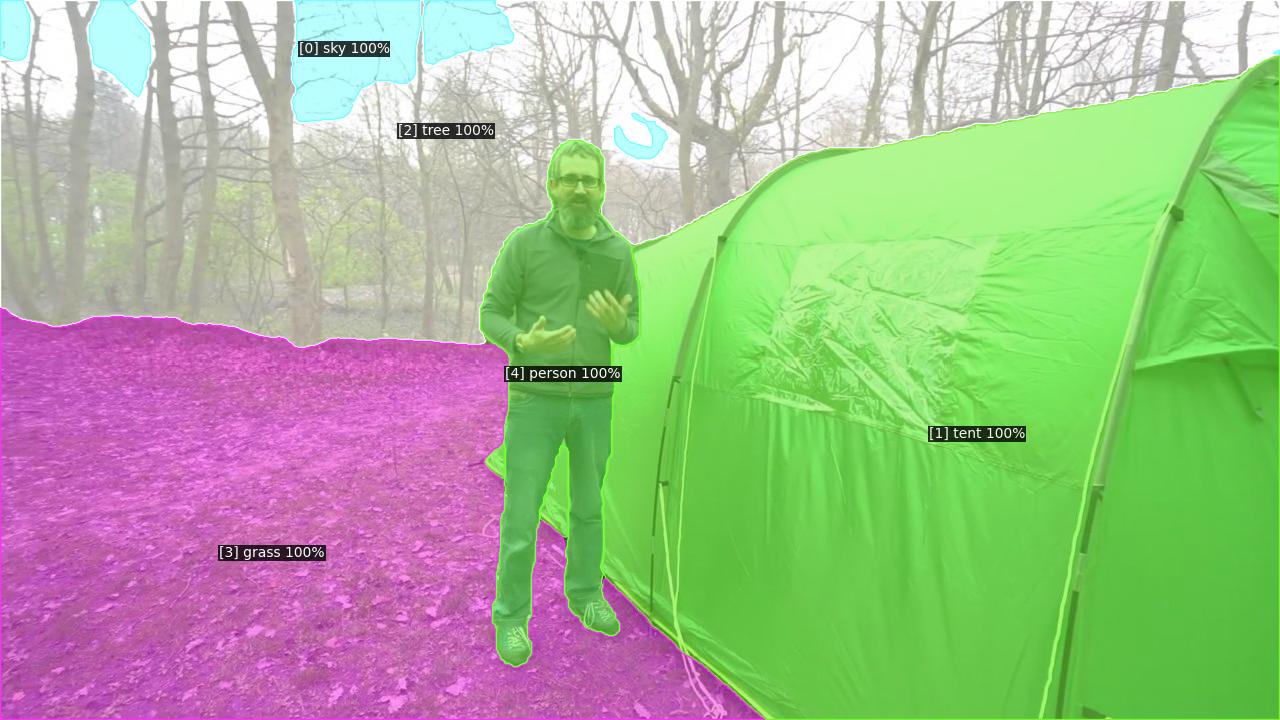}
\includegraphics[width=0.163\linewidth]{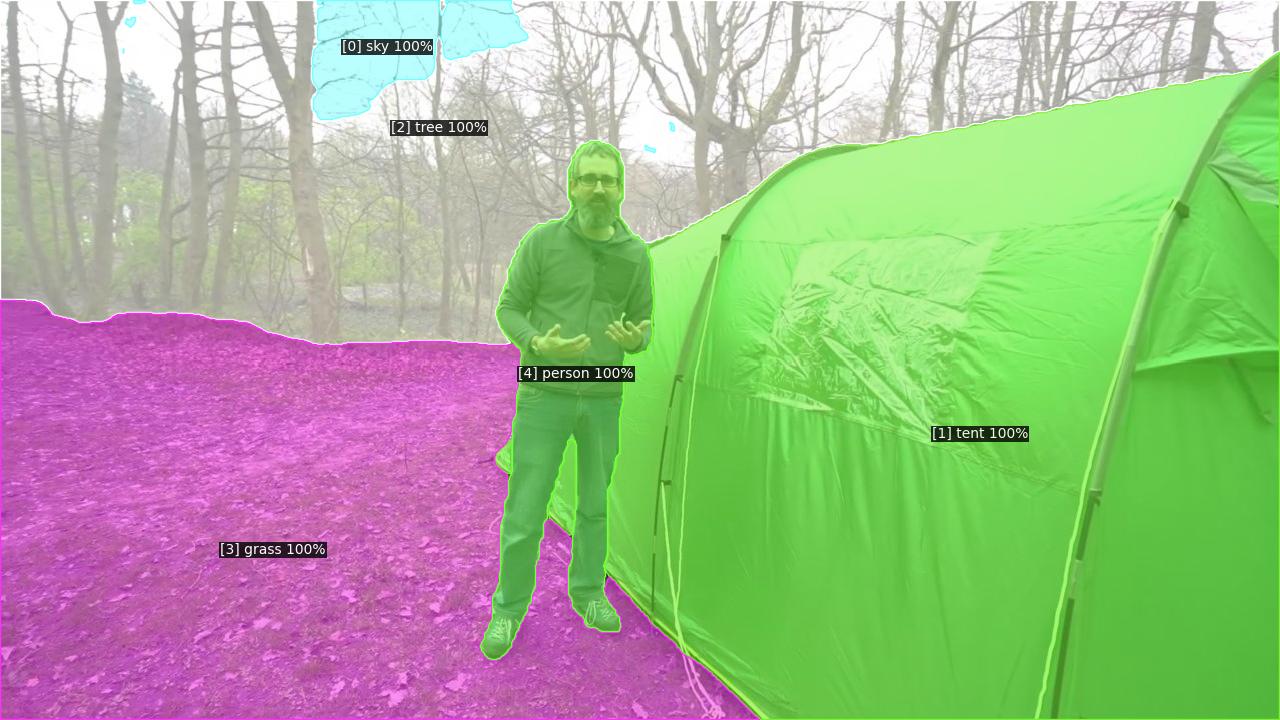}
\end{minipage}\hfill\vspace{1mm}

\begin{minipage}[c]{1.00\linewidth}
\includegraphics[width=0.163\linewidth]{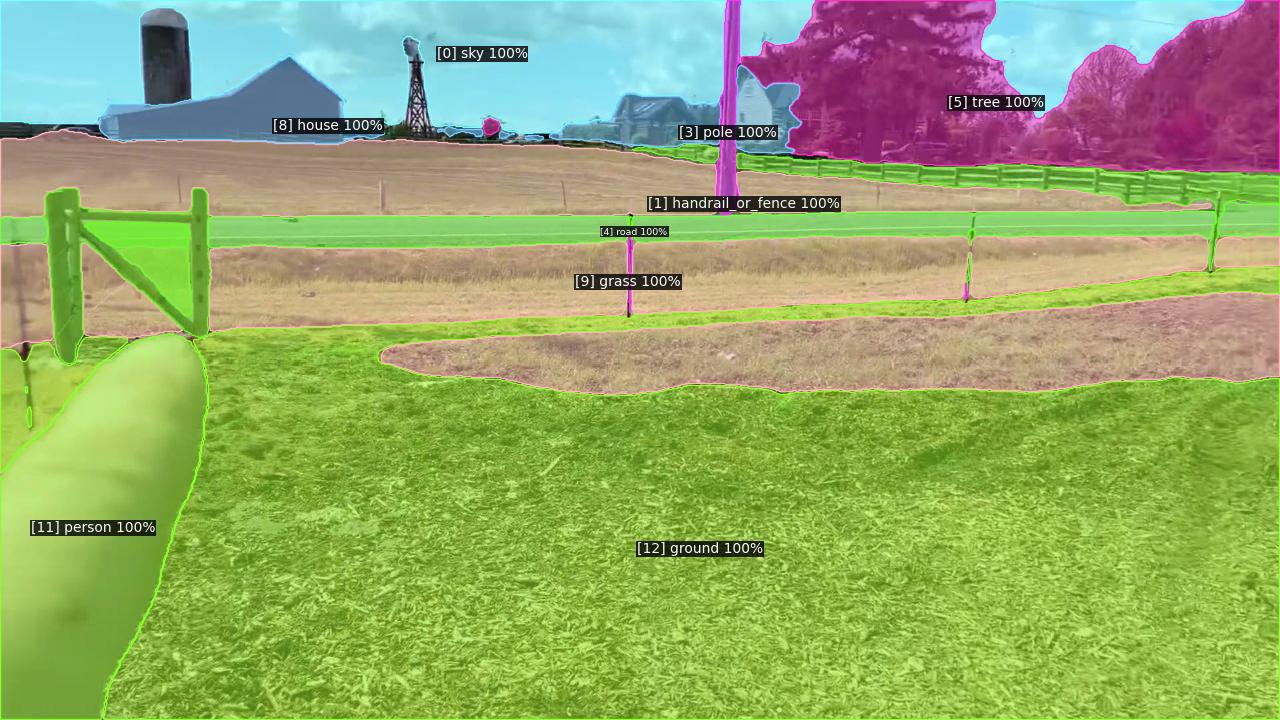}
\includegraphics[width=0.163\linewidth]{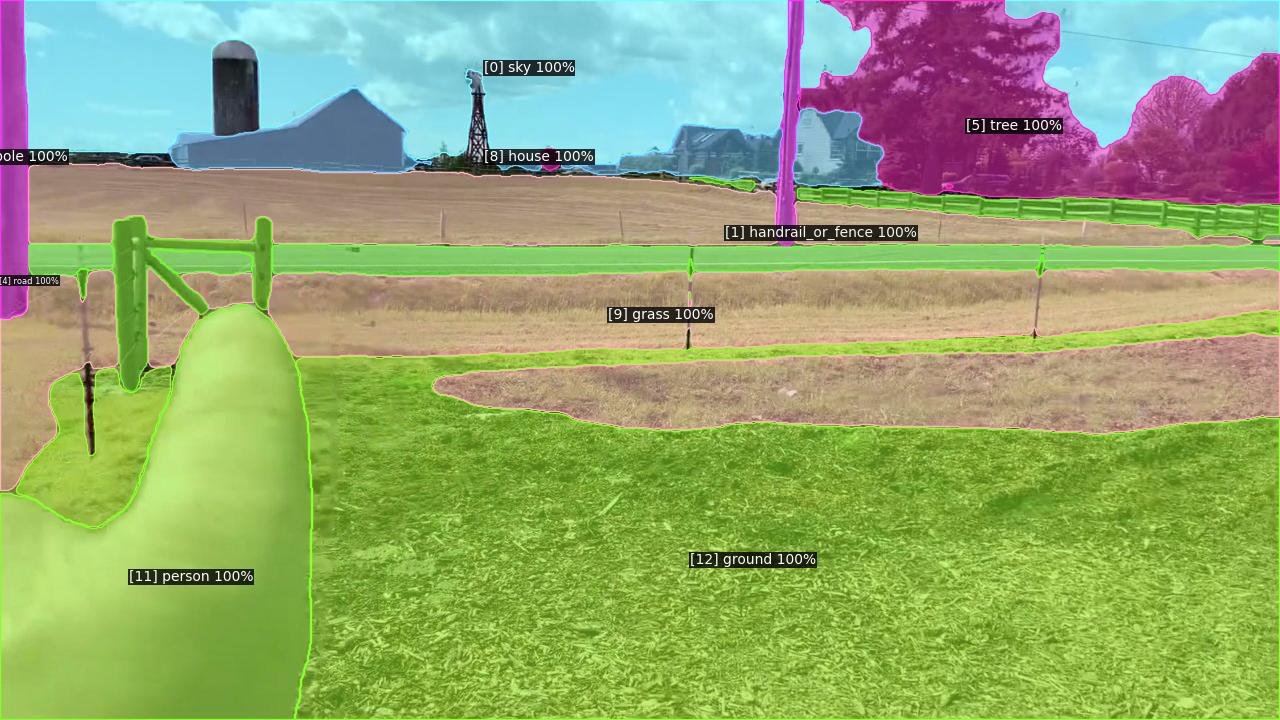}
\includegraphics[width=0.163\linewidth]{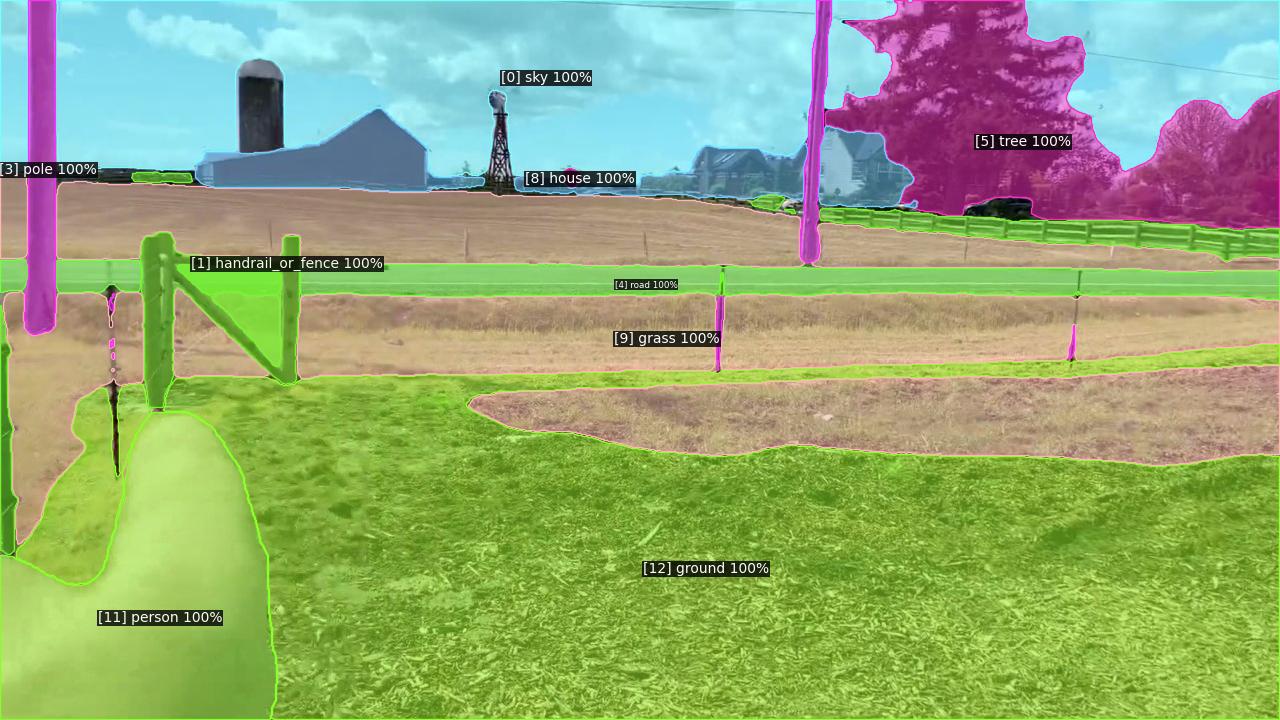}
\includegraphics[width=0.163\linewidth]{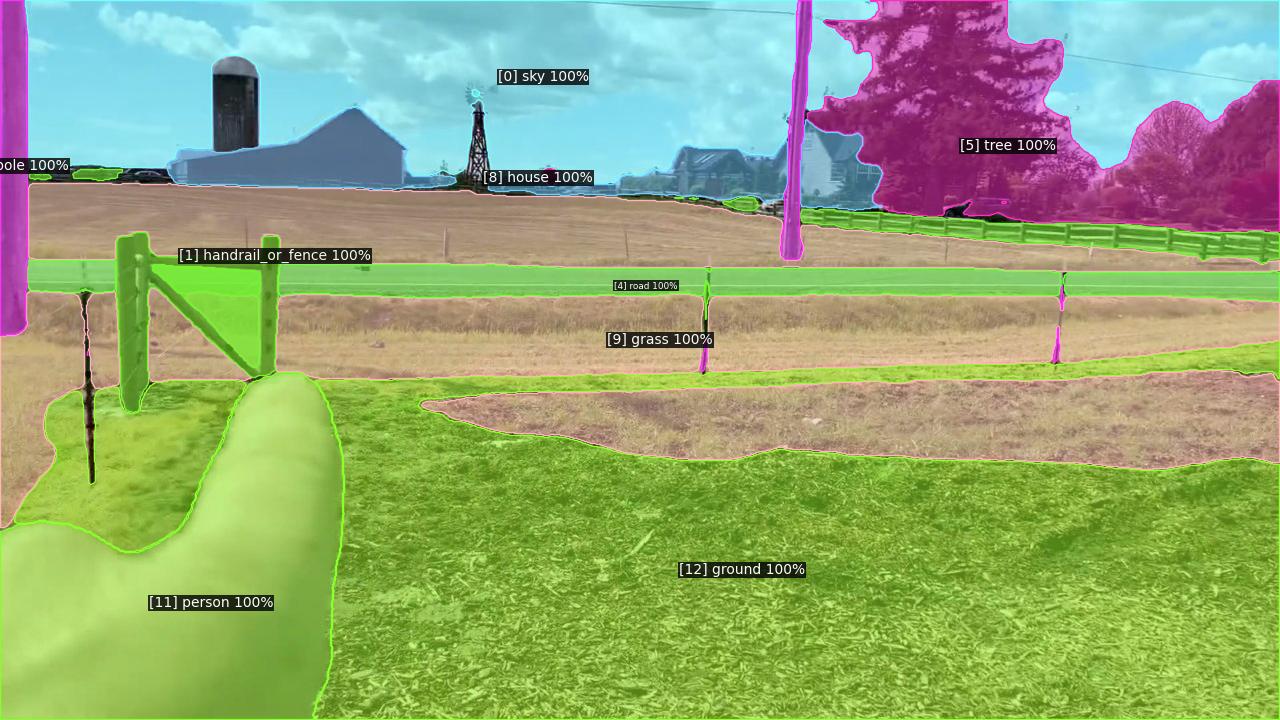}
\includegraphics[width=0.163\linewidth]{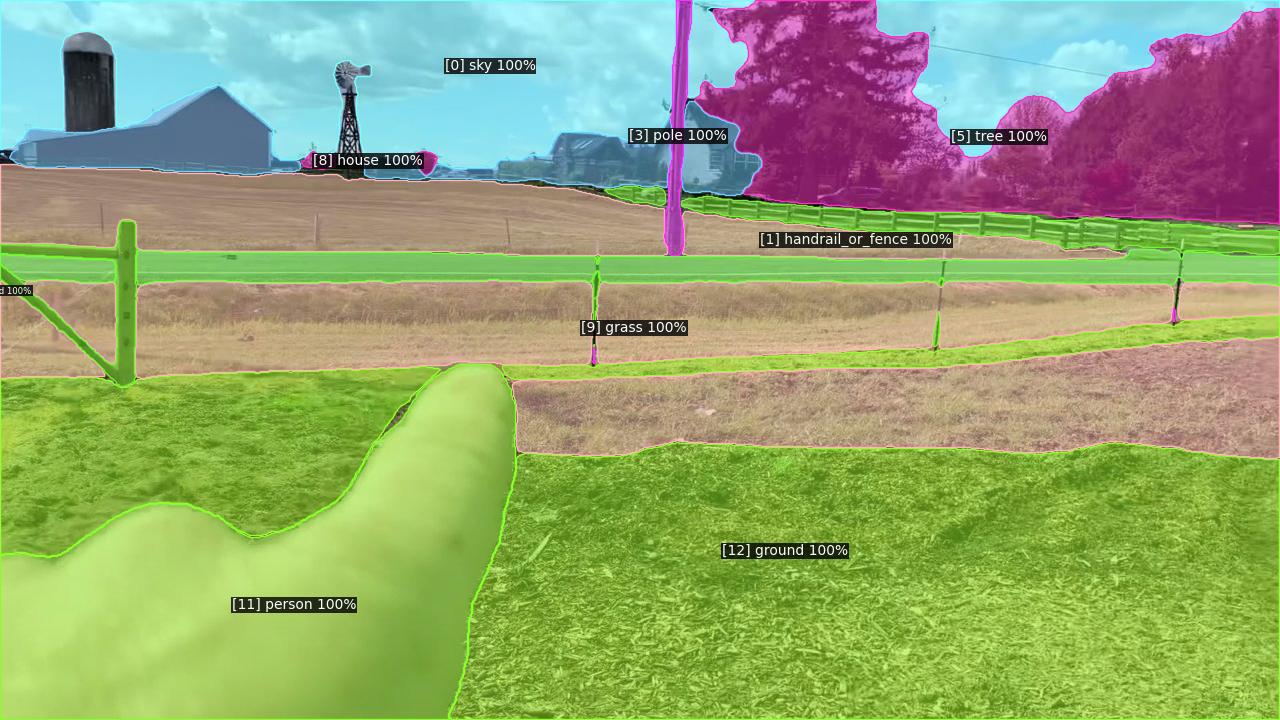}
\includegraphics[width=0.163\linewidth]{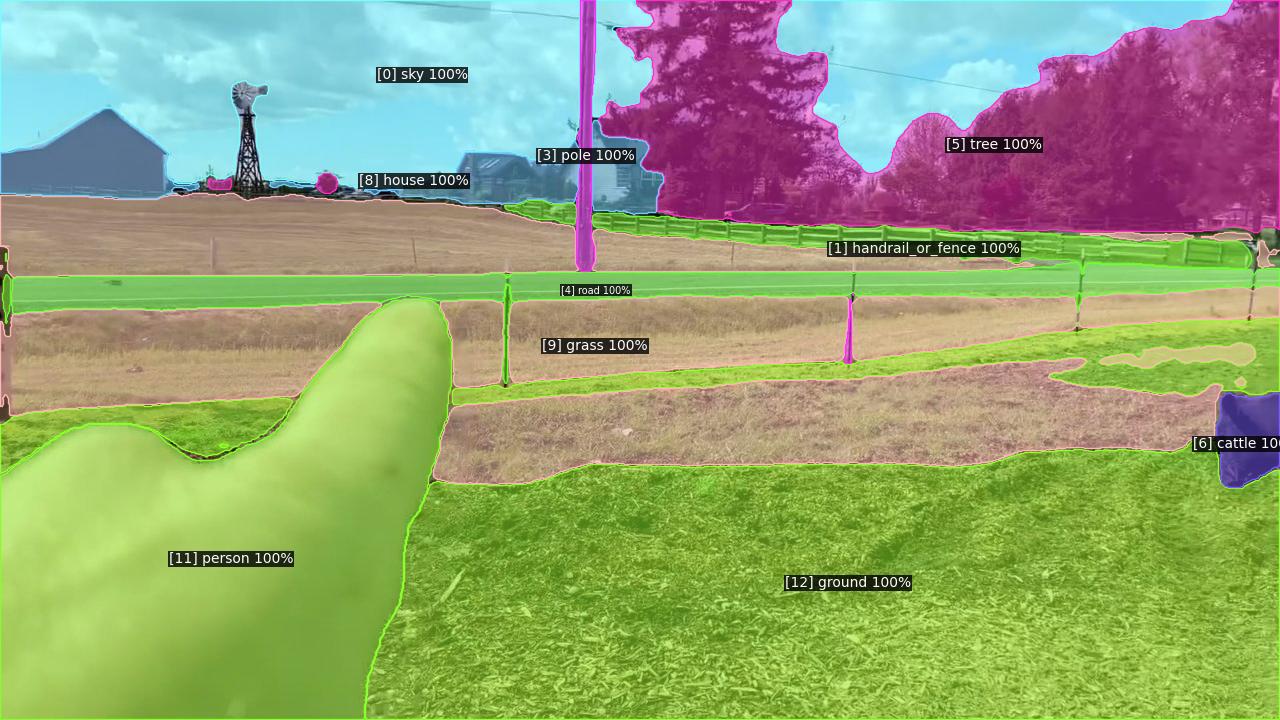}
\end{minipage}\hfill
\begin{minipage}[c]{1.0\linewidth}
\includegraphics[width=0.163\linewidth]{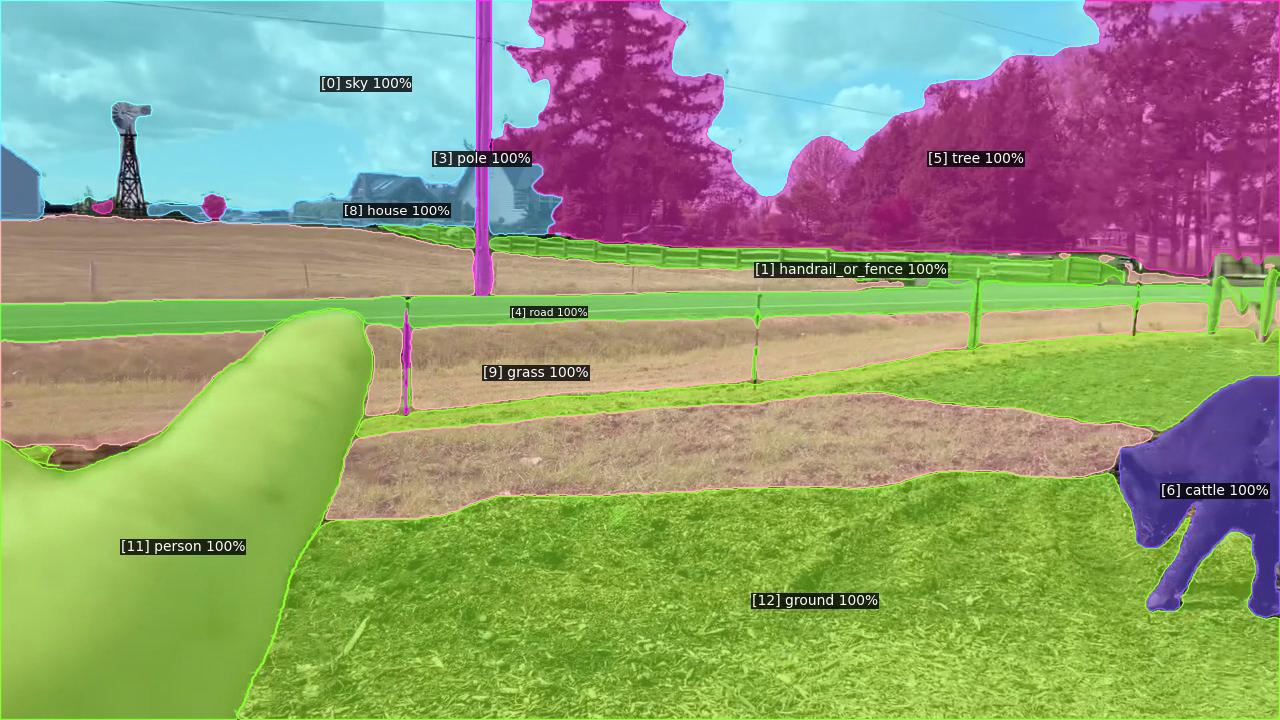}
\includegraphics[width=0.163\linewidth]{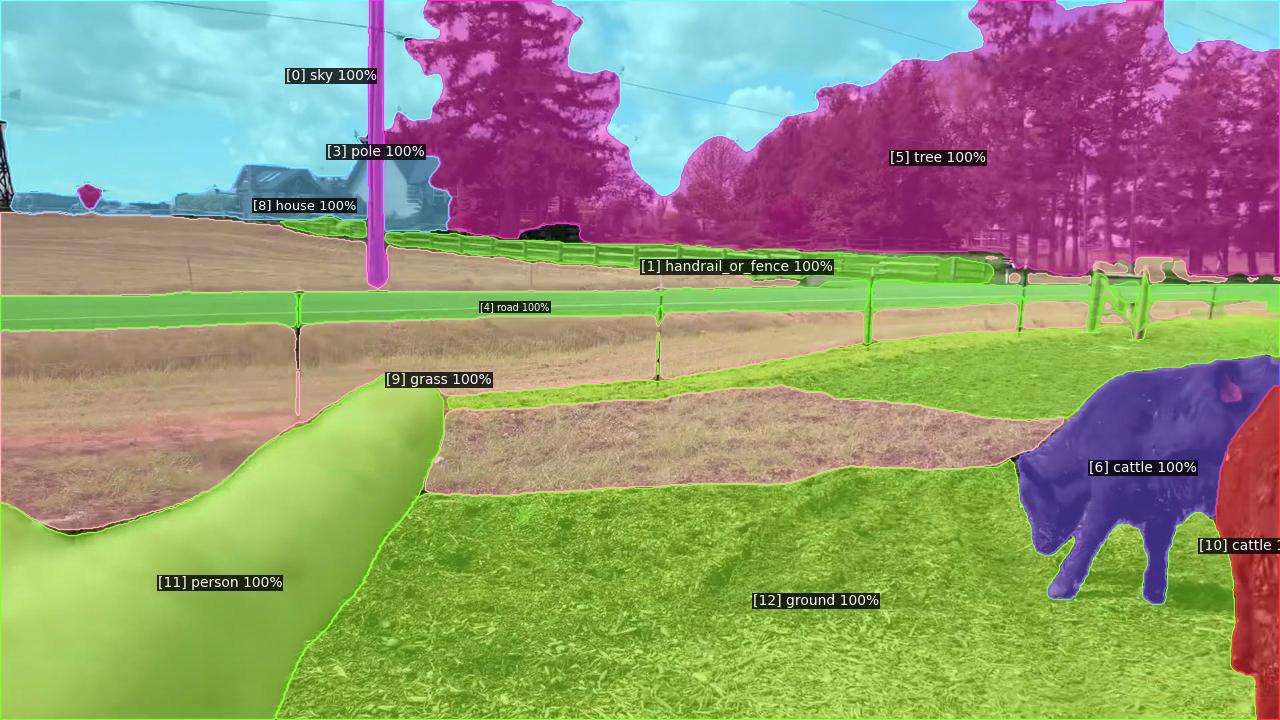}
\includegraphics[width=0.163\linewidth]{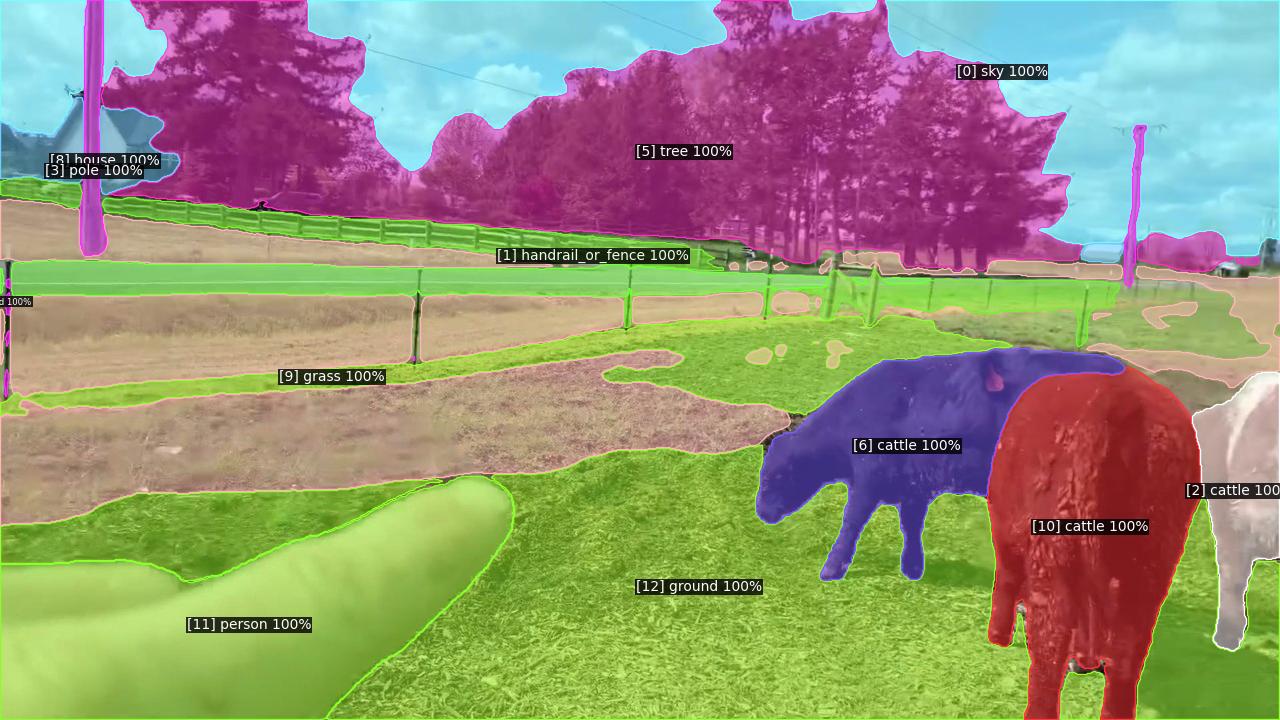}
\includegraphics[width=0.163\linewidth]{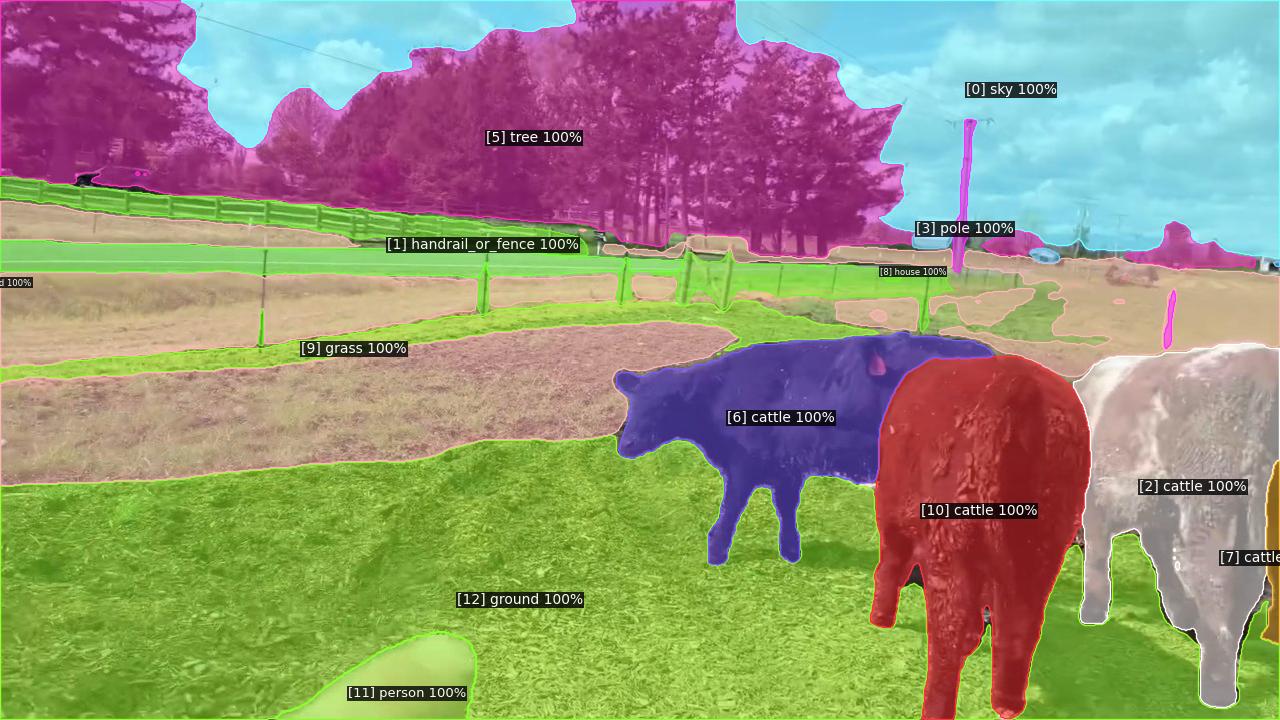}
\includegraphics[width=0.163\linewidth]{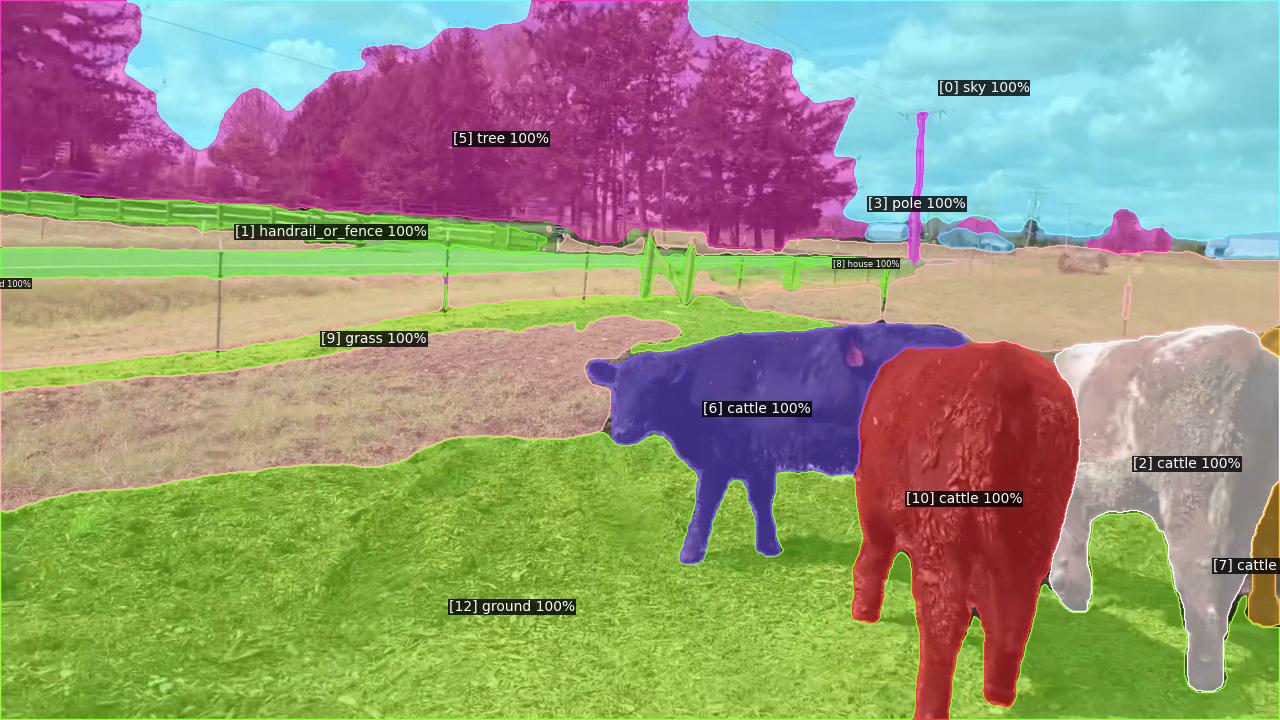}
\includegraphics[width=0.163\linewidth]{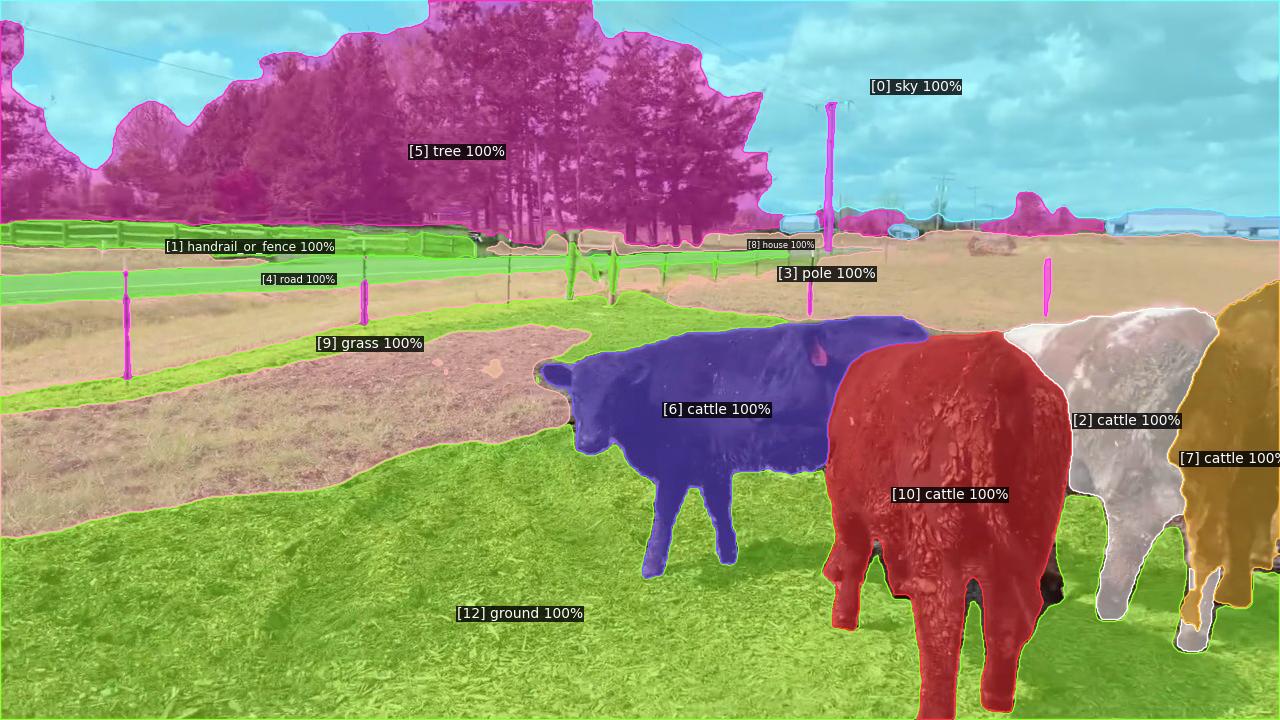}
\end{minipage}\hfill\vspace{1mm}

\begin{minipage}[c]{1.00\linewidth}
\includegraphics[width=0.163\linewidth]{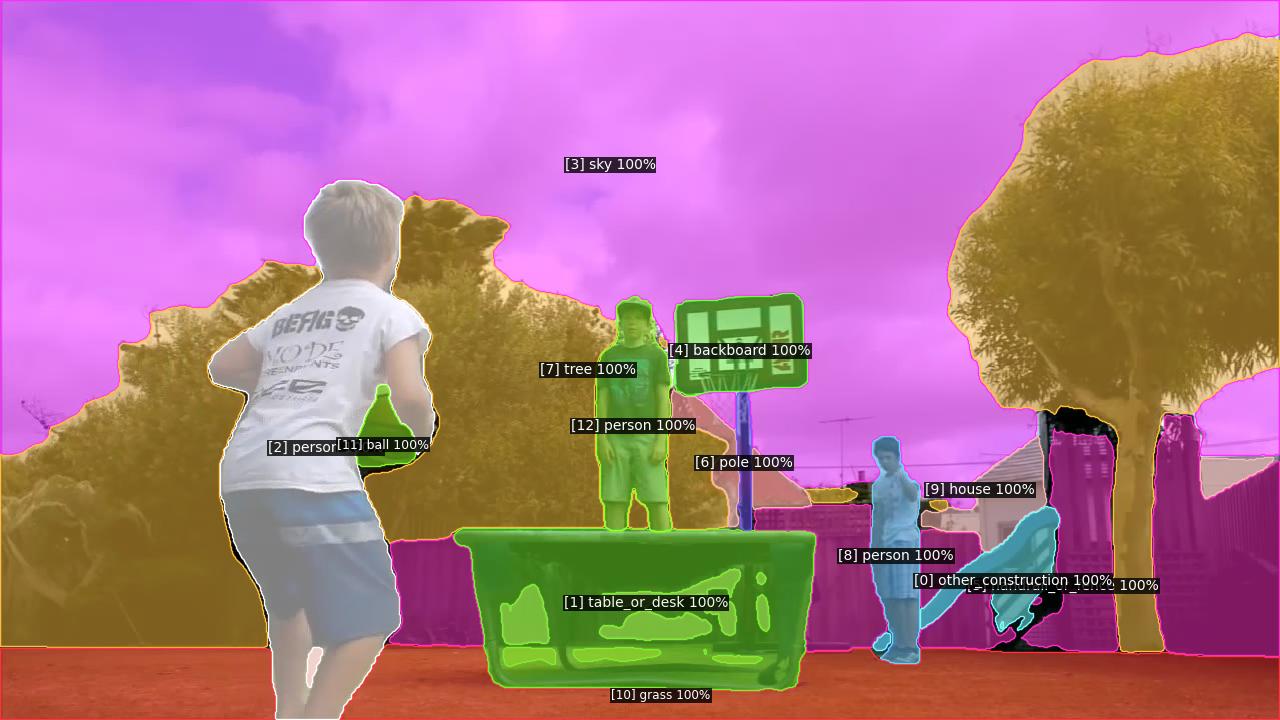}
\includegraphics[width=0.163\linewidth]{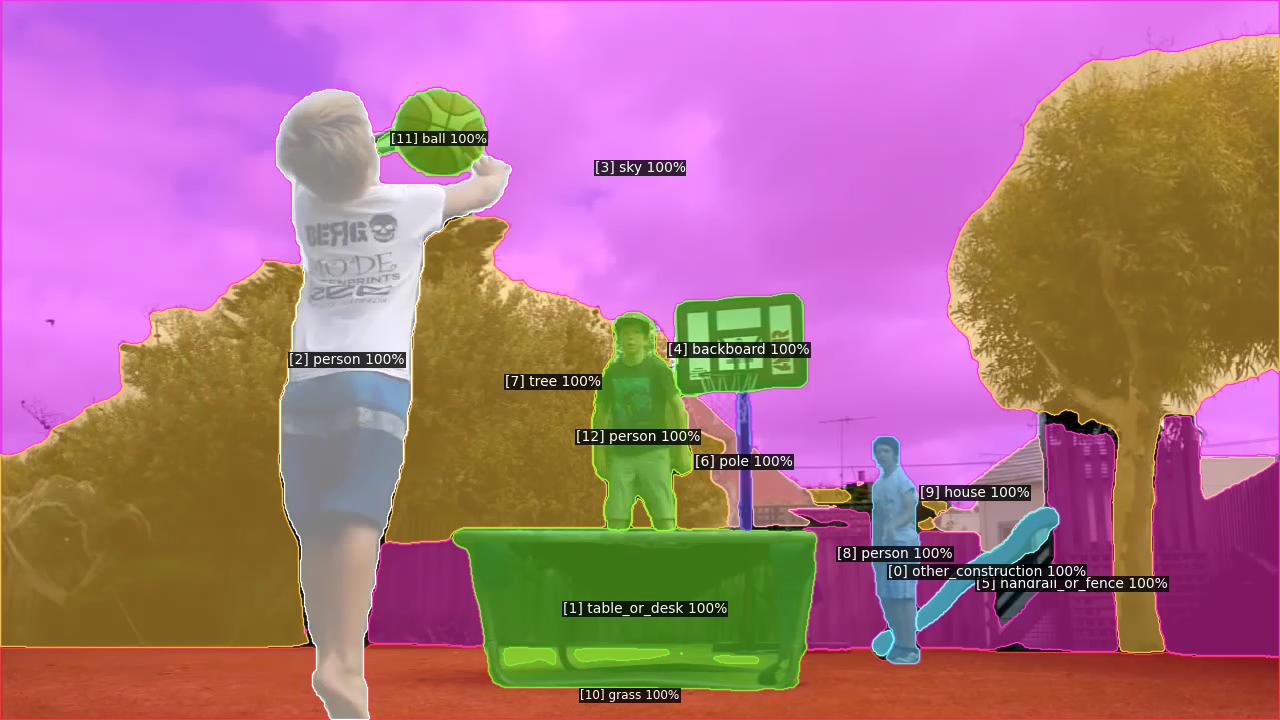}
\includegraphics[width=0.163\linewidth]{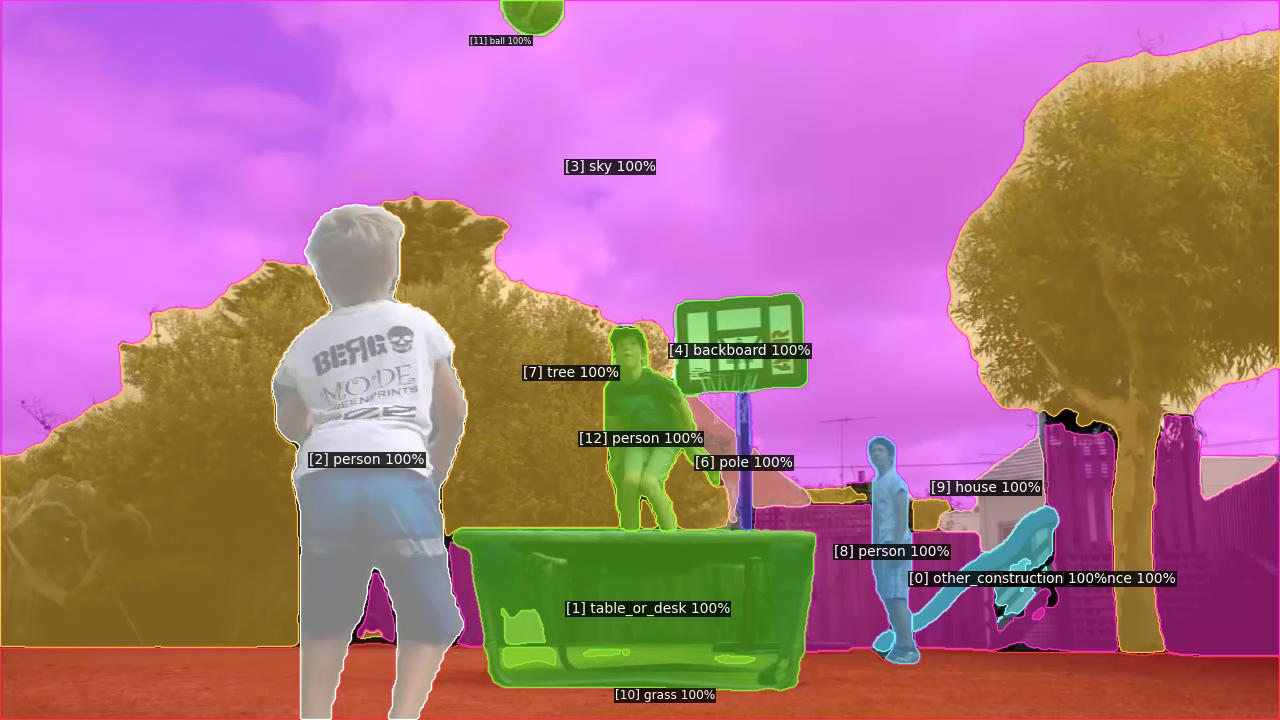}
\includegraphics[width=0.163\linewidth]{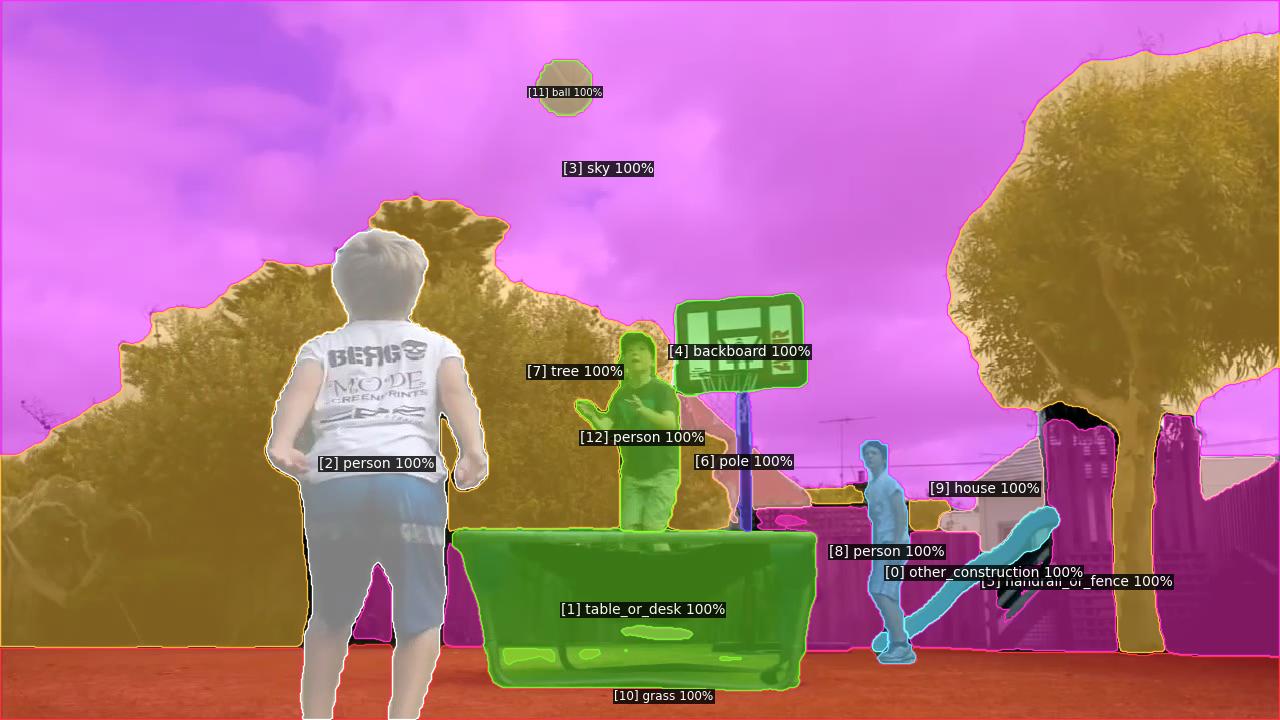}
\includegraphics[width=0.163\linewidth]{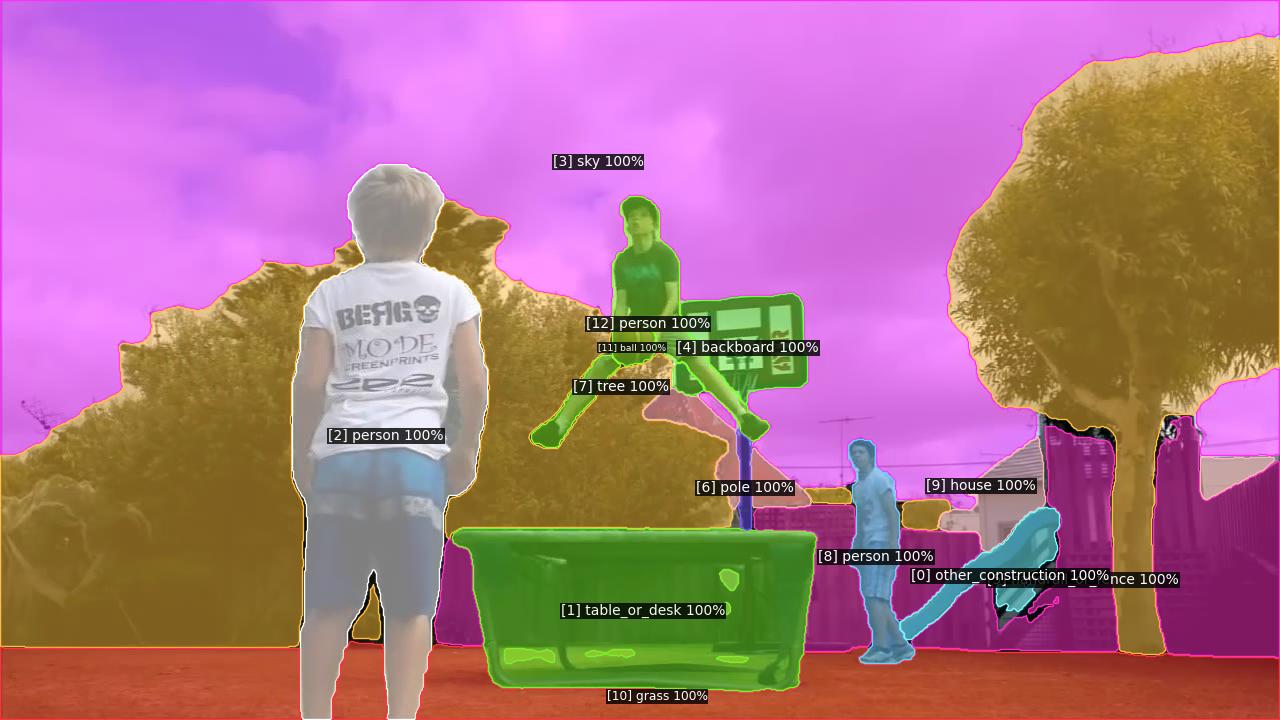}
\includegraphics[width=0.163\linewidth]{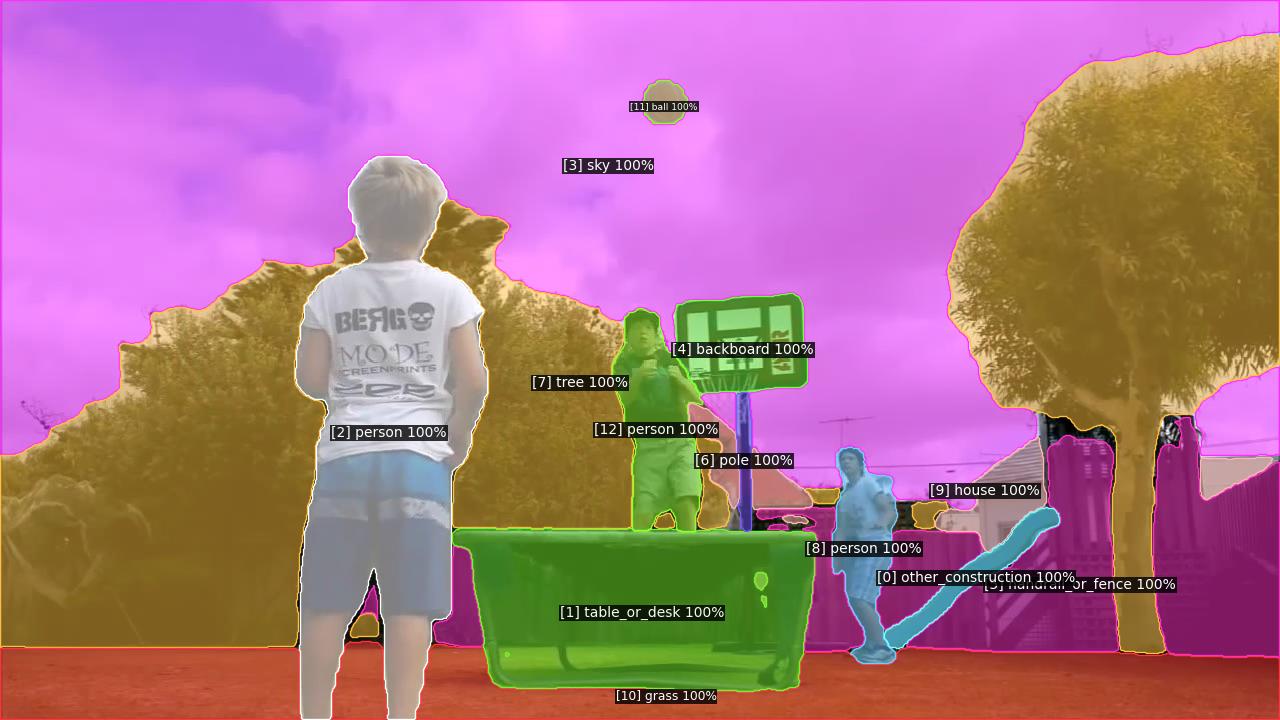}
\end{minipage}\hfill
\begin{minipage}[c]{1.0\linewidth}
\includegraphics[width=0.163\linewidth]{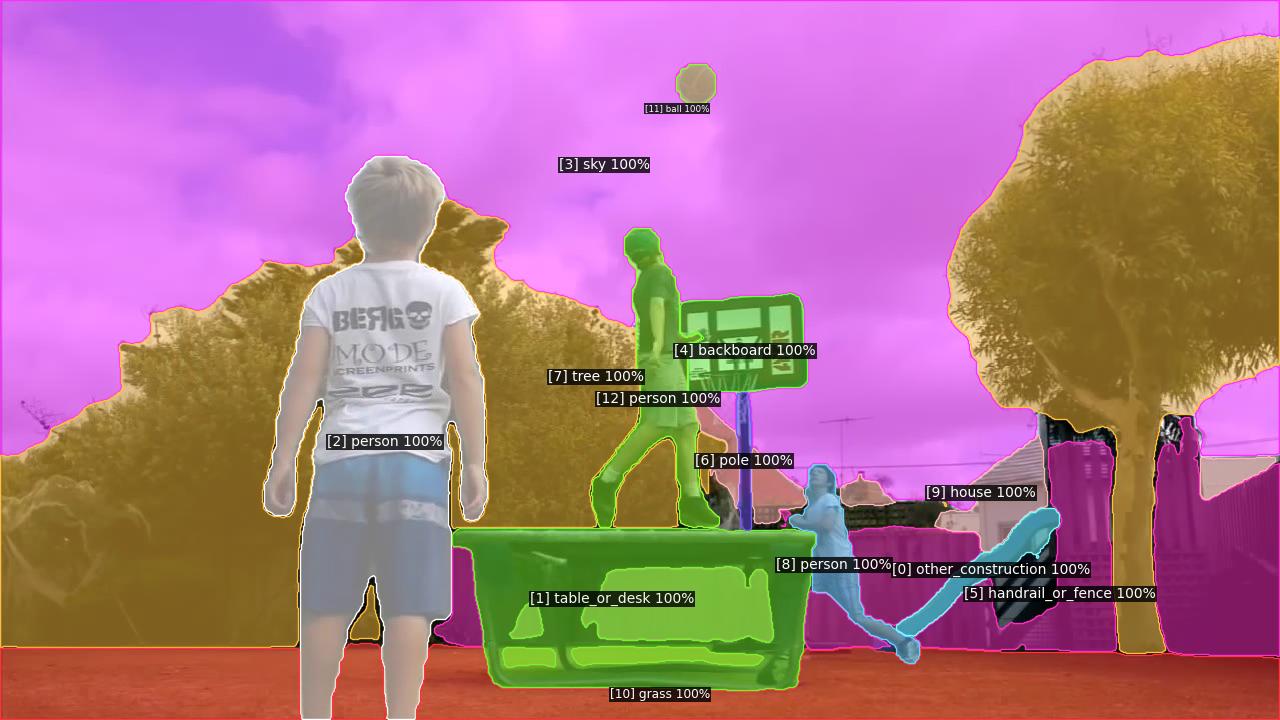}
\includegraphics[width=0.163\linewidth]{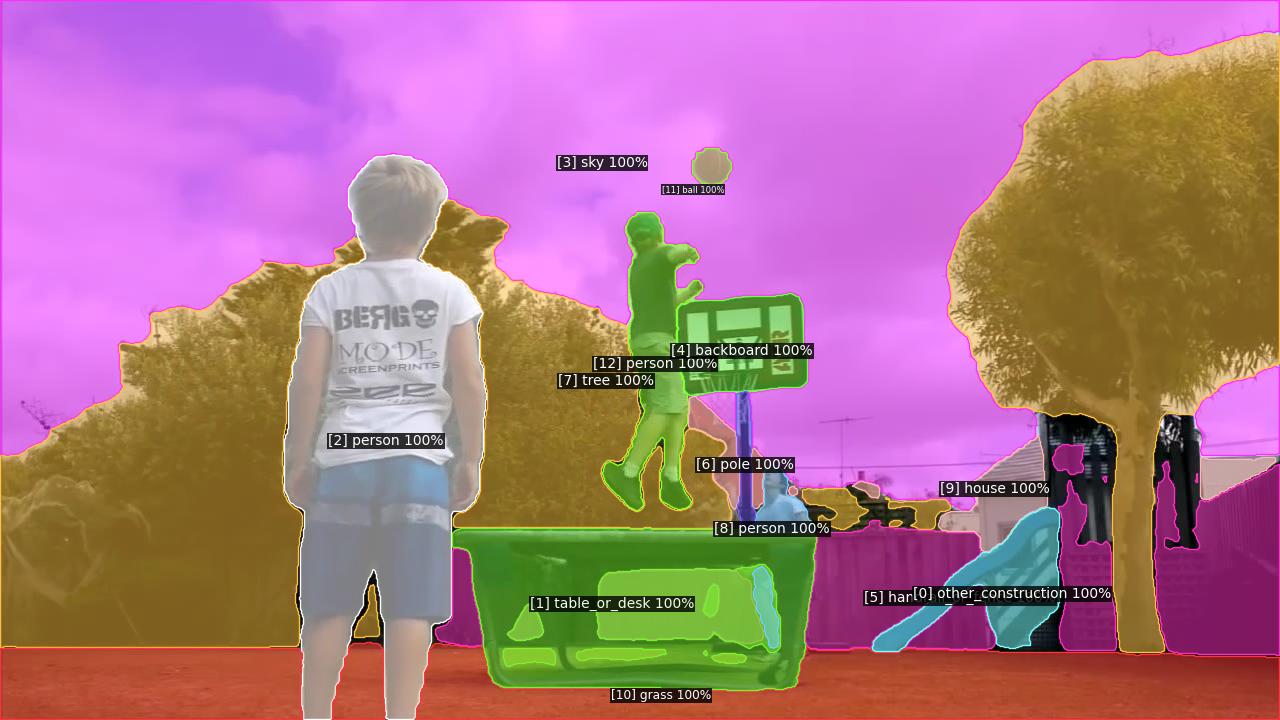}
\includegraphics[width=0.163\linewidth]{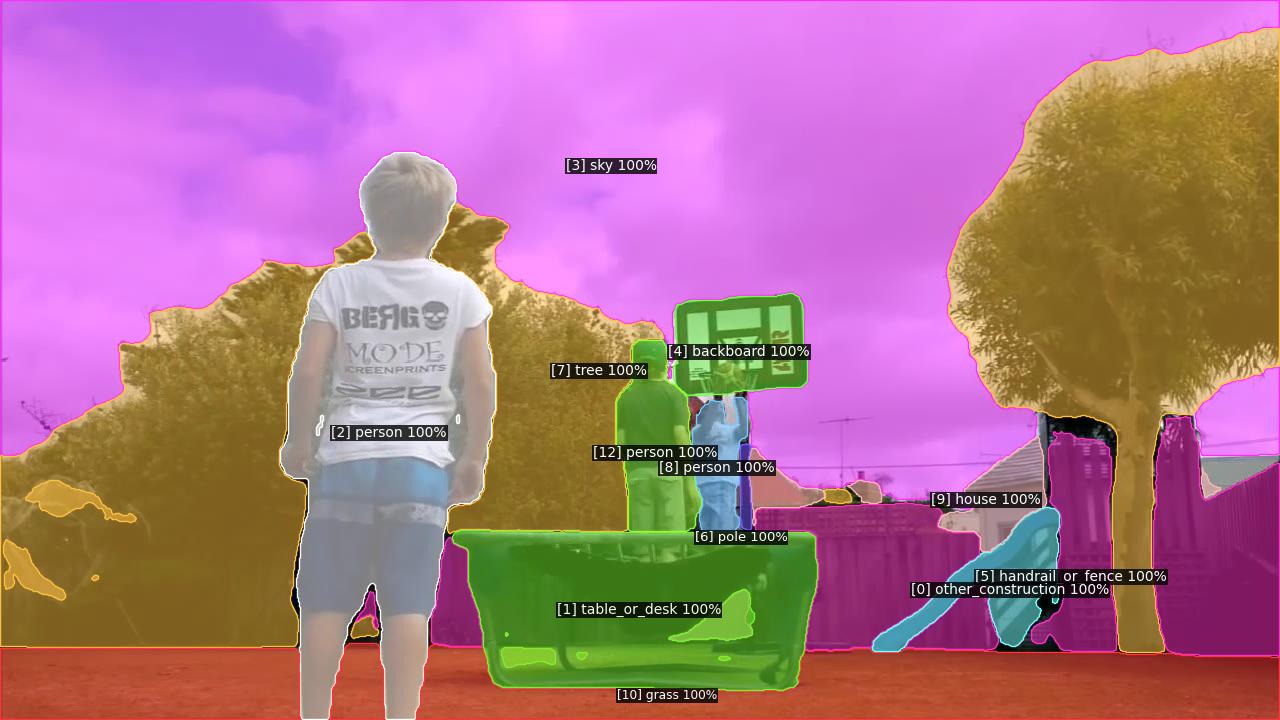}
\includegraphics[width=0.163\linewidth]{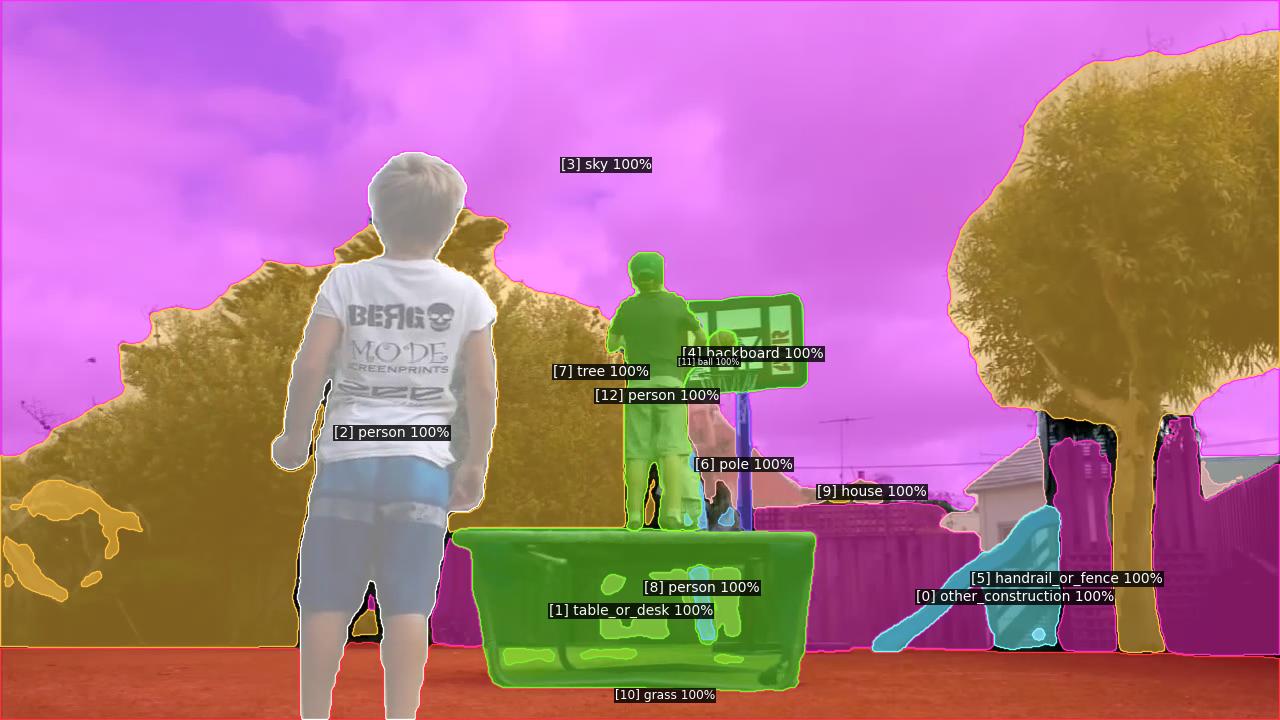}
\includegraphics[width=0.163\linewidth]{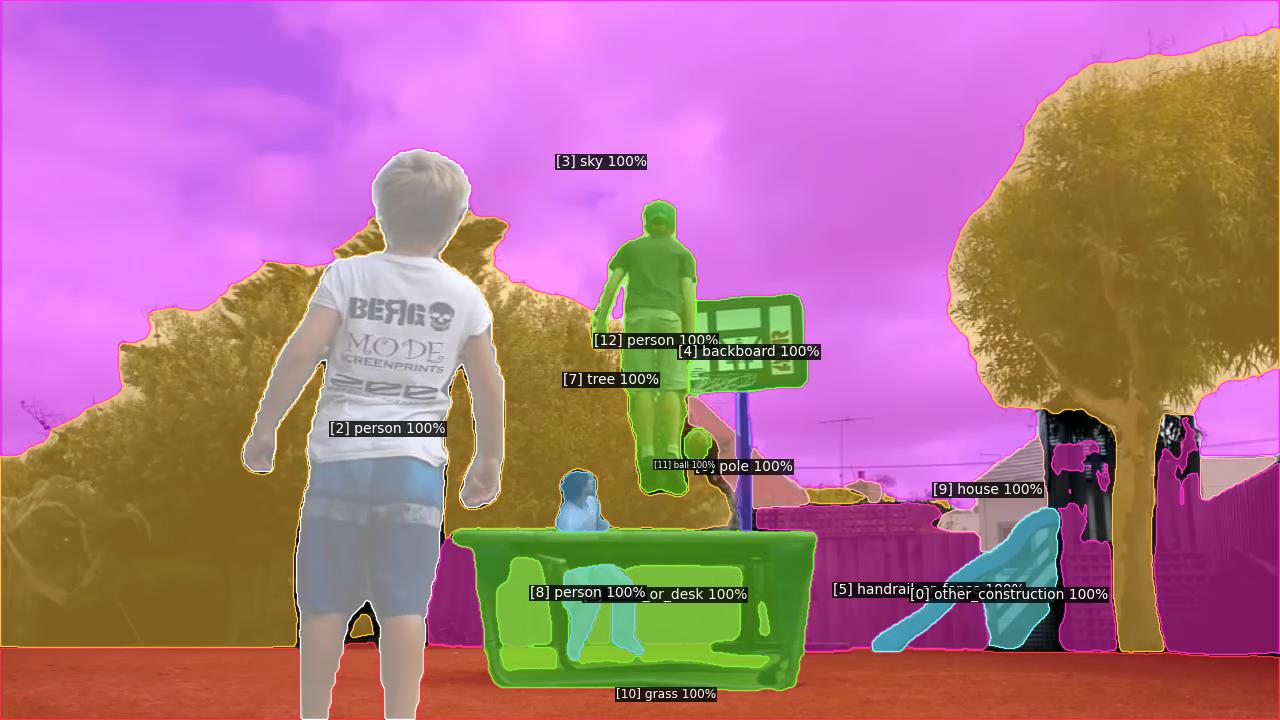}
\includegraphics[width=0.163\linewidth]{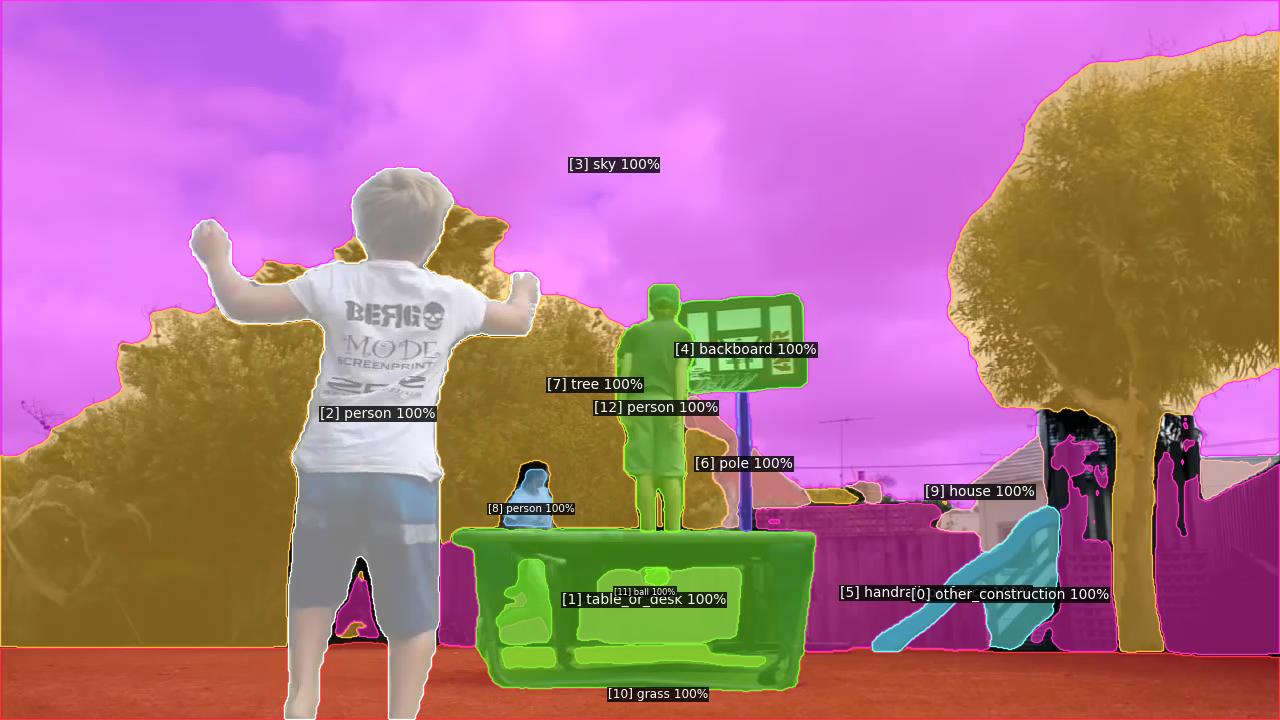}
\end{minipage}\hfill\vspace{1mm}

\begin{minipage}[c]{1.00\linewidth}
\includegraphics[width=0.163\linewidth]{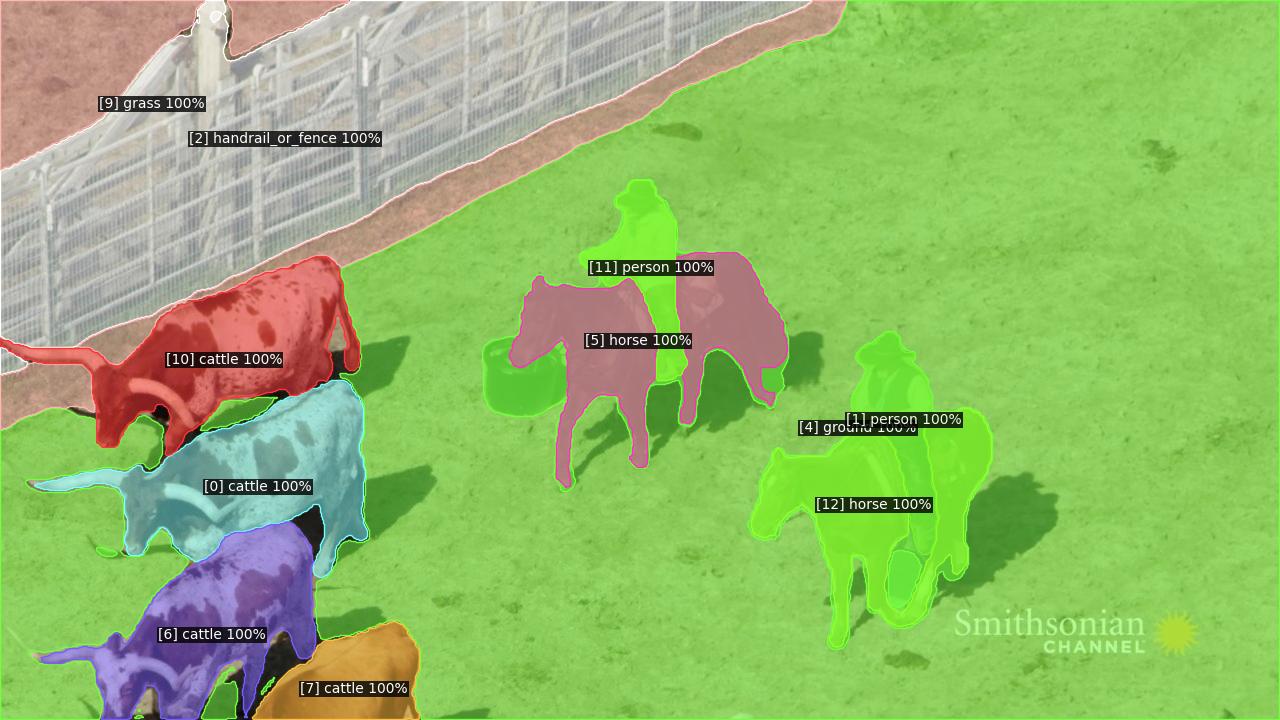}
\includegraphics[width=0.163\linewidth]{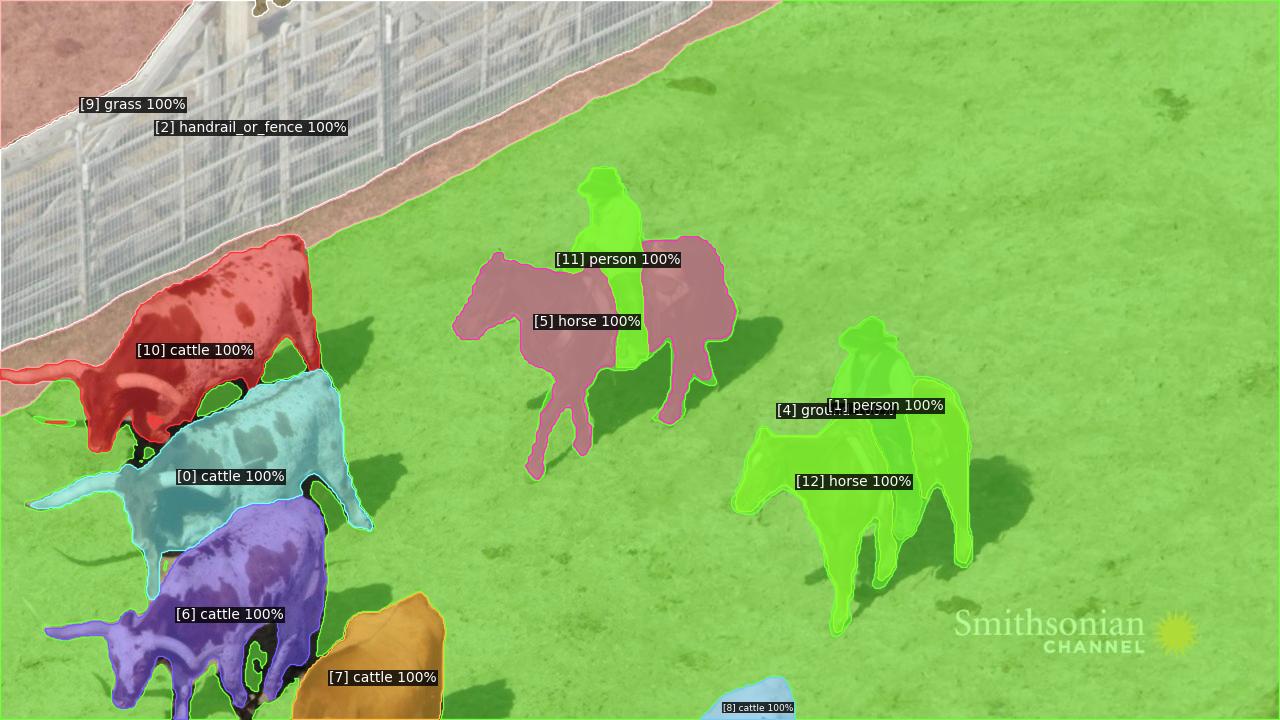}
\includegraphics[width=0.163\linewidth]{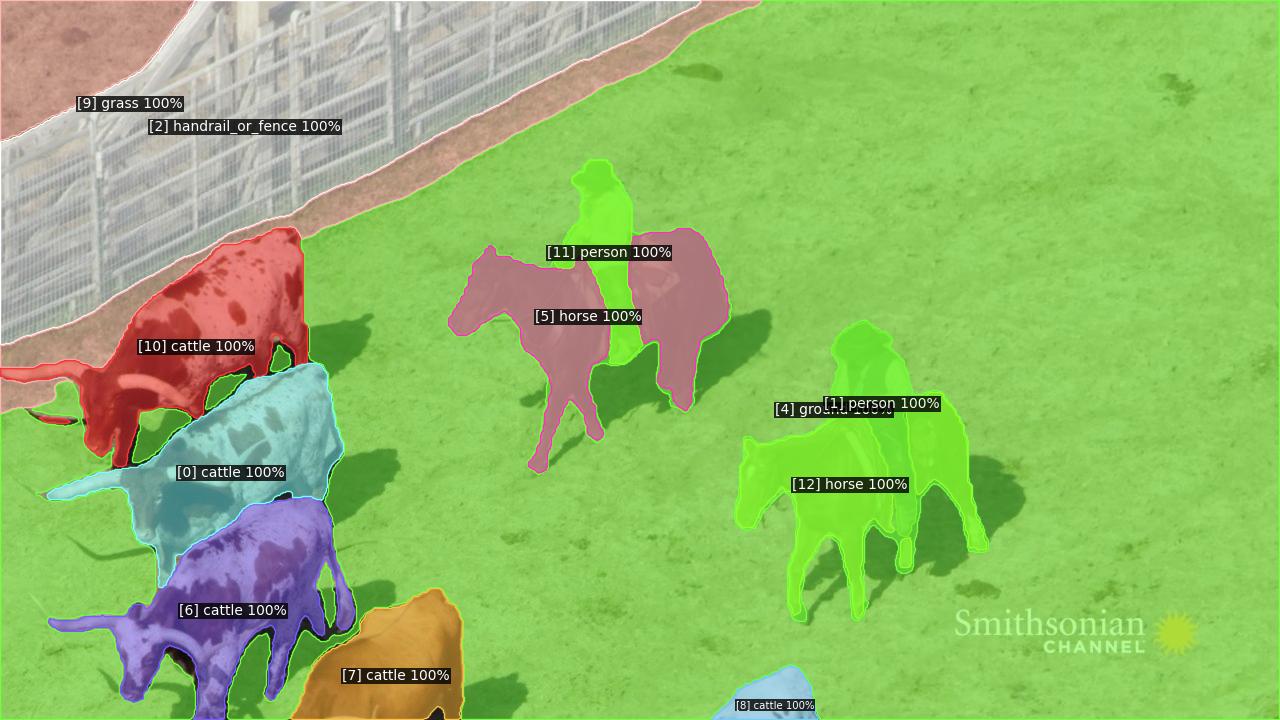}
\includegraphics[width=0.163\linewidth]{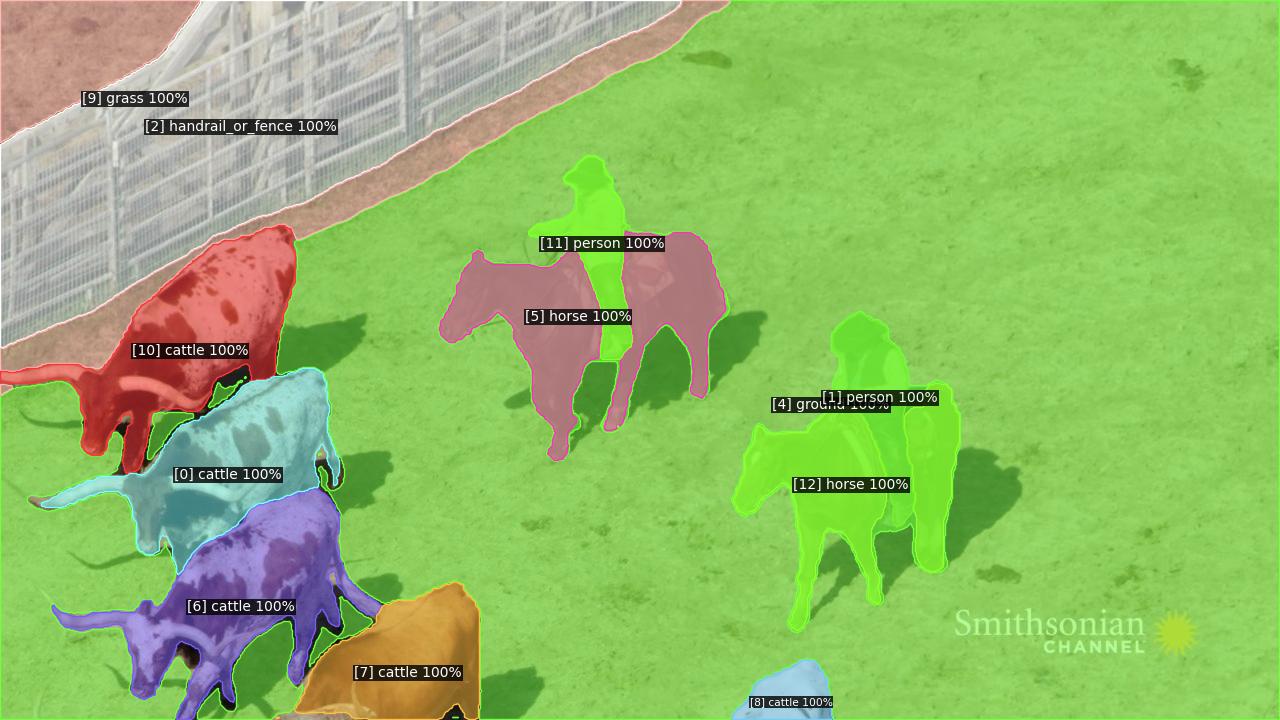}
\includegraphics[width=0.163\linewidth]{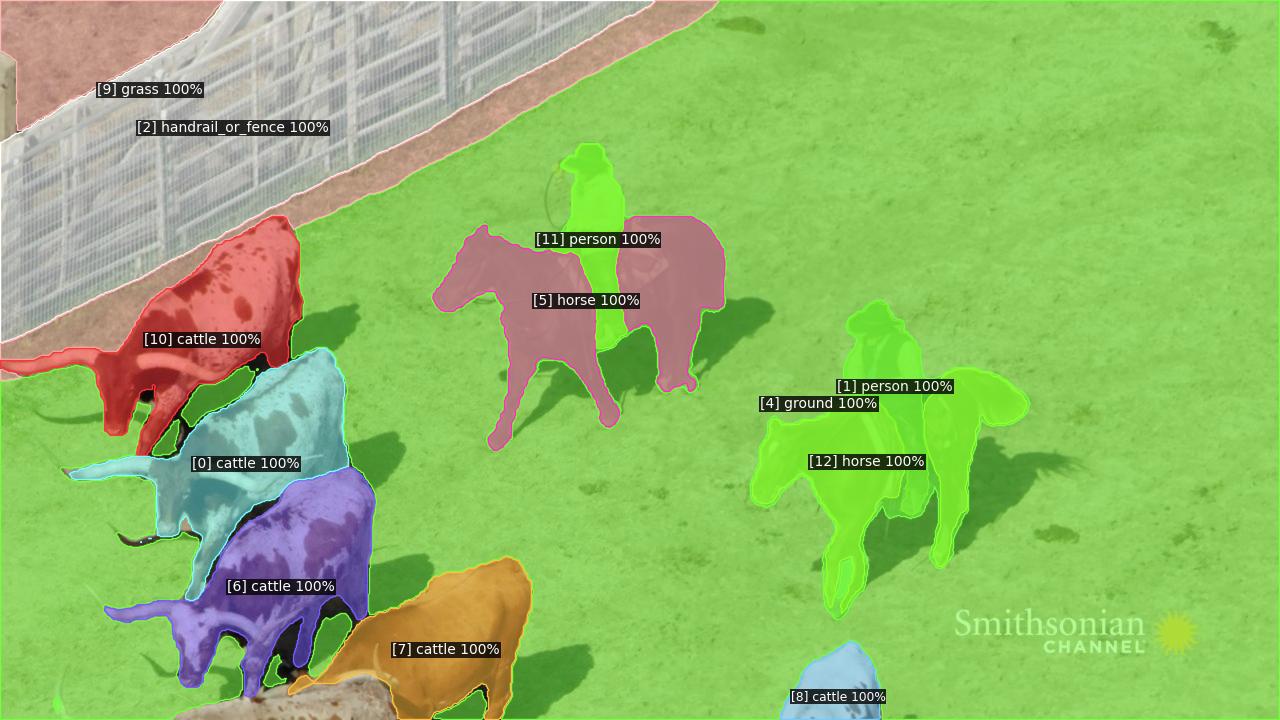}
\includegraphics[width=0.163\linewidth]{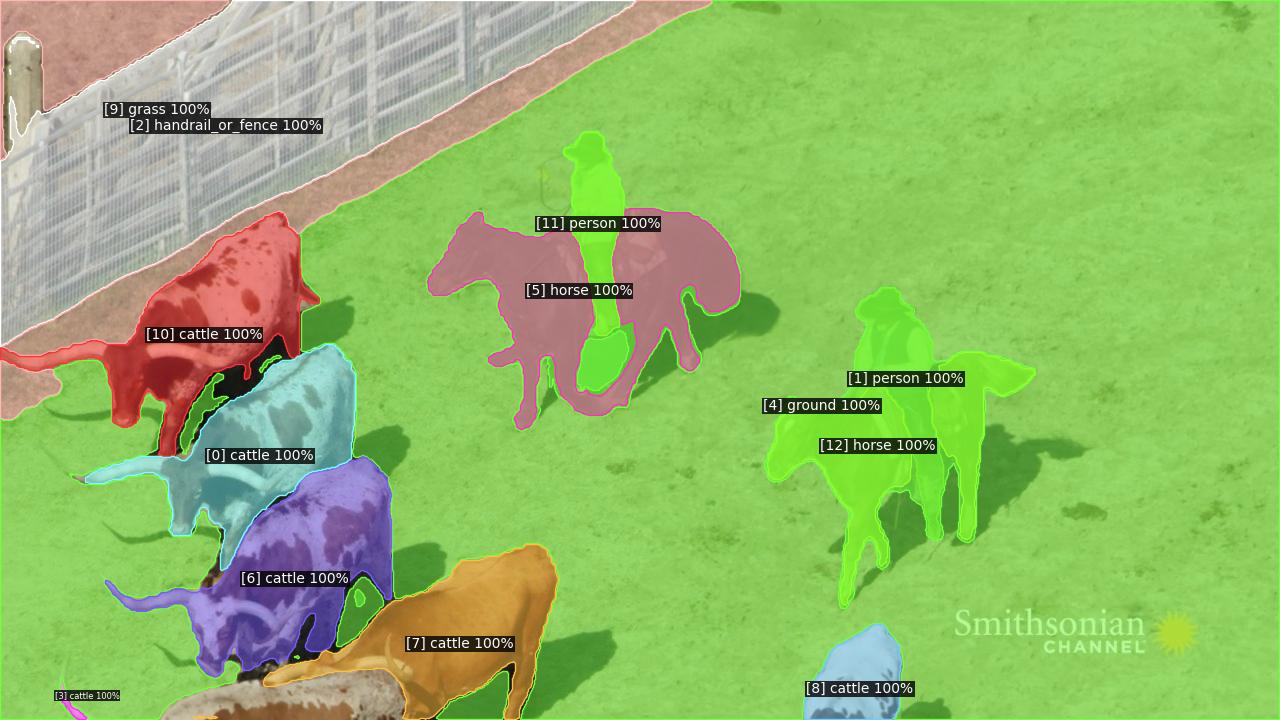}
\end{minipage}\hfill
\begin{minipage}[c]{1.0\linewidth}
\includegraphics[width=0.163\linewidth]{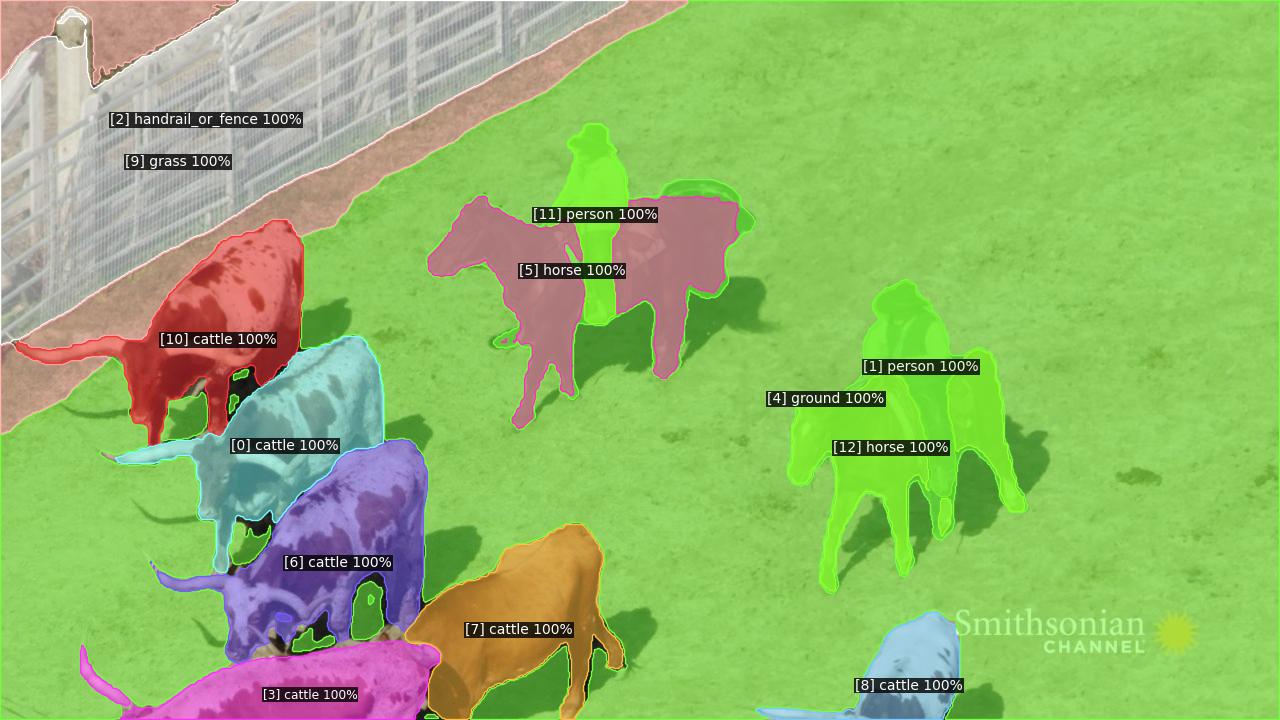}
\includegraphics[width=0.163\linewidth]{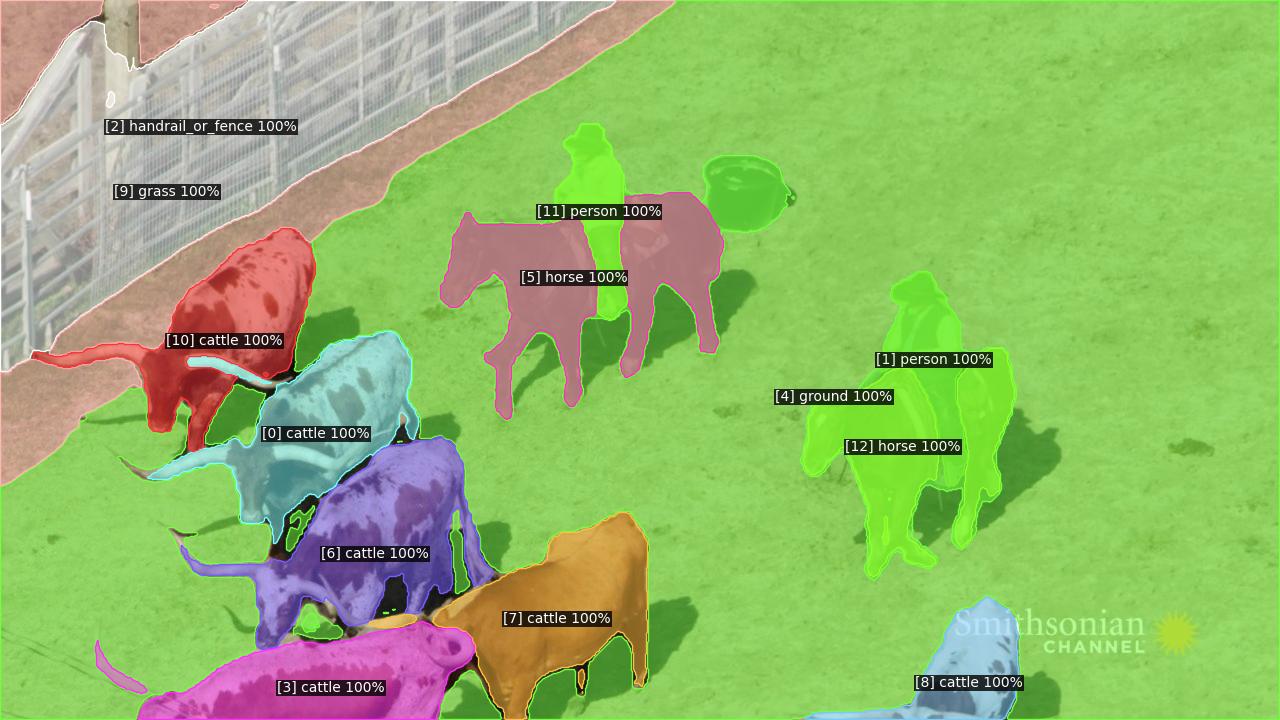}
\includegraphics[width=0.163\linewidth]{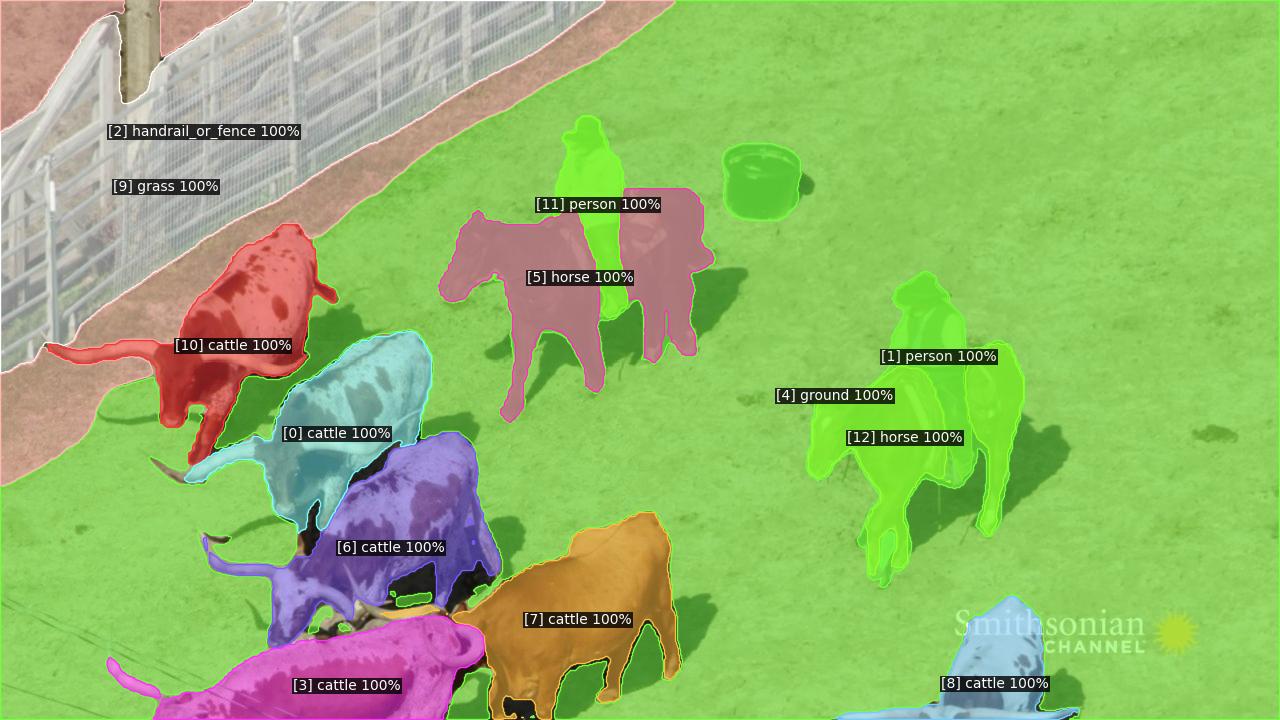}
\includegraphics[width=0.163\linewidth]{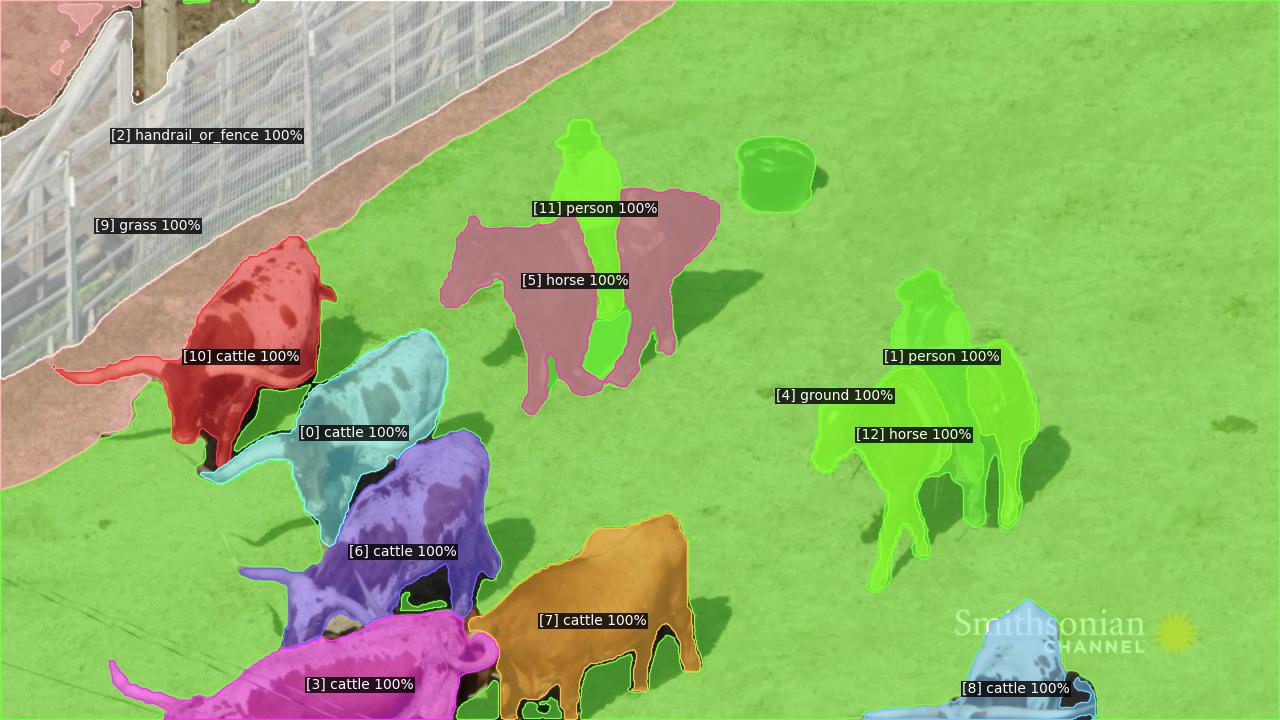}
\includegraphics[width=0.163\linewidth]{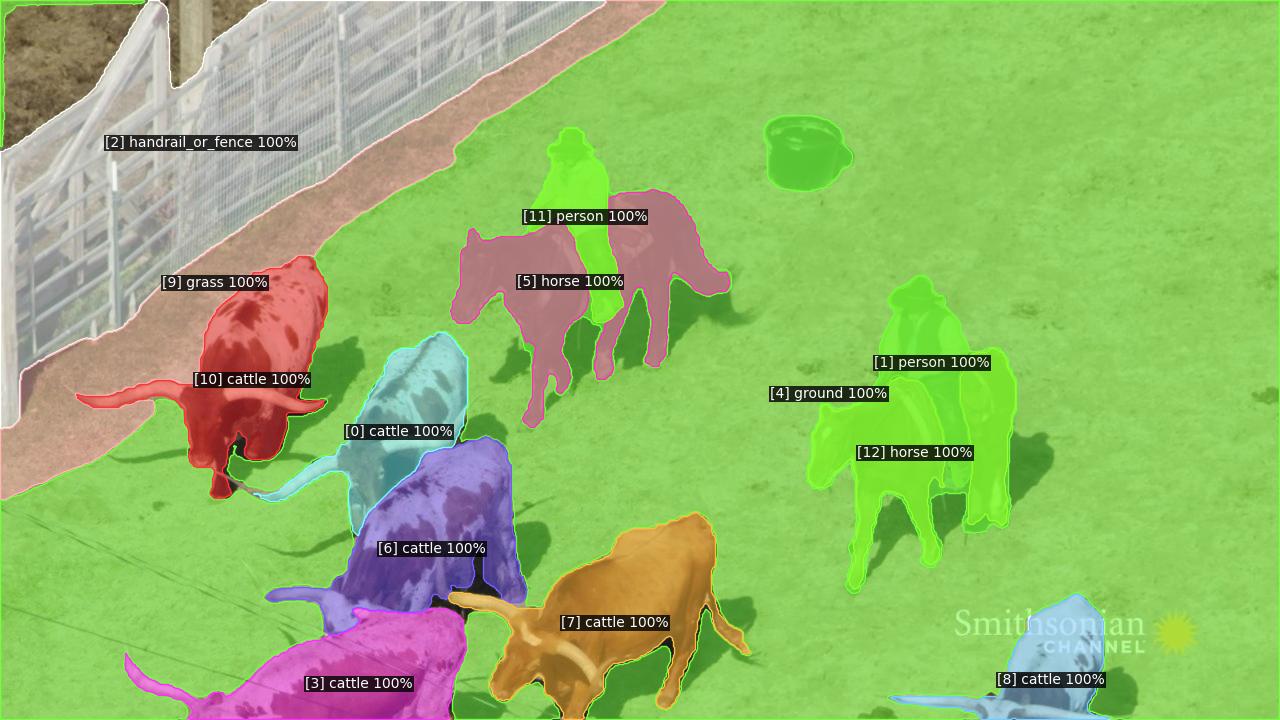}
\includegraphics[width=0.163\linewidth]{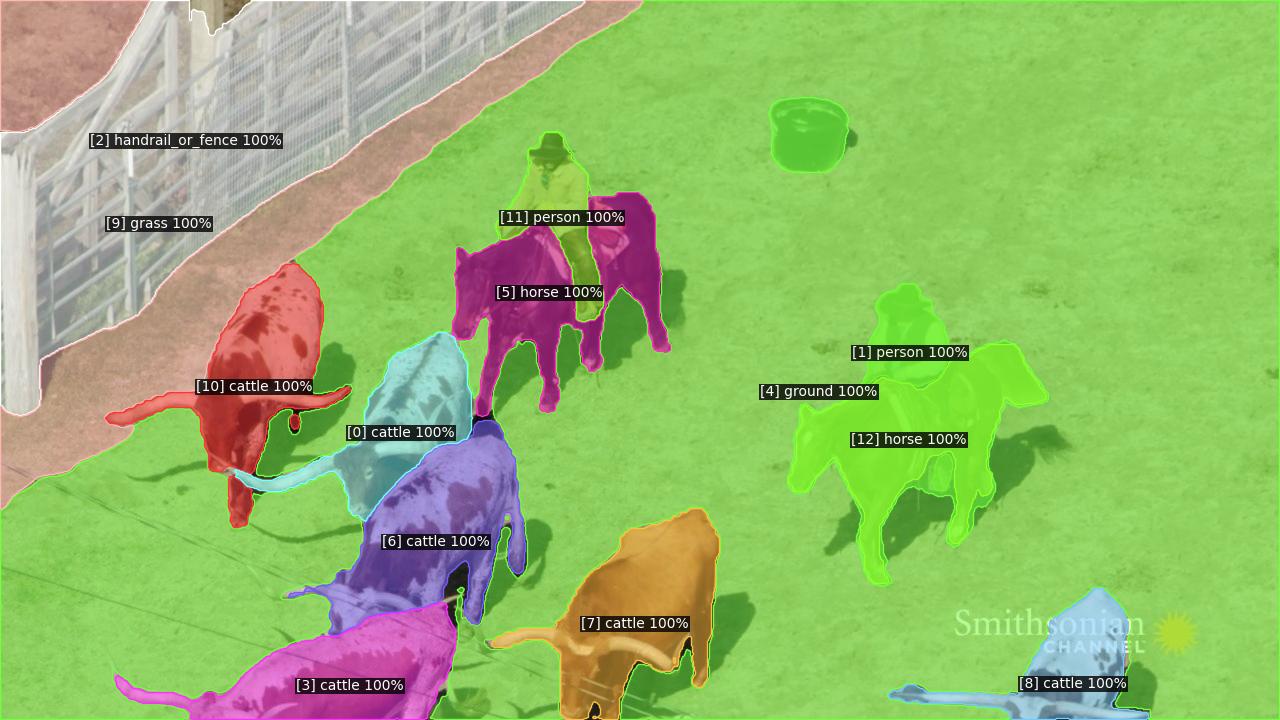}
\end{minipage}\hfill\vspace{1mm}

\caption{\textbf{Visualization results obtained on the VIPSeg dataset.}}
\label{fig:vipseg demo}
\end{figure*}

\end{document}